%% file: discaffkernels-arXiv-v5.tex
\documentclass{siamart1116}

\input{discaffkernels-siam-subm-shared}

\begin{document}

\maketitle

\begin{abstract}
  The affine Gaussian derivative model can in several respects be
regarded as a canonical model for receptive fields over a spatial
image domain: (i)~it can be derived by necessity from scale-space
axioms that reflect structural
properties of the world, (ii)~it constitutes an excellent model
for the receptive fields of simple cells in the primary
visual cortex and
(iii)~it is covariant under affine image deformations,
which enables
more accurate modelling of image measurements under the local image deformations caused by the
perspective mapping, compared to the more commonly used Gaussian
derivative model based on derivatives of the rotationally symmetric Gaussian kernel.


This paper presents a theory for discretizing the affine Gaussian
scale-space concept underlying the affine Gaussian derivative model, so that scale-space properties hold
also for the discrete implementation. 

Two ways of discretizing spatial smoothing with affine Gaussian kernels are presented:
(i)~by solving semi-discretized affine diffusion equation, which has been
derived by necessity from the requirements of a semi-group structure
over scale as parameterized by a family of spatial covariance matrices 
and obeying non-creation of new structures from any finer to
any coarser scale in terms of non-enhancement of local extrema and
(ii)~approximating these semi-discrete affine receptive fields by 
parameterized families of $3 \times 3$-kernels as obtained from an additional
discretization along the scale direction. 
The latter discrete approach
can be optionally complemented by spatial subsampling at coarser scales, leading to
the notion of affine hybrid pyramids.

For the first approach, we show how the solutions can
be computed reasonably efficiently from a closed form expression for the Fourier transform, and 
analyse how a remaining degree of freedom in the theory 
can be explored to ensure a positive discretization and optionally
achieve higher-order discrete approximation of the angular dependency
of the discrete affine Gaussian receptive fields.
For the second approach, we analyse how the step
length in the scale direction can be determined, given the requirements of
a positive discretization and other scale-space properties.

We do also show how discrete directional derivative approximations can
be efficiently implemented to approximate affine Gaussian derivatives.
Using these theoretical results,
we outline hybrid architectures for discrete approximations of affine covariant
receptive field families,
to be used as a first processing layer for affine covariant and affine
invariant visual operations at higher levels.
\end{abstract}



\section{Introduction}

A basic fact when interpreting image information is that a pointwise measurement of the
image intensity $f(x, y)$ at a single image point does usually not
carry any relevant information, since it is dependent on external and
unknown illumination. The essential information is instead
mediated by the relations between the image intensities at
neighbouring points. To handle this issue, the notion of receptive
fields has been developed in both biological vision and computer
vision, by performing local image measurements over neighbourhoods of
different spatial extent
(Hubel and Wiesel \cite{HubWie59-Phys,HubWie05-book};
 DeAngelis {\em et al.\/}\
 \cite{DeAngOhzFre95-TINS,deAngAnz04-VisNeuroSci};
 Conway and Livingstone \cite{ConLiv06-JNeurSci};
 Johnson  {\em et al.\/}\ \cite{JohHawSha08-JNeuroSci};
 Shapley and Hawken {\em et al.\/}\ \cite{ShaHaw11-VisRes};
 Koenderink and van Doorn \cite{KoeDoo87-BC,KoeDoo92-PAMI};
 Young {\em et al.\/}\ \cite{You87-SV,YouLesMey01-SV};
 Lindeberg \cite{Lin93-Dis,Lin13-BICY,Lin13-PONE};
 Schiele and Crowley \cite{SchCro00-IJCV};
 Lowe \cite{Low04-IJCV}; 
 Dalal and Triggs \cite{DalTri05-CVPR};
 Serre {\em et al.\/}\ \cite{SerWolBilRiePog07-PAMI};
 Bay {\em et al.\/}\ \cite{BayEssTuyGoo08-CVIU};
 Burghouts and Geusebroek \cite{BurGeu09-CVIU};
 van de Sande {\em et al.\/}\ \cite{SanGevSno10-PAMI};
 Tola {\em et al.\/}\ \cite{TolLepFua10-PAMI};
 Linde and Lindeberg \cite{LinLin12-CVIU};
 Larsen {\em et al.\/}\ \cite{LarDarDahPed12-ECCV};
 Krizhevsky {\em et al.\/}\ \cite{KriSutHin12-NIPS};
 Simonyan and Zisserman \cite{SimZis15-ICLR};
 Szegedy {\em et al.\/}\ \cite{SzeLiuJiaSerReeAngErhVanRab15-CVPR};
 He {\em et al.\/}\ \cite{HeZhaRenSun16-CVPR}).

A fundamental problem in this context concerns what shapes of
receptive field profiles are meaningful. Would any shape of the
receptive field do? This problem has been addressed from an
axiomatical viewpoint in the area of scale-space theory, by deriving families of
receptive fields that obey scale-space axioms that reflect structural
properties of the world. Over an isotropic spatial domain, arguments
and axiomatic derivations by several researchers have shown that the
rotationally symmetric Gaussian kernel and Gaussian derivatives derived from it constitute a
canonical family of kernels to model spatial receptive fields
(Iijima \cite{Iij62-TR}; Koenderink \cite{Koe84-BC};
 Koenderink and van Doorn \cite{KoeDoo87-BC,KoeDoo92-PAMI};
 Lindeberg \cite{Lin93-Dis,Lin10-JMIV};  Florack \cite{Flo97-book}; 
 ter Haar Romeny \cite{Haa94-GDDbook}; 
 Weickert {\em et al.\/}\ \cite{WeiIshImi99-JMIV}).
Such Gaussian derivatives can in turn be used as primitives for
expressing a large class of visual modules in computer vision,
including feature detection, feature classification, surface shape,
image matching and object recognition.

The underlying assumption about spatial isotropy that underlies the part of the scale-space theory
that leads to receptive fields based on rotational symmetric Gaussian
kernels is, however, neither necessary nor always desirable. 
For images of objects that are subject to variations in the viewing
direction, the perspective mapping leads to non-isotropic perspective
transformations that are not within the group of image transformations
covered by rotationally symmetric Gaussian kernels.
If we locally at every image point approximate the non-linear perspective
mapping by its derivative, then the effect of the perspective mapping can
for locally smooth surfaces to first order of approximation be
approximated by local affine image deformations.
If we would like to base a vision system on a receptive field model
that is closed under such affine transformations, we should therefore
replace the rotationally symmetric Gaussian kernels by affine Gaussian
kernels (Lindeberg \cite{Lin93-Dis}; Lindeberg and G{\aa}rding \cite{LG94-ECCV,LG96-IVC}).
Interestingly, the family of affine Gaussian kernel and affine
Gaussian derivatives can also be uniquely derived from basic
scale-space axioms that reflect structural properties of the world in
combination with a requirement about internal consistency between
image representations at multiple spatial scales
(Lindeberg \cite{Lin10-JMIV}).

The subject of this article is to address the problem of how to
discretize receptive fields based on affine Gaussian derivatives, while
preserving scale-space properties also in a discrete sense.
This work constitutes a continuation of a previously developed
scale-space theory for discrete signals and images 
(Lindeberg \cite{Lin90-PAMI,Lin93-Dis,Lin93-JMIV}), which was
conceptually extended to a discrete affine scale-space in a conference
publication (Lindeberg \cite{Lin97-ICSSTCV}), while leaving out
several details that are needed when implementing the theory in practice. 
The work is also closely related to other work on discretizing
scale-space operations as done by 
Deriche \cite{Der87-IJCV},
Weickert {\em et al.\/} \cite{Wei98-book,WeiRomHaaVie98-TIP}, 
van~Vliet {\em et al.\/}\ \cite{VliYouVer98-PR},
Wang \cite{Wan00-TIP},
Florack \cite{Flo00-PAMI},
Almansa and Lindeberg \cite{AL00-IP},
Lim and Stiehl \cite{LimSti03-ScSp},
Geusebroek {\em et al.\/}\ \cite{GeuSmeWei03-TIP},
Farid and Simoncelli \cite{FarSim04-TIP},
Bouma {\em et al.\/}\ \cite{BouVilBesRomGer07-ScSp},
Doll{\'a}r {\em et al.\/} \cite{DolAppBelPer14-PAMI},
Tschirsich and Kuijper \cite{TscKui15-JMIV},
Slav{\'\i}k and Stehl{\'\i}k \cite{SlaSte15-JMathAnalAppl},
Lindeberg \cite{Lin16-JMIV} and others.

We will show that by axiomatic arguments it is possible to derive a
three-parameter family of semi-discrete affine Gaussian receptive
fields, where the receptive fields are discretized over image space, 
while being continuous over scale as parameterized by a family
of spatial covariance matrices that represent receptive fields of
different sizes and orientations in the image domain as well as different
eccentricities in terms of the ratio between the eigenvalues of the
spatial covariance matrix.
The resulting family of discrete affine Gaussian receptive fields
obeys discrete analogues of several of the properties
that define the uniqueness of the continuous affine Gaussian kernel
over a continuous spatial domain.

For both theoretical and practical purposes, we will show that these receptive fields can
up to numerical errors in an actual implementation be computed exactly
in terms of a closed-form expression of the Fourier transform, thus
without explicit need for solving the PDE that describes the evolution
over scale using approximate methods.
If convolution by FFT can be regarded sufficiently computationally
efficient, then this discretization method gives the numerically most accurate
approximation of the continuous affine scale-space concept,
and also allowing for a full transfer of the underlying scale-space
properties to the discrete implementation.

It will also be shown that if this theory is additionally discretized
over scale, then this leads to families of compact $3 \times 3$-kernels 
that can be interpreted as discretizations of affine Gaussian kernels,
and which are to be applied repeatedly.
Such kernels can be efficiently implemented on parallel architectures
such as GPUs. This model will be developed in two versions:
(i)~a conceptually simpler model that preserves the same resolution at
all levels of scale and (ii)~a hybrid pyramid model that enables
higher computational efficiency by combining the iterative spatial
smoothing operations with subsampling stages at coarser levels of
scale, implying that less computations will be needed because of the
resulting lower number of image pixels at coarser resolutions and the ability to take larger
scale steps.

The latter type of model does specifically
have an interesting structural relationship to recent developments 
in deep learning, where deep networks, such as VGG-Net 
(Simonyan and Zisserman \cite{SimZis15-ICLR}) and
ResNet (He {\em et al.\/}\ \cite{HeZhaRenSun16-CVPR}), are based
on automated learning of compact $3 \times 3$ kernels that are
applied in a repeated layered structure to the image data.

With the theory presented in this paper, we aim at opening up for
relating such kernels to scale-space operations, with the possibilities of
building closer links between the empirical developments in deep
learning and axiomatically derived scale-space theory with its in turn
close relations to receptive fields in biological vision.
A long term goal is to open up for hybrid architectures that
combine ideas and concepts from scale-space theory, pyramid
representation and deep learning,
to allow for cross-fertilization between these areas.

A specific aim is to replace
the training of the earliest
layers in deep networks by theoretically derived scale-space filters that obey the algebra of
affine Gaussian receptive fields, to enable affine covariance of the
deep network, to in turn enable provably affine invariant recognition 
methods based on deep learning that are able to
work more robustly under local image deformations caused by
perspective effects.

Specifically, we propose that the computational architectures developed in this paper, in
terms of directional derivatives of different orders of spatial differentiation, in different
orientations in image space and over variations in the sizes and the
shapes (eccentricities) of affine Gaussian kernels, with their conceptually very
large similarities to the spatial receptive fields of simple cells in 
the primary visual cortex (V1), can be used as the first
layers of input to both classical type computer vision/image analysis methods and deep networks.

\subsection{Structure of this article}

This article is organized as follows:
Section~\ref{sec-aff-rec-fields-overview} gives an overview of the
affine Gaussian derivative model for spatial receptive fields, with
its properties of:
(i)~being determined by necessity from scale-space axioms that reflect
structural properties of the world,
(ii)~being
able to handle local perspective image deformations in a covariant
manner and (iii)~having close similarities to receptive fields in
biological vision.

Section~\ref{sec-disc-aff-gauss-rec-fields} gives an overview of basic
approaches for discretizing the continuous model for affine Gaussian receptive fields over a
spatial domain, with specific mention of the different types of
properties that different spatial discretizations lead to.
Section~\ref{sec-theory-disc-aff-rec-fields} then describes a genuine
discrete theory for defining discrete analogues of affine Gaussian
receptive fields in such a way that the discrete receptive fields exactly
obey discrete analogues of several of the special properties that the
continuous affine Gaussian kernels obey.

Section~\ref{sec-theory-disc-aff-rec-fields-2D} gives a deeper
treatment of this topic for the specific case of a two-dimensional
image domain, based on a semi-discrete affine diffusion equation
that defines semi-discrete affine Gaussian kernels over a discretized
spatial image domain, while maintaining a continuum of affine
scale-space representations over a three-parameter family of affine
covariance matrices that represent the shapes of spatial kernels with different
sizes and orientations in the image domain and with different eccentricities in terms
of the ratios between the eigenvalues of the covariance matrix.

Section~\ref{sec-theory-disc-aff-rec-fields-3x3} then discretizes the
above semi-discrete theory additionally over scale, in terms of a discrete set of
spatial covariance matrices for which receptive fields are to be
computed, and shows that this leads to families of compact 
$3 \times 3$-kernels, which allow for efficient implementation on
parallel architectures such as GPUs.

Section~\ref{sec-aff-hybr-pyr} in the appendix additionally combines that theory with
the notion of pyramid representation, and shows how families of affine hybrid
pyramid representations can be defined from the discrete approach,
with the additional constraint that receptive field responses at coarser
spatial scales may be represented at coarser spatial resolution, to
reduce the computational work and the memory requirements.

Section~\ref{sec-sc-norm-ders-disc} shows how scale-normalized
derivatives can be defined for the presented discrete approaches.
Finally, Section~\ref{sec-summ-disc} gives a summary and discussion
about some of the main results and describes how the presented theory
opens up for different types of architectures for affine covariant and
affine invariant visual receptive fields.

To not load the main flow of article with too much technical
material, we have put the axiomatic derivation of the discrete affine
scale-space concept, which constitutes the theoretical foundation for
the treatment in Section~\ref{sec-theory-disc-aff-rec-fields}
as well as the later sections that build upon this material,
in Appendix~\ref{sec-app-disc-scsp}.
A condensed summary of the necessary material resulting from the axiomatic treatment
is given in Section~\ref{sec-summ-axiom-disc-scsp}.

\section{The affine Gaussian derivative model for receptive fields over a continuous image domain}
 \label{sec-aff-rec-fields-overview}

This section gives an overview of the affine Gaussian derivative model regarding:
(i)~its axiomatic derivation from assumptions that reflect
structural properties of the world in combination with internal
consistency requirements between image representations at different
scales,
(ii)~how it leads to families of spatial and spatio-chromatic
receptive fields,
(iii)~its relations to receptive field profiles in biological vision
and (iv)~its affine covariance properties as highly desirable for
computing receptive field responses under perspective image
deformations.
This material 
constitutes the theoretical and conceptual background
for the genuine discrete theory that will be developed in 
Sections~\ref{sec-theory-disc-aff-rec-fields}--\ref{sec-sc-norm-ders-disc}
and Appendix~\ref{sec-aff-hybr-pyr}.

\subsection{Spatial affine Gaussian derivative based receptive fields}

Let us assume that the spatial smoothing operation in a spatial scale-space representation should obey:
(i)~linearity, (ii)~spatial shift invariance, (iii)~a semi-group structure
over spatial scales and (iv)~a requirement of not introducing new
structures from finer to coarser scales formalized in terms of non-enhancement of
local extrema, meaning that the value at a local maximum must not
increase from finer to coarser scales and correspondingly the value at
a local minimum must not decrease. Then, it can be shown that affine
Gaussian kernels 
\begin{equation}
  \label{eq-aff-scsp-conv-kernel}
  g(x;\; \Sigma_s) 
   = \frac{1}{2 \pi \sqrt{\det \Sigma_s}} \,
      e^{- {x^T \Sigma_s^{-1} x}/{2}},
\end{equation}
are uniquely determined by necessity (Lindeberg \cite{Lin10-JMIV}). 
Directional derivatives of these kernels do in turn constitute a canonical model for
spatial receptive fields over a two-dimensional image domain (Lindeberg \cite{Lin13-BICY,Lin13-PONE}).

For the specific parameterization of the spatial covariance matrix
$\Sigma_s = s \, \Sigma$, convolution with these kernels obeys the affine diffusion equation
  \begin{equation}
    \label{eq-aff-scsp-diff-eq}
    \partial_s L 
    = \frac{1}{2} \nabla^T \left( \Sigma \, \nabla L \right)
  \end{equation}
with initial condition $L(x, y;\; 0) = f(x, y)$. Convolution with
affine Gaussian kernels also obeys the cascade
smoothing property
\begin{equation}
    g(\cdot, \cdot;\; \Sigma_1) * L(\cdot, \cdot;\; \Sigma_2) =
    L(\cdot, \cdot;\; \Sigma_1 + \Sigma_2).
\end{equation}
To parameterize the affine Gaussian kernels, let us in the two-dimensional case 
consider the covariance matrix determined by two eigenvalues $\lambda_1$, $\lambda_2$
and one orientation $\alpha$.
Then, the covariance matrix can be written 
\begin{align}
  \begin{split}
  \label{eq-aff-cov-mat-2D}
  \Sigma' & =
  \left(
    \begin{array}{ccc}
      \lambda_1 \cos^2 \alpha + \lambda_2 \sin^2 \alpha \quad
        &  (\lambda_1 - \lambda_2) \cos \alpha \, \sin \alpha 
        \\
      (\lambda_1 - \lambda_2) \cos \alpha \, \sin \alpha \quad
        & \lambda_1 \sin^2 \alpha + \lambda_2 \cos^2 \alpha
    \end{array}
  \right).
  \end{split}
\end{align}

\begin{figure}[hbtp]
  \begin{center}
    \begin{tabular}{cccccc}
     \hspace{-4mm}
     \includegraphics[width=0.15\textwidth]{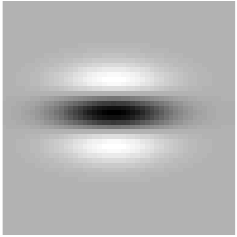} \hspace{-4mm} &
      \includegraphics[width=0.15\textwidth]{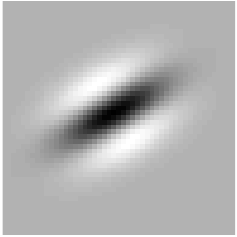} \hspace{-4mm} &
      \includegraphics[width=0.15\textwidth]{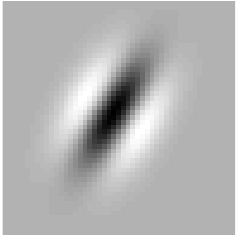} \hspace{-4mm} &
      \includegraphics[width=0.15\textwidth]{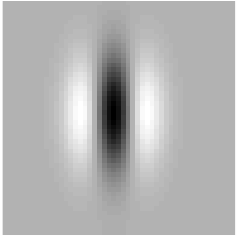} \hspace{-4mm} &
      \includegraphics[width=0.15\textwidth]{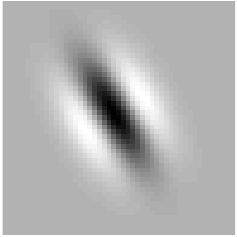} \hspace{-4mm} &
      \includegraphics[width=0.15\textwidth]{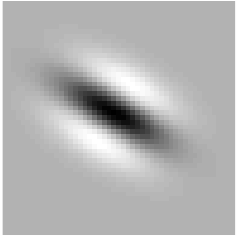} \hspace{-4mm} \\
     \hspace{-4mm}
     \includegraphics[width=0.15\textwidth]{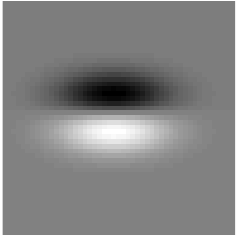} \hspace{-4mm} &
      \includegraphics[width=0.15\textwidth]{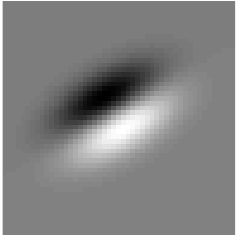} \hspace{-4mm} &
      \includegraphics[width=0.15\textwidth]{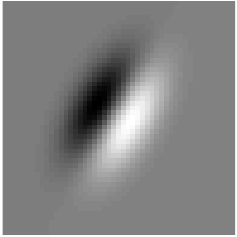} \hspace{-4mm} &
      \includegraphics[width=0.15\textwidth]{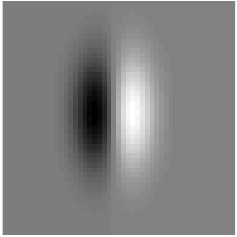} \hspace{-4mm} &
      \includegraphics[width=0.15\textwidth]{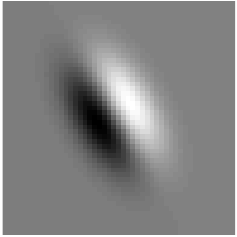} \hspace{-4mm} &
      \includegraphics[width=0.15\textwidth]{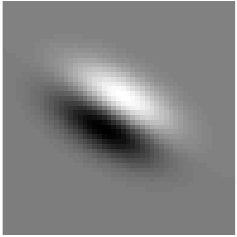} \hspace{-4mm} \\
     \hspace{-4mm}
     \includegraphics[width=0.15\textwidth]{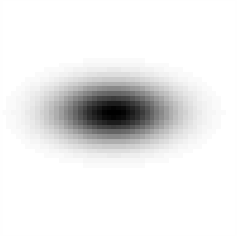} \hspace{-4mm} &
      \includegraphics[width=0.15\textwidth]{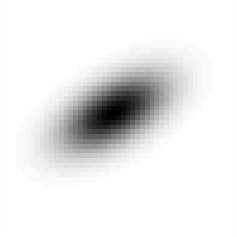} \hspace{-4mm} &
      \includegraphics[width=0.15\textwidth]{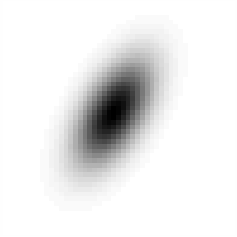} \hspace{-4mm} &
      \includegraphics[width=0.15\textwidth]{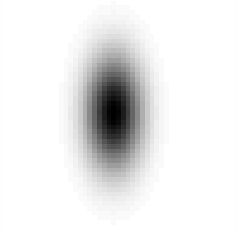} \hspace{-4mm} &
      \includegraphics[width=0.15\textwidth]{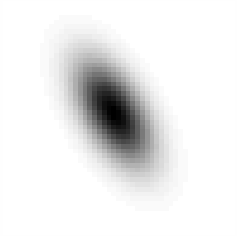} \hspace{-4mm} &
      \includegraphics[width=0.15\textwidth]{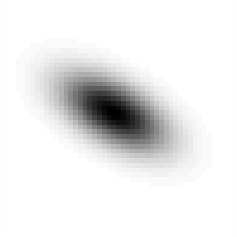} \hspace{-4mm} \\
    \end{tabular} 
  \end{center}
  \vspace{-4mm}
  \caption{Examples of affine Gaussian kernels $g(x, y;\; \Sigma)$ and their directional
    derivatives $\partial_{\orth \varphi}g(x, y;\; \Sigma)$ and
    $\partial_{\orth \varphi \orth \varphi}g(x, y;\; \Sigma)$ 
    up to order two in the two-dimensional case,
           here for $\lambda_1 = 64$, $\lambda_2=16$ and
            $\alpha = 0, \pi/6, \pi/3, \pi/2, 2\pi/3, 5\pi/6$.
   (Horizontal axis: $x \in [-24, 24]$. Vertical axis: $y \in [-24, 24]$.)}
  \label{fig-aff-elong-filters-dir-ders}
\end{figure}

\begin{figure}[hbtp]
  \begin{center}
    \begin{tabular}{cccccc}
     \hspace{-4mm}
     \includegraphics[width=0.15\textwidth]{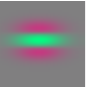} \hspace{-4mm} &
      \includegraphics[width=0.15\textwidth]{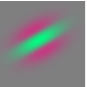} \hspace{-4mm} &
      \includegraphics[width=0.15\textwidth]{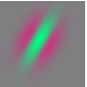} \hspace{-4mm} &
      \includegraphics[width=0.15\textwidth]{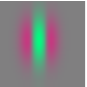} \hspace{-4mm} &
      \includegraphics[width=0.15\textwidth]{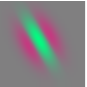} \hspace{-4mm} &
      \includegraphics[width=0.15\textwidth]{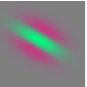}
                                                                                                               \hspace{-4mm} \\
     \hspace{-4mm}
     \includegraphics[width=0.15\textwidth]{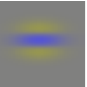} \hspace{-4mm} &
      \includegraphics[width=0.15\textwidth]{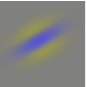} \hspace{-4mm} &
      \includegraphics[width=0.15\textwidth]{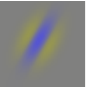} \hspace{-4mm} &
      \includegraphics[width=0.15\textwidth]{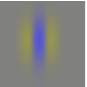} \hspace{-4mm} &
      \includegraphics[width=0.15\textwidth]{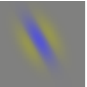} \hspace{-4mm} &
      \includegraphics[width=0.15\textwidth]{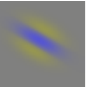}
                                                                                                               \hspace{-4mm} \\
     \hspace{-4mm}
     \includegraphics[width=0.15\textwidth]{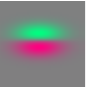} \hspace{-4mm} &
      \includegraphics[width=0.15\textwidth]{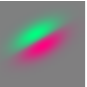} \hspace{-4mm} &
      \includegraphics[width=0.15\textwidth]{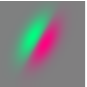} \hspace{-4mm} &
      \includegraphics[width=0.15\textwidth]{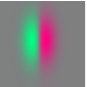} \hspace{-4mm} &
      \includegraphics[width=0.15\textwidth]{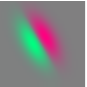} \hspace{-4mm} &
      \includegraphics[width=0.15\textwidth]{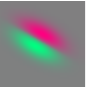}
                                                                                                               \hspace{-4mm} \\
     \hspace{-4mm}
     \includegraphics[width=0.15\textwidth]{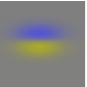} \hspace{-4mm} &
      \includegraphics[width=0.15\textwidth]{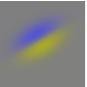} \hspace{-4mm} &
      \includegraphics[width=0.15\textwidth]{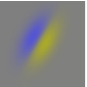} \hspace{-4mm} &
      \includegraphics[width=0.15\textwidth]{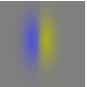} \hspace{-4mm} &
      \includegraphics[width=0.15\textwidth]{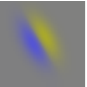} \hspace{-4mm} &
      \includegraphics[width=0.15\textwidth]{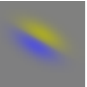} \hspace{-4mm} \\
    \end{tabular} 
  \end{center}
  \vspace{-4mm}
  \caption{Examples of affine Gaussian colour-opponent directional
    derivatives according to (\ref{eq-col-opp-aff-gauss-dir-der}) and 
    (\ref{eq-col-opp-space-uv-from-rgb}) up to order two in the two-dimensional case,
           here for $\lambda_1 = 64$, $\lambda_2=16$ and
            $\alpha = 0, \pi/6, \pi/3, \pi/2, 2\pi/3, 5\pi/6$.
   (Horizontal axis: $x \in [-24, 24]$. Vertical axis: $y \in [-24, 24]$.)}
  \label{fig-aff-elong-filters-dir-ders-col-opp}
\end{figure}

\noindent
Correspondingly, we can for two orthogonal directions $\varphi$ and $\orth \varphi$ aligned 
to the eigendirections of the spatial covariance matrix parameterize
the directional derivative operators in these directions according to
\begin{equation}
  \label{eq-dir-ders}
  \partial_{\varphi} = \cos \varphi \, \partial_x + \sin \varphi \, \partial_y
  \quad\quad
  \partial_{\orth\varphi} = -\sin \varphi \, \partial_x + \cos \varphi \, \partial_y.
\end{equation}
This leads to the following canonical model for affine Gaussian
receptive fields over a spatial domain  (Lindeberg \cite{Lin10-JMIV})
\begin{equation}
  \label{eq-aff-scsp-conv-kernel-dir-der}
  g_{\varphi^m \orth \varphi^n}(x, y;\; \Sigma_s) =
  \partial_{\varphi}^m \, \partial_{\orth \varphi}^n \, g(x;\; \Sigma_s),
\end{equation}
preferably with the angle $\varphi$ in (\ref{eq-dir-ders}) and 
(\ref{eq-aff-scsp-conv-kernel-dir-der}) chosen equal to the angle
$\alpha$ in (\ref{eq-aff-cov-mat-2D}).

Figure~\ref{fig-aff-elong-filters-dir-ders} shows a few examples of affine
Gaussian kernels and their directional derivatives up to order two for
one combination of the two eigenvalues determined such that the ratio
between the scale parameters in the two eigendirections of the
covariance matrix is equal to two, with the scale parameters 
$\sigma_1 = \sqrt{\lambda_1}$ and $\sigma_2 = \sqrt{\lambda_2}$
expressed in terms of dimension $[\mbox{length}]$.

\subsection{Spatio-chromatic affine Gaussian derivative based receptive fields}

Consider next the colour channels of colour-opponent space
defined from an RGB colour representation (Hall {\em et al.\/}\ \cite{HalVerCro00-ECCV})
\begin{equation}
  \label{eq-col-opp-space-uv-from-rgb}
  \left(
    \begin{array}{c}
      f\\
      u\\
      v
    \end{array}
   \right)
   =
  \left(
    \begin{array}{c}
      f\\
      c^{(1)}\\
      c^{(2)}
    \end{array}
   \right)
   =
   \left(
     \begin{array}{ccc}
       \tfrac{1}{3} &   \tfrac{1}{3} & \tfrac{1}{3} \\
       \tfrac{1}{2} & - \tfrac{1}{2} & 0 \\
       \tfrac{1}{2} &   \tfrac{1}{2} & -1 \\
    \end{array}
   \right)
   \left(
     \begin{array}{c}
       R \\
       G\\
       B
     \end{array}
   \right),
\end{equation}
where $c^{(1)}$ represents the red/green colour-opponent channel and 
$c^{(2)}$ the yellow/blue colour-opponent channel,
with yellow approximated by the average of the $R$ and $G$
channels and $f$ denoting the channel of pure intensity information.

Then, {\em affine Gaussian colour-opponent receptive fields\/} 
$(U, V)^T = (C^{(1)}, C^{(2)})^T$ can be defined by applying affine
Gaussian receptive fields of the form (\ref{eq-aff-scsp-conv-kernel-dir-der})
to the colour-opponent channels $(c^{(1)}, c^{(2)})$:
\begin{align}
  \begin{split}
     \label{eq-col-opp-aff-gauss-dir-der}
     U = C^{(1)}(\cdot, \cdot;\; \Sigma_s) 
     & = g_{\varphi^m \orth \varphi^n}(\cdot, \cdot;\; \Sigma_s) * c^{(1)}(\cdot),
  \end{split}\\
  \begin{split}
     V = C^{(2)}(\cdot, \cdot;\; \Sigma_s) 
     & = g_{\varphi^m \orth \varphi^n}(\cdot, \cdot;\; \Sigma_s) * c^{(2)}(\cdot).
  \end{split}
\end{align}
Figure~\ref{fig-aff-elong-filters-dir-ders-col-opp} shows examples of
such affine Gaussian spatio-chromatic receptive fields
up to order two over red/green and yellow/blue colour-opponent space.

\subsection{Relations to biological receptive fields}

In the most ideal form of the affine Gaussian receptive field model,
one should at any image point $(x, y)$ consider affine receptive fields as being
present for all positive definite covariance matrices $\Sigma$, as
parameterized by their eigenvalues $\lambda_1 > 0$ and $\lambda_2 > 0$
and for all directions $\alpha$.

\begin{figure}[hbtp]
   \begin{center}
     \begin{tabular}{cc}
        & {\small$\partial_x g(x, y;\; \Sigma)$} \\
       \includegraphics[height=0.17\textheight]{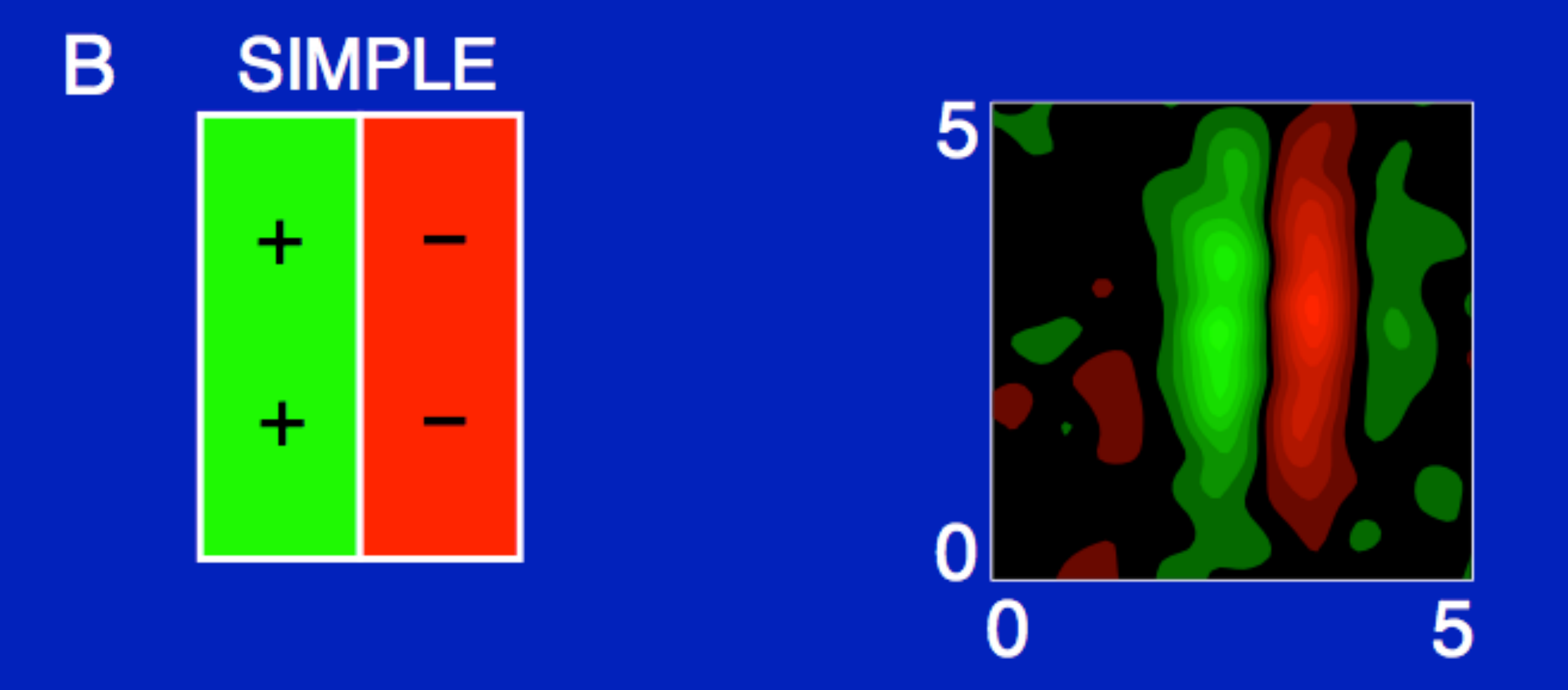}
       & \includegraphics[height=0.17\textheight]{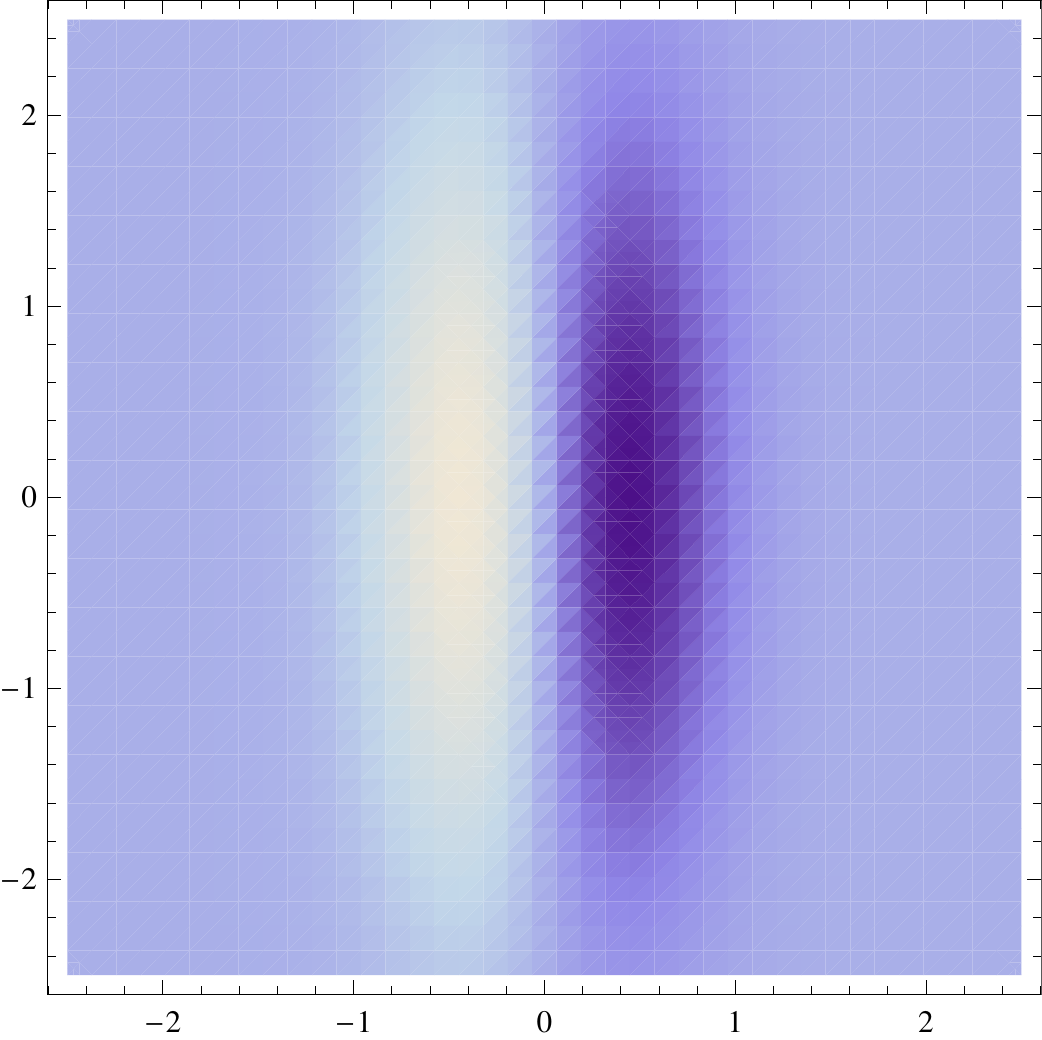}
     \end{tabular}
   \end{center}
  \caption{Example of a receptive field profile over the spatial domain
           in the primary visual cortex (V1) as reported by DeAngelis {\em et al.\/}\
           \cite{DeAngOhzFre95-TINS,deAngAnz04-VisNeuroSci}.
          (middle) Receptive field profile of a simple cell over image
          intensities as reconstructed from cell
          recordings, with positive weights 
          represented as green and negative weights by red. (left) Stylized simplification of the receptive
          field shape. (right) Idealized model of the receptive field
          from a first-order directional derivative of an affine
          Gaussian kernel $\partial_x g(x, y;\; \Sigma) = 
            \partial_x g(x, y;\; \lambda_x, \lambda_y) 
            = - \frac{x}{\lambda_x} 1/(2 \pi \sqrt{\lambda_x \lambda_y}) 
                 \exp(-x^2/2 m\lambda_x -y^2/2\lambda_y)$,
           here with $\lambda_x = 0.2$ and $\lambda_y = 2$ in units of
           degrees of visual angle, and with positive weights with
           respect to image intensities represented by white and
           negative values by violet.}
  \label{fig-simple-cell-aff-gauss-model}

   \begin{center}
     \begin{tabular}{ccc}
        & & {\small $\partial_{\orth \varphi}g(x, y;\; \Sigma)$} \\
       \includegraphics[height=0.17\textheight]{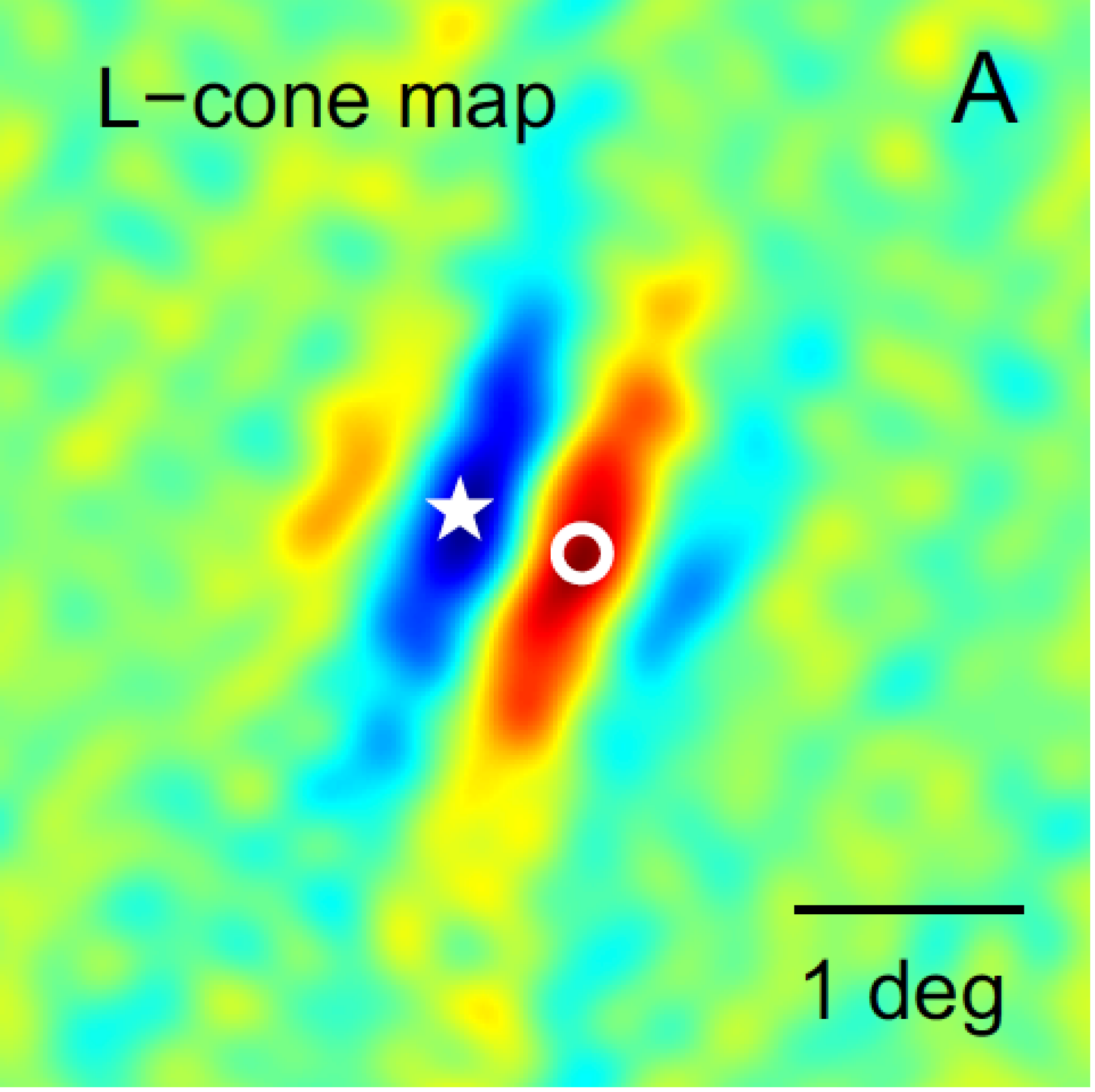}
       & \includegraphics[height=0.17\textheight]{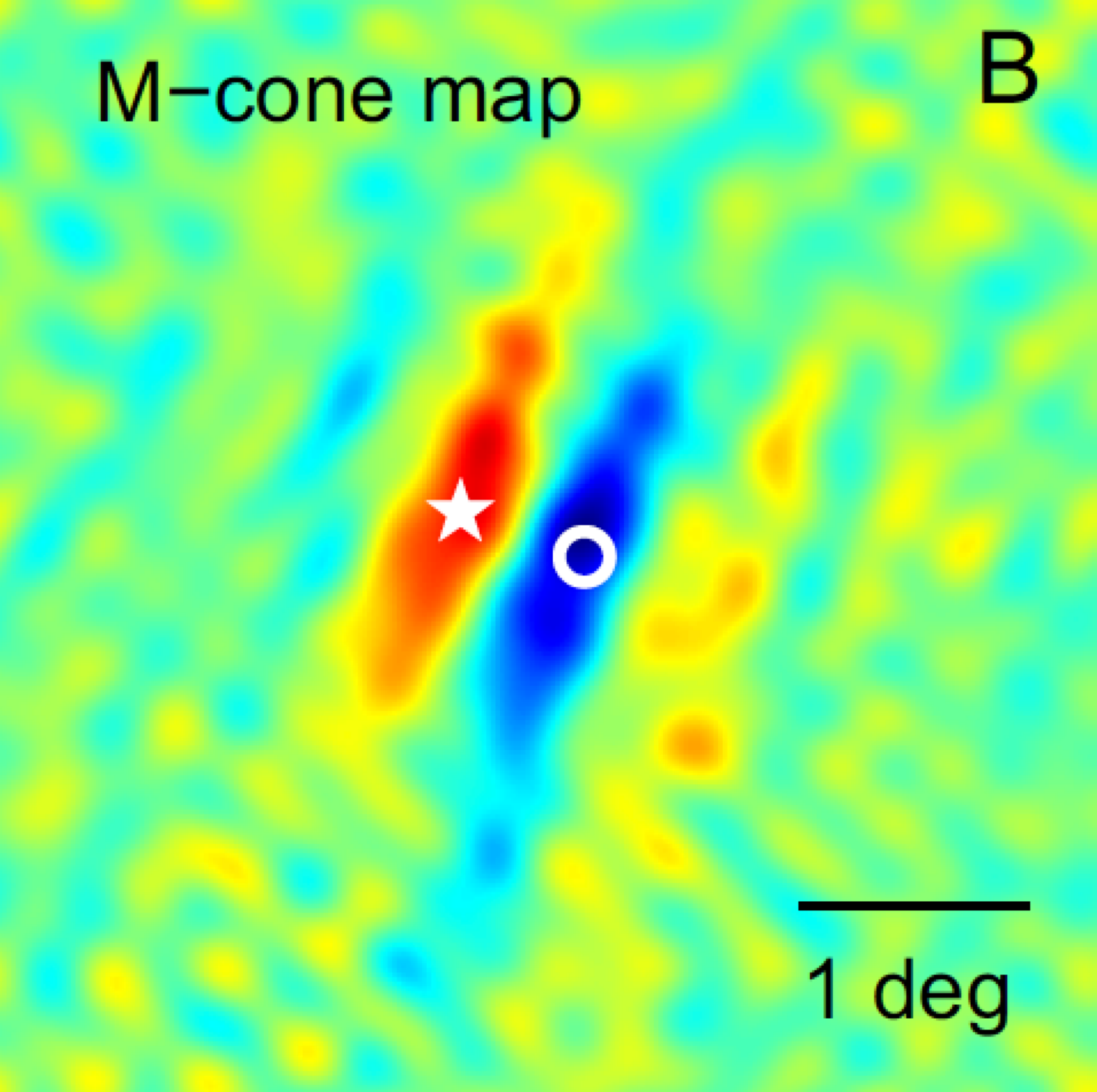}
       & \includegraphics[height=0.17\textheight]{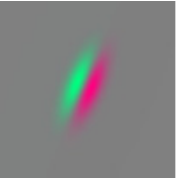}
     \end{tabular}
   \end{center}
  \caption{Example of a colour-opponent receptive field profile over the spatial domain
           for a double-opponent simple cell in the primary visual
           cortex (V1) as measured by Johnson {\em et al.\/}\
           \cite{JohHawSha08-JNeuroSci}.
          (left) Responses to L-cones corresponding to long wavelength
          red cones, with positive weights
          represented by red and negative weights by blue. 
          (middle) Responses to M-cones corresponding to medium wavelength
          green cones, with positive weights
          represented by red and negative weights by blue. 
          (right) Idealized model of the receptive field
          from a first-order directional derivative of an affine
          Gaussian kernel $\partial_{\orth \varphi}g(x, y;\; \Sigma)$ 
          according to (\ref{eq-aff-scsp-conv-kernel-dir-der}),
          (\ref{eq-dir-ders}), (\ref{eq-aff-scsp-conv-kernel}) and
          (\ref{eq-aff-cov-mat-2D}) for $\sigma_1 = \sqrt{\lambda_1} = 0.6$,
         $\sigma_2 = \sqrt{\lambda_2} = 0.2$ in units of
           degrees of visual angle, $\alpha = 67~\mbox{degrees}$ and with positive
           weights for the red-green colour-opponent channel
           $U$ according to (\ref{eq-col-opp-aff-gauss-dir-der}) and
           (\ref{eq-col-opp-space-uv-from-rgb}) represented by red and
           negative values by green.}
  \label{fig-simple-cell-aff-gauss-model-col-opp}
\end{figure}

This model is proposed in Lindeberg \cite[Section~6]{Lin10-JMIV} \cite[Section~6.3]{Lin13-BICY}) as a
model for the spatial component of simple cells in the primary visual
cortex (V1) and is in good agreement with neural cell recordings by 
DeAngelis {\em et al.\/} \cite{DeAngOhzFre95-TINS,deAngAnz04-VisNeuroSci} and 
Johnson {\em et al.\/}\ \cite{JohHawSha08-JNeuroSci}.
Figure~\ref{fig-simple-cell-aff-gauss-model} shows an example of the
spatial dependency of a simple cell that can be well modelled by a
first-order affine Gaussian derivative over image intensities. 
Figure~\ref{fig-simple-cell-aff-gauss-model-col-opp} shows
corresponding results for a color-opponent receptive field of a simple
cell in V1 that can be modelled as a first-order affine Gaussian spatio-chromatic
derivative over an R-G colour-opponent channel.

\begin{figure}
   \begin{center}
     \begin{tabular}{cc}
       \includegraphics[width=0.38\textwidth]{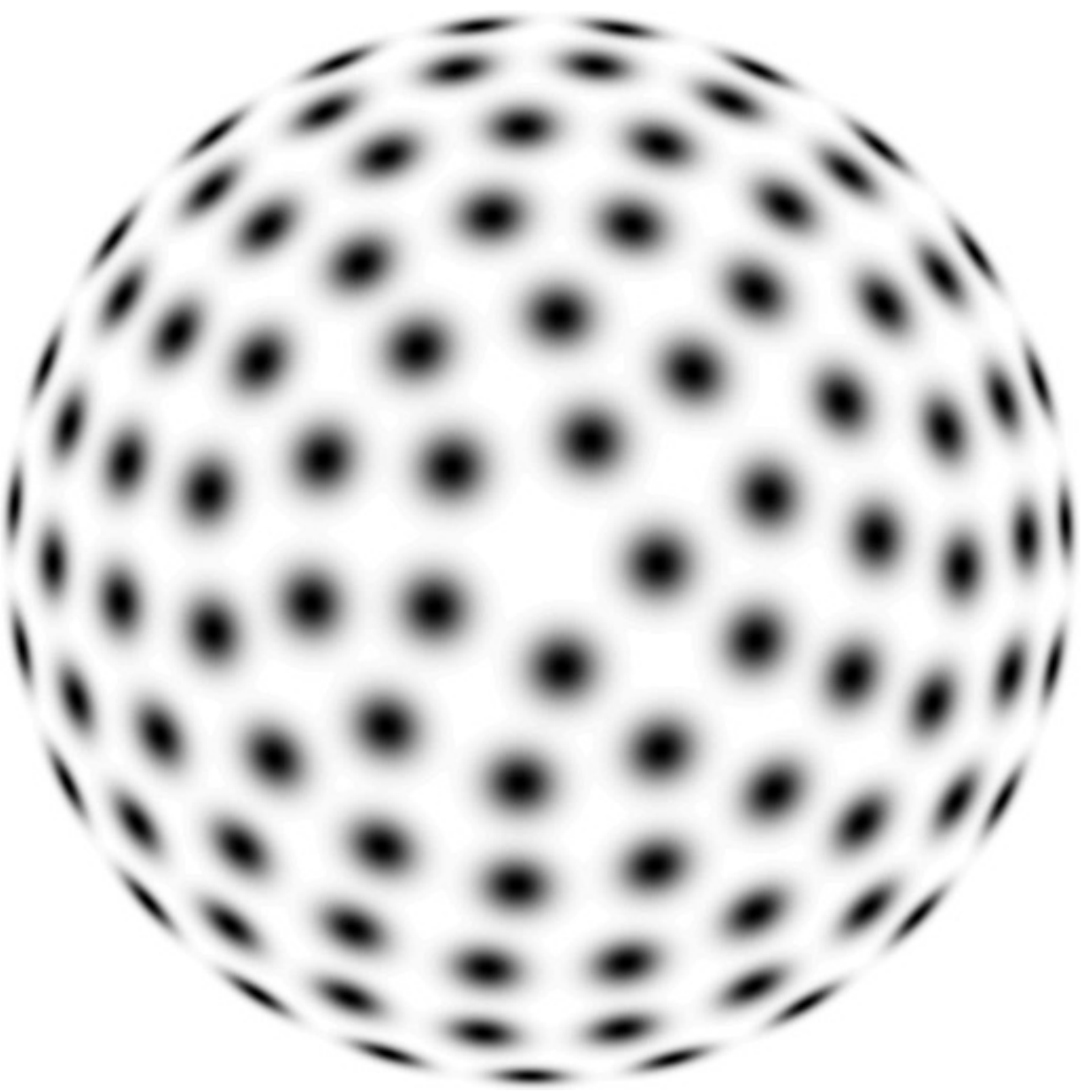} &
       \includegraphics[width=0.38\textwidth]{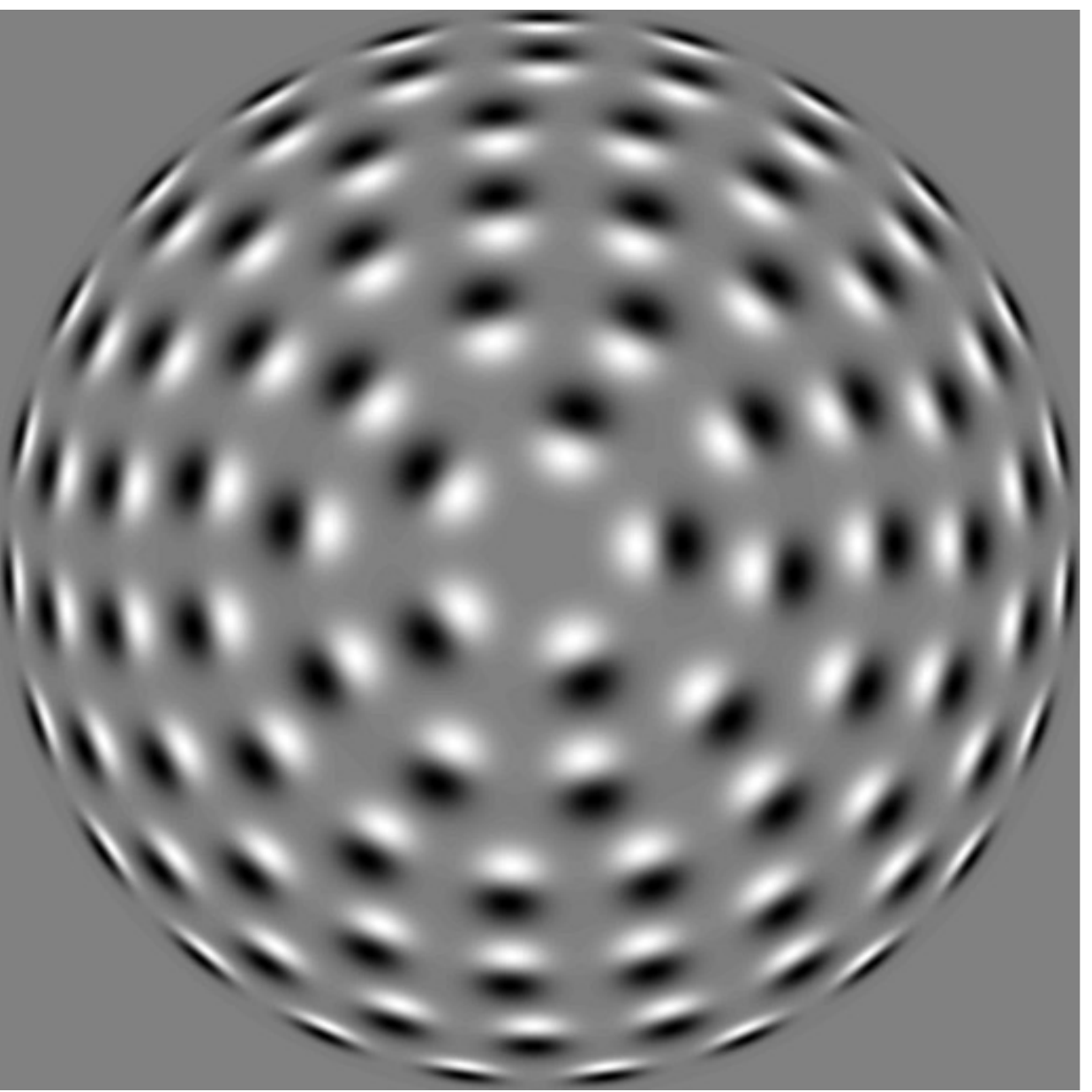}
     \end{tabular}
   \end{center}

  \caption{Distributions of affine Gaussian receptive fields
    corresponding to a uniform distribution on a hemisphere regarding
    (left) zero-order smoothing kernels and (right) first-order
    derivatives.}
  \label{fig-distr-aff-rec-fields}
\end{figure}
 
\subsection{Affine covariance}

A theoretically very attractive property of the affine Gaussian model
for spatial receptive fields is that this family of receptive fields
is closed under affine transformations of the spatial image domain.
If we consider two images $f_L$ and $f_R$ that with vector notation
for the image coordinates
$\xi = (\xi_1, \xi_2)^T$ and $\eta = (\eta_1, \eta_2)^T$ are related by an
affine image deformation
\begin{equation}
  \label{eq-aff-trans-two-imgs-aff-scsp}
    f_L(\xi) = f_R(\eta) \quad \mbox{where} \quad \eta = A \, \xi + b
\end{equation}
and define corresponding scale-space representations according to
\begin{equation}
    L(\cdot;\; \Sigma_L) = g(\cdot;\; \Sigma_L) * f_L(\cdot), \quad
       R(\cdot;\; \Sigma_R) = g(\cdot;\; \Sigma_R) * f_R(\cdot),
\end{equation}
then these scale-space representations will be related according to 
(Lindeberg \cite{Lin93-Dis}; Lindeberg and G{\aa}rding \cite{LG94-ECCV,LG96-IVC})
\begin{equation}
  \label{eq-aff-transf-prop-aff-scsp}
    L(x;\; \Sigma_L) = R(y;\; \Sigma_R) \quad \mbox{where} \quad
     \Sigma_R = A \, \Sigma_L \, A^T
\quad \mbox{and} \quad y = A \, x + b
\end{equation}
with $x = (x_1, x_2)^T$ and $y = (y_1, y_2)^T$.
In other words, given that an image $f_L$ is affine transformed into
an image $f_R$, it will always be possible to find a transformation
between the spatial covariance matrices $\Sigma_L$ and $\Sigma_R$ in the two domains
that makes it possible to match the corresponding derived 
 internal representations $L(\cdot;\; s_L)$ and $R(\cdot;\, s_R)$ perfectly.
If we in turn locally approximate the non-linear perspective image
deformation between two images of the same scene by local affine
deformations (first-order derivatives), then this means that it will to first-order of
approximation be possible to match the
receptive field responses computed from different views of the same
scene.

This idea was originally proposed for reducing the shape distortions that
arise when computing estimates of local surface orientation by shape-from-texture and
shape-from-disparity-gradients (Lindeberg and G{\aa}rding \cite{LG94-ECCV,LG96-IVC})
and has later been explored for image-based matching and recognition
(Baumberg \cite{Bau00-CVPR};
 Schaffalitzky and Zisserman \cite{SchZis01-ICCV};
 Sivic and Zisserman \cite{SivZis03-ICCV};
 Mikolajczyk and Schmid \cite{MikSch04-IJCV};
 Tuytelaars and van Gool \cite{TuyGoo04-IJCV};
 Mikolajczyk {\em et al.\/}\ \cite{MikTuySchZisMatSchKadGoo05-IJCV};
 Lazebnik {\em et al.\/}\ \cite{LazSchPon05-PAMI};
 Rothganger {\em et al.\/}\ \cite{RotLazSchPon06-IJCV};
 Tuytelaars and Mikolajczyk \cite{TuyMik08-Book};
 Burghouts and Geusebroek \cite{BurGeu09-CVIU};
 Lia {\em et al.\/}\ \cite{LiaLiuHui13-PRL}).

Similar ideas of affine invariance as underlying these approaches
based on affine shape adaptation (Lindeberg \cite{Lin13-PONE}) have also been used for designing
smoothing methods with a larger amount of smoothing along local image
structures than across them (Weickert \cite{Wei99-IJCV}; Almansa and
Lindeberg \cite{AL00-IP}),
for performing affine invariant segmentation 
(Ballester and Gonz{\'a}lez \cite{BalGon98-JMIV};
 Rothganger {\em et al.\/}\ \cite{RotLazSchPon07-PAMI}),
for constructing affine SIFT descriptors 
(Morel and Yu \cite{MorGuo09-SIAM-JIS,YuMor09-ASSP}; Sadek {\em et
  al.\/}\ \cite{SadConMeiBalCas12-SIM-JIS}),
for affine invariant tracking
 (Giannarou {\em et al.\/}\ \cite{GiaVisYan13-PAMI}),
for formulating affine covariant metrics (Fedorov {\em et al.\/} \cite{FedAriSadFacBal15-SIAM})
and for affine invariant inpainting (Fedorov {\em et al.\/}\ \cite{FedAriFacBal16-VISAPP}).

When implementing such affine covariance or affine invariance in practice, there are two
main approaches to follow, either (i)~by adapting the shape of the affine
Gaussian kernel to the local image structure as proposed in the affine shape adaptation methodology
proposed in (Lindeberg and G{\aa}rding \cite{LG94-ECCV,LG96-IVC}) or 
(ii)~by computing receptive field responses for all affine covariance
matrices alternatively for some sparse sampling of the resulting family of receptive fields.

Figure~\ref{fig-distr-aff-rec-fields} shows the resulting
distributions of affine Gaussian receptive fields of different
orientations and degrees of orientation as they arise from local
linearizations of the perspective projection model, if we
assume that (i)~the set of surface directions in the world is on average
uniformly distributed in the world and that (ii)~the distributions of the
local surface patterns on these object surfaces are in turn without dominant
directional bias and uncoupled to the orientations of the local surface patches.

In our idealized model of receptive fields, all these receptive fields
can be thought of as being present at every position in image space,
and corresponding to a uniform distribution on a hemisphere.
For practical purposes, and to obtain an affine covariant basis to be
used as input to later stage processes on a format where the
rotational angles are represented in the same way for all shapes of
the affine covariance matrices $\Sigma$, it is alternatively natural to parameterize these kernels
using: (i)~a self-similar (geometric) distribution over the overall size of the
kernel as represented by {\em e.g.\/}\ the maximum eigenvalue
$\lambda_{max}$ of the spatial covariance matrix $\Sigma$,
(ii)~a self-similar (geometric) distribution over the eccentricity as
parameterized by the ratio $\lambda_{min}/\lambda_{max}$ between the minimum and the maximum
eigenvalues $\lambda_{min}$ and $\lambda_{max}$ of the spatial spatial covariance matrix $\Sigma$ and
(iii)~a uniform distribution over the directions in image space as parameterized by
{\em e.g.\/}\ the orientation $\varphi$ of the eigenvector corresponding to the
maximum eigenvalue of the spatial covariance matrix $\Sigma$.

\section{Discretizing continuous affine Gaussian receptive fields}
\label{sec-disc-aff-gauss-rec-fields}

When implementing these affine Gaussian receptive fields computationally,
there are different approaches to follow.
Two basic alternatives
consist of
(i)  sampling the affine Gaussian kernel or its derivatives at the
grid points of the image domain or
(ii)~integrating these kernels over the support region of each image
  pixel.
It is, however, known that sampling the Gaussian kernel is not the best way of
implementing the regular scale-space concept based on the rotationally
symmetric Gaussian kernel, where local integration of the scale-space
kernel is a better choice that remedies some of the artefacts
(Lindeberg \cite{Lin90-PAMI}). 

More seriously, with respect to computational efficiency, the resulting kernels will
in general be inseparable and will therefore for a truncated filter
size of $M$ along each dimension require $M^2$ operations as 
opposed to $2M$ operations for separable filtering.%
\footnote{For the class of compact $3 \times 3$-kernels that we will
  arrive at later in Section~\ref{sec-theory-disc-aff-rec-fields-3x3},
  the situation is different, since the difference between $M^2$ and $2 M$ is much lower for $M = 3$.}
At moderate scale levels, where $M$ may typically of the order of
around $10$, $20$ or $50$ or
even more at coarser scales or for a numerically more accurate implementation, the computational
work for performing non-separable spatial convolution with the full
affine Gaussian filter may therefore be far too much for a computationally efficient implementation.

Implementing the affine Gaussian kernels in the Fourier domain is
then an alternative option.
Besides the technical need for additional extensions when handling
image sizes that do not match well with powers of 2,
there are, however, then also other discretization issues to consider.
If we first discretize the spatial affine Gaussian kernel by local
integration, and then implement the subsequent discrete convolution by FFT,
then that should in principle be a numerically reasonable implementation.
If we on the other hand perform the discretization in the Fourier
domain, there are other discretization and trade-off issues to consider,
see Florack \cite{Flo00-PAMI} for a treatment of some of those for the
specific case of rotationally symmetric Gaussian kernels.

An alternative approach is by exploring the affine covariance property of the
continuous affine Gaussian scale-space concept
(\ref{eq-aff-transf-prop-aff-scsp}) by first warping the original
image by an affine transformation implemented in terms of spline
interpolation, performing separable discrete scale-space smoothing in
the warped domain and then warping the smoothed image back to the
original image domain. Such an operation may be faster than first
generating a filter corresponding to the sampled affine Gaussian
kernel or better using a locally integrated affine Gaussian kernel over
the support region of each pixel and then performing inseparable
convolution with the resulting discrete kernel. This form of warping-based
discretization  has also been used in most implementations
of affine shape adaptation for interest point detection
(Lindeberg and G{\aa}rding \cite{LG96-IVC};
 Mikolajczyk and Schmid \cite{MikSch04-IJCV};
Tuytelaars and Mikolajczyk \cite{TuyMik08-Book}
 Lindeberg \cite{Lin08-EncCompSci}).

Instead of performing a full affine warp, the image the affine
covariance property could instead also be explored by rotating
the image by an angle corresponding to the main orientation of the affine
Gaussian kernel using spline interpolation, performing separable convolution
with scale values determined from the eigenvalues of the covariance
matrix of the
affine Gaussian kernel, and then rotating the smoothed image back again using spline
interpolation. For highly anisotropic affine Gaussian kernels, such an
approach can specifically be expected to be computationally more
efficient compared to the
above full affine warp, since the separable spatial smoothing stage
can be based on a 1-D kernel with smaller spatial extent for the
smoothing operation in the direction corresponding to the smallest
eigenvalue of the covariance matrix $\Sigma$.

Another alternative proposed by Geusebroek {\em et al.\/}\
\cite{GeuSmeWei03-TIP} is to approximate the inseparable affine
Gaussian kernel with a set of separable recursive filters, which may,
however, lead to artefacts, specifically a lack of rotational and
affine covariance.

Yet a wider class of alternative implementation methods is to consider 
numerical methods for solving the affine diffusion equation
(\ref{eq-aff-scsp-diff-eq}) that determines the evolution properties 
of the affine Gaussian receptive fields over scale.

For implementations of these types, discrete scale-space properties
are, however, not guaranteed to hold, so the relation to the
underlying aims of scale-space
theory rests primarily on the degree of accuracy by which the equations
obtained from the continuous scale-space theory are numerically approximated.

If additionally computing a set of affine Gaussian receptive field
responses at every image point, corresponding to different
combinations of directional derivative operators according to the
model (\ref{eq-aff-scsp-conv-kernel-dir-der}),
there is also an issue of whether the
computationally more demanding smoothing operation has to be repeated
for each receptive field response, or if it could be shared between
different combinations of directional derivative operators applied to
the affine Gaussian scale-space representation.

\section{Genuine discrete theory for affine Gaussian receptive fields}
\label{sec-theory-disc-aff-rec-fields}

In this and following sections, we will develop a spatial discretization approach
based on discrete scale-space theory that has been specifically
designed to preserve scale-space properties in the discretization
(Lindeberg \cite{Lin97-ICSSTCV,CVAP257}). 

We will start by formulating a genuine discrete affine scale-space
theory over a continuum of spatial covariance matrices that represent
receptive fields with different sizes, orientations and eccentricities 
in the image domain.
Then, we will discretize that semi-discrete scale-space 
further to compact $3 \times 3$-kernels that are to be
applied repeatedly.

For both of these concepts, we will obtain discrete scale-space kernels whose
covariance matrices are exactly equal to the covariance matrices that
would be obtained from the corresponding continuous theory. 
Additionally, receptive field responses corresponding to different
combinations of directional derivatives applied to the affine
scale-space representation can be computed by applying local 
$3 \times 3$ derivative approximation kernels to the output of the
discrete affine Gaussian scale-space representation, implying that
the most computationally component of a discrete derivative approximation
kernel, the spatial smoothing step, can be shared between
directional derivative approximations for different
orders of differentiation.

We will start by presenting a general theoretical results on which the
subsequent treatment will be based.

\subsection{Continuous scale parameter scale space over a discrete image
  domain}
\label{sec-summ-axiom-disc-scsp}

In (Lindeberg \cite{Lin97-ICSSTCV,CVAP257}) different generalizations are presented of the discrete
scale-space theory originally proposed in (Lindeberg \cite{Lin90-PAMI,Lin93-Dis})
for an isotropic spatial image domain to:
(i)~non-isotropic spatial domains, where the
scale-space kernels are no longer required to be rotationally
symmetric over image space, as well as to
(ii)~spatio-temporal image domains, where the temporal dimension is treated
in a conceptually different way than the spatial dimensions.

In Appendix~\ref{sec-app-disc-scsp}, we do after a set of necessary
formal definitions give necessity and
sufficiency results in
Theorem~\ref{thm-nD-scale-space-family-nec-disc} and
Theorem~\ref{thm-nD-scale-space-family-suff-disc} of applying such
discrete scale-space theory to the notion of affine scale space.
In summary, 
Theorems~\ref{thm-nD-scale-space-family-nec-disc} and~\ref{thm-nD-scale-space-family-suff-disc}  
state that if we assume 
(i)~a convolution structure corresponding to linearity and spatial
shift invariance,
(ii)~a semi-group property over spatial scales,
(iii)~certain regularity assumptions to ensure differentiability over
scale and
(iv)~non-enhancement of local extrema meaning that the
value at a local maximum must not increase and that the value at a
local minimum must not decrease,
then for a signal defined over a $D$-dimensional discrete domain, 
these scale-space axioms together imply that the scale-space family
$L \colon \bbbz^D \times \bbbr^+ \rightarrow \bbbr$ of any
discrete signal $f \colon \bbbz^D \rightarrow \bbbr$
must by necessity and sufficiency satisfy the semi-discrete differential equation
\begin{equation}
  \label{gen-nD-lin-op}
  (\partial_s L)(x;\; s) 
  = ({\cal A} L)(x;\;s) 
  = \sum_{\xi \in \bbbz^D} a_{\xi} \, L(x-\xi; \;s)
\end{equation}
for some {\em infinitesimal scale-space generator\/} ${\cal A}$
characterized by
\begin{itemize}
    \item
    the {\em locality\/} condition 
    $a_{\xi} = 0$ if $|\xi|_{\infty} > 1$,
    \item
    the {\em positivity\/} constraint
    $a_{\xi} \geq 0$ if $\xi \neq 0$ and
    \item
    the {\em zero sum\/} condition 
    $\sum_{\xi \in \bbbz^D} a_{\xi} = 0$.
\end{itemize}
Specifically, these necessity and sufficiency results state that the
spatial discretization should be performed by local $3 \times 3$-kernels over
the spatial domain.

If we additionally require the spatial image domain to be mirror
symmetric over any line through the origin, then we should additionally require
\begin{itemize}
    \item
    the {\em symmetry\/} condition 
    $a_{-\xi} = a_{\xi}$.
 \end{itemize}

\subsection{Methodology}

To investigate the solutions of Equation~(\ref{gen-nD-lin-op}), a general approach
that we shall follow in this paper will consist of computing the 
generating function 
\begin{equation}
  \label{eq-gen-fcn-def}
  \varphi(z;\; s) 
  = \sum_{n \in \bbbz^D} c_{n} \, z^n
\end{equation}
of the set of filter coefficients $c_n$ in the filter for
computing the scale-space representation $L$ from
the input signal $f$
\begin{equation}
  L(x;\; s) = \sum_{n \in \bbbz^D} c_n \, f(x-n).
\end{equation}
By formally transforming (\ref{gen-nD-lin-op}) into generating
functions
\begin{equation}
  \label{eq-gen-fcn-diff-eq}
  \partial_s \left( \varphi_C(z;\; s) \right) 
  = \left(\sum_{\xi \in \bbbz^D} a_{\xi} \, z^{\xi}\right) \varphi_C(z;\; s),
\end{equation}
where $a_{\xi} z^{\xi}$ should be interpreted as multi-index
notation for $a_{\xi_1, \dots, \xi_D} z_1^{\xi_1} \dots z_D^{\xi_D}$,
we obtain 
\begin{equation}
  \label{eq-gen-fcn}
  \varphi(z;\; s) 
  = e^{s \sum_{\xi \in \bbbz^D} a_{\xi} z^{\xi}}.
\end{equation}
In compact operator notation, the solution of (\ref{gen-nD-lin-op})
at scale $s$ can equivalently be written 
\begin{equation}
  L = e^{s \, {\cal A}} f
\end{equation}
and this expression can be regarded as a general parameterization
of scale-space kernels on discrete image domains.
Alternatively, we obtain the Fourier transform
of the kernel by substituting
$z = e^{-i \omega}$ into the generating function
with the multi-index interpretation 
$z^{\xi} = z_1^{\xi_1} \dots z_D^{\xi_D} = 
e^{-i \omega_1 \xi_1 } \dots e^{-i \omega_D \xi_D } = e^{-i \omega^T \xi}$,
which gives
\begin{equation}
  \label{eq-FT-nD}
  \psi(\omega;\; s) 
  = \sum_{n \in \bbbz^D} c_n e^{-i \omega^T n}
  = e^{s \sum_{\xi \in \bbbz^D} a_{\xi} e^{-i \omega^T \xi}}.
\end{equation}
A main subject of this article is to interpret this result over a
two-dimensional image domain and to develop a theory for discrete affine
scale space.

The methodology we shall follow is to reparameterize the filter
class in terms of  the following basic difference operators, here over
the dimension $x$ and analogously for the dimension $y$:
\begin{align}
    \begin{split}
      \label{eq-def-1st-central-diff}
    (\delta_x f)(x) 
        & 
    = \left( f(x+1) - f(x-1) \right)/2,
    \end{split}\\
    \begin{split}
      \label{eq-def-2nd-central-diff}
    (\delta_{xx} f)(x) 
        &
       = f(x+1) - 2 f(x) + f(x-1),
    \end{split}
  \end{align}
which will be used as a basis for expressing the 
degrees of freedom in the coefficients $a_{\xi}$.
 
\section{Discrete 2-D affine Gaussian scale space}
\label{sec-theory-disc-aff-rec-fields-2D}

For a two-dimensional spatial domain, the discrete counterpart of the affine Gaussian scale-space
in (\ref{eq-aff-scsp-conv-kernel}) and (\ref{eq-aff-scsp-diff-eq})
is obtained if we require the discrete filter kernels in (\ref{gen-nD-lin-op})
to be mirror symmetric through the origin,
{\em i.e.\/}, $a_{i,j} = a_{-i,-j}$.
Then, the computational molecule of the infinitesimal operator ${\cal A}$
in (\ref{gen-nD-lin-op}) can be written as
\begin{equation}
  {\cal A} =
  \left(
    \begin{array}{ccc}
       a_{-1, 1} & a_{ 0, 1} & a_{ 1, 1} \\
       a_{-1, 0} & a_{ 0, 0} & a_{ 1, 0} \\
       a_{-1,-1} & a_{ 0,-1} & a_{ 1,-1} 
    \end{array}
  \right)
  =
  \left(
    \begin{array}{ccc}
       D & C & B \\
       A & -E & A \\
       B & C       & D 
    \end{array}
  \right)
\end{equation}
for some $A, B, C, D \geq 0$ and $E = A + B + C + D$.
In terms of the previously mentioned difference operators,
and with $\delta_{xy} = \delta_x \, \delta_y$ and
 $\delta_{xxyy} = \delta_{xx} \, \delta_{yy}$,
this computational molecule can with
\begin{align}
  \begin{split}
    \label{def-A-aff-gauss-scsp}
    A = \frac{C_{xx}}{2} - \frac{C_{xxyy}}{2},
  \end{split}\\
  \begin{split}
    B = \frac{C_{xy}}{4} + \frac{C_{xxyy}}{4},
  \end{split}\\
  \begin{split}
    C = \frac{C_{yy}}{2} - \frac{C_{xxyy}}{2},
  \end{split}\\
  \begin{split}
    \label{def-D-aff-gauss-scsp}
    D = - \frac{C_{xy}}{4} + \frac{C_{xxyy}}{4},
  \end{split}
\end{align}
be reparameterized as
\begin{equation}
  \label{eq-inf-gen-disc-aff-scsp}
  {\cal A} =
  \frac{1}{2}
  \left(
    \begin{array}{ccc}
      -C_{xy}/2 &          C_{yy}      & C_{xy}/2 \\
       C_{xx}   & -2 (C_{xx} + C_{yy}) & C_{xx} \\
       C_{xy}/2 &          C_{yy}      & - C_{xy}/2 
    \end{array}
  \right)
  +
  \frac{C_{xxyy}}{4}
  \left(
    \begin{array}{rrr}
       1 & -2 &  1 \\
      -2 &  4 & -2 \\
       1 & -2 &  1 
    \end{array}
  \right).
\end{equation}

\subsection{Semi-discrete affine diffusion equation}

With the parameterization in terms of $C_{xx}$, $C_{xy}$, $C_{yy}$
and $C_{xxyy}$,
the corresponding {\em discrete affine Gaussian scale-space\/}
is given as the solution of the 
semi-discrete differential equation
\begin{equation}
  \label{eq-diff-eq-disc-aff-scsp}
  \partial_{s} L 
  = \frac{1}{2} \, (C_{xx} \, \delta_{xx} L 
		    + 2 C_{xy} \, \delta_{xy} L 
		    + C_{yy} \, \delta_{yy} L)
    + \frac{C_{xxyy}}{4} \, \delta_{xxyy} L,
\end{equation}
which in turn can be interpreted as a second-order
discretization of the diffusion equation
associated with the continuous affine Gaussian scale-space (\ref{eq-aff-scsp-diff-eq})
\begin{equation}
  \label{eq-diff-eq-cont-aff-scsp}
  \partial_s L = 
  \tfrac{1}{2} (C_{xx} \, L_{xx} + 2 C_{xy} \, L_{xy} + C_{yy} \, L_{yy}),
\end{equation}
where $C_{xx}, C_{yy} > 0$ and $C_{xx} \, C_{yy} - C_{xy}^2 > 0$ 
are necessary conditions for the infinitesimal operator ${\cal A}$
to be positive definite.

\subsection{Generating function and Fourier transform}

With the transform variable for the generating function over the $x$-dimension
denoted by $z$ and the transform variable for the generating function over the
$y$-dimension denoted by $w$, the generating function
(\ref{eq-gen-fcn}) as parameterized by the coefficients 
$C_{xx}$, $C_{xy}$, $C_{yy}$ and $C_{xxyy}$ becomes
\begin{align}
  \begin{split}
    \label{eq-gen-fcn-aff-scsp}
    \varphi(z, w) = \exp( 
      & C_{xx} \, (z - 2 + z^{-1})/2 + 
        C_{yy} \, (w - 2 + w^{-1})/2 +
  \end{split}\\
  \begin{split}
      & C_{xy} \, (z - z^{-1})(w - w^{-1})/4 +
        C_{xxyy} \, (z - 2 + z^{-1}) \, (w - 2 + w^{-1})/4).
  \end{split}\nonumber
\end{align}
Let $c_{x,y}$ denote the (non-negative) filter coefficients of the discretized
spatial filter, which are guaranteed to sum to one because $\varphi(1,
1) = \sum_{(x, y) \in \bbbz^2} c_{x,y} = 1$, and let $E()$ denote the averaging operator over the
spatial domain using these filter coefficients as weights:
\begin{equation}
   E(h) = \sum_{(x, y) \in \bbbz^2} c_{x,y} \, h_{x,y}.
\end{equation}
From well-known properties of the generating function
\begin{align}
  \begin{split}
     m_x = E(x) = \varphi_z(1, 1),
  \end{split}\\
  \begin{split}
     m_y = E(y) = \varphi_w(1, 1),
  \end{split}\\
 \begin{split}
     \Sigma_{xx} = E(x^2) - (E(x))^2 = \varphi_{zz}(1, 1) + \varphi_{z}(1, 1) - \varphi_z^2(1, 1),
  \end{split}\\
  \begin{split}
     \Sigma_{xy} = E(x \, y) - E(x) \, E(y) = \varphi_{zw}(1, 1) - \varphi_z(1, 1) \, \varphi_w(1, 1),
  \end{split}\\
  \begin{split}
     \Sigma_{yy} = E(y^2) - (E(y))^2 = \varphi_{ww}(1, 1) + \varphi_{w}(1, 1) - \varphi_w^2(1, 1),
  \end{split}
\end{align}
which can be derived by differentiating the definition of the
generating function
\begin{equation}
   \varphi(z, w) = \sum_{(x, y) \in \bbbz^2} c_{x,y} \, z^x \, w^y
\end{equation}
with respect to the transform variables $z$ and $w$ and then setting
$z = 1$ and $w = 1$,
it follows that the corresponding discrete kernels have spatial mean vector
\begin{equation}
  \label{eq-mean-semi-disc-kernels}
     m 
     = \left( \begin{array}{c} m_x \\ m_y \end{array} \right)
     = \left( \begin{array}{c} 0 \\ 0 \end{array} \right)
\end{equation}
and spatial covariance matrix 
\begin{equation}
  \label{eq-cov-semi-disc-kernels}
     \Sigma
     = \left(
           \begin{array}{cc} 
               \Sigma_{xx} & \Sigma_{xy} \\
               \Sigma_{xy} & \Sigma_{yy} 
           \end{array} 
         \right)
     = \left(
           \begin{array}{cc} 
               C_{xx} & C_{xy} \\
               C_{xy} & C_{yy} 
           \end{array} 
         \right).
\end{equation}
Thus, by this definition of discrete affine Gaussian scale space,
provided that the filter parameters $C_{xx}$, $C_{xy}$, $C_{yy}$ and 
$C_{xxyy}$ are chosen such that the semi-discrete affine diffusion
equation (\ref{eq-diff-eq-disc-aff-scsp}) corresponds to a non-negative
discretization of the continuous affine diffusion equation (\ref{eq-diff-eq-cont-aff-scsp}), 
the mean vectors and the covariance matrices of the corresponding discrete affine
Gaussian kernels are {\em exactly equal\/} to the mean values and
covariance matrices of the corresponding continuous affine Gaussian
kernels. 

Such a property would, for example, not hold if we instead
would perform the discrete spatial smoothing by convolution with the
sampled affine Gaussian kernel or an FFT-implementation based on
sampling the continuous Fourier transform of the continuous affine
Gaussian kernel.

\begin{figure}[hbtp]
  \begin{center}
    \begin{tabular}{cccccc}
      \hspace{-4mm}
\includegraphics[width=0.15\textwidth]{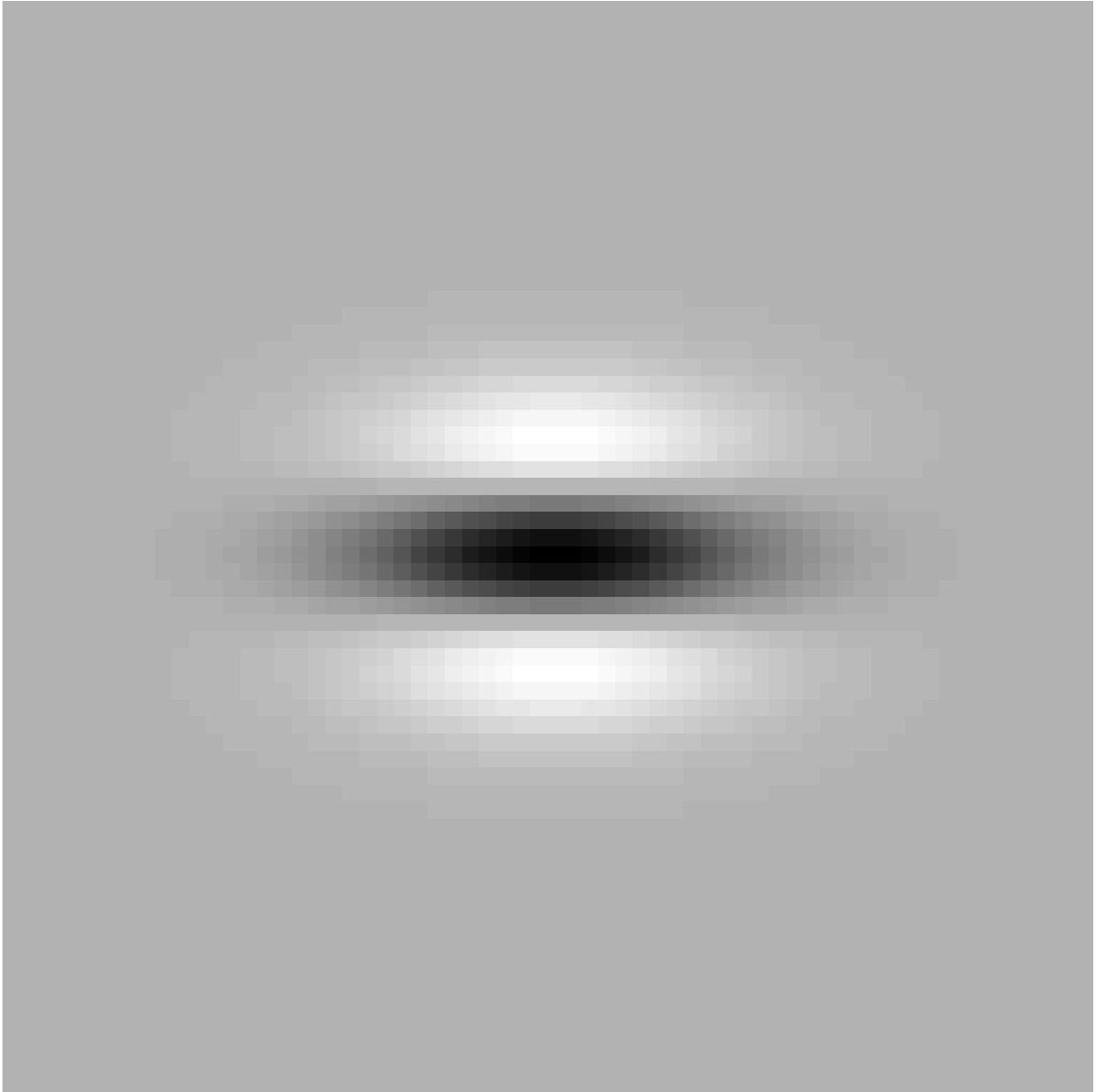} \hspace{-4mm} &
      \includegraphics[width=0.15\textwidth]{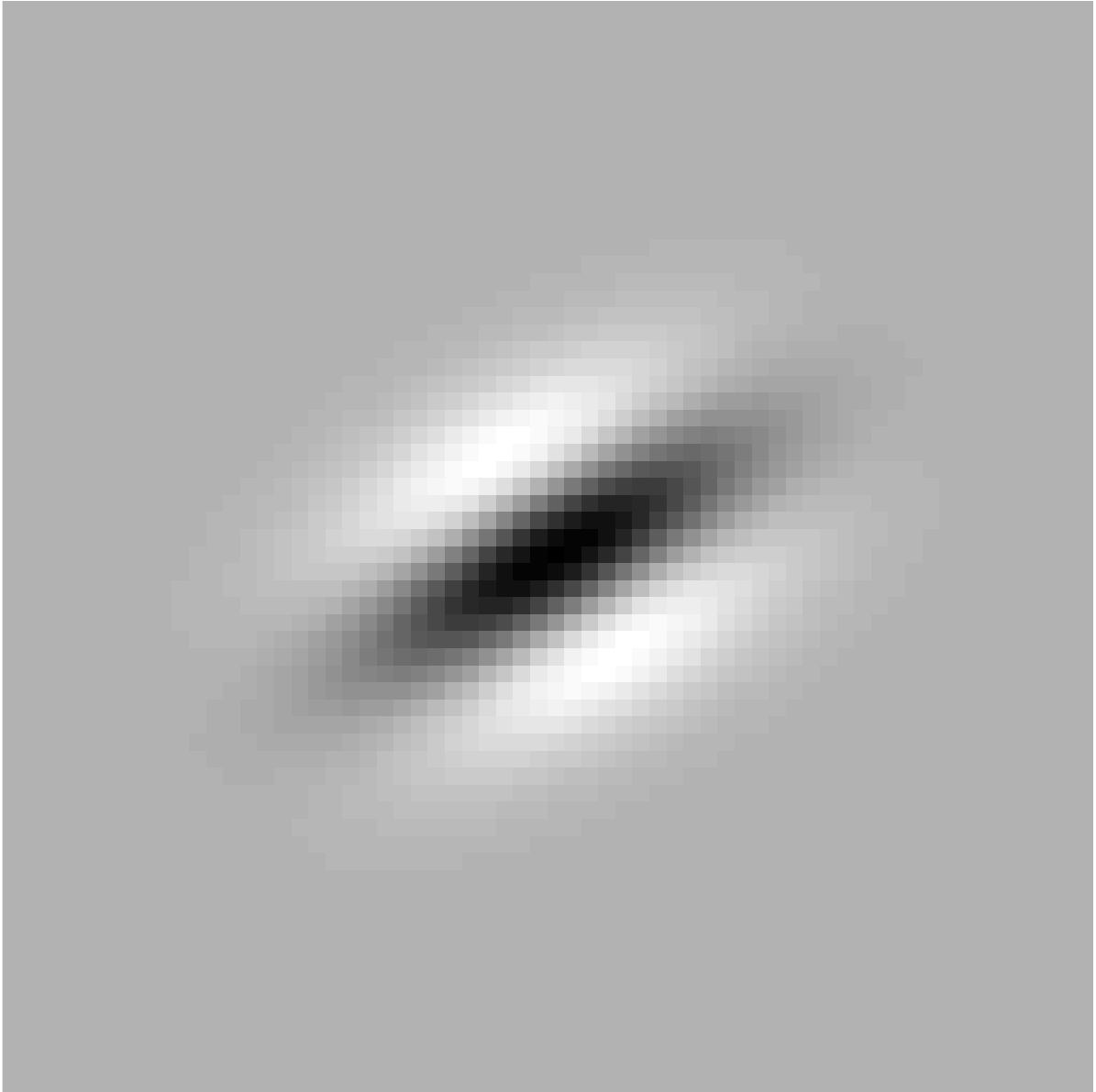} \hspace{-4mm} &
      \includegraphics[width=0.15\textwidth]{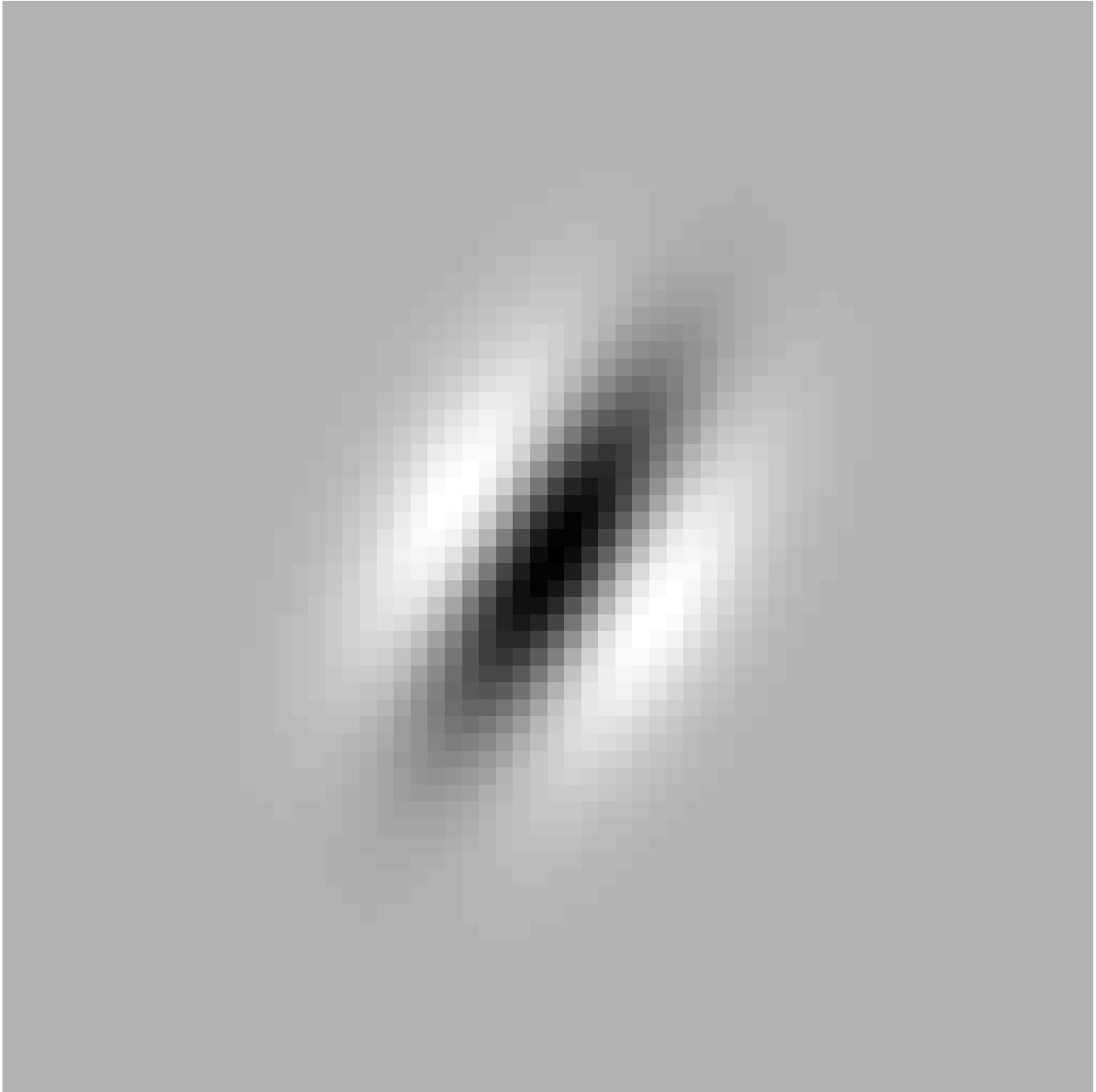} \hspace{-4mm} &
      \includegraphics[width=0.15\textwidth]{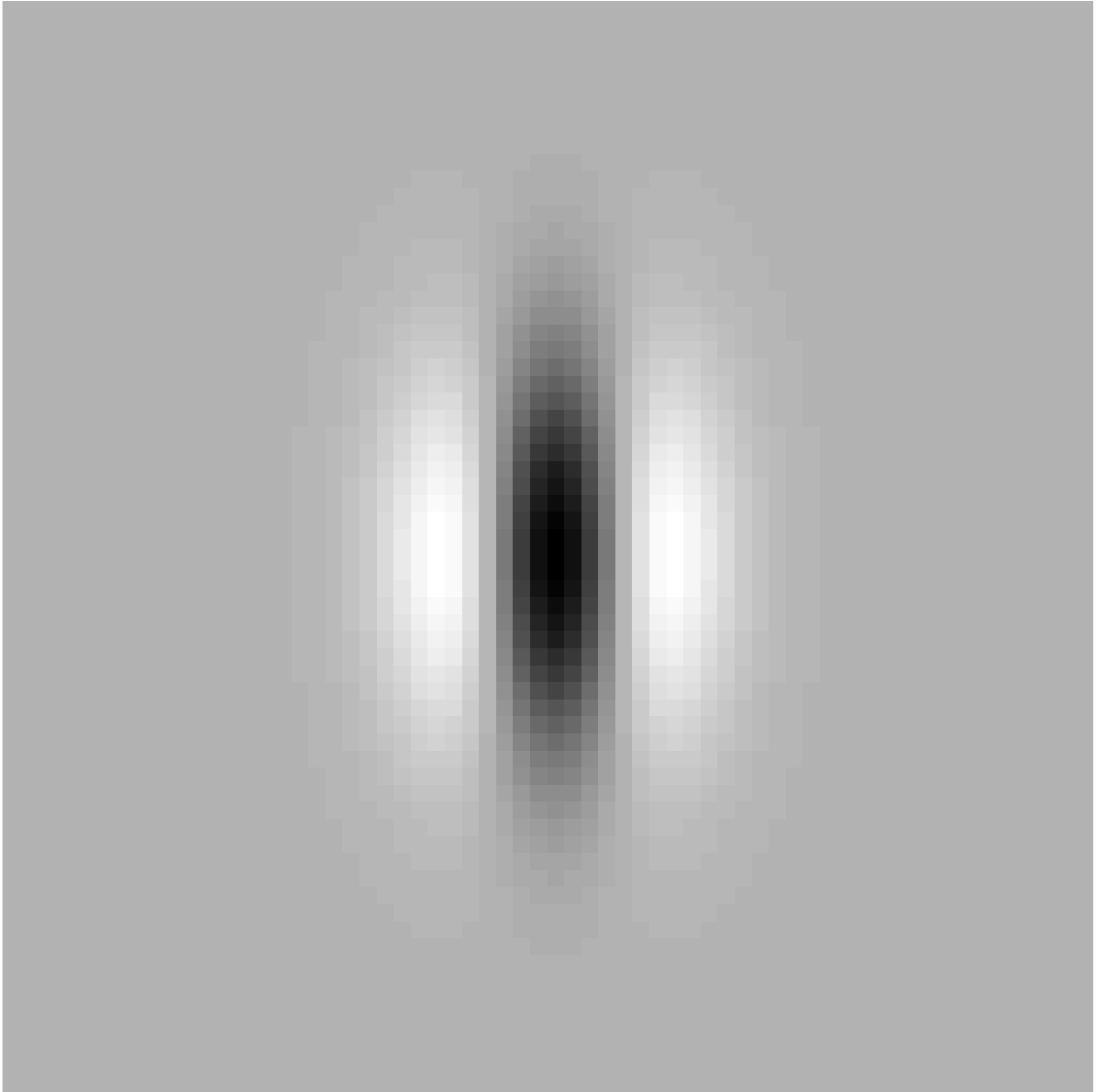} \hspace{-4mm} &
      \includegraphics[width=0.15\textwidth]{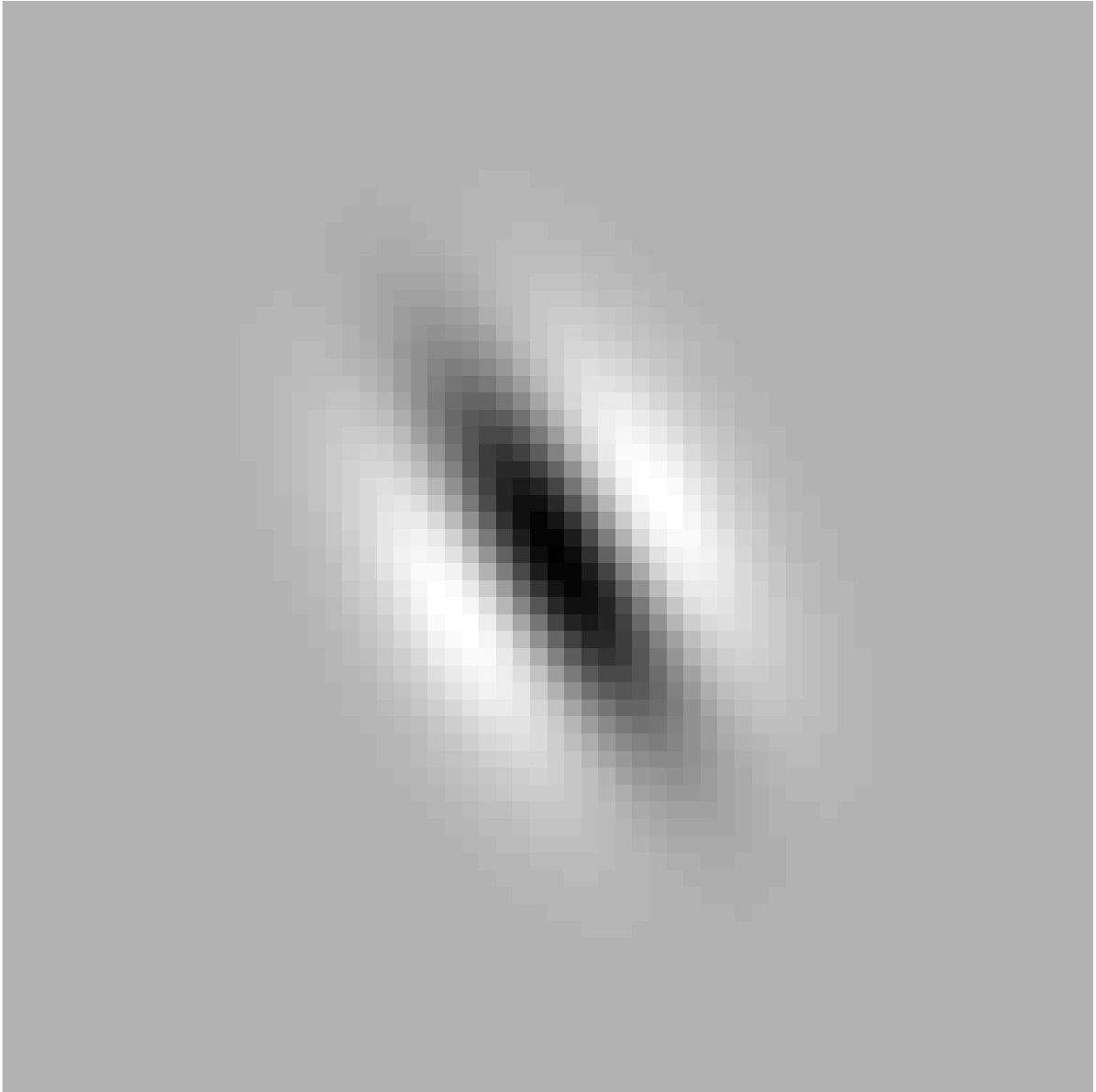} \hspace{-4mm} &
      \includegraphics[width=0.15\textwidth]{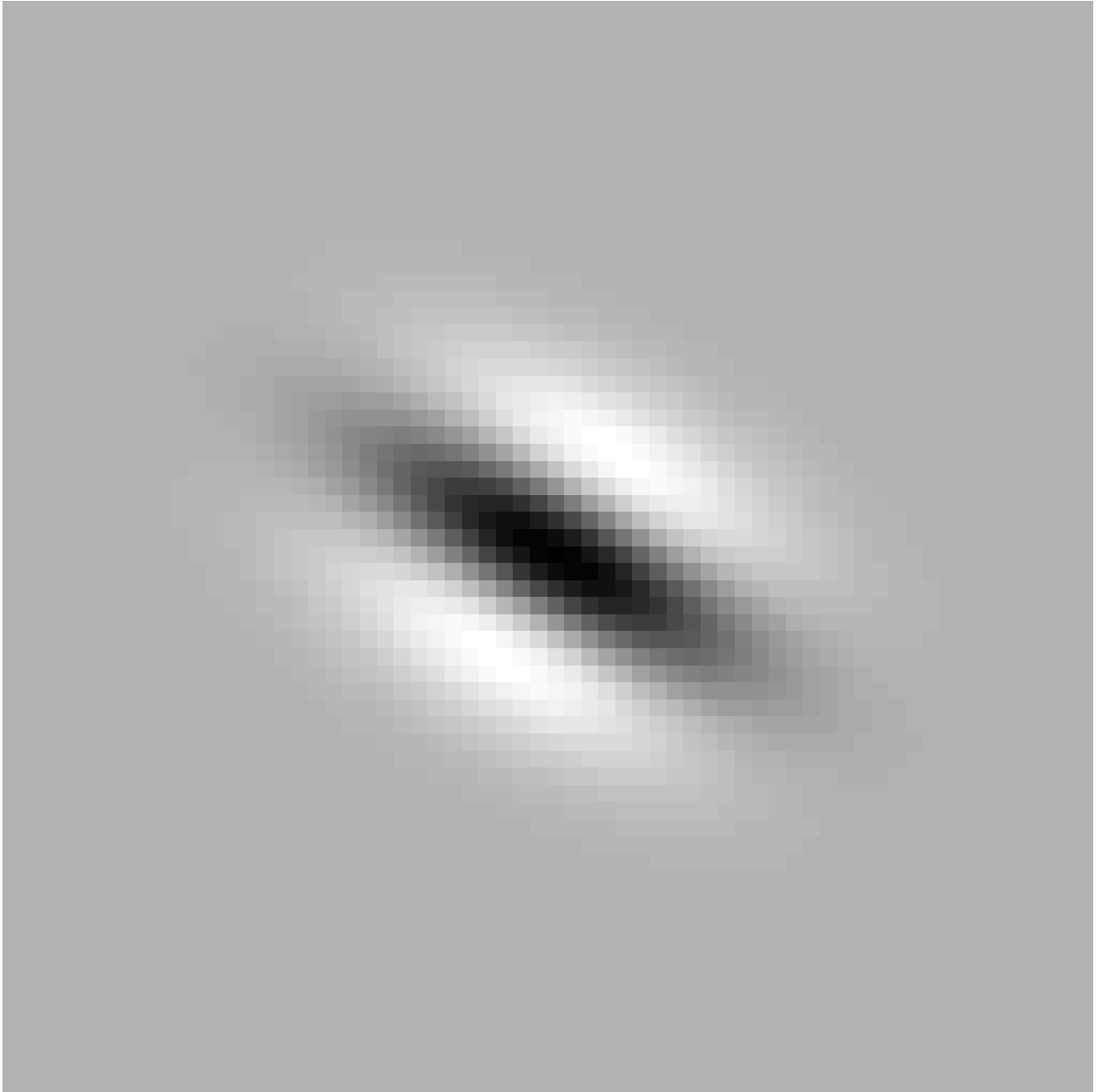} \hspace{-4mm} \\
      \hspace{-4mm}
\includegraphics[width=0.15\textwidth]{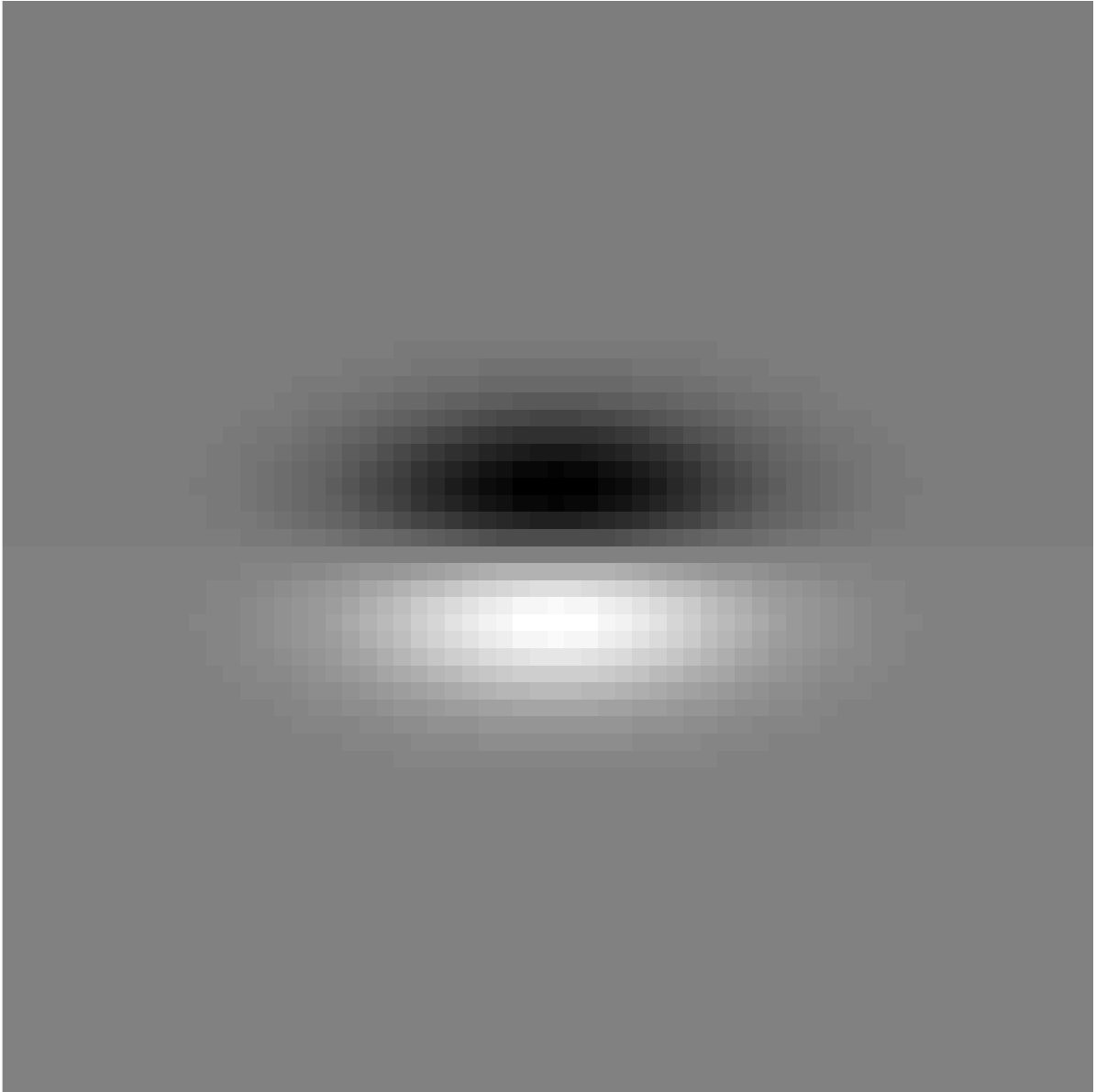} \hspace{-4mm} &
      \includegraphics[width=0.15\textwidth]{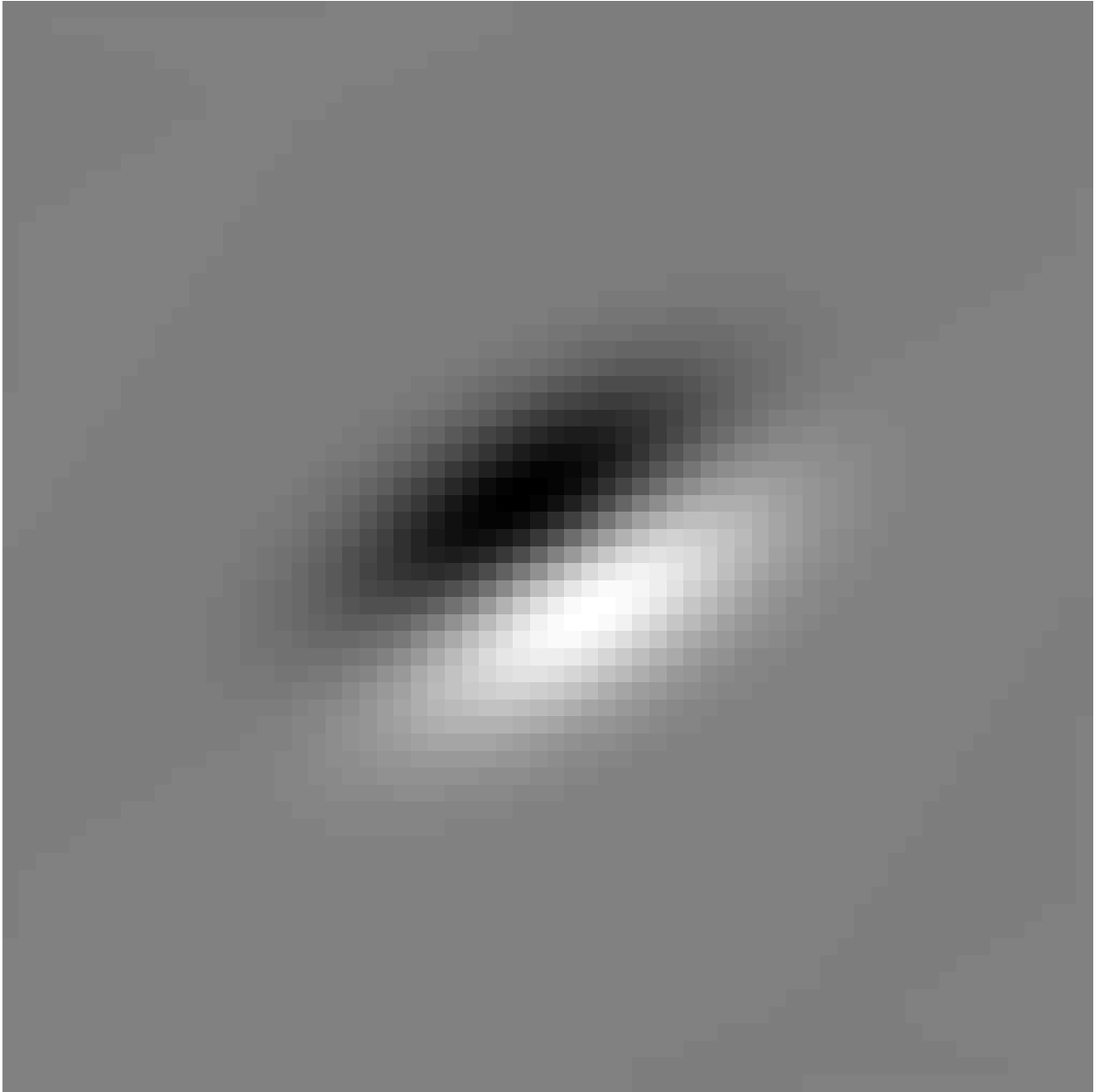} \hspace{-4mm} &
      \includegraphics[width=0.15\textwidth]{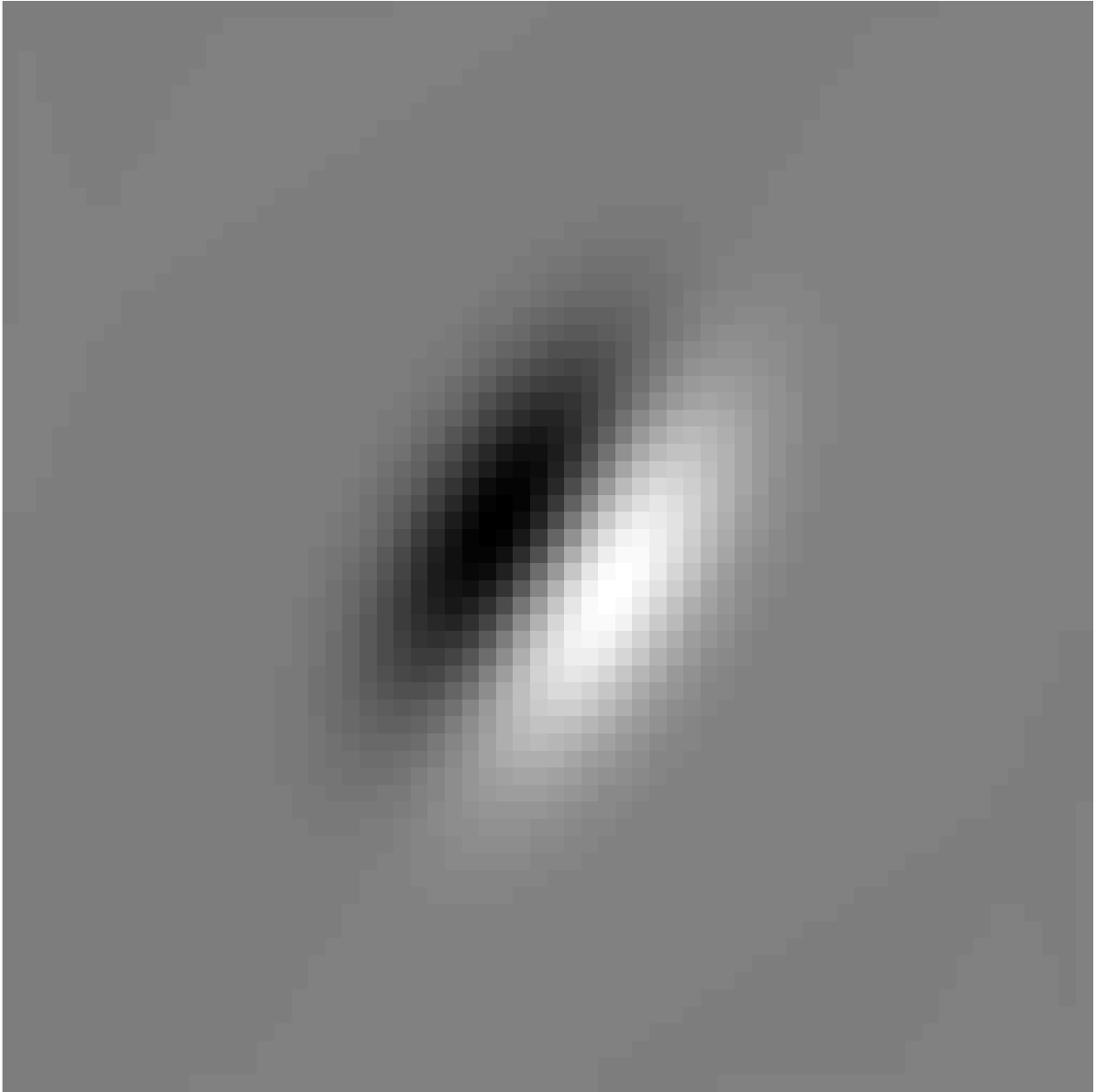} \hspace{-4mm} &
      \includegraphics[width=0.15\textwidth]{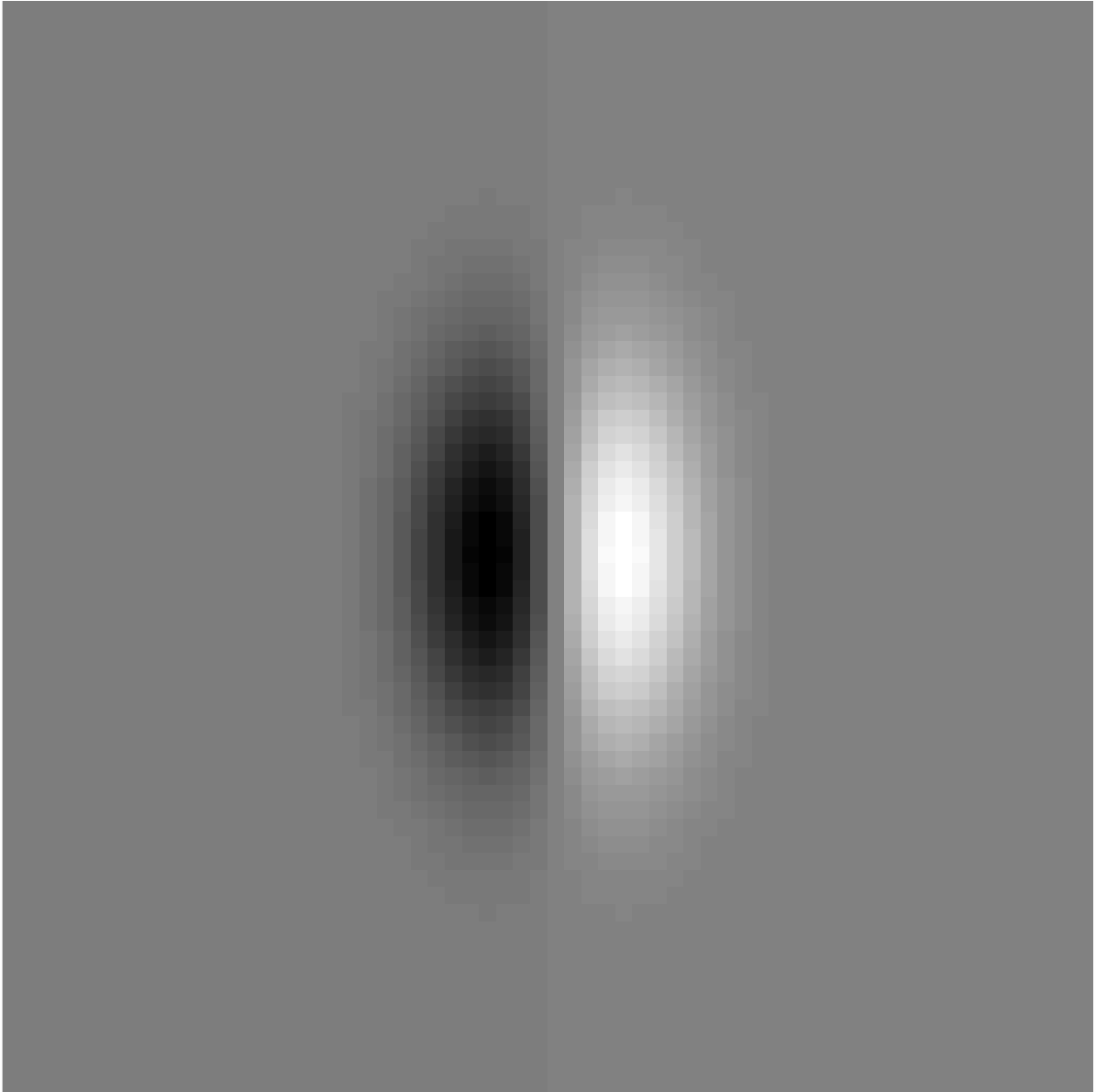} \hspace{-4mm} &
      \includegraphics[width=0.15\textwidth]{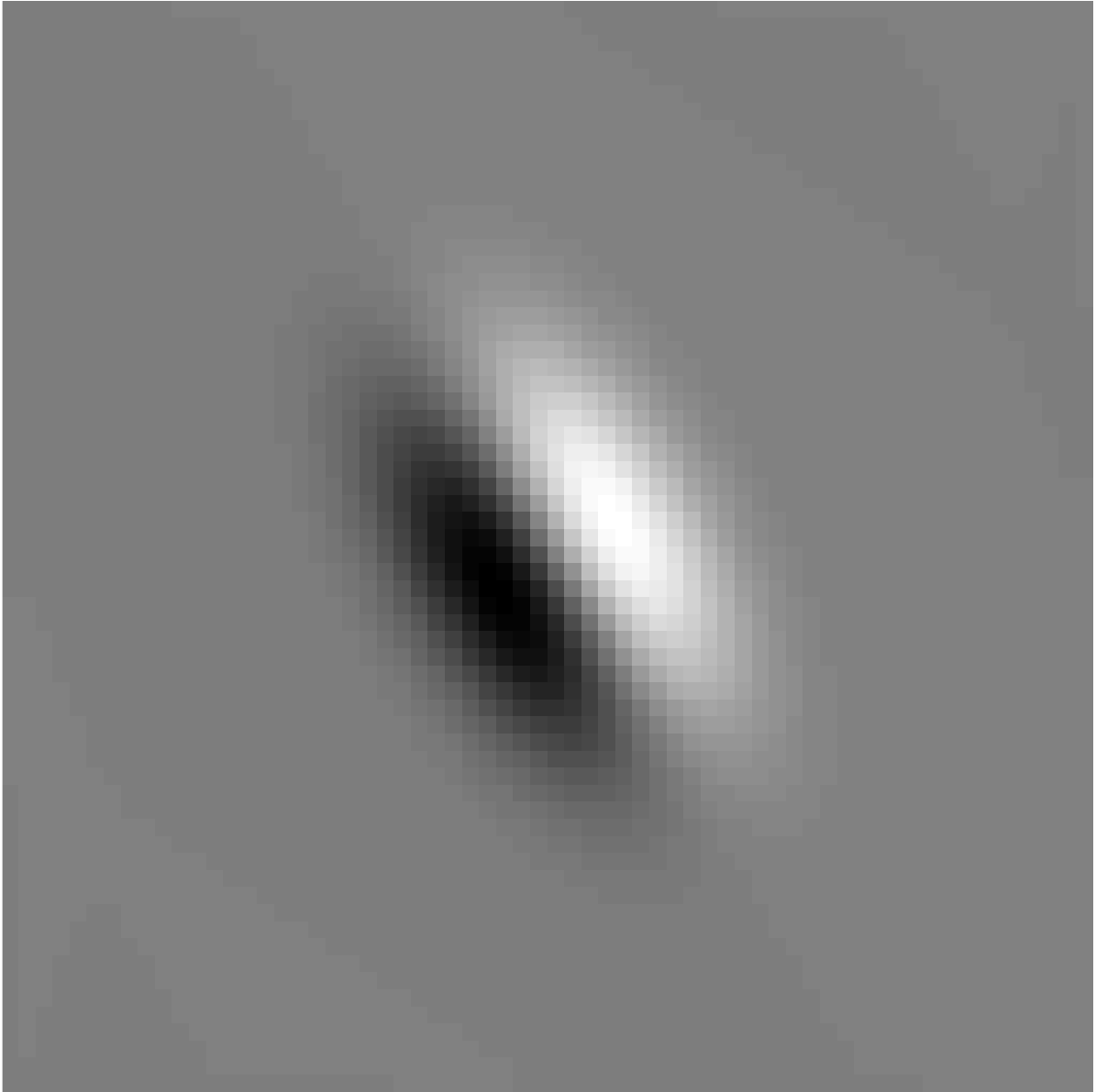} \hspace{-4mm} &
      \includegraphics[width=0.15\textwidth]{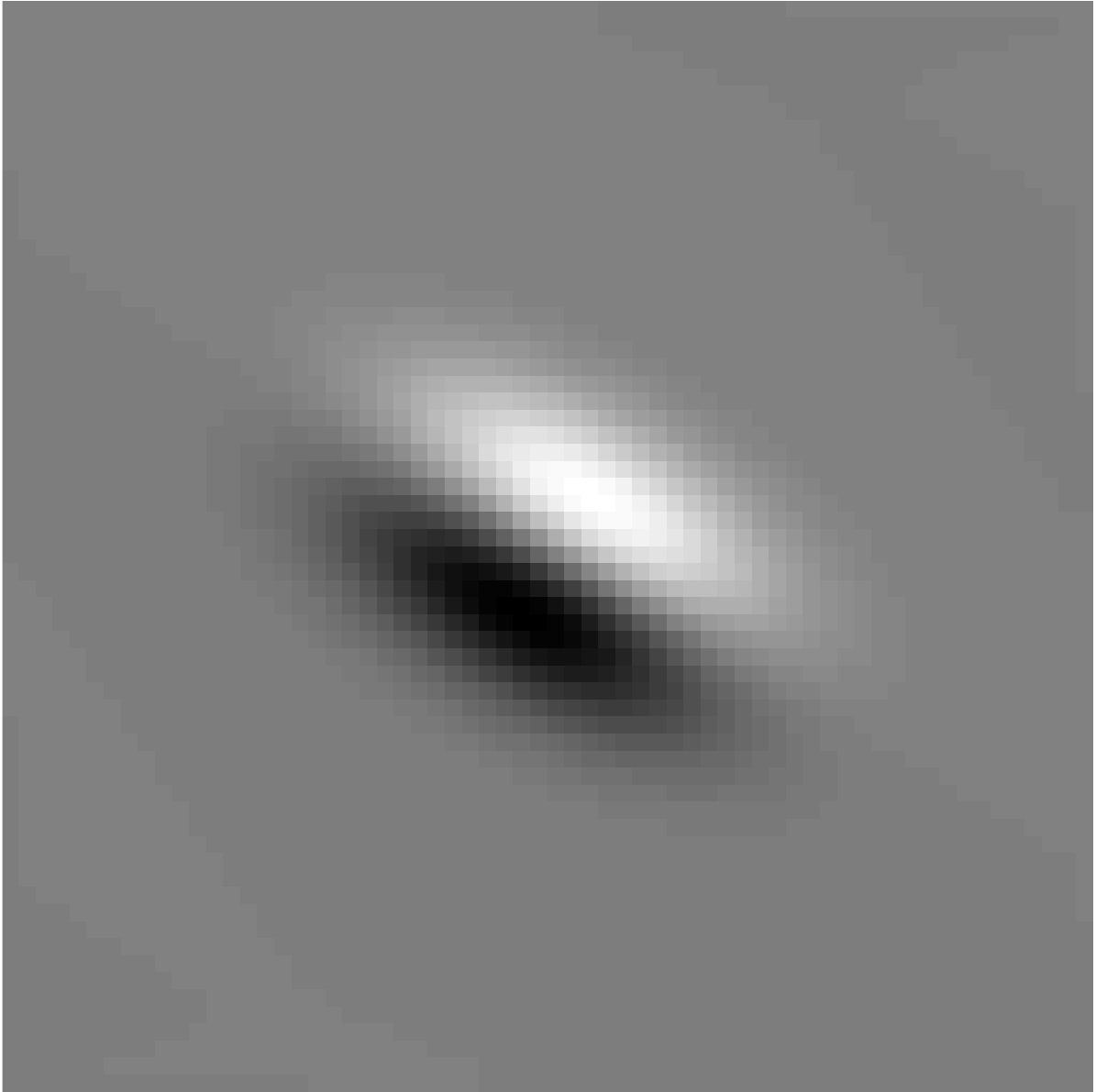} \hspace{-4mm} \\
      \hspace{-4mm}
\includegraphics[width=0.15\textwidth]{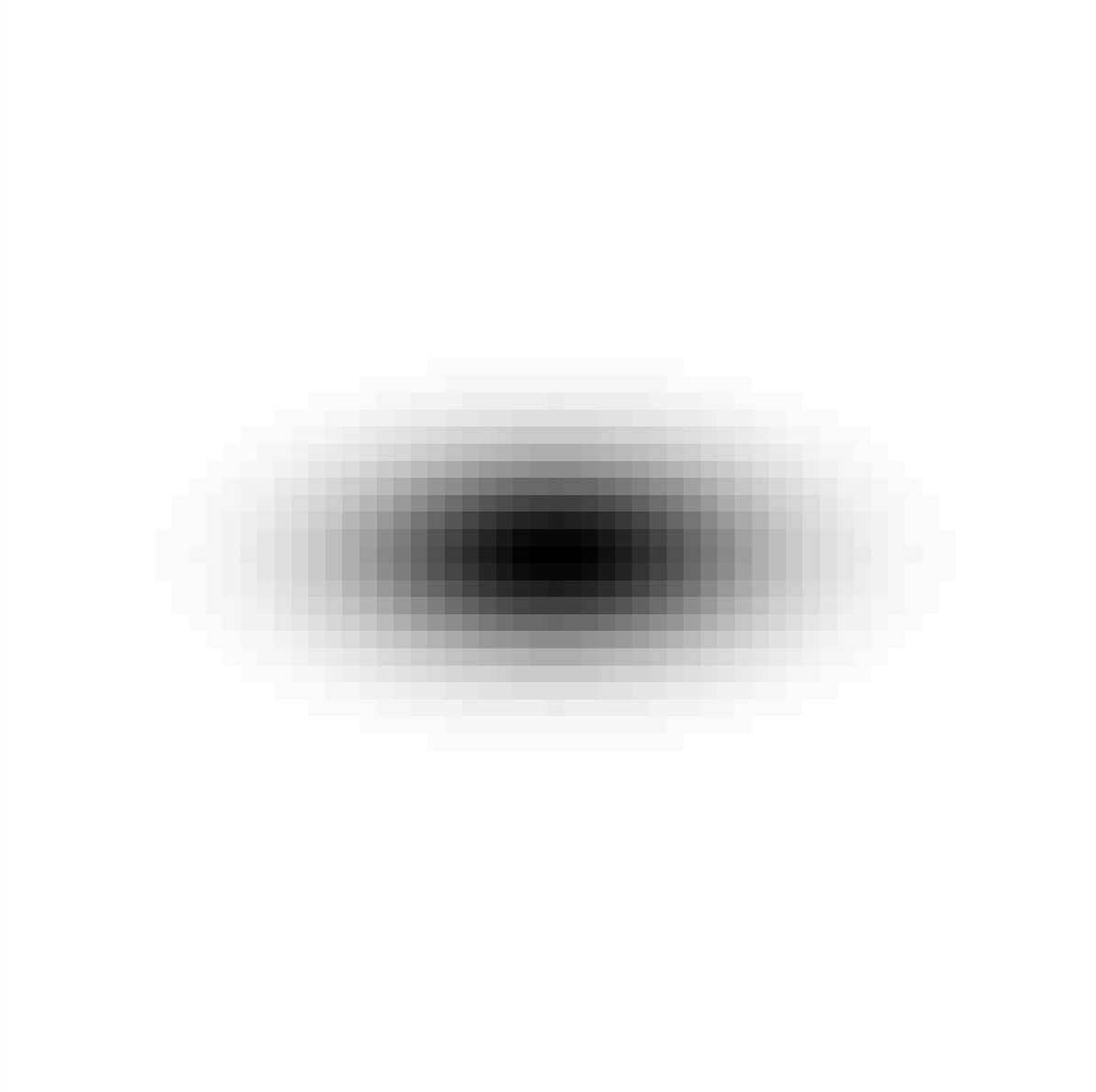} \hspace{-4mm} &
      \includegraphics[width=0.15\textwidth]{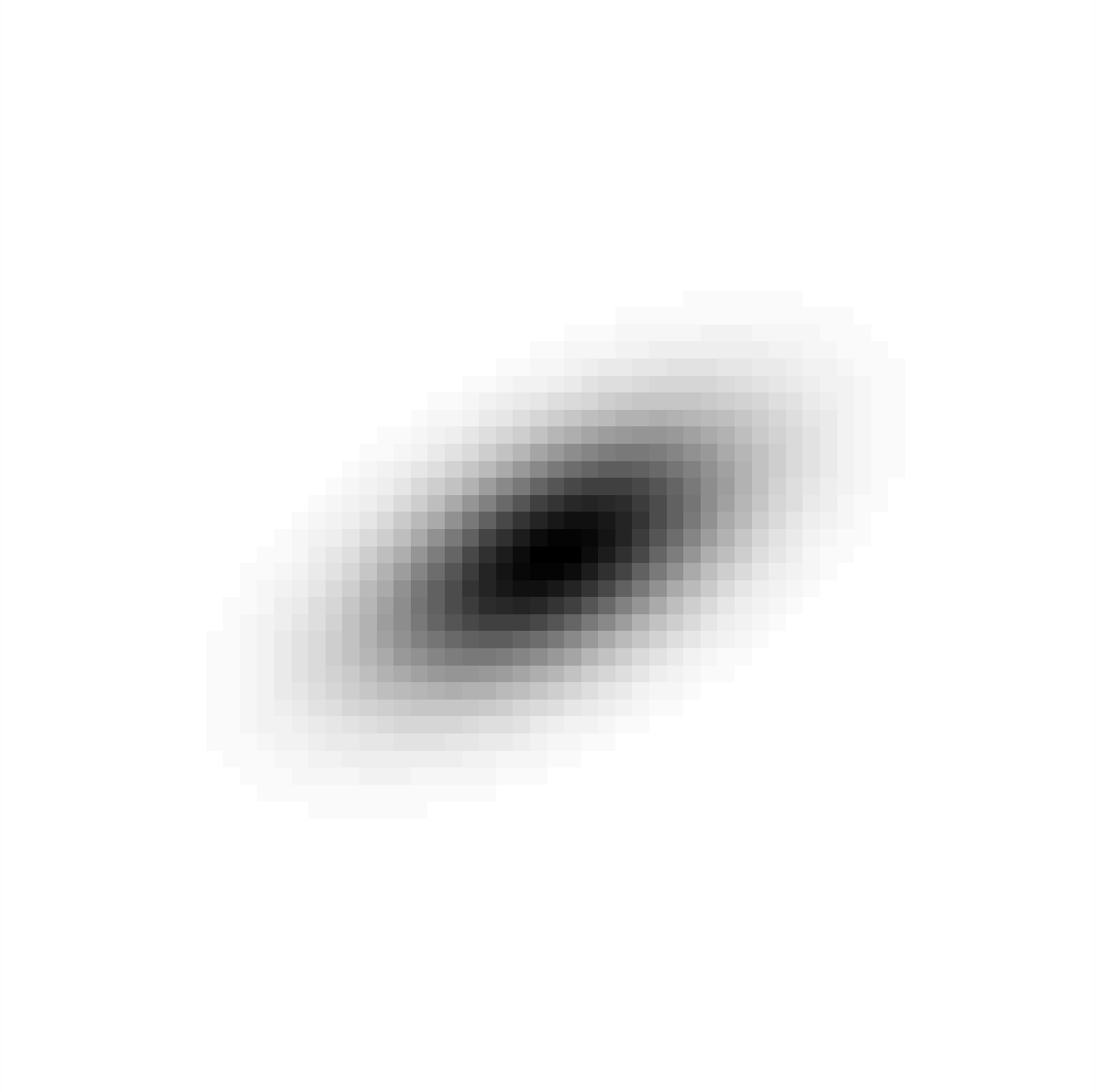} \hspace{-4mm} &
      \includegraphics[width=0.15\textwidth]{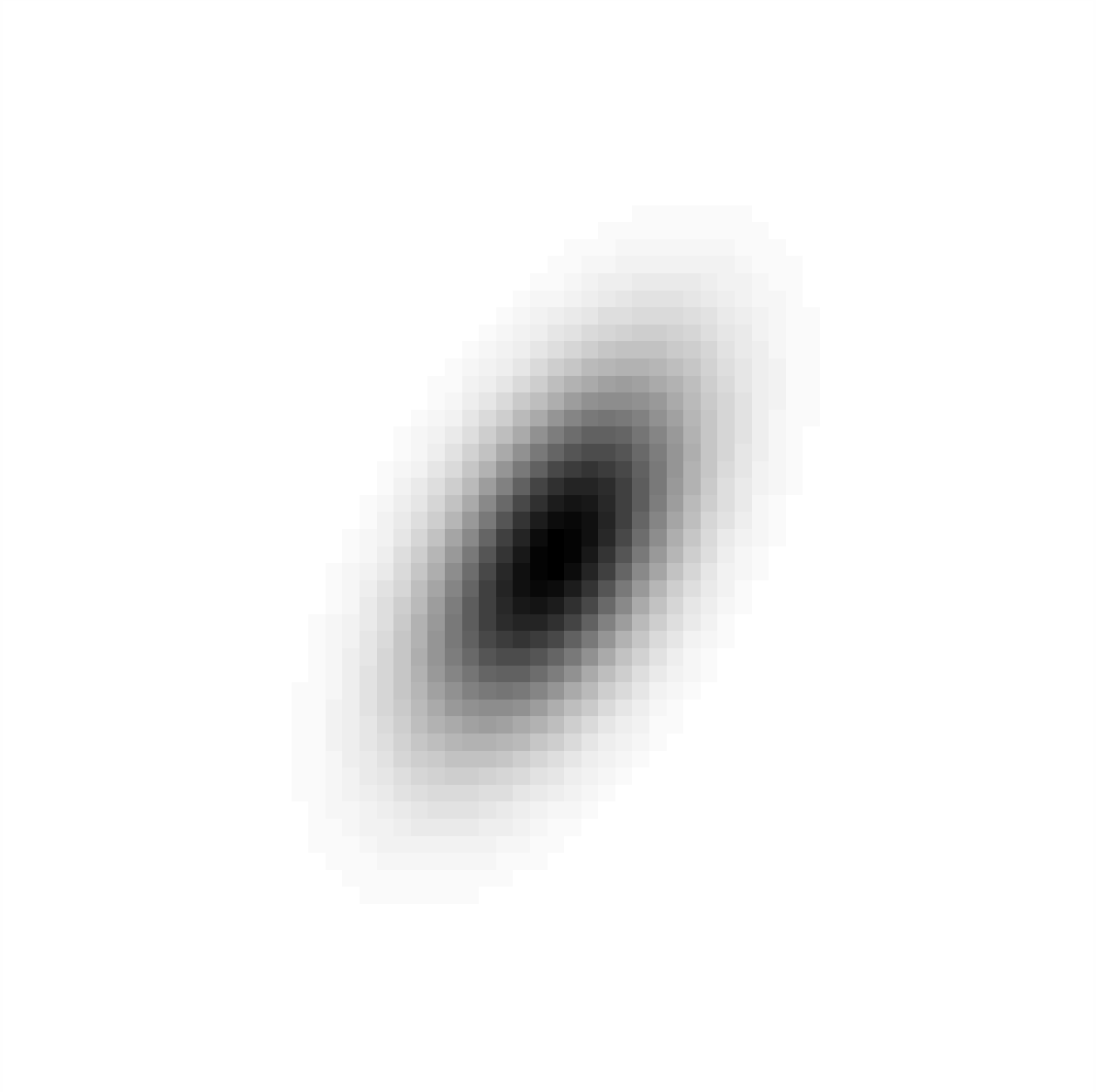} \hspace{-4mm} &
      \includegraphics[width=0.15\textwidth]{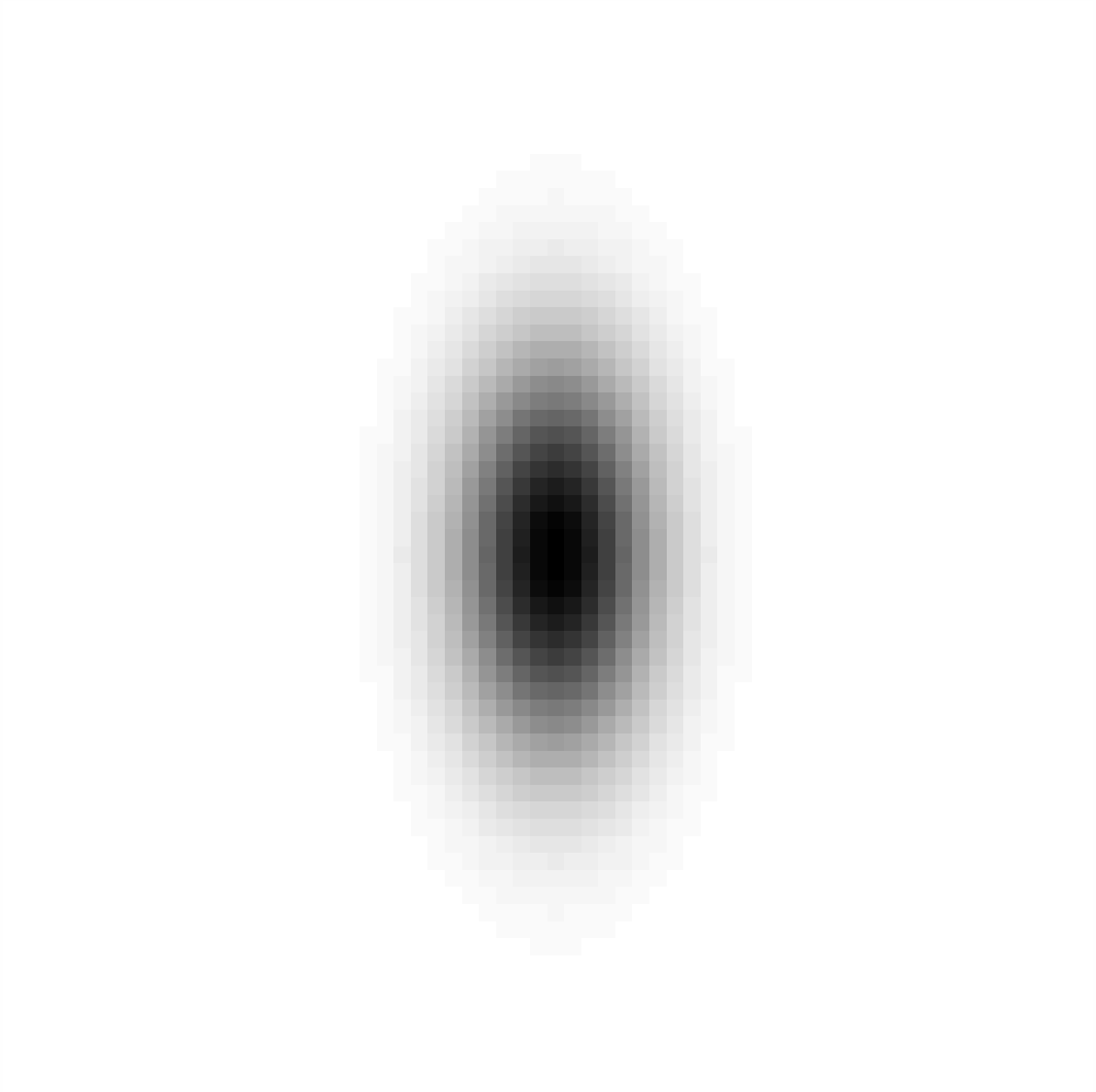} \hspace{-4mm} &
      \includegraphics[width=0.15\textwidth]{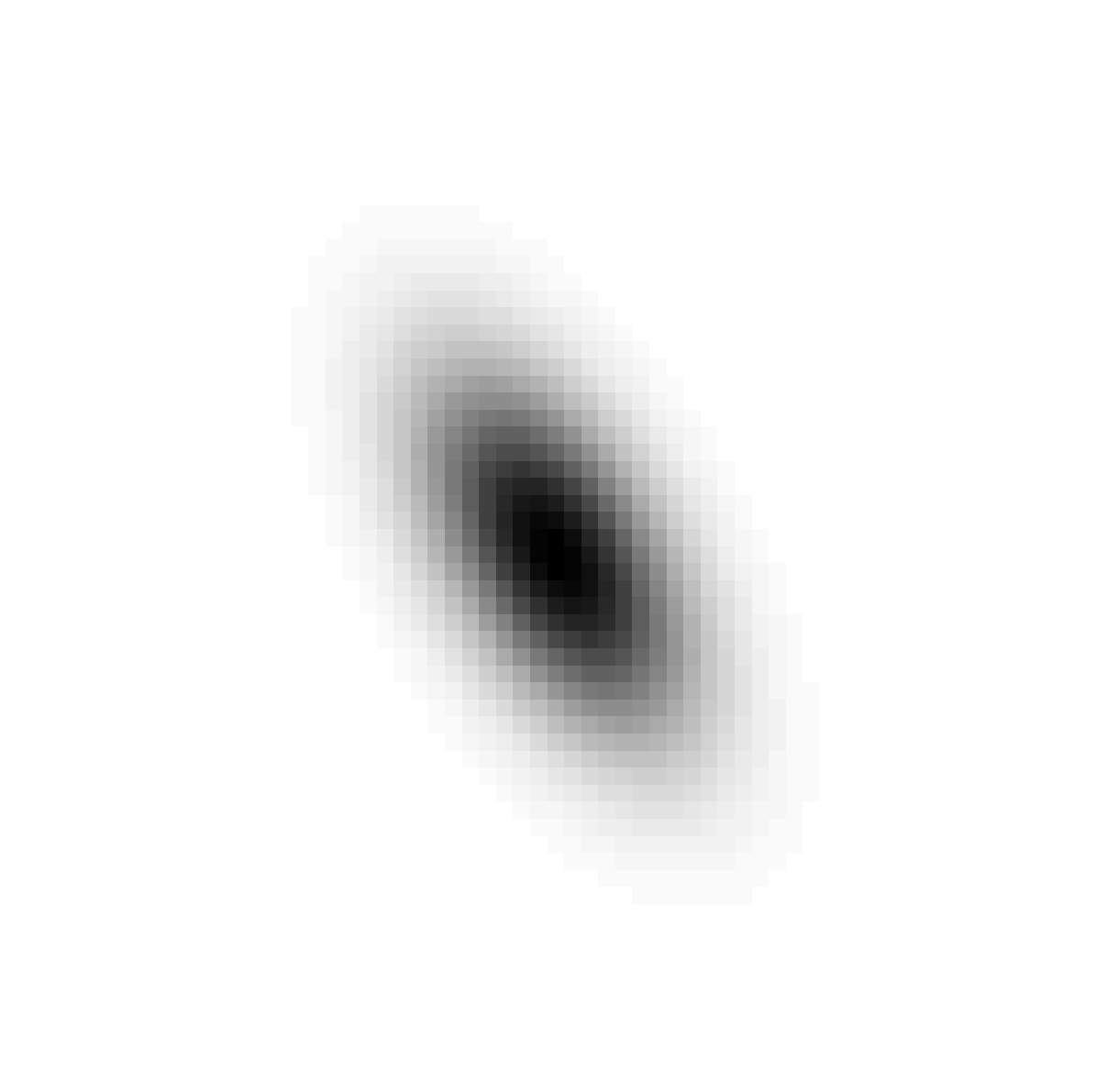} \hspace{-4mm} &
      \includegraphics[width=0.15\textwidth]{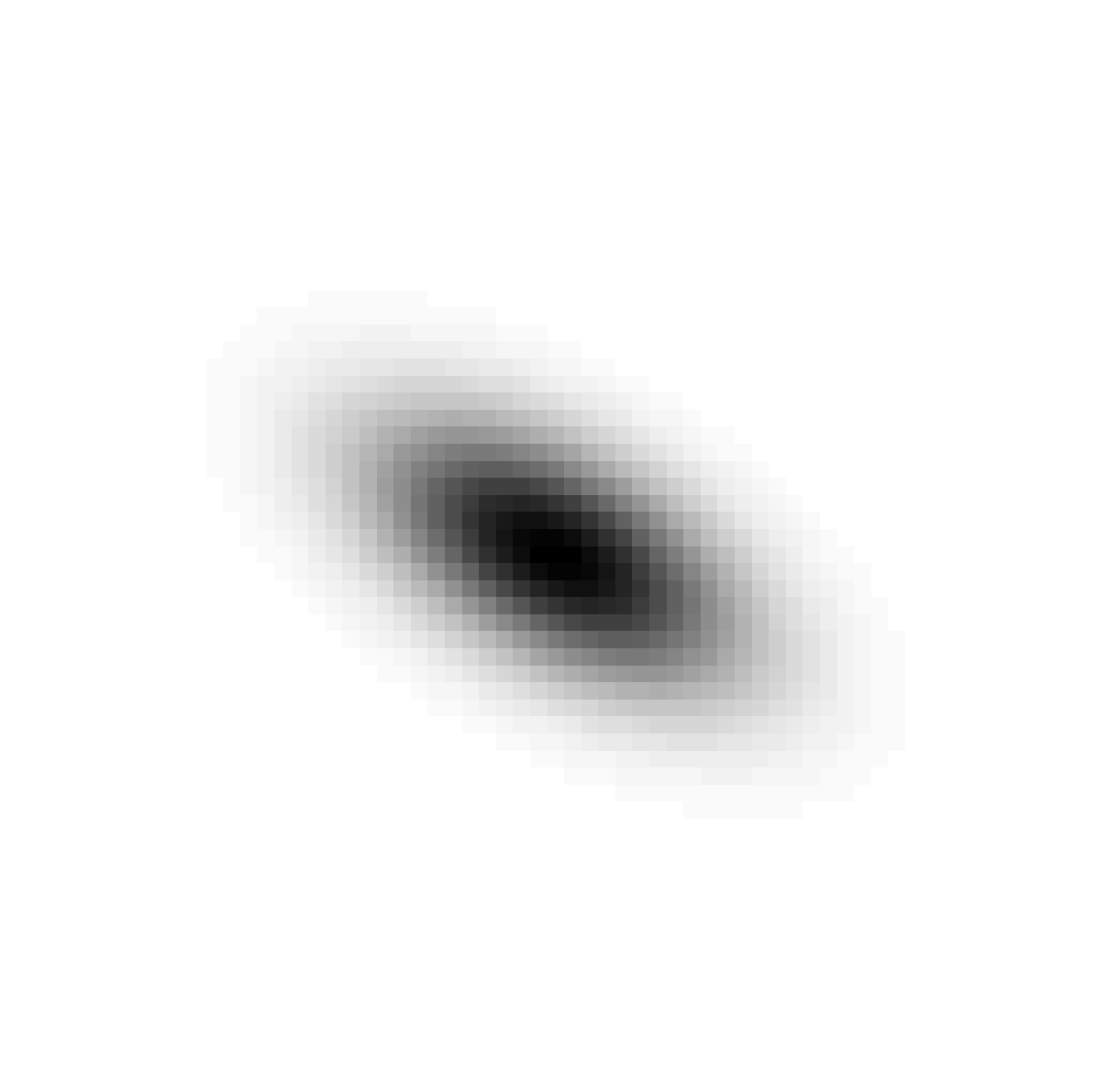} \hspace{-4mm} \\
    \end{tabular} 
  \end{center}
  \vspace{-4mm}
  \caption{Examples of discrete affine Gaussian kernels $h(x, y;\; \Sigma)$ and their equivalent directional
    derivative approximation kernels $\delta_{\orth \varphi}h(x, y;\; \Sigma)$ and
    $\delta_{\orth \varphi \orth \varphi}h(x, y;\; \Sigma)$ 
    up to order two in the two-dimensional case, here as generated
    from the explicit expression of the Fourier transform (\ref{eq-FT-semi-disc-aff-scsp})
    for $\lambda_1 = 64$, $\lambda_2=16$ and
            $\alpha = 0, \pi/6, \pi/3, \pi/2, 2\pi/3, 5\pi/6$ and with
          $C_{xxyy} = |C_{xy}|$ according to
          (\ref{eq-choice-Cxxyy-minimal-pos-discr}).
          (Kernel size: $64 \times 64$ pixels.)}
  \label{fig-aff-elong-filters-dir-ders-fft}
\end{figure}

From the explicit expression for the Fourier transform, we can in turn obtain the continuous Fourier
transform over an infinite discrete spatial domain by letting $(z, w) = (e^{i u}, e^{i v})$, which gives
\begin{align}
  \begin{split}
    \psi(u, v) 
      & = \varphi(e^{iu}, e^{iv})
  \end{split}\nonumber\\
  \begin{split}
      & = \exp(- C_{xx} \, (1 - \cos u) - C_{yy} \, (1 - \cos v) 
          + C_{xy} \sin u \sin v 
  \end{split}\nonumber\\
  \begin{split}
      \label{eq-gen-fcn-semi-disc-aff-scsp}
      & \phantom{= \exp(}  
        + C_{xxyy} \, (1 - \cos u) \, (1 - \cos v)).
  \end{split}
\end{align}
The corresponding discrete Fourier transform over a finite image of
size $M \times N$ can then be obtained by setting $u = 2m\pi/M$ and
$v = 2n\pi/N$ for $m = 0 \dots M-1$ and $n = 0 \dots N-1$, which gives
\begin{align}
  \begin{split}
    \psi(m, n) 
      & = \varphi(e^{\frac{i2m\pi}{M}}, e^{\frac{i2n\pi}{N}})
  \end{split}\nonumber\\
  \begin{split}
      & = \exp(- C_{xx} \, (1 - \cos \tfrac{2m\pi}{M} ) - C_{yy} \, (1 - \cos \tfrac{2n\pi}{N}) 
          + C_{xy} \sin \tfrac{2m\pi}{M} \sin \tfrac{2n\pi}{N}
  \end{split}\nonumber\\
  \begin{split}
      \label{eq-FT-semi-disc-aff-scsp}
      & \phantom{= \exp(}  
        + C_{xxyy} \, (1 - \cos \tfrac{2m\pi}{M}) \, (1 - \cos \tfrac{2n\pi}{N})).
  \end{split}
\end{align}
Up to numerical errors in an FFT algorithm, we can thereby from
these explicit expressions for the Fourier transform of the discrete
affine Gaussian kernels generate discrete affine Gaussian kernels and
compute convolutions with them with spatial covariance matrices that
are exactly equal to the corresponding spatial covariance matrices of
the continuous theory. Hence, there is no need for resolving to an
approximate numerical method for computing the discrete affine
Gaussian scale-space representation. What remains is to determine the free parameter
$C_{xxyy}$ in the spatial discretization as function of the parameters
$C_{xx}$, $C_{xy}$ and $C_{yy}$ that determine the spatial covariance
matrix of the discrete affine Gaussian kernel.

Note, however, that although being a much better spatial
discretization, in the sense of preserving scale-space properties for
the discrete implementation compared to (i)~discretizing affine
Gaussian receptive fields by spatial sampling, (ii)~sampling the Fourier
transform of the affine Gaussian kernel or by using (iii)~affine
warping alternatively 
(iv)~an implementation in terms of recursive filters
as outlined in Section~\ref{sec-disc-aff-gauss-rec-fields},
the primary type of implementation that we are
aiming at for purposes of practical implementation is not 
only a Fourier-based
implementation of discrete affine Gaussian receptive fields.
This Fourier-based model is intended for two purposes:
(i)~as an idealized theoretical model to be used also for practical
implementation in situations where an FFT-implementation of
the spatial smoothing operation can be regarded as affordable given
the external conditions,
to make it possible to compute maximally accurate discrete receptive
fields, and
(ii)~a reference and benchmark to be used for comparisons in relation
to the additional discretization
in the scale direction that will be treated later in Section~\ref{sec-theory-disc-aff-rec-fields-3x3}.

\subsection{Free parameter and positivity constraint}
\label{sec-free-Cxxyy-semi-disc-aff-scsp}

There is one free parameter
$C_{xxyy} \geq 0$ in 
(\ref{eq-diff-eq-disc-aff-scsp}),
which controls the addition of a discretization of 
the mixed fourth-order derivative $L_{xxyy}$.
By combining (\ref{def-A-aff-gauss-scsp})--(\ref{def-D-aff-gauss-scsp}) with
the positivity condition $A, B, C, D \geq 0$, it follows
that a necessary and sufficient condition for obtaining a
discretization with non-negative filter coefficients is that the free
parameter $C_{xxyy}$ must satisfy
\begin{equation}
  \label{eq-pos-req-aff-scsp-2D}
  |C_{xy}| \leq C_{xxyy} \leq \min(C_{xx}, C_{yy}).
\end{equation}
The feasibility condition
$|C_{xy}| \leq \min(C_{xx}, C_{yy})$ arising from
this positivity constraint is always more restrictive
than the condition $|C_{xy}| \leq \sqrt{C_{xx} \, C_{yy}}$
for the infinitesimal generator in (\ref{eq-diff-eq-cont-aff-scsp}) to be positive semi-definite.
This implies that highly eccentric affine Gaussian kernels 
cannot be represented by non-negative discrete scale-space kernels 
on a square discrete grid,
unless the orientation of the filter is approximately aligned 
to the coordinate directions.

To analyse for which positive definite covariance matrices the positivity
condition may be violated, let us reparameterize the covariance matrix
in terms of its eigenvalues $\lambda_{1,2} > 0$ as well as the orientation
$\alpha$ of the eigenvectors
\begin{align}
  \begin{split}
    \label{eq-def-Cxx-aff-lambda}
    C_{xx} = \lambda_1 \cos^2 \alpha + \lambda_2 \sin^2 \alpha,
  \end{split}\\
  \begin{split}
    \label{eq-def-Cxy-aff-lambda}
    C_{xy} = (\lambda_1 - \lambda_2) \cos \alpha \, \sin \alpha,
  \end{split}\\
  \begin{split}
    \label{eq-def-Cyy-aff-lambda}
    C_{yy} = \lambda_1 \sin^2 \alpha + \lambda_2 \cos^2 \alpha.
  \end{split}
\end{align}
Then, the positivity requirement $|C_{xy}| \leq \min(C_{xx}, C_{yy})$ assumes the form
\begin{equation}
  |\lambda_1 - \lambda_2| \, |\cos \alpha \, \sin \alpha|
  \leq \min(\lambda_1 \cos^2 \alpha + \lambda_2 \sin^2 \alpha,
            \lambda_1 \sin^2 \alpha + \lambda_2 \cos^2 \alpha),
\end{equation}
which can be rewritten as
\begin{equation}
  |\lambda_1 - \lambda_2| \, |\sin 2 \alpha|
  \leq \min(\lambda_1 + \lambda_2 + (\lambda_1 - \lambda_2) \cos 2 \alpha,
            \lambda_1 + \lambda_2 - (\lambda_1 - \lambda_2) \cos 2 \alpha)
\end{equation}
or equivalently
\begin{equation}
  |\lambda_1 - \lambda_2| \, 
  \left( |\cos 2 \alpha| + |\sin 2 \alpha| \right)
  \leq \lambda_1 + \lambda_2.
\end{equation}
The ``worst case orientation'' is given by $\alpha = \tfrac{\pi}{8}$,
and violations of the positivity requirement start to occur when
$\lambda_1$ and $\lambda_2$ obey the relation
\begin{equation}
  \sqrt{2} \, |\lambda_1 - \lambda_2| =  \lambda_1 + \lambda_2,
\end{equation}
corresponding to a ratio of the eigenvalues equal to
\begin{equation}
  \label{eq-bound-ecc-pos-constr-disc-aff-scsp}
  \epsilon_{pos-bound} = \frac{\lambda_{max}}{\lambda_{min}} = 3 + 2\sqrt{2} \approx 5.8.
\end{equation}
As long as the eccentricity of the affine Gaussian kernels is lower
than this value, we can, however, represent the affine Gaussian
scale-space kernels by a non-negative discretization.

This result was stated in Lindeberg \cite[Section~3.3]{Lin97-ICSSTCV}
with the derivation left out because of space constraints.
A similar bound on the positivity requirement has been derived by
Weickert \cite{Wei98-book} in the context of non-negative discretizations
of non-linear diffusion equations.
In relation to Weickert's analysis, the derivation here is shorter.

If we consider the task of representing the distribution of affine
Gaussian kernels as parameterized by spatial covariance matrices on
the hemisphere as illustrated in
Figure~\ref{fig-distr-aff-rec-fields}, then the bound
(\ref{eq-bound-ecc-pos-constr-disc-aff-scsp}) on the eccentricity
$\epsilon$ as obtained from a positivity constraint corresponds to an
angle on the hemisphere relative to the north pole of
\begin{equation}
   \theta = \arccos \frac{1}{\sqrt{\epsilon_{pos-bound}}} = 65.5~\mbox{degrees}.
\end{equation}

\subsection{Behaviour for low frequencies}
\label{sec-determ-Cxxyy-low-freq}

By transforming the Fourier transform (\ref{eq-FT-semi-disc-aff-scsp})
of the discrete affine Gaussian kernel to polar coordinates
$(u, v) = \omega (\cos \beta, \sin \beta)$, and performing
a Taylor expansion for low angular frequencies $\omega$
\begin{align}
  \begin{split}
    \label{eq-aff-scsp-FT-taylor}
    \psi(\omega \cos \beta, \omega \sin \beta) =
      & - \frac{1}{2} 
        \left( 
          C_{xx} \cos^2 \beta 
          + 2 C_{xy} \cos \beta \sin \beta
          + C_{yy} \sin^2 \beta 
        \right) \, \omega^2
  \end{split}\nonumber\\
  \begin{split}
      & + \frac{1}{24} 
        \left( 
           C_{xx} \cos^4 \beta  
           + 4 C_{xy} \cos^3 \beta \sin \beta
           + 6 C_{xxyy} \cos^2 \beta \sin^2 \beta
        \right.
  \end{split}\nonumber\\
  \begin{split}
      & \phantom{\frac{1}{24} ( }
        \left.
           + 4 C_{xy} \cos \beta \sin^3 \beta
           + C_{yy} \sin^4 \beta  
         \right) 
         \, \omega^4 
         + {\cal O}(\omega^6),
  \end{split}
\end{align}
we can with
  $C_{xxyy} = (C_{xx} + C_{yy})/6 + \rho/3$
rewrite this expression as
\begin{align}
  \begin{split}
    \label{eq-aff-scsp-FT-taylor-2}
    \psi(\omega \cos \beta, \omega \sin \beta) =
      & - \frac{1}{2} 
        \left( 
          C_{xx} \cos^2 \beta 
          + 2 C_{xy} \cos \beta \sin \beta
          + C_{yy} \sin^2 \beta 
        \right) \, \omega^2
  \end{split}\nonumber\\
  \begin{split}
      & + \frac{1}{24} 
        \left( 
          C_{xx} \cos^2 \beta 
          + 2 C_{xy} \cos \beta \sin \beta
          + C_{yy} \sin^2 \beta 
        \right) \, \omega^4
  \end{split}\nonumber\\
  \begin{split}
      & + \frac{1}{12} 
        \left(
           C_{xy} \cos \beta \sin \beta
           + \rho \cos^2 \beta \sin^2 \beta
         \right) 
         \, \omega^4 
         + {\cal O}(\omega^6).
  \end{split}
\end{align}
In the specific case when $C_{xy} = 0$, we can note that the choice $\rho = 0$
implies that
\begin{equation}
  \label{eq-choice-Cxxyy-when-Cxy0}
  C_{xxyy} = \frac{C_{xx} + C_{yy}}{6},
\end{equation}
and we obtain a Fourier transform in which the fourth-order terms have
a similar angular dependency, here an elliptic shape, as the 
second-order terms.
Indeed, this elliptic shape also coincides with the shape of the
corresponding Fourier transform of the Gaussian kernel
on a continuous spatial domain.

In the specific case when $C_{xx} = C_{yy} = 1$ and $C_{xy} = 0$, 
the choice of $C_{xxyy} = (C_{xx} + C_{yy})/3$ does in turn correspond
to the following discretization of the isotropic diffusion equation
(Lindeberg \cite{Lin90-PAMI,Lin93-Dis})
\begin{equation}
  \label{eq-disc-isotrop-diff-eq-gen-gamma}
  \partial_s L  
  = \frac{1}{2} 
    \left(
      (1 - \gamma) \, \nabla_5^2 L + \gamma \, \nabla_{\times^2}^2 L
    \right)
\end{equation}
with $\nabla_5^2$ and $\nabla_{\times}^2$ denoting the
numerical five-point and the cross-operators, respectively, defined by 
(Dahlquist {\em et al.\/} \cite{DBA74})
\begin{align}
  \begin{split}
      (\nabla_5^2 f)_{0,0} 
     & 
     = f_{-1, 0} + f_{+1, 0} + f_{0, -1} + f_{0, +1} - 4f_{0, 0}
    \end{split}\\
    \begin{split}
      (\nabla_{\times}^2 f)_{0, 0} 
      & 
      = \frac{1}{2} \, 
          (f_{-1, -1} + f_{-1, +1} + f_{+1, -1} + f_{+1, +1} - 4f_{0, 0}),
    \end{split}
  \end{align}
for $\gamma = \tfrac{1}{3}$ and corresponding to the following composed computational
molecule 
\begin{equation}
  \frac{2}{3} \, \nabla_5^2 + \frac{1}{3} \, \nabla_{\times}^2
  =
  \frac{1}{6}
    \left( 
      \begin{array}{ccc} 
        {1}  &  {4}   &  {1} \vspace{1mm} \\
        {4}  & -{20}  &  {4}  \vspace{1mm} \\
        {1}  &  {4}   &  {1}   
      \end{array}
    \right),
\end{equation}
which gives the $3 \times 3$ discrete approximation of the Laplacian
operator with highest degree of rotational symmetry (Dahlquist {\em et al.\/} \cite{DBA74}).

\subsection{Choice of the free parameter $C_{xxyy}$}
\label{sec-choice-Cxxyy-semi-disc-aff-scsp}

In view of the above analysis, specifically with the condition
(\ref{eq-pos-req-aff-scsp-2D}) on 
necessary relations between the free parameter $C_{xxyy}$ in relation
to the parameters $C_{xx}$, $C_{xy}$ and $C_{yy}$ of the covariance
matrix, we can choose
\begin{equation}
  \label{eq-choice-Cxxyy-minimal-pos-discr}
  C_{xxyy} = |C_{xy}|.
\end{equation}
This is the minimal choice and leads
to the discrete kernel with lowest fourth-order moments.
(Note that for continuous Gaussian kernels, the fourth-order moments
as well as all other moments of order higher than two should be zero.)

The bottom row in figure~\ref{fig-aff-elong-filters-dir-ders-fft}
shows examples of discrete affine kernels generated in this way from
the explicit expression of the Fourier transform
(\ref{eq-FT-semi-disc-aff-scsp}) and with $C_{xxyy} = |C_{xy}|$ 
according to (\ref{eq-choice-Cxxyy-minimal-pos-discr}).


\subsection{Discrete approximation of scale-space derivatives}

Given the above discrete affine Gaussian kernels, we can in turn
compute discrete approximations of directional derivatives according
to 
\begin{align}
  \begin{split}
    \label{eq-dir-der-phi}
    \delta_{\varphi} & = \cos \varphi \, \delta_x + \sin \varphi \, \delta_y,
  \end{split}\\
  \begin{split}
\label{eq-dir-der-phiorth}
    \delta_{\orth\varphi} & = -\sin \varphi \, \delta_x + \cos \varphi \, \delta_y,
  \end{split}\\
  \begin{split}
    \label{eq-dir-der-phiphi}
    \delta_{\varphi\varphi} 
    & = \cos^2 \varphi \, \delta_{xx} + 2 \cos \varphi \, \sin \varphi \, \delta_{xy} + \sin^2 \varphi \, \delta_{yy},
  \end{split}\\
  \begin{split}
    \label{eq-dir-der-phiorthphi}
    \delta_{\varphi\orth\varphi} 
    & = - \cos \varphi \, \sin \varphi \, \delta_{xx} - (\cos^2 \varphi - \sin^2 \varphi) \, \delta_{xy} + \cos \varphi \, \sin \varphi \, \delta_{yy},
  \end{split}\\
  \begin{split}
    \label{eq-dir-der-phiorthphiorth}
    \delta_{\orth\varphi\orth\varphi} 
    & = \sin^2 \varphi \, \delta_{xx} - 2 \cos \varphi \, \sin \varphi \, \delta_{xy} + \cos^2 \varphi \, \delta_{yy},
  \end{split}
\end{align}
where $\delta_{x}$, $\delta_{y}$, $\delta_{xx}$, $\delta_{xy}$ and $\delta_{yy}$ denote discrete
approximations to the partial derivative operators $\partial_{x}$, $\partial_{y}$, $\partial_{xx}$, 
$\partial_{xy}$ and $\partial_{yy}$, respectively, and we can
preferably choose these of the same types as the discrete approximation operators 
(\ref{eq-def-1st-central-diff}) and (\ref{eq-def-2nd-central-diff})
used for discretizing the continuous affine diffusion equation
(\ref{eq-aff-scsp-diff-eq}) into the semi-discrete affine diffusion 
equation (\ref{eq-diff-eq-disc-aff-scsp}) as well as in the
generator of its solution (\ref{eq-gen-fcn-aff-scsp}) and (\ref{eq-FT-semi-disc-aff-scsp}).

The middle and the top rows in Figure~\ref{fig-aff-elong-filters-dir-ders-fft}
show examples of discrete affine derivative approximation kernels generated in this way from
discrete affine Gaussian kernels generated from the explicit expression of the Fourier transform
(\ref{eq-FT-semi-disc-aff-scsp}) and with $C_{xxyy} = |C_{xy}|$ 
according to (\ref{eq-choice-Cxxyy-minimal-pos-discr}).

By applying corresponding discrete derivative approximation kernels to
the colour-opponent channels $u$ and $v$ according to a
colour-opponent representation (\ref{eq-col-opp-space-uv-from-rgb}),
we obtain discrete affine Gaussian colour-opponent directional
derivative approximations analogous to the colour-opponent receptive fields
shown in figure~\ref{fig-aff-elong-filters-dir-ders-col-opp}.

Note that when we apply these affine Gaussian derivative approximation
receptive fields in practice, we do never compute these kernels
explicitly. Instead, we apply the compact support $3 \times 3$ discrete directional
derivative approximation kernels
(\ref{eq-dir-der-phi})--(\ref{eq-dir-der-phiorthphiorth}) directly
to the output of the discrete affine Gaussian smoothing as
obtained after the FFT implementation of discrete affine Gaussian
kernels.
In this way, the spatial smoothing operation, which constitutes the
computationally most expensive part of a discrete derivative
approximation computation, can be shared between directional
derivative operations of different spatial orders, thus improving the
computational efficiency.

\section{Discretization over scale $s$ into $3 \times 3$ iteration kernels}
\label{sec-theory-disc-aff-rec-fields-3x3}

In this section, we will show how the theory presented in previous
section leads to an implementation scheme based on compact 
$3 \times 3$-kernels, when complemented by a discretization in the
scale direction.

\subsection{Compact $3 \times 3$-kernel for discrete affine Gaussian
  smoothing}

By further discretizing the semi-discrete affine diffusion equation
(\ref{eq-diff-eq-disc-aff-scsp}) with respect to the scale parameter
$s$ using Euler's forward method (Dahlquist {\em et al.\/} \cite{DBA74})
\begin{equation}
  \label{eq-disc-Euler-forward}
  \partial_s L = \frac{L^{k+1} - L^k}{\Delta s},
\end{equation}
we obtain the discrete iteration formula
\begin{align}
  \begin{split}
    L^{k+1} 
    & = L^{k} + \Delta s \, \partial_s L^{k}  
  \end{split}\nonumber\\
  \begin{split}
     \label{eq-disc-Euler-forward-aff-diff-eq}
     & = 1 + \Delta s \, 
            \left( 
               \frac{1}{2} \, (C_{xx} \, \delta_{xx} L^{k}  
	 	                    + 2 C_{xy} \, \delta_{xy} L^{k}  
		                    + C_{yy} \, \delta_{yy} L^{k} )
               + \frac{C_{xxyy}}{4} \, \delta_{xxyy} L^{k} 
           \right)
 \end{split}
\end{align}
with the following computational molecule for forward iteration by a
$3 \times 3$ filter kernel:
\begin{multline}
  \label{eq-disc-3x3-kernel}
  k(C_{xx}, C_{xy}, C_{yy}, C_{xxyy}, \Delta_s) = \\
  \left(
     \begin{array}{ccc}
        \frac{1}{4} (-C_{xy} + C_{xxyy}) \Delta s
        & \frac{1}{2} (C_{yy} - C_{xxyy}) \Delta s
        & \frac{1}{4} (+C_{xy} + C_{xxyy}) \Delta s \\ $\,$ \\
        \frac{1}{2} (C_{xx} - C_{xxyy}) \Delta s
        & 1 - (C_{xx} + C_{yy} - C_{xxyy}) \Delta s
        & \frac{1}{2} (C_{xx} - C_{xxyy}) \Delta s \\ $\,$ \\
        \frac{1}{4} (+C_{xy} + C_{xxyy}) \Delta s
        & \frac{1}{2} (C_{yy} - C_{xxyy}) \Delta s
        & \frac{1}{4} (-C_{xy} + C_{xxyy}) \Delta s
    \end{array}
  \right).
\end{multline}
This kernel has spatial mean vector
\begin{equation}
   \label{eq-mean-semi-disc-3x3-kernels}
     m 
     = \left( \begin{array}{c} m_x \\ m_y \end{array} \right)
     = \left( \begin{array}{c} 0 \\ 0 \end{array} \right)
\end{equation}
and spatial covariance matrix 
\begin{equation}
   \label{eq-cov-semi-disc-3x3-kernels}
     \Sigma
     = \left(
           \begin{array}{cc} 
               \Sigma_{xx} & \Sigma_{xy} \\
               \Sigma_{xy} & \Sigma_{yy} 
           \end{array} 
         \right)
     = \left(
           \begin{array}{cc} 
               C_{xx} \Delta s & C_{xy} \Delta s \\
               C_{xy} \Delta s & C_{yy} \Delta s 
           \end{array} 
         \right).
\end{equation}
Thereby, in analogy with the semi-discrete affine Gaussian scale-space
concept over a continuous scale parameter $s$, also the covariance
matrices obtained from repeated forward iteration of the $3 \times 3$ kernel
by $K$ steps will be exactly equal to the covariance matrices of the
corresponding continuous affine Gaussian kernels
\begin{equation}
     \Sigma_{composed}
     = \left(
           \begin{array}{cc} 
               \Sigma_{xx} & \Sigma_{xy} \\
               \Sigma_{xy} & \Sigma_{yy} 
           \end{array} 
         \right)
     = \left(
           \begin{array}{cc} 
               C_{xx} \, K \, \Delta s & C_{xy} \, K \, \Delta s \\
               C_{xy} \, K \, \Delta s & C_{yy} \, K \, \Delta s 
           \end{array} 
         \right).
\end{equation}
In Equation~(\ref{eq-disc-Euler-forward-aff-diff-eq}), the fully discrete
affine scale-space representations $L^k(x)$ will for $s = k \, \Delta s$ 
be numerical approximations 
\begin{equation}
  L^k(x) \approx L(x;\; k \, \Delta s, \Sigma)
\end{equation}
of the exact semi-discrete affine
scale-space representation $L(x;\; s, \Sigma)$ according to (\ref{eq-diff-eq-disc-aff-scsp}).

\subsection{Choice of scale step $\Delta s$}
\label{sec-choice-scale-step}

In this section, we will analyse how large scale steps $\Delta s$ can
be taken with the discrete iteration kernel (\ref{eq-disc-3x3-kernel})
while preserving scale-space properties or transfers thereof.

\subsubsection{Normalization of the parameters of the covariance matrix}

Since the scale step $\Delta s$ interacts with the parameters 
$C_{xx}$, $C_{xy}$, $C_{yy}$, $C_{xxyy}$ along a cone, we shall
specifically choose to normalize the parameters 
$C_{xx}$, $C_{xy}$, $C_{yy}$ and $C_{xxyy}$ such that the maximum
eigenvalue of the covariance matrix according to the parameterization
(\ref{eq-def-Cxx-aff-lambda}), (\ref{eq-def-Cxy-aff-lambda}) and
(\ref{eq-def-Cyy-aff-lambda}) is given by 
\begin{equation}
  \label{eq-par-Sigma-max-eigenval-1}
  \lambda_{max} = \max(\lambda_1, \lambda_2) = 1
\end{equation}
and thus the
second eigenvalue is 
\begin{equation} 
  \label{eq-par-Sigma-max-eigenval-1-second-eigenval}
  \lambda_{min} = \min(\lambda_1, \lambda_2) \in [0, 1].
\end{equation}
With regard to our goal of discretizing the manifold of affine
Gaussian receptive fields over all covariance matrices $\Sigma$ as
illustrated in Figure~\ref{fig-distr-aff-rec-fields}, this
parameterization has a very natural geometric interpretation.
Assume that the central point in a first image $f$ of a scene shows a local
surface pattern viewed from a
point along the direction of the surface normal at some depth $Z$,
and that we compute isotropic Gaussian receptive field responses at
scale $s$ with the parameters of the covariance matrix given by 
$C_{xx} = C_{yy} = 1$ and $C_{xy} = 0$.
Next, assume that we view the same local surface pattern from the same
distance $Z$ while from some other oblique direction with slant angle
$\theta$ relative to the surface normal and corresponding to a local foreshortening factor of
$\epsilon = \cos \theta$. 
Then, the second view should in an affine covariant manner be
represented at scale $s$ for a covariance matrix with eigenvalues 
$\lambda_{max} = 1$ and $\lambda_{min} = \epsilon^2 \in [0, 1]$ with the parameter $\alpha$
that determines the eigendirections of the covariance matrix $\Sigma$
directly related to the local tilt direction.

In the following, we will throughout make use of such a
parameterization of the covariance matrix $\Sigma$ for expressing the
magnitude of the scale step $\Delta s$.

\subsubsection{Analysis for the case of the isotropic diffusion equation}

To illustrate how the choice of the complementary filter parameter $C_{xxyy}$
in (\ref{eq-inf-gen-disc-aff-scsp}) interrelates with how large scale steps one may take, let us initially
consider a discretization of the isotropic diffusion equation for 
$C_{xx} = C_{yy} = 1$ and $C_{xy} = 0$ for the specific choice of 
$\gamma = 0$ in (\ref{eq-disc-isotrop-diff-eq-gen-gamma})
and corresponding to $C_{xxyy} = 0$.
This leads to a forward iteration kernel of the form
\begin{equation}
  \label{eq-disc-3x3-kernel-isotropic-Cxxyy0}
  k(1, 0, 1, 0, \Delta s) =
\left(
     \begin{array}{ccc}
        0
        & \frac{1}{2} \Delta s
        & 0 \\ $\,$ \\
        \frac{1}{2} \Delta s
        & 1 - 2 \Delta s
        & \frac{1}{2} \Delta s \\ $\,$ \\
        0
        & \frac{1}{2} \Delta s
        & 0
    \end{array}
  \right).
\end{equation}
In (Lindeberg \cite{Lin90-PAMI,Lin93-Dis}) a complete scale-space
theory is developed for one-dimensional discrete signals and for isotropic
higher-dimensional images over a symmetric spatial domain.
In the one-dimensional version of this theory, spatial discretizations
of the one-dimensional diffusion equation 
$\partial_s L = \tfrac{1}{2} \partial_{xx} L$ as obtained from
Euler's forward method lead to kernels of the form
\begin{equation}
  (\frac{\Delta s}{2}, 1-\Delta s, \frac{\Delta s}{2}).
\end{equation}
Based on the complete classification of discrete scale-space kernels
over a 1-D domain, it is furthermore 
shown that such discretizations are true discrete
scale-space kernels in the one-dimensional sense of guaranteeing
non-creation of new local extrema or new zero-crossings from finer to
coarser temporal scales if and only if 
\begin{equation}
  \Delta s \leq \frac{1}{2}.
\end{equation}
Specifically, the boundary case $\Delta s = 1/2$ corresponds to
forward iteration with the binomial kernel 
\begin{equation}
  \left( \frac{1}{4}, \frac{1}{2}, \frac{1}{4} \right),
\end{equation}
in which the
central kernel value $1/2$ is twice as large as its nearest
neighbours having values equal to $1/4$.
If we apply a corresponding criterion to the two-dimensional forward
iteration kernel (\ref{eq-disc-3x3-kernel-isotropic-Cxxyy0}),
we obtain $1 - 2 \Delta s \geq 2 \Delta s/2$ leading to the
requirement
\begin{equation}
   \Delta s \leq \frac{1}{3}.
\end{equation}
If we next again for the isotropic case of $C_{xx} = C_{yy} = 1$ and
$C_{xy} = 0$, consider a discretization in the scale direction of
(\ref{eq-disc-isotrop-diff-eq-gen-gamma}) for a general value of the
parameter $\gamma = C_{xxyy}$, we obtain a forward iteration kernel of the form
(Lindeberg \cite[Equation~(4.39)]{Lin93-Dis})
\begin{equation}
  \label{eq-disc-3x3-kernel-isotropic-gengamma}
  k(1, 0, 1, \gamma, \Delta s) =
   \left(
     \begin{array}{ccc}
        \frac{1}{4} \gamma \Delta s
        & \frac{1}{2} (1 - \gamma) \Delta s
        & \frac{1}{4} \gamma \Delta s\\ $\,$ \\
        \frac{1}{2} (1 - \gamma) \Delta s
        & 1 - (2 - \gamma) \Delta s
        & \frac{1}{2} (1 - \gamma) \Delta s \\ $\,$ \\
        \frac{1}{4} \gamma \Delta s
        & \frac{1}{2} (1 - \gamma) \Delta s
        & \frac{1}{4} \gamma \Delta s
    \end{array}
  \right).
\end{equation}
This kernel can be shown to be separable only if $\gamma = \Delta s$
(Lindeberg \cite[Proposition~4.15]{Lin93-Dis}). The corresponding
one-dimensional kernel $(a, 1-2a, a)$ for
$a = \frac{1}{2} \sqrt{\gamma \Delta s}$ is then a one-dimensional
discrete scale-space kernel only if
\begin{equation}
  \label{eq-upper-bound-scale-step-isotropic-scsp}
  \Delta s \leq \frac{1}{2}
\end{equation}
where the boundary case $\gamma = \Delta s = 1/2$ corresponds to
forward iteration with the kernel
\begin{equation}
  \label{eq-disc-3x3-kernel-isotropic-ds-0p5}
  k(1, 0, 1, \tfrac{1}{2}, \tfrac{1}{2}) =
  \left(
     \begin{array}{ccc}
        \frac{1}{16} 
        & \frac{1}{8} 
        & \frac{1}{16} \\ $\,$ \\
        \frac{1}{8}
        & \frac{1}{4}
        & \frac{1}{8} \\ $\,$ \\
        \frac{1}{16} 
        & \frac{1}{8} 
        & \frac{1}{16} 
    \end{array}
  \right).
\end{equation}
This kernel is a common iteration kernel for computing pyramid representations 
(Crowley \cite{Cro81}; Crowley and Parker
\cite{Cro84-dolp,Cro84-peaks}; Lindeberg \cite{Lin90-PAMI,Lin93-Dis};
Lindeberg and Bretzner \cite{LinBre03-ScSp}). Specifically, applying
the underlying one-dimensional kernel $(1/4, 1/2, 1/4)$ for separable
convolution twice in the same direction gives a binomial kernel of the
form $(1/16, 4/16, 6/16, 4/16, 1/16)$ within the family of kernels of
length five derived by Burt and Adelson \cite{BA83-COM}
$(\tfrac{1}{4} - \tfrac{a}{2}, \tfrac{1}{4}, a, \tfrac{1}{4}, \tfrac{1}{4} - \tfrac{a}{2})$ 
for $a = 3/8 = 0.375$, which is in very good agreement with the numerical value 
$a \approx 0.4$ that they derived empirically.

Note that if we would apply the same value of $\Delta s = 1/2$ to the
kernel (\ref{eq-disc-3x3-kernel-isotropic-Cxxyy0}) that arises from
a spatial discretization of the isotropic diffusion equation
for $\gamma = 0$ and corresponding to 
$C_{xxyy} = 0$, we would obtain a forward iteration kernel of the form
\begin{equation}
  \label{eq-disc-3x3-kernel-isotropic-Cxxyy0-ds-0p5}
  k(1, 0, 1, 0, \tfrac{1}{2}) =
  \left(
     \begin{array}{ccc}
        0
        & \frac{1}{4} 
        & 0 \\ $\,$ \\
        \frac{1}{4} 
        & 0
        & \frac{1}{4}  \\ $\,$ \\
        0
        & \frac{1}{4}
        & 0
    \end{array}
  \right),
\end{equation}
which would clearly not be a desirable kernel for scale-space filtering.
For example, this kernel is not even unimodal.
Thus, by the use of a non-zero value of $\gamma$ and corresponding to
a non-zero value of $C_{xxyy}$, in this specific case corresponding to
$C_{xxyy} = C_{xx}/2 = C_{yy}/2$, we can increase the value at the central
position of the $3 \times 3$ kernel in relation to its nearest
neighbours along the coordinate axes and are thereby be able to take longer steps
$\Delta s$ in the scale direction.

\subsubsection{Analysis for the non-isotropic diffusion equation}
\label{sec-anal-delta-s-non-aniso-diff-eq}

In this section, we will consider discretizations of the affine Gaussian
diffusion equation for general non-zero values of $C_{xy}$ and
$C_{xxyy}$ and also without assuming $C_{xx} = C_{yy}$.
Specifically, with a normalization of the parameters of the covariance
matrix $\Sigma$ to the maximum eigenvalue equal to one, we will consider
the possibility of taking scale steps up to $\Delta s = 1/2$ so that
the discrete binomial kernel
(\ref{eq-disc-3x3-kernel-isotropic-ds-0p5}) that represents the
maximum possible scale step based on isotropic discrete scale-space
theory is regarded as an upper bound on how much spatial smoothing is
allowed in different spatial directions.

\paragraph{Requirement on the relations between the kernel values
  along the coordinate axes and the central point}
If we inspired by the ratio of two between the kernel value $1/4$ at
the center and the value $1/8$ at the nearest neighbouring points for
the kernel (\ref{eq-disc-3x3-kernel-isotropic-ds-0p5}) that
represents the maximum possible smoothing for the isotropic diffusion
equation, apply a corresponding criterion that the kernel values at the
nearest four-neighbours to the central point having indices 
$(i, j) \in \{ (-1, 0), (1, 0), (0, -1), (0, 1)\}$ should have values that
are not greater than half the value of the central point with index $(0, 0)$, we 
obtain the conditions
\begin{align}
  \begin{split}
     \frac{1}{2} (C_{xx} - C_{xxyy}) \, \Delta s
     & \leq \frac{1}{2} (1 - (C_{xx} C_{yy} - C_{xxyy})) \, \Delta s,
  \end{split}\\
  \begin{split}
     \frac{1}{2} (C_{yy} - C_{xxyy}) \, \Delta s
     & \leq \frac{1}{2} (1 - (C_{xx} C_{yy} - C_{xxyy})) \, \Delta s,
  \end{split}
\end{align}
which can be summarized into
\begin{equation}
   (C_{xx} + C_{yy} + \max(C_{xx}, C_{yy}) - 2 C_{xxyy}) \, \Delta s \leq 1.
\end{equation}
If we inspired by the above theoretical analysis for the isotropic case
would like to be able to take scale steps up to $\Delta s = 1/2$ for
general anisotropic discrete affine Gaussian kernels, 
we obtain the following condition for the values of the nearest
four-neighbours to not exceed half the value at the central point
\begin{equation}
  \label{eq-cond-Cxxyy-from-central-val-gt-2times-nearest-neighbours}
   C_{xxyy} \geq \frac{C_{xx} + C_{yy} + \max(C_{xx}, C_{yy}) - 2}{2},
\end{equation}
where specifically the case $C_{xx} = C_{yy} = 1$ corresponds to
$C_{xxyy} \geq 1/2$. 

\paragraph{Requirement on the relations between the kernel values
  at the corners and the central point}
If we apply a condition that the kernel values at the corners having
indices $(i, j) \in \{ (-1, -1), (-1, 1), (1, -1), (1, 1)\}$ should
have values that are not greater than one fourth of the value at the
central point, we obtain the condition
\begin{equation}
  \frac{1}{4} (|C_{xy}| + C_{xxyy}) \, \Delta s 
  \leq \frac{1}{4} (1 - (C_{xx} + C_{yy} - C_{xxyy}) \, \Delta s)
\end{equation}
which can be simplified to to
\begin{equation}
  (C_{xx} + C_{yy} + |C_{xy}|) \, \Delta s \leq 1.
\end{equation}
If we again inspired by the theoretical analysis for the isotropic
case would like to take scale steps up to $\Delta s = 1/2$ for
general anisotropic discrete affine Gaussian kernels, 
then we obtain the following condition for the values at the corners
should not exceed one quarter of the value at the central point
\begin{equation}
  \label{eq-cond-corner-1over4-central}
  (C_{xx} + C_{yy} + |C_{xy}|)  \leq 2.
\end{equation}
For the specific assumption about the parameters $C_{xx}$, $C_{xy}$
and $C_{yy}$ corresponding to a covariance matrix with maximum
eigenvalue $\lambda_{max} = 1$ with the other eigenvalue 
$\lambda_{min} \in [0, 1]$, this condition can according to
(\ref{eq-def-Cxx-aff-lambda}), (\ref{eq-def-Cxy-aff-lambda}) and
(\ref{eq-def-Cyy-aff-lambda}) be written
\begin{equation}
  1 + \lambda_{min} + (1 - \lambda_{min}) \, | \sin \alpha \, \cos \alpha |  \leq 2.
\end{equation}
The worst case condition arises when $|\sin \alpha| = |\cos \alpha|$,
for which the inequality can be reduced to 
\begin{equation}
  \lambda_{min} \leq 1
\end{equation}
and which is consistent with the assumptions, thus showing that the 
condition (\ref{eq-cond-corner-1over4-central})
is guaranteed to hold for all relevant combinations of $C_{xx}$, $C_{xy}$ and
$C_{yy}$.

\subsection{Choice of the free parameter $C_{xxyy}$}
\label{sec-choice-Cxxyy-semi-disc-aff-scsp-3x3} 
In view of the above analysis, specifically with the conditions
(\ref{eq-pos-req-aff-scsp-2D}) and
(\ref{eq-cond-Cxxyy-from-central-val-gt-2times-nearest-neighbours}) on
necessary relations between the free parameter $C_{xxyy}$ in relation
to the parameters $C_{xx}$, $C_{xy}$ and $C_{yy}$ of the covariance
matrix, we can given the a priori choice of scale step $\Delta s = 1/2$
choose
\begin{equation}
  \label{eq-choice-Cxxyy-discaff-3x3-iter-minimal-factor0p5center}
  C_{xxyy} = \max\left(|C_{xy}|, \frac{C_{xx} + C_{yy} + \max(C_{xx}, C_{yy}) - 2}{2}\right).
\end{equation}
This is the minimal choice and leads
to the discrete kernel with lowest fourth-order moment.
(Note again that for continuous Gaussian kernels, the fourth-order moment
as well as all other moments of order higher than two should be zero.)

\begin{figure}[hbtp]
  \begin{center}
    \begin{tabular}{cccccc}
      \hspace{-4mm}
      \includegraphics[width=0.15\textwidth]{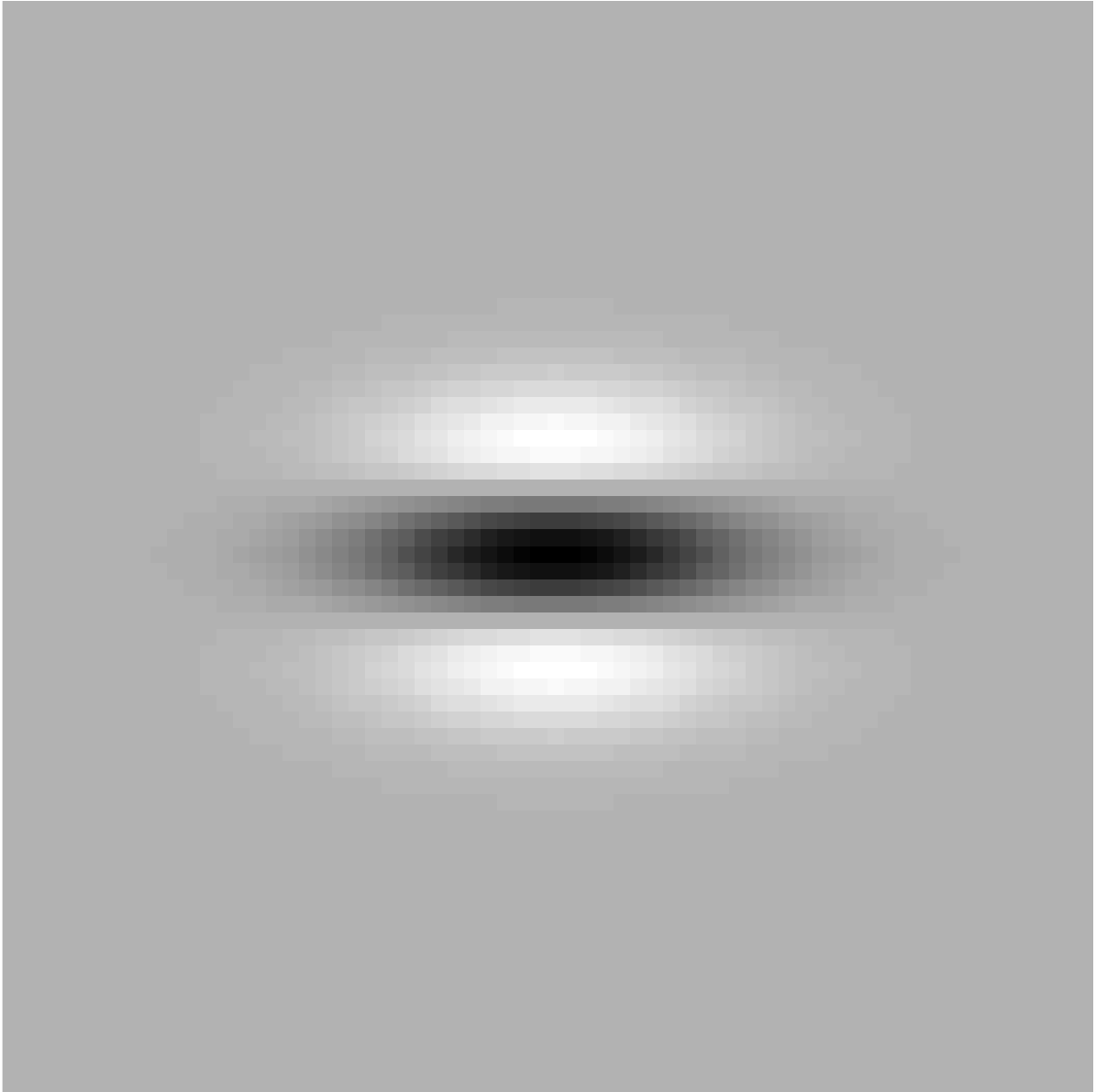} \hspace{-4mm} &
      \includegraphics[width=0.15\textwidth]{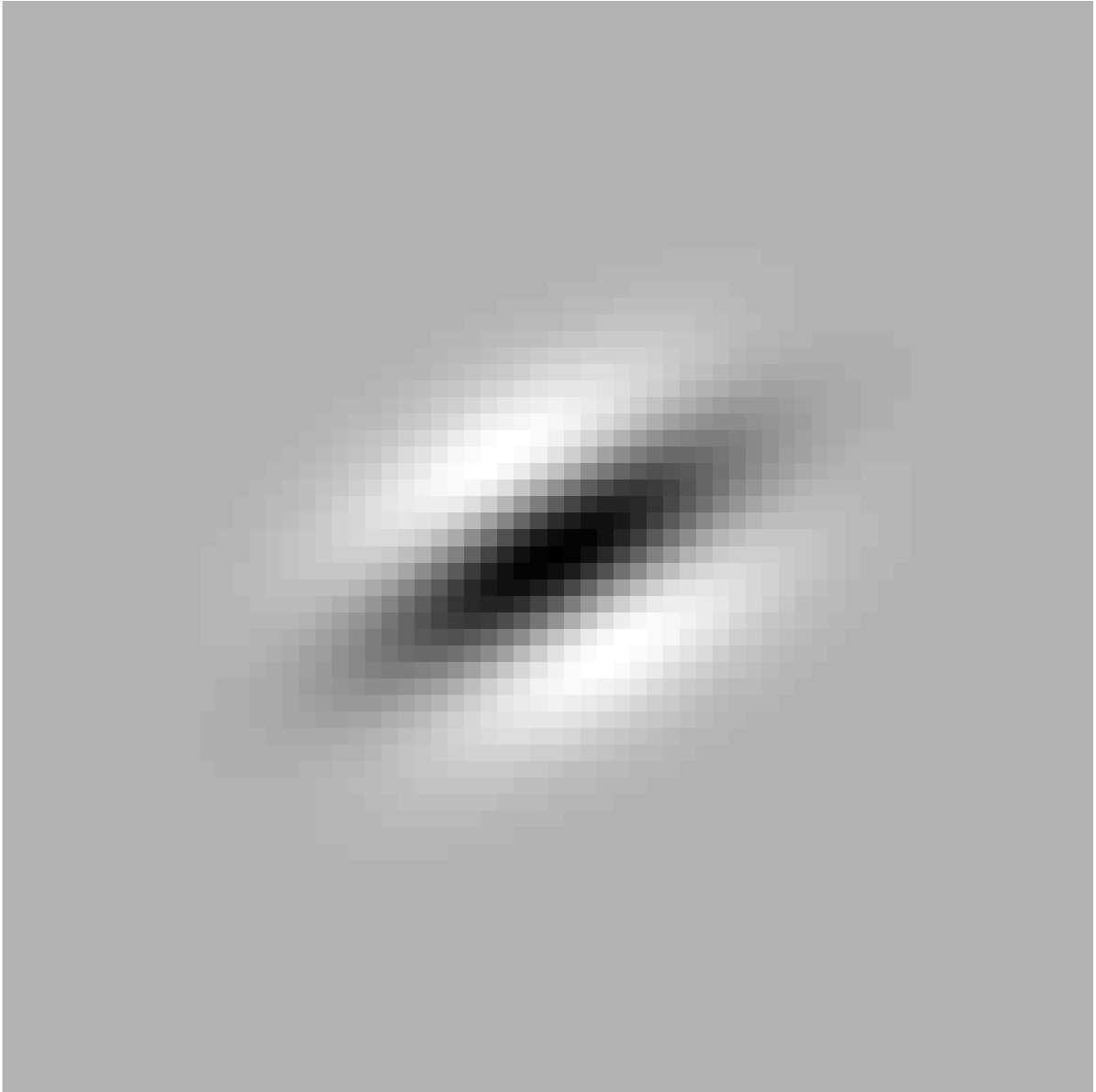} \hspace{-4mm} &
      \includegraphics[width=0.15\textwidth]{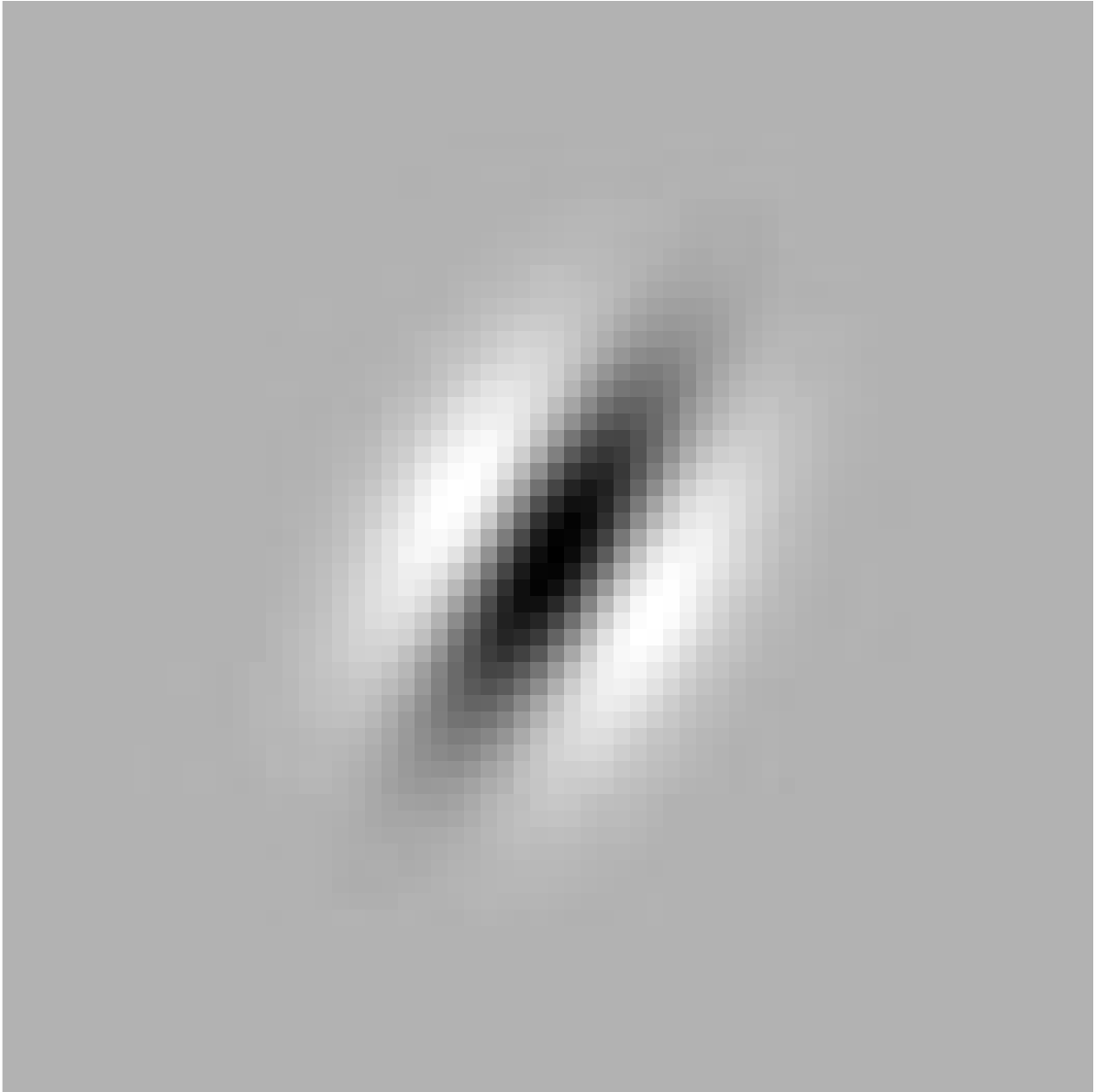} \hspace{-4mm} &
      \includegraphics[width=0.15\textwidth]{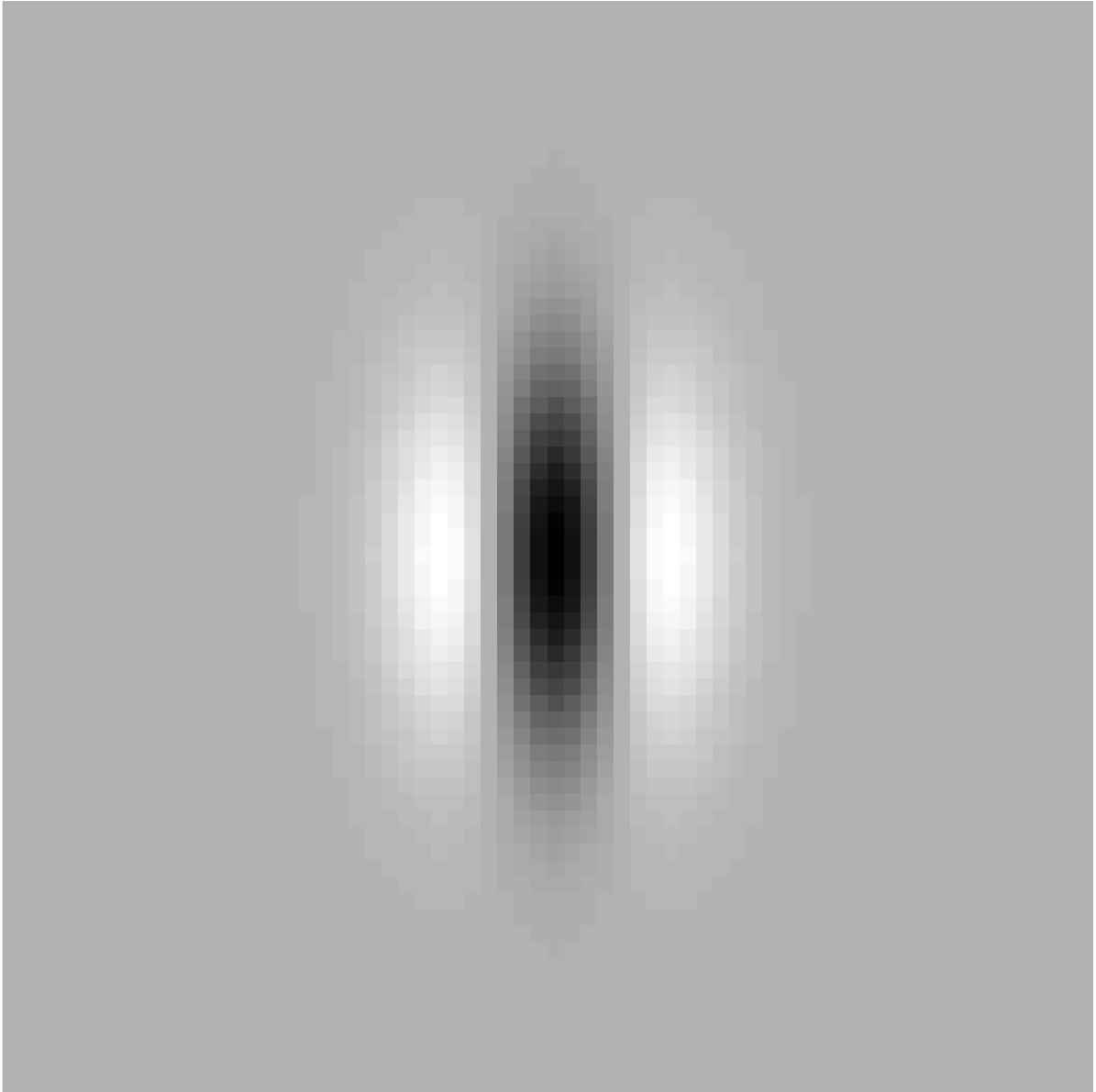} \hspace{-4mm} &
      \includegraphics[width=0.15\textwidth]{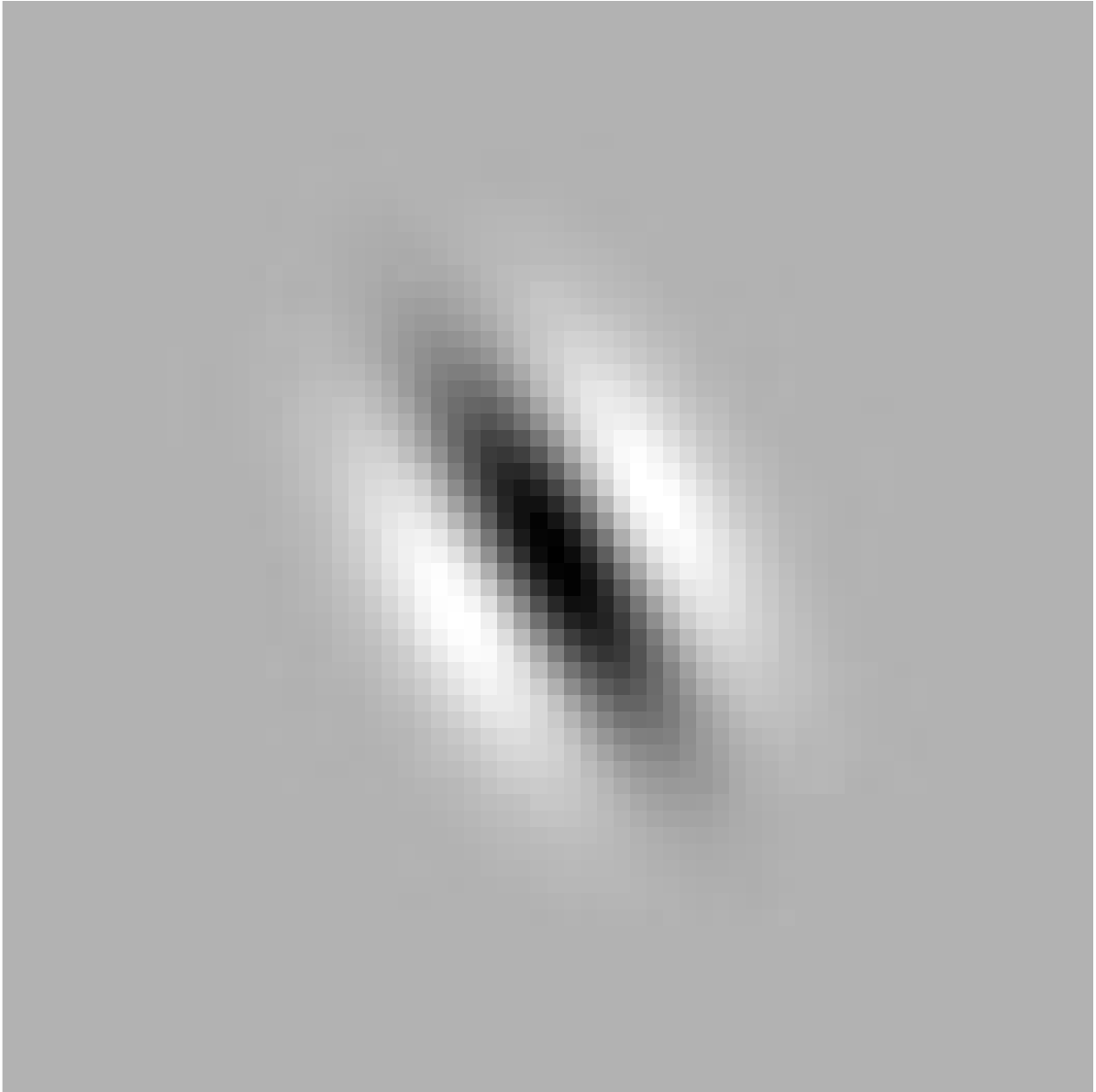} \hspace{-4mm} &
      \includegraphics[width=0.15\textwidth]{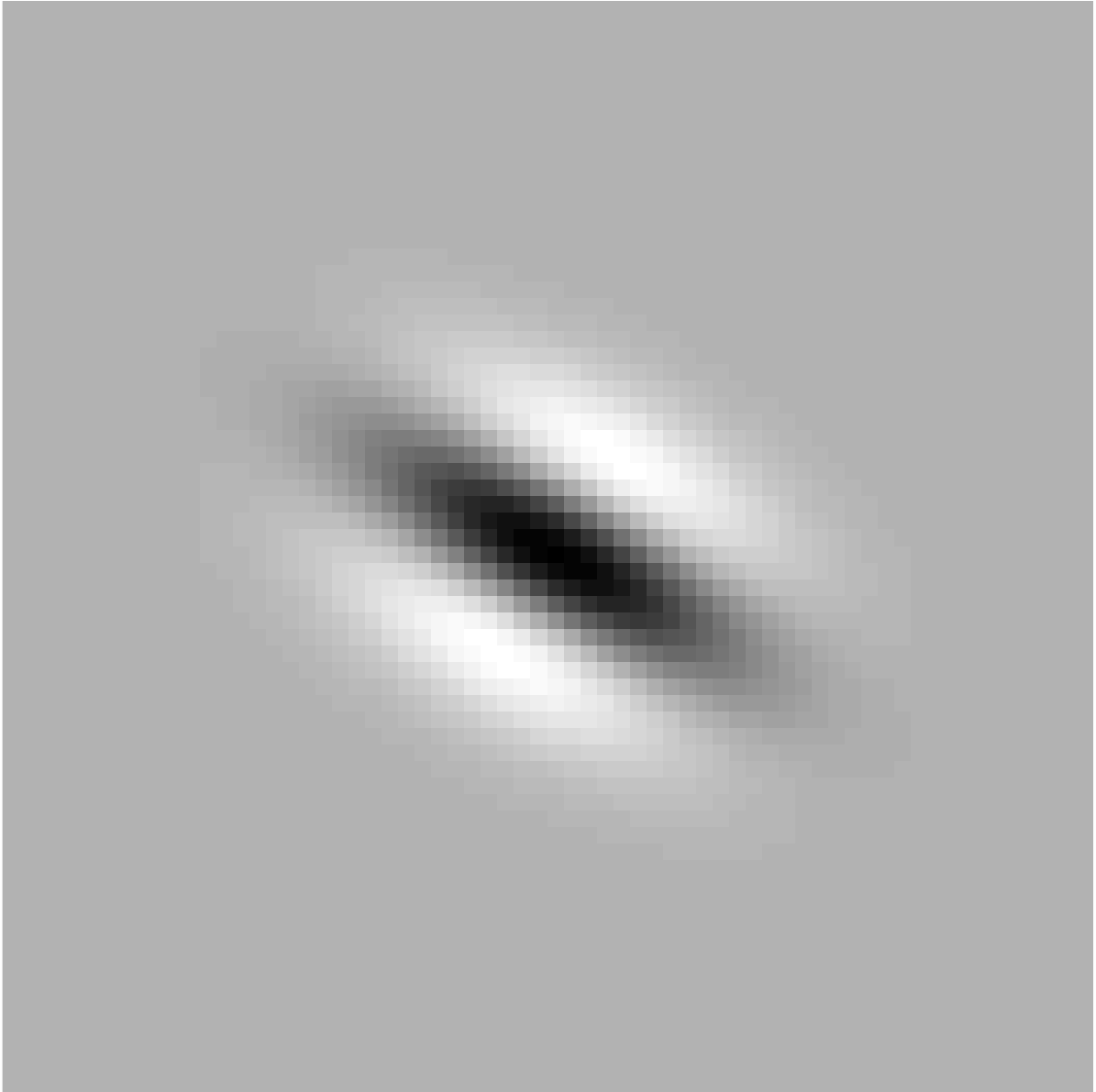}
                                                                                                              \hspace{-4mm} \\
    \hspace{-4mm}
\includegraphics[width=0.15\textwidth]{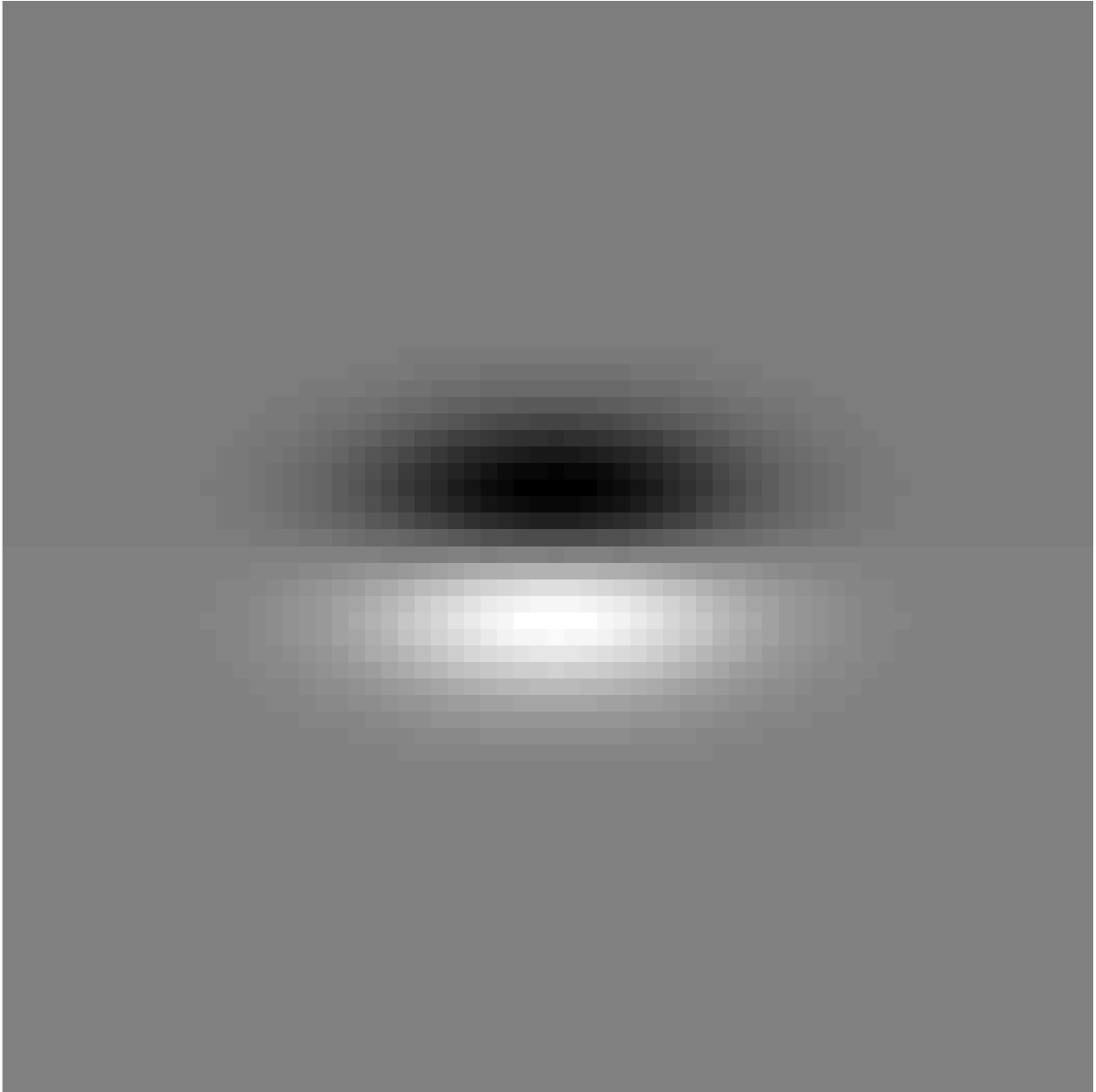} \hspace{-4mm} &
      
\includegraphics[width=0.15\textwidth]{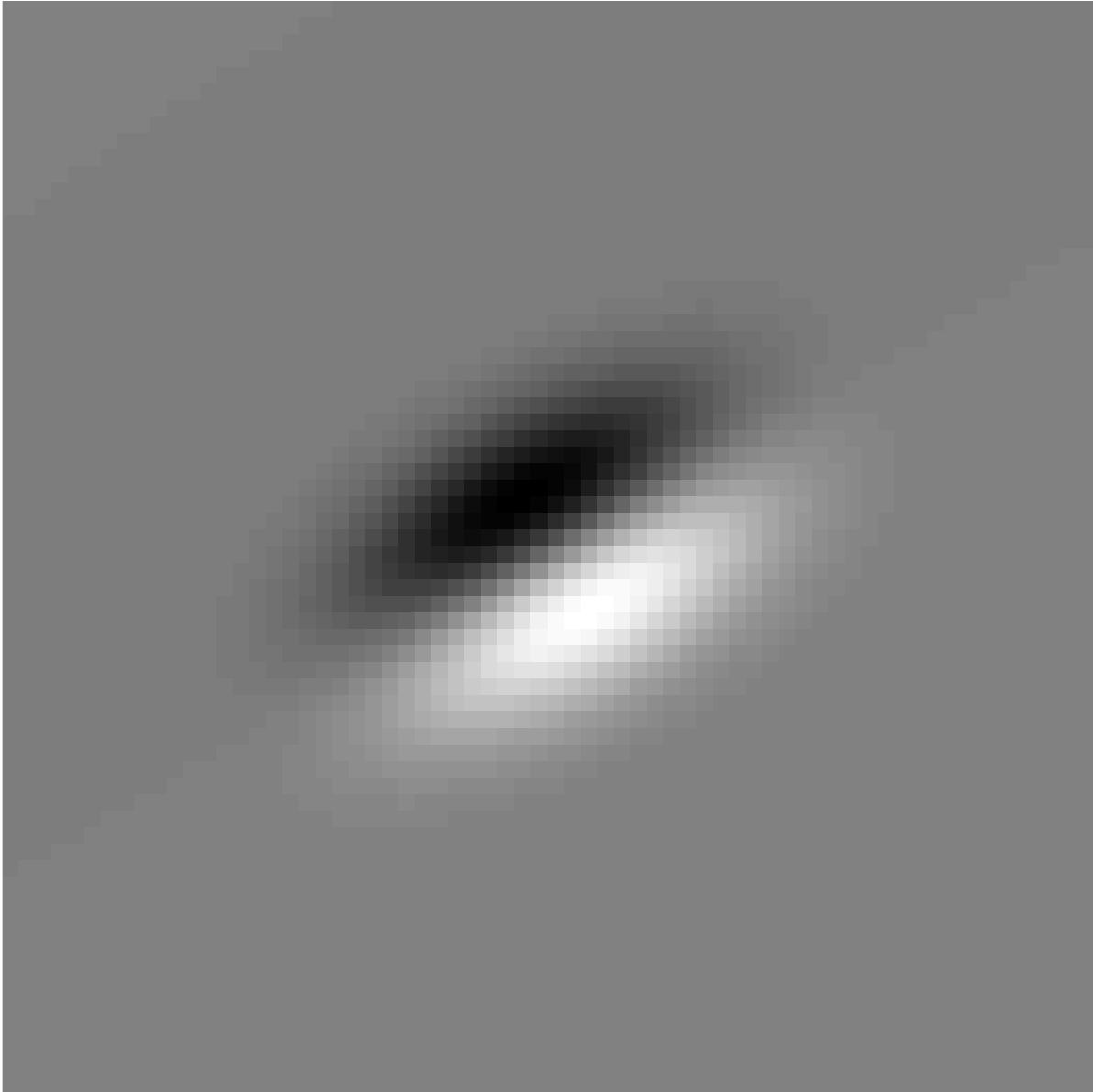} \hspace{-4mm} &
      \includegraphics[width=0.15\textwidth]{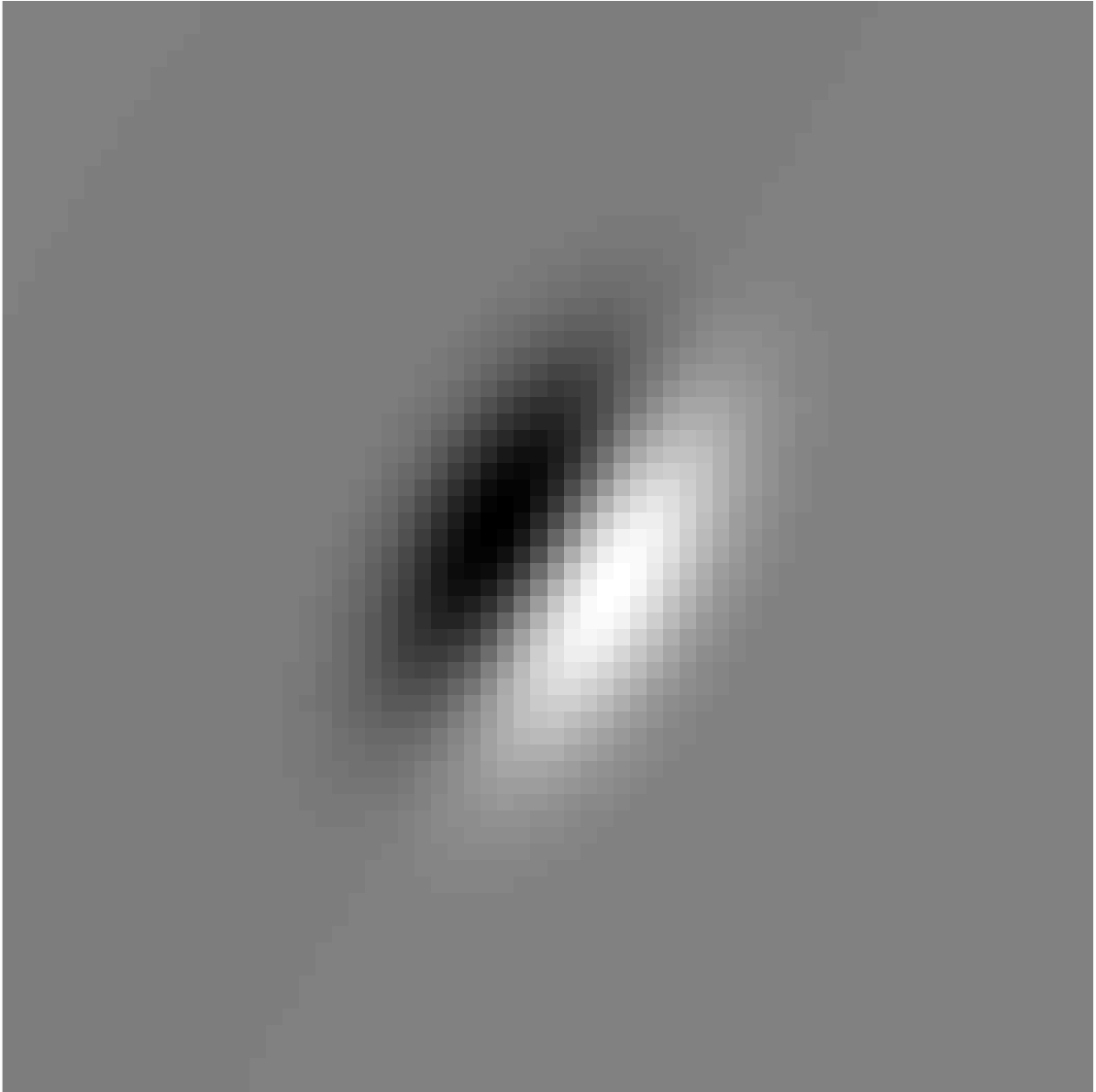} \hspace{-4mm} &
      \includegraphics[width=0.15\textwidth]{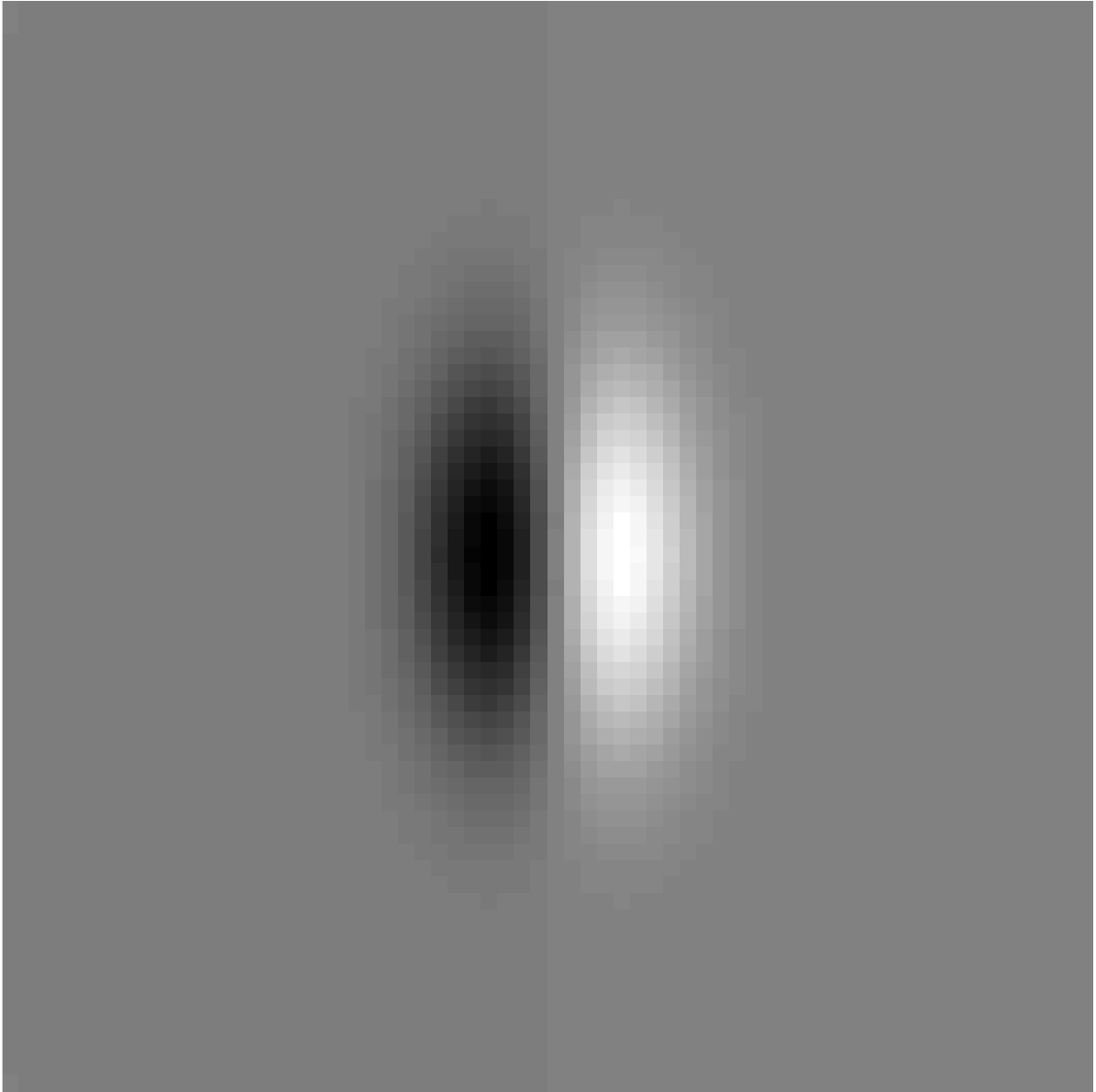} \hspace{-4mm} &
      \includegraphics[width=0.15\textwidth]{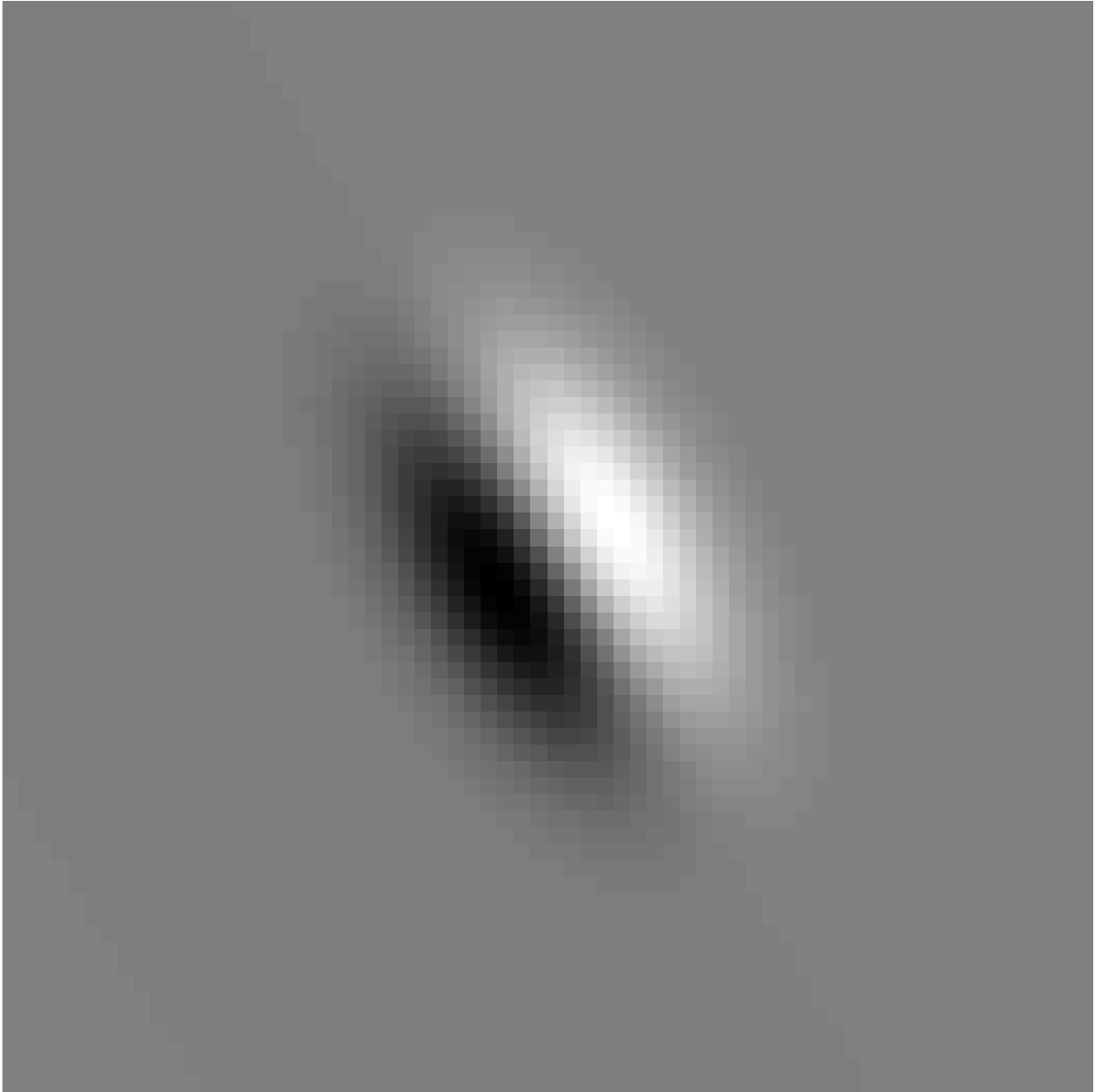} \hspace{-4mm} &
      \includegraphics[width=0.15\textwidth]{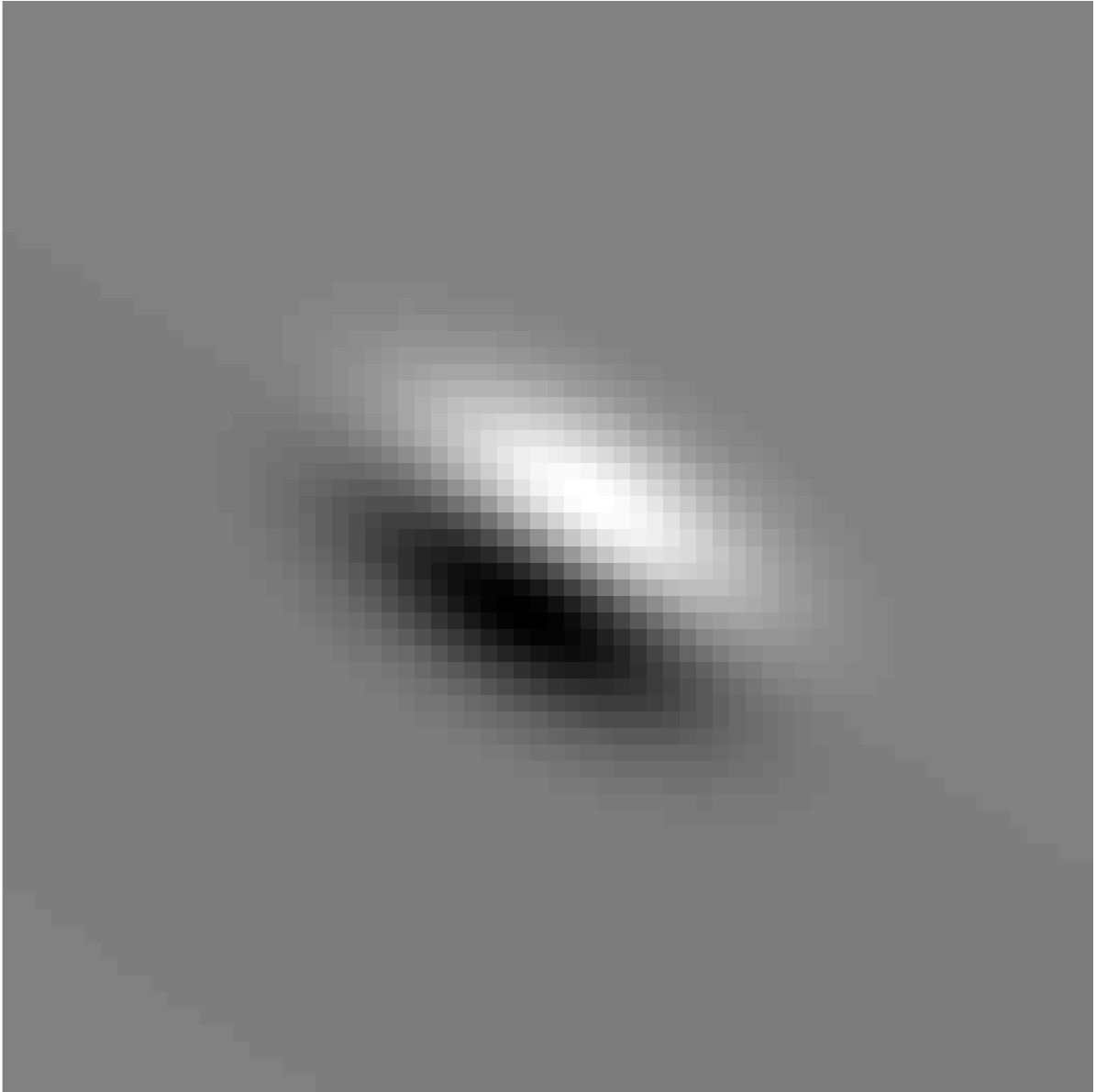} \hspace{-4mm} \\
      \hspace{-4mm}
\includegraphics[width=0.15\textwidth]{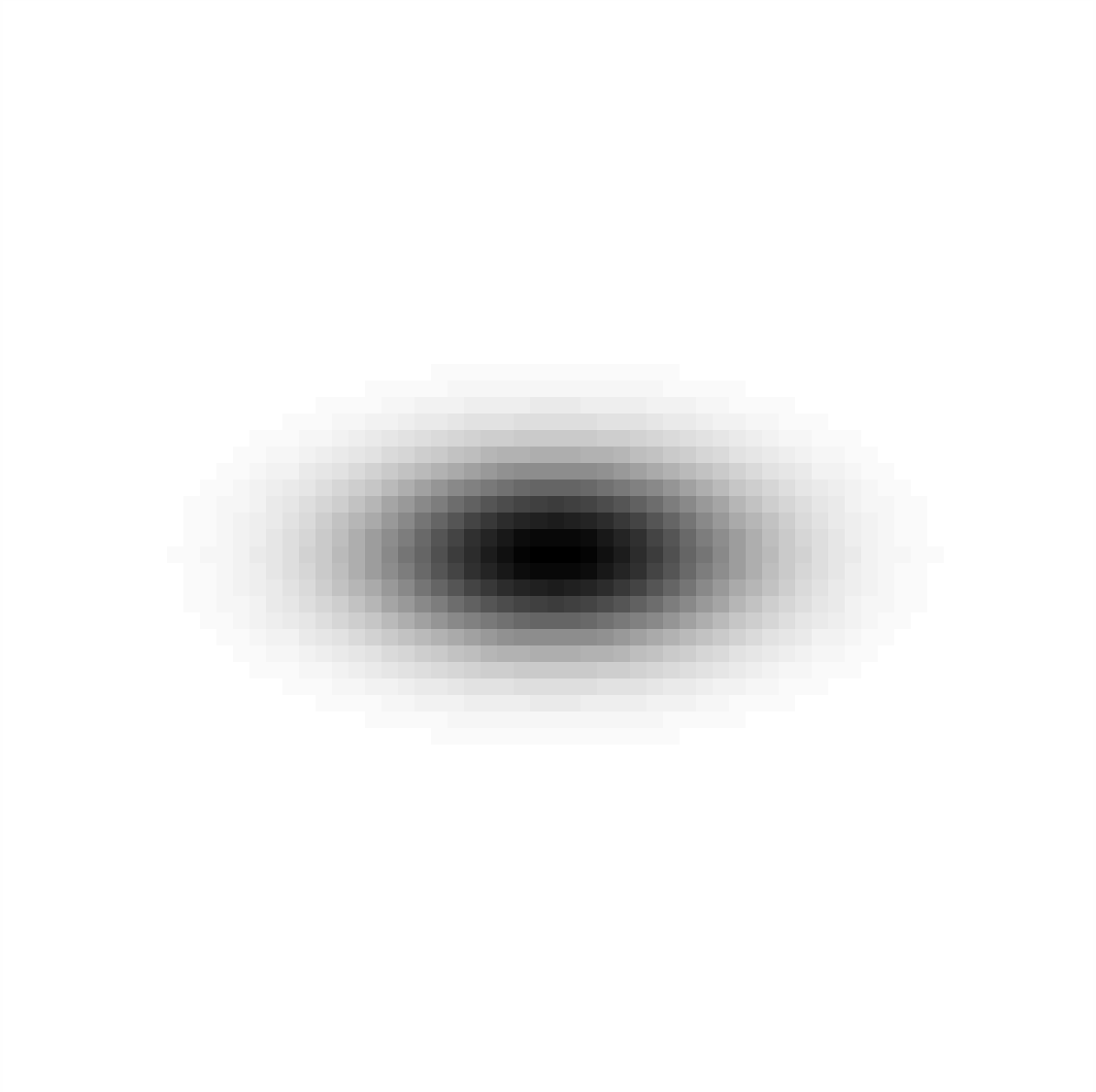} \hspace{-4mm} &
      \includegraphics[width=0.15\textwidth]{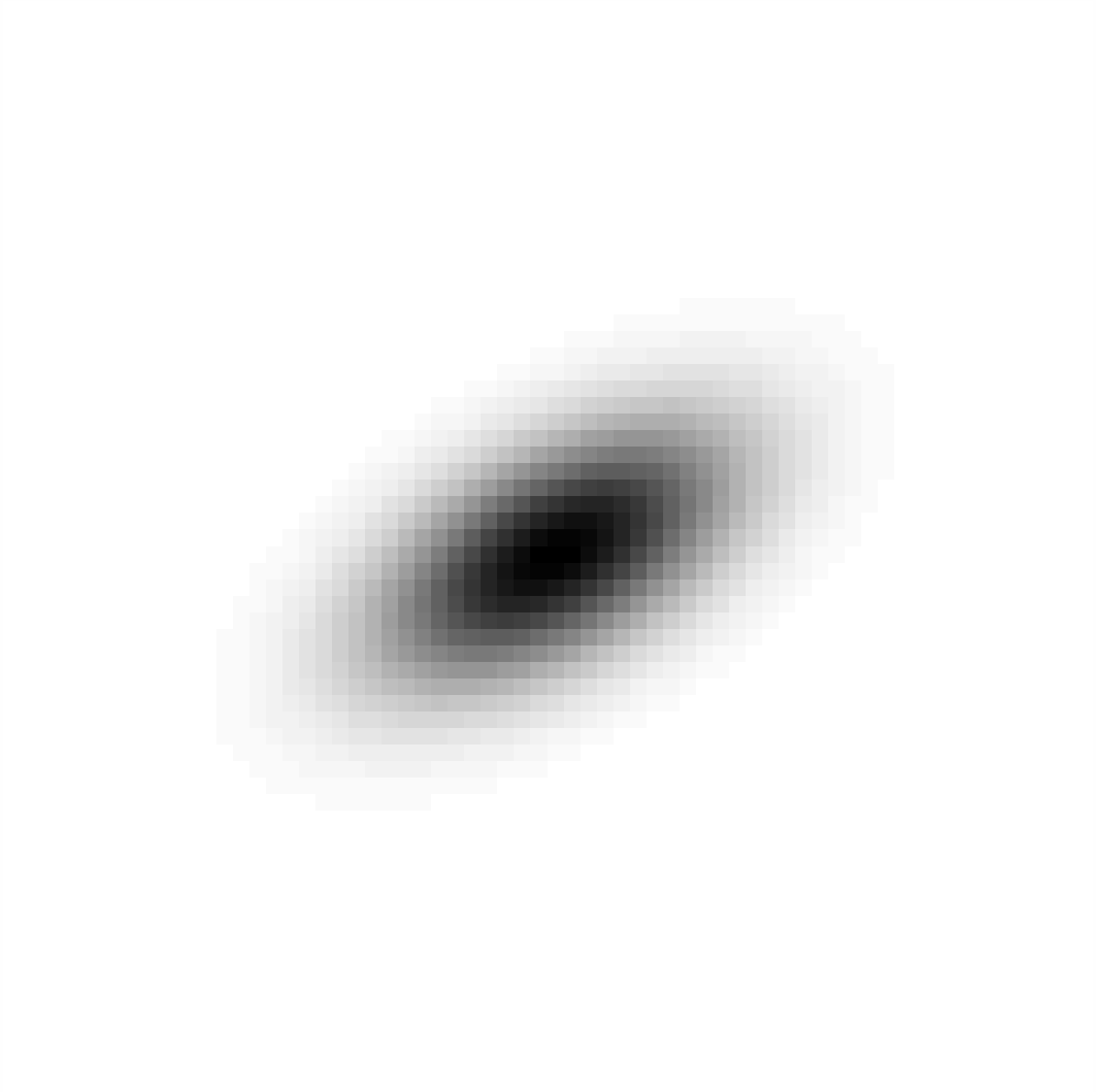} \hspace{-4mm} &
      \includegraphics[width=0.15\textwidth]{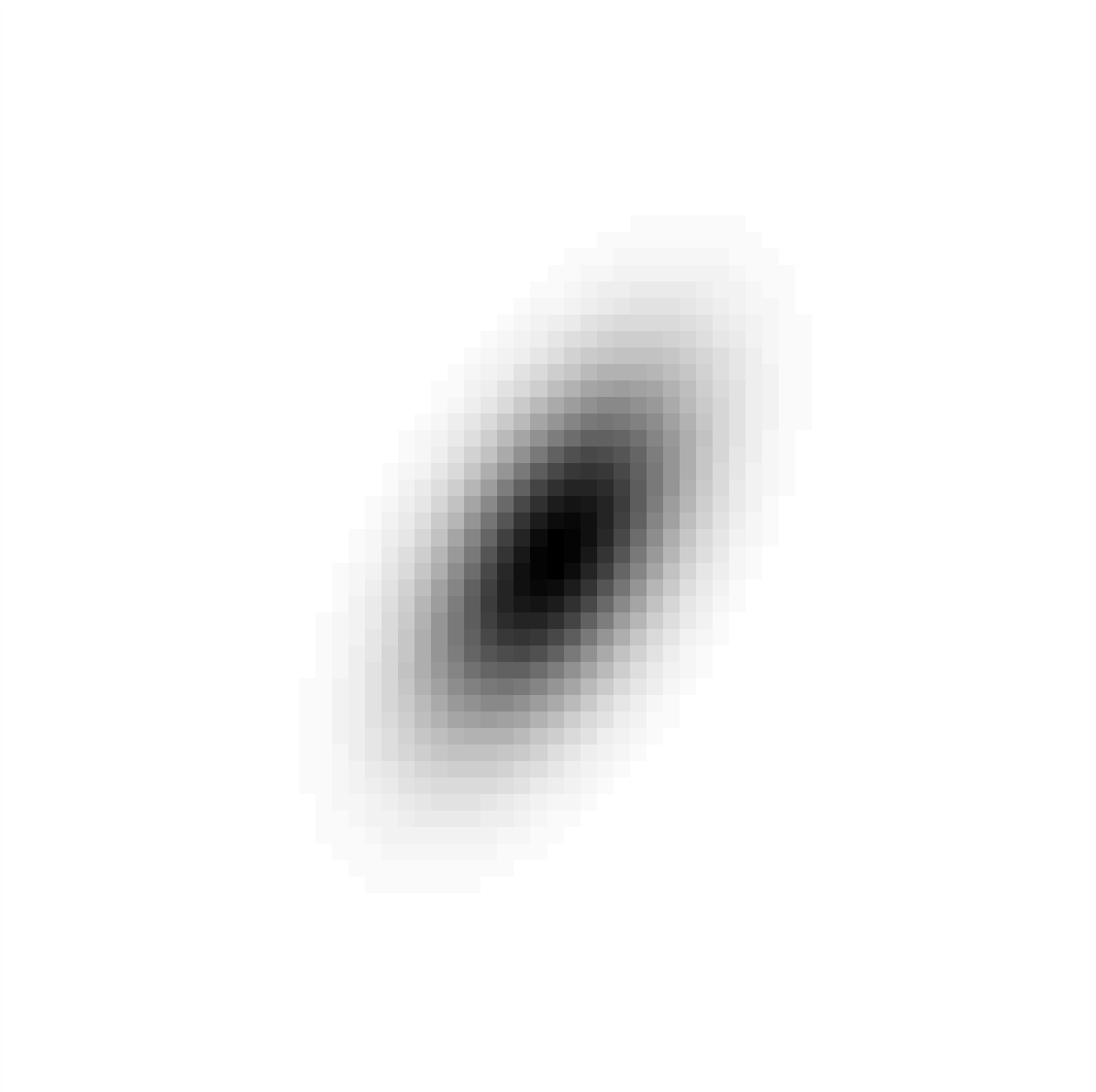} \hspace{-4mm} &
      \includegraphics[width=0.15\textwidth]{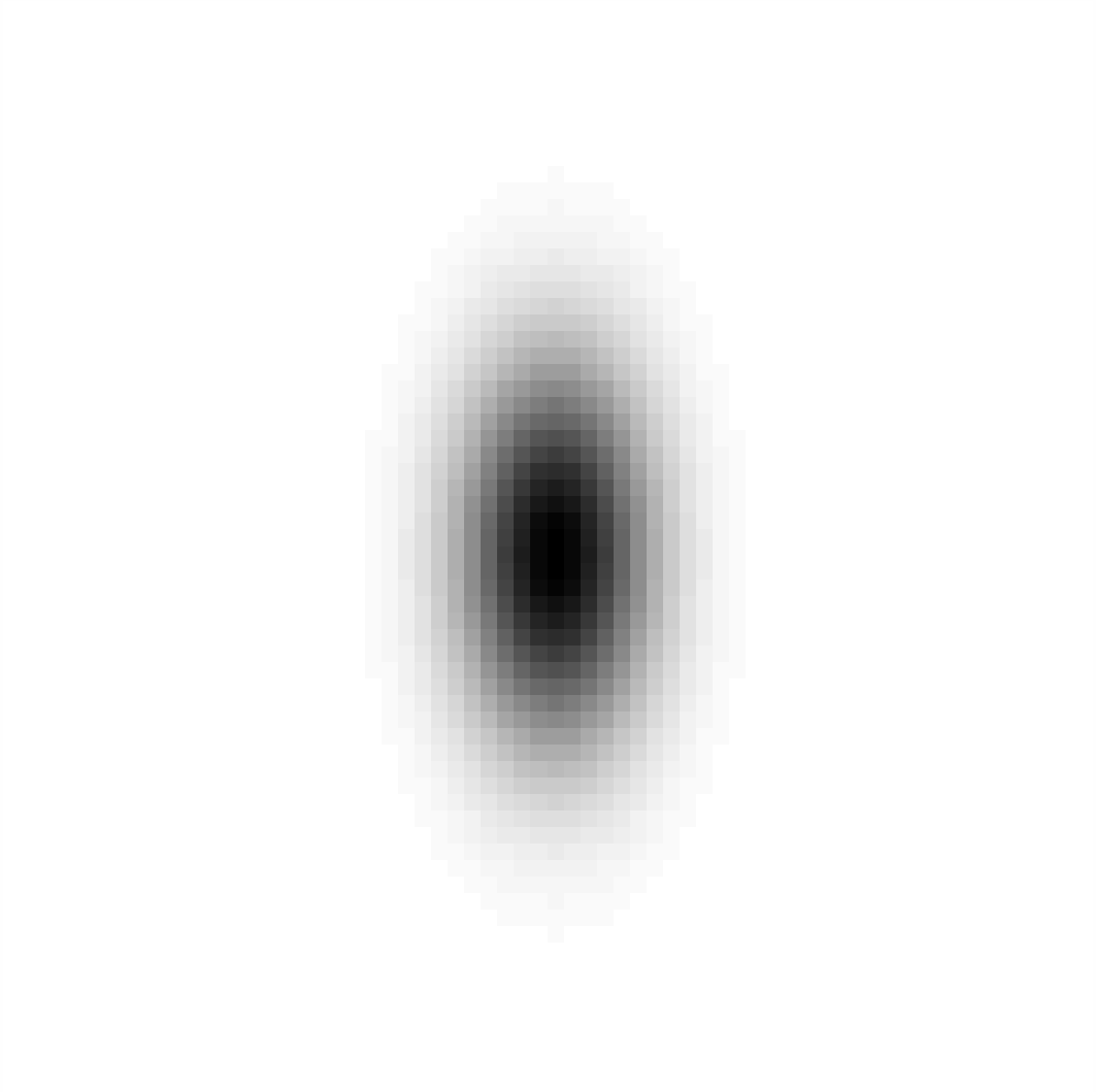} \hspace{-4mm} &
      \includegraphics[width=0.15\textwidth]{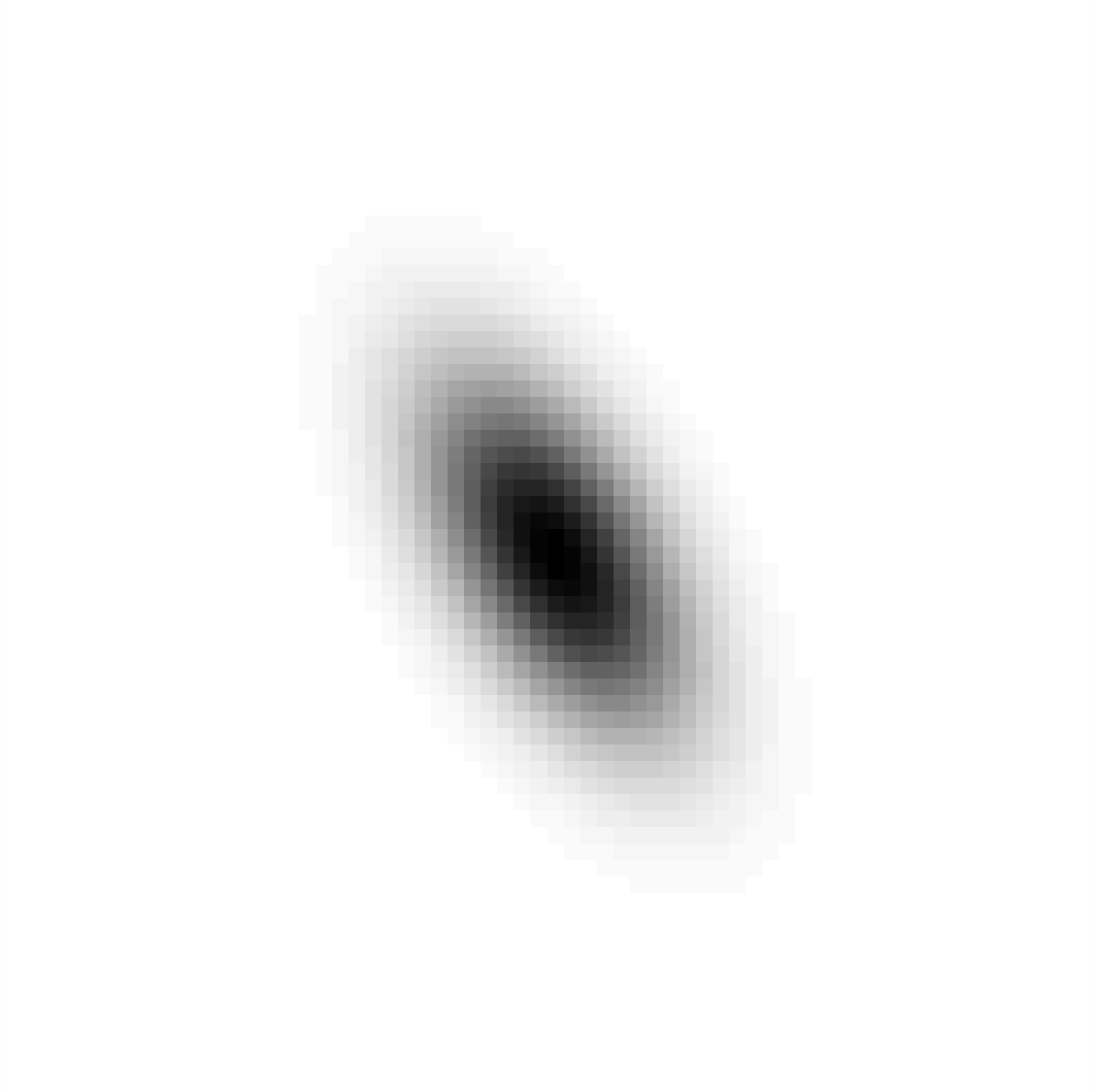} \hspace{-4mm} &
      \includegraphics[width=0.15\textwidth]{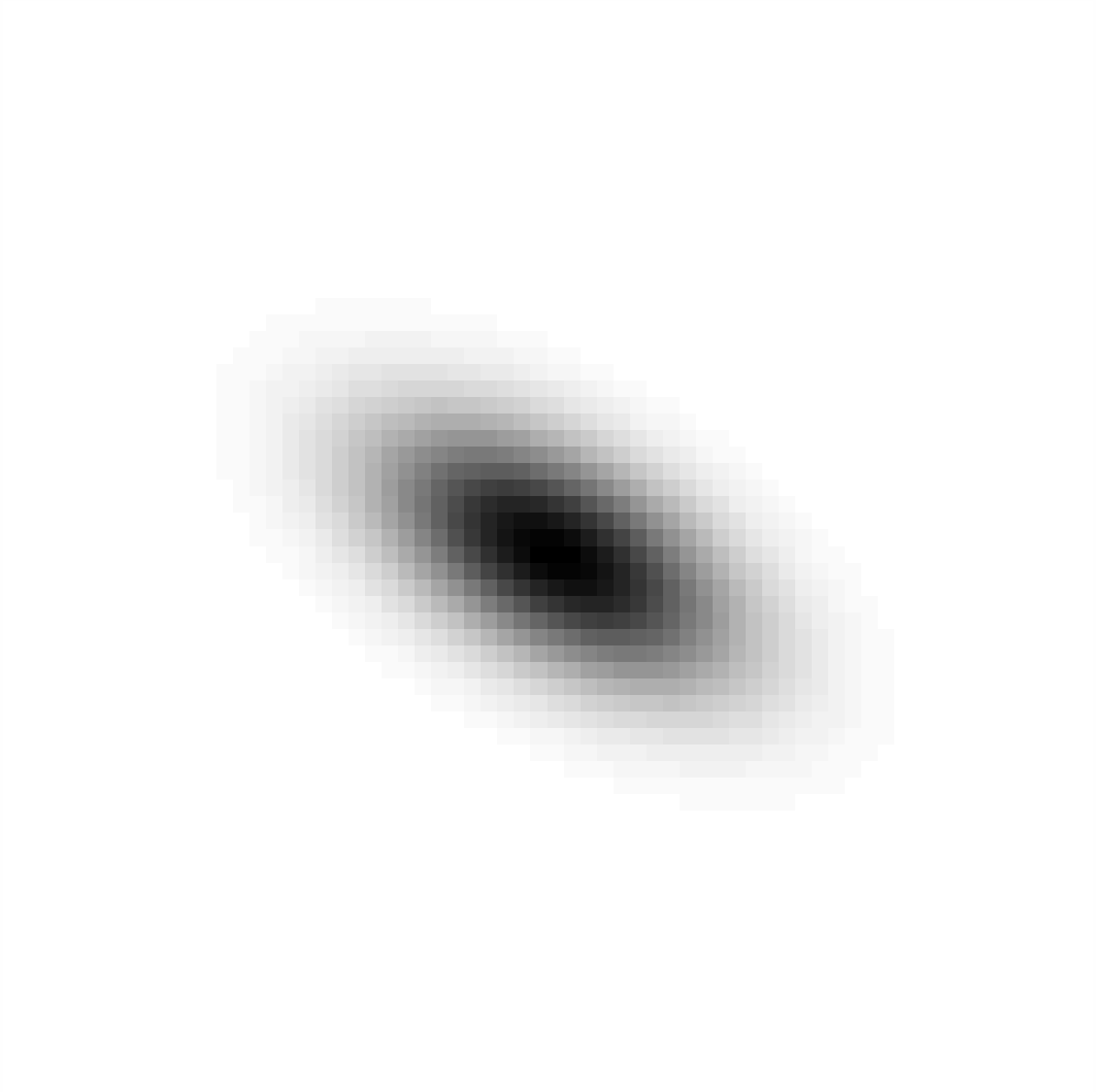} \hspace{-4mm} \\
    \end{tabular} 
  \end{center}
  \vspace{-4mm}
  \caption{Examples of discrete affine Gaussian kernels $h(x, y;\; \Sigma)$ and their equivalent directional
    derivative approximation kernels $\delta_{\orth \varphi}h(x, y;\; \Sigma)$ and
    $\delta_{\orth \varphi \orth \varphi}h(x, y;\; \Sigma)$ 
    up to order two in the two-dimensional case, here as generated
    from repeated iteration of $3 \times 3$-kernels of the form
    (\ref{eq-disc-3x3-kernel}) 
    for $\lambda_1 = 64$, $\lambda_2=16$
            $\alpha = 0, \pi/6, \pi/3, \pi/2, 2\pi/3, 5\pi/6$,
   $\Delta s = 1/2$ and with $C_{xxyy}$ according to
   (\ref{eq-choice-Cxxyy-discaff-3x3-iter-minimal-factor0p5center}). (Kernel size: $65 \times 65$ pixels.)}
  \label{fig-aff-elong-filters-dir-ders-iter-3x3}
\end{figure}

The bottom row in figure~\ref{fig-aff-elong-filters-dir-ders-iter-3x3}
shows examples of discrete affine kernels generated in this way from
repeated application of $3 \times 3$-kernels of the form
(\ref{eq-disc-3x3-kernel}) for $\Delta s = 1/2$ and with $C_{xxyy}$ 
according to (\ref{eq-choice-Cxxyy-discaff-3x3-iter-minimal-factor0p5center}).

Alternatively, one can choose the degrees of freedom within the
constraint
\begin{equation}
  \max\left(|C_{xy}|, \frac{C_{xx} + C_{yy} + \max(C_{xx}, C_{yy}) - 2}{2}\right)
  \leq C_{xxyy} \leq \min(C_{xx}, C_{yy})
\end{equation}
to minimize the deviations of the fourth-order terms in the Fourier
transform (\ref{eq-aff-scsp-FT-taylor-2}) from an elliptic behaviour or some other measure of the
deviations from affine isotropy. Since such an analysis does, however,
become technically more complicated, we will not
develop that theory further in this treatment.

\subsection{Discrete approximation of scale-space derivatives}

When computing discrete approximations to directional derivatives of
these discrete affine Gaussian kernels, we can proceed in a similar
way as for the discrete affine Gaussian kernels obtained via an FFT
and define discrete directional derivative operators $\delta_{\varphi}$,
$\delta_{\orth\varphi}$, $\delta_{\varphi\varphi}$, $\delta_{\varphi\orth\varphi}$
and $\delta_{\orth\varphi\orth\varphi}$ according to
equations~(\ref{eq-dir-der-phi})--(\ref{eq-dir-der-phiorthphiorth}).

The middle and the top rows in figure~\ref{fig-aff-elong-filters-dir-ders-iter-3x3}
show examples of discrete affine derivative approximation kernels
generated in this way from repeated iteration of $3 \times 3$-kernels of the form
(\ref{eq-disc-3x3-kernel}) for $\lambda_1 = 64$, $\lambda_2=16$
$\alpha = 0, \pi/6, \pi/3, \pi/2, 2\pi/3, 5\pi/6$, $\Delta s = 1/2$
and with $C_{xxyy}$ according to (\ref{eq-choice-Cxxyy-discaff-3x3-iter-minimal-factor0p5center}).

By applying corresponding discrete derivative approximation kernels to
the colour-opponent channels $u$ and $v$ according to a
colour-opponent representation (\ref{eq-col-opp-space-uv-from-rgb}),
we obtain discrete affine Gaussian colour-opponent directional
derivative approximations analogous to the colour-opponent receptive fields
shown in figure~\ref{fig-aff-elong-filters-dir-ders-col-opp}.

Note that when we apply these affine Gaussian derivative approximation
receptive fields in practice, we do never generate these kernels
explicitly. Instead, we apply compact support $3 \times 3$ discrete directional
derivative approximation kernels
(\ref{eq-dir-der-phi})--(\ref{eq-dir-der-phiorthphiorth}) directly
to the output of iterative application of the $3 \times 3$-kernel
(\ref{eq-disc-3x3-kernel}).


In Section~\ref{sec-aff-hybr-pyr}, we show how this discrete affine
scale-space concept can be complemented by spatial subsampling,
to generate affine hybrid pyramids, which allow for computationally
more efficient smoothing operations at coarser spatial scales,
by allowing for both coarser steps in the scale direction and
computations over a lower number of image pixels at coarser levels of
resolution as depending on the scale level.

\section{Scale-normalized derivatives for affine Gaussian scale space}
\label{sec-sc-norm-ders-disc}

\subsection{Continuous affine Gaussian derivative kernels}

When defining scale-normalized derivatives for the continuous affine
Gaussian derivative based receptive fields of the form $g_{\varphi^m \orth \varphi^n}(x, y;\; \Sigma_s)$, 
we use the eigenvalues $\lambda_1$ and $\lambda_2$ of the spatial covariance matrix $\Sigma_s$
as scale parameters
\begin{equation}
  L_{\varphi^m \orth \varphi^n,norm}(x, y;\; \Sigma_s) 
  = \lambda_1^{m\gamma_1/2} \, \lambda_2^{n\gamma_2/2} \, 
      \partial_{\varphi}^m \partial_{\orth \varphi}^n \; L(x, y;\; \Sigma_s),
\end{equation}
where $\gamma_1$ and $\gamma_2$ denote possibly different scale
normalization powers for the two orthogonal directions $\varphi$ and
$\orth \varphi$ and specifically the choice $\gamma_1 = \gamma_2 = 1$
implies maximal scale invariance.

\subsection{Discrete affine Gaussian derivative approximation kernels}

For discrete approximations of affine Gaussian derivatives obtained by
applying the discrete directional derivative approximations according
to (\ref{eq-dir-der-phi})--(\ref{eq-dir-der-phiorthphiorth}) to either
the semi-discrete affine scale-space concept according to
(\ref{eq-diff-eq-disc-aff-scsp}) or the scale-discretized scale-space
concept corresponding to repeated application of the 
$3 \times 3$-kernel (\ref{eq-disc-3x3-kernel}), we can in a corresponding
manner define {\em variance-normalized\/} affine Gaussian scale-space
derivative approximations according to
\begin{equation}
  L_{\varphi^m \orth \varphi^n,norm}(x, y;\; \Sigma_s) 
  = \lambda_1^{m\gamma_1/2} \, \lambda_2^{n\gamma_2/2} \, 
      \delta_{\varphi^m \orth \varphi^n} \, L(x, y;\; \Sigma_s),
\end{equation}
where $\delta_{\varphi^m \orth \varphi^n}$ denotes the directional
derivative approximation operator generalizing the definitions in
(\ref{eq-dir-der-phi})--(\ref{eq-dir-der-phiorthphiorth}) and
with the implicit understanding that $\delta_{\varphi^2} = \delta_{\varphi\varphi}$
and $\delta_{\orth\varphi^2} = \delta_{\orth\varphi\orth\varphi}$, etc.

Alternatively, we can define scale-normalized affine Gaussian
derivative approximations from $l_p$-normalization by requiring that the discrete
$l_p$-norm of the discrete derivative approximation kernel 
being equal to
the continuous $L_p$-norm of the corresponding continuous Gaussian
derivative operator
\begin{equation}
  \label{eq-sc-norm-der-raw-def-alpha-Lp-norm}
  L_{\varphi^m \orth \varphi^n,norm}(x, y;\; \Sigma_s) 
  = \alpha(\lambda_1, \gamma_1, \lambda_2, \gamma_2) \,
      \delta_{\varphi^m \orth \varphi^n} \, L(x, y;\; \Sigma_s),
\end{equation}
with the affine scale normalization factor $\alpha(\lambda_1, \gamma_1, \lambda_2, \gamma_2)$ determined
such that
\begin{equation}
  \label{eq-sc-norm-der-def-alpha-Lp-norm}
   \alpha(\lambda_1, \gamma_1, \lambda_2, \gamma_2) \, 
   \| \delta_{\varphi^m \orth \varphi^n} \, h(x, y;\; \Sigma_s)  \|_p
  = \lambda_1^{m\gamma_1/2} \, \lambda_2^{n\gamma_2/2} \,
    \| \partial_{\varphi}^m \partial_{\orth \varphi}^n \; g(x, y;\;  \Sigma_s) \|_p.
\end{equation}
In the absence of further information, the most scale-invariant choice
is to use $p = 1$ for $\gamma_1 = \gamma_2 = 1$.

\section{Summary and discussion}
\label{sec-summ-disc}

The affine Gaussian derivative model constitutes a canonical model for
visual receptive fields over a spatial image domain.
As we described in Section~\ref{sec-aff-rec-fields-overview},
this model can be derived by necessity from assumptions that reflect
structural properties of the world in combination with requirements to
ensure internal consistency between image representations at different
scales.
The receptive field profiles predicted by this theory do furthermore well
model the spatial shapes of the receptive fields of simple cells in the primary visual
cortex (V1).
Thus, this model can be motivated both mathematically/physically and
from a biological viewpoint. The close similarity between the results
obtained from mathematical/physical arguments {\em vs.\/}\
the properties of biological receptive fields, as resulting from ages
of evolutionary pressure, strongly support that this should be an
appropriate computational model for spatial visual receptive fields.

Compared to the more commonly used Gaussian derivative model based on
spatial derivatives of the rotationally symmetric Gaussian kernel, 
the family of affine Gaussian derivative kernels based on directional
derivatives of affine Gaussian kernels is additionally
covariant under local affine transformations of the spatial domain.
This property allows for more accurate modelling of receptive field
measurements under perspective transformations, which in turn
enables the development of more robust computer vision algorithms
under the geometric transformations that arise when observing a 3-D
world using a 2-D image sensor. 

From these properties together, it is natural to propose a design
strategy for a vision system, where affine Gaussian derivative kernels
are used as a first layer of visual receptive fields applied to the
raw input images, and then higher-layer visual operations use 
the output from these affine Gaussian derivatives as input.
The theory for affine Gaussian derivatives is, however, fully continuous,
assuming image functions defined over a continuous spatial domain.
This raises the question about
how to discretize this theory for a discrete implementation, while still preserving as many as
possible of the desirable properties that define the uniqueness of the affine
Gaussian derivative kernels in the continuous case.

To address this problem, we have presented a theory for discretizing the affine Gaussian
scale-space concept underlying the affine Gaussian derivative model, 
so that scale-space properties hold also in 
a discrete sense for the discrete implementation.
Specifically, we have presented two types of spatial discretizations
of the continuous affine diffusion equation that describes the effect
of convolution with affine Gaussian kernels: 
(i)~a semi-discretization over space only while leaving the continuous
scale parameter in terms of a family of spatial covariance matrices defined over a
continuum and
(ii)~an additional discretization in the scale direction that leads to
compact $3 \times 3$ kernels to be applied repeatedly, to compute
discrete affine Gaussian scale-space representations at successively coarser
scales, for families of discrete affine Gaussian kernels of different
size, orientation and eccentricity.

The first theory is defined uniquely from similar assumptions as
constitute the scale-space axioms that define the uniqueness of the
affine Gaussian kernel over a continuous spatial domain.
A fully axiomatic derivation of this model is given in Appendix~\ref{sec-app-disc-scsp}.
For the second theory, the restriction to local kernels of size 
$3 \times 3$ is implied by necessity from the locality property
implied by the assumption of non-enhancement of local extrema,
which constitutes our formalization of the requirement that new
structures must not be created from finer to coarser scales.

A third option is to combine the smoothing process in the second model 
with spatial subsampling operations at coarser spatial scales,
leading to the notion affine hybrid pyramids for improving both the
computational efficiency and the memory requirements at coarser
spatial scales.
 
These three families of discrete affine Gaussian kernels can in turn be combined with discretizations of
spatial directional derivative operators for computing discrete
approximations of directional derivatives, leading to families of
discrete affine Gaussian derivative approximations, which are intended to serve as a
basis for expressing later stage visual processes.

For these three types of spatial discretization approaches, we have additionally
solved a number of technical problems:
\begin{itemize}
\item
For the first two models, we have shown that the spatial covariance matrices
of the resulting discrete kernels are (up to numerical errors) exactly
equal to the corresponding covariance matrices in the continuous affine
scale-space theory
(Equations~(\ref{eq-cov-semi-disc-kernels}) and (\ref{eq-cov-semi-disc-3x3-kernels})).
\item
For the first semi-discrete model, we have analysed the influence of a
remaining free parameter $C_{xxyy}$ in the theory, not determined by the primary
assumptions 
(Sections~\ref{sec-free-Cxxyy-semi-disc-aff-scsp}--\ref{sec-choice-Cxxyy-semi-disc-aff-scsp}).
 We have stated necessary and sufficient conditions to
ensure a positive discretization 
(Equation~(\ref{eq-pos-req-aff-scsp-2D})) and shown how this parameter in
specific cases can be optimized for achieving better numerical
approximation of rotational symmetry or affine angular dependency of
the Fourier transform 
(Section~\ref{sec-determ-Cxxyy-low-freq}) or minimizing the
fourth-order moments of the discrete smoothing kernel
(Section~\ref{sec-choice-Cxxyy-semi-disc-aff-scsp}).
\item
For the second fully discrete model, while without spatial subsampling, we have derived an explicit
parameterization of compact $3 \times 3$-kernels 
in terms of the contribution they give to the spatial covariance of
the composed discrete affine Gaussian kernel
(Equation~(\ref{eq-disc-3x3-kernel}))
and analysed how the scale step in the scale direction can be
determined when applying these $3 \times 3$-kernels repeatedly
(Section~\ref{sec-choice-scale-step}). We have also analyzed
how the free parameter $C_{xxyy}$ can be determined within 
the positivity constraints
(Section~\ref{sec-choice-Cxxyy-semi-disc-aff-scsp-3x3}) and proposed the
specific choice
(\ref{eq-choice-Cxxyy-discaff-3x3-iter-minimal-factor0p5center}) for
choosing this parameter as function of the elements $C_{xx}$,
$C_{xy}$ and $C_{yy}$ in the affine covariance matrix $\Sigma$.
\item
For the third fully discrete model, also involving spatial subsampling
at coarser spatial scales, we have proposed an algebra for defining
reduction operators from finer to coarser spatial scales 
(Section~\ref{sec-reduct-op-aff-hybr-pyr}) as well as a
criterion for measuring how much spatial smoothing needs to be applied
(depending on the eccentricity of the anisotropic spatial smoothing kernels) before
the affine hybrid pyramid representation may be spatially subsampled
(Equation~(\ref{eq-crit-choose-k-for-subsampl-aff-hybr-pyr})).
We have specifically proposed ways of measuring the scale
levels and the subsampling rate in the presence of spatial subsampling
and shown how equivalent discrete derivative computation kernels can
be defined to in turn allow for the definition of appropriate scale
normalization factors 
(\ref{eq-sc-norm-der-def-alpha-Lp-norm-aff-hybr-pyr}) in
the presence of spatial subsampling
(Sections~\ref{eq-equiv-conv-der-kern-aff-hybr-pyr}--\ref{sec-sub-sampl-rate-aff-hybr-pyr}).
\end{itemize}
Out of the three presented spatial discretization models, the first semi-discrete model allows
for the most accurate discrete implementation, in the sense of mimicking the
corresponding continuous scale-space properties.
This model can be implemented with reasonable efficiency using a
closed-form expression for the discrete Fourier transform of the 
discrete affine Gaussian kernel (\ref{eq-FT-semi-disc-aff-scsp}).
If an inverse Fourier transform for each layer in the affine
scale-space representation is regarded as
requiring too much computational work, a discrete implementation based on repeated
application of the $3 \times 3$-kernels (\ref{eq-disc-3x3-kernel})
according to the second discretization model
may be computationally more efficient, at least at finer scale levels,
specifically on massively parallel architectures such as GPUs.
From such a conceptual background, the semi-discrete Fourier-based implementation
can be regarded as a reference, to optimize the implementation based
on repeated application of $3 \times 3$-kernels against.
If an implementation based on repeated application of 
$3 \times 3$-kernels using the same resolution at all scale levels is
still regarded as requiring too much computational work with respect
to the task at hand, then an affine hybrid pyramid
implementation should be chosen, with the minimum amount of 
necessary smoothing before a subsampling stage as determined by the spatial 
smoothing parameter $K$ in
(\ref{eq-crit-choose-k-for-subsampl-aff-hybr-pyr}) and leading to the
subsampling rate according to
(\ref{eq-def-subsampl-rate-rho}) optimized to lead to an
appropriate trade-off between numerical accuracy and computational
efficiency. If on the other hand, coarser scale representations are to
be computed frequently, and if a Fourier-based implementation is
affordable, then the first semi-discrete Fourier-based method can be chosen, since
it is the most accurate of these three methods. Thus, the presented
 theory offers three main alternatives, depending on the complexity
 and the accuracy requirements of
the visual task {\em vs.\/}\ the available computational resources.

A limitation of the presented genuinely discrete affine Gaussian
scale-space concepts, however, is that the requirement of a positive
discretization, as implied by the discrete scale-space axioms
underlying the formulation of the theory, implies a bound on the ratio
between the maximum and minimum eigenvalues of the spatial covariance
matrix that determines the shapes of the affine Gaussian kernels.
The theoretical analysis in
Section~\ref{sec-free-Cxxyy-semi-disc-aff-scsp} gives an upper bound
on the eccentricity $\epsilon = \lambda_{max}/\lambda_{min} = 3 + 2\sqrt{2} \approx 5.8$
corresponding to the amount of foreshortening caused by a slant angle of 
$65.5~\mbox{degrees}$.
While not covering extreme amounts of foreshortening, this bound 
still enables handling of basic use cases
for expressing affine covariant image operations between multiple
views of the same object.
If higher amounts of foreshortening are to be handled, alternatives
beyond the {\em ad hoc\/} solution of just imposing an upper bound on
the eccentricity according to the stated criterion, are to investigate
numerical approximations of the continuous
model using alternative discretizations as outlined in 
Section~\ref{sec-disc-aff-gauss-rec-fields}.

We propose that the presented theory can be used for generating
different architectures for early visual operations based on affine
Gaussian derivative kernels or extensions thereof.
The most straightforward type of architecture consists of extending
the use of a regular Gaussian scale-space representation, based on
rotationally symmetric Gaussian kernels over
multiple spatial scales, to an affine Gaussian scale-space
representation over both spatial scale levels, orientations in image space and
eccentricities in terms of ratios between the scale values in the two
orthogonal eigen-directions of the affine Gaussian kernel.
Thus, in a first step a set of covariance matrices representing affine Gaussian kernels of different orientations
and different eccentricities may be generated analogous to the sample
tesselation on a hemisphere as shown in
Figure~\ref{fig-distr-aff-rec-fields} or a rectangular tesselation
over the sizes, orientations and eccentricities of the affine Gaussian kernels.
Then, for each orientation and each eccentricity, a set of affine
Gaussian scale-space representations with associated affine Gaussian
derivatives can be computed for a set of
scale values, according to either of the three presented
discretization models. This representation can in turn be used as
input for a vision system addressing one or several visual tasks under
substantial image deformations caused by significant perspective effects.

\appendix

\section{Non-isotropic scale-space for discrete signals: Necessity and sufficiency}
\label{sec-app-disc-scsp}

This appendix presents a general theoretical result concerning linear
shift-invariant scale-space representations of discrete signals over an
anisotropic $D$-dimensional domain.
The result is a generalization of earlier results presented in
(Lindeberg \cite{Lin90-PAMI,Lin93-Dis}) for isotropic discrete scale spaces.

\subsection{Definitions}

Let us state this (minimal) set of basic properties
a family should satisfy to be a candidate family
for generating a linear scale-space.

\begin{definition}
  {\tlthmheadfont (Discrete pre-scale-space family of kernels)}
  \label{def-pre-scsp-kern-disc}
  \newline
  A one-parameter family of kernels 
  $T \colon \bbbz^D \times \bbbr_+ \rightarrow \bbbr$ 
  is said to be a discrete pre-scale-space family of kernels
  if it satisfies
  \begin{itemize}
  \item
  $T(\cdot;\;0) = \delta(\cdot)$,
  \item
    the semi-group property
    $T(\cdot;\;s_1) \ast T(\cdot;\;s_2) = T(\cdot;\;s_1+s_2)$,
  \item
    the continuity requirement
    $\parallel T(\cdot;\;s) - \delta(\cdot) \parallel_1
    \rightarrow 0$ when $s \downarrow 0$.
  \end{itemize}
\end{definition}
\begin{definition}
  {\tlthmheadfont (Discrete pre-scale-space representation)}
  \label{def-pre-scsp-repr-disc}
  Let $f \colon \bbbz^D \rightarrow \bbbr$ be a discrete signal
  and let $T \colon \bbbz^D \times \bbbr_+ \rightarrow \bbbr$ be a 
  discrete pre-scale-space family
  of kernels. Then, the one-parameter family of signals 
  $L \colon \bbbz^D \times \bbbr_+ \rightarrow \bbbr$ given by
  \begin{equation}
    \label{eq-post-form-smoothing-disc}
    L(x;\;s) = 
      \sum_{\xi \in \bbbz^D} T(\xi ;\;s) \, f(x-\xi).
  \end{equation}
  is said to be the discrete {pre-scale-space representation} of 
  $f$ generated by $T$.
\end{definition}
\begin{lemma}
  {\tlthmheadfont (A discrete pre-scale-space representation is differentiable)}
  \label{lem-pre-scsp-differentiable-disc}
  \newline
  Let $L \colon \bbbz^D~\times~\bbbr_+~\rightarrow~\bbbr$ be the 
  discrete pre-scale-space representation of a signal 
  $f \colon \bbbz^D \rightarrow \bbbr$ in $l_1$. 
  Then, $L$ satisfies the differential equation
  \begin{equation}
    \label{gen-nD-diff-eq-disc}
    \partial_s L = {\cal A} L 
  \end{equation}
  for some linear and shift-invariant operator ${\cal A}$.
\end{lemma}
\begin{myproof}
  If $f$ is sufficiently regular, {\em e.g.\/}, if $f \in L_1$, 
  define a family of operators $\{ {\cal T}_s, s>0 \}$, here from from $L_1$
  to $L_1$, by ${\cal T}_s f = T(\cdot;\;s) \ast f$. 
  Due to the conditions imposed on
  the kernels, the family satisfies the relation
  \begin{equation}
    \lim_{s \rightarrow s_0}
    \parallel ({\cal T}_s - {\cal T}_{s_0}) f \parallel_1 =
    \lim_{s \rightarrow s_0}
    \parallel ({\cal T}_{s-s_0} - {\cal I}) ({\cal T}_{s_0} f) \parallel_1 = 0,
  \end{equation}
  where ${\cal I}$ is the identity operator.
  Such a family is called a strongly continuous semigroup of operators
  (Hille and Phillips \cite[pages~58--59]{HilPhi57})
  A semi-group is often characterized by its 
  {\em {infinitesimal generator}\/}
  ${\cal A}$ defined by
  \begin{equation}
    {\cal A}f = \lim_{h \downarrow 0} \frac{{\cal T}_h f - f}{h}.
  \end{equation}
  The set of elements $f$ for which ${\cal A}$ exists is denoted
  ${\cal D}({\cal A})$.
  This set is not empty and never reduces to the zero element.
  Actually, it is even dense in $L_1$ 
  (Hille and Phillips \cite[page~307]{HilPhi57}).
  If this operator exists then
    \begin{multline}
      \lim_{h \downarrow 0} 
        \frac{L(\cdot,\cdot;\;s+h) - L(\cdot,\cdot;\;s)}{h} 
      = \lim_{h \downarrow 0} \frac{{\cal T}_{s+h} f - {\cal T}_s f}{h} \\ 
      = \lim_{h \downarrow 0} 
        \frac{{\cal T}_h ({\cal T}_s f) - ({\cal T}_s f)}{h} 
      = {\cal A} ({\cal T}_s f) = {\cal A} L(\cdot;\;s).
    \end{multline}
  According to a theorem in Hille and Phillips \cite[page~308]{HilPhi57},
  strong continuity implies
  $\partial_s ({\cal T}_s f) = {\cal A} {\cal T}_s f =
   {\cal T}_s {\cal A} f$ 
  for all $f \in {\cal D}({\cal A})$.
  Hence, the scale-space family $L$
  must obey the differential equation
  $\partial_s L = {\cal A} L$
  for some linear operator ${\cal A}$.
  Since $L$ is generated from $f$ by a convolution operation, it follows
  that ${\cal A}$ must be shift-invariant.
\end{myproof}
\begin{definition}
  \label{def-pre-scsp-prop-disc}
  {\tlthmheadfont (Pre-scale-space property: Non-enhancement of local extrema)}
  \newline
  A discrete pre-scale-space representation 
  $L \colon \bbbz^D \times \bbbr_+ \rightarrow \bbbr$ 
  of a discrete signal
  is said to possess 
  pre-scale-space properties, or equivalently not to enhance
  local extrema,
  if for every value of the scale parameter $s_0 \in\bbbr_+$ it holds that
  if $x_0 \in \bbbz^D$ is a local maximum or a local minimum for the
  mapping $x \mapsto L(x;\;s_0)$, then the derivative
  of $L$ with respect to $s$ satisfies
  \begin{align}
    \begin{split}
      \label{max-point-cond}
      \partial_s L \leq 0 \quad\quad \mbox{at any local maximum},
    \end{split}\\
    \begin{split}
      \label{min-point-cond}
      \partial_s L \geq 0 \quad\quad \mbox{at any local minimum}.
    \end{split}
  \end{align}
\end{definition}
\begin{definition}
  {\tlthmheadfont (Discrete scale-space family of kernels)}
  \label{def-sc-sp-fam-disc}
  A one-parameter family of discrete pre-scale-space kernels 
  $T \colon \bbbz^D \times \bbbr_+ \rightarrow \bbbr$
  is said to be a discrete scale-space family of kernels if
  for any discrete signal
  $f \colon \bbbz^D \rightarrow \bbbr \in l_1$ the discrete pre-scale-space
  representation
  of $f$ generated by $T$ possesses pre-scale-space properties,
  {\em i.e.\/}, if {for any} signal local extrema are never enhanced.
\end{definition}
\begin{definition}
  {\tlthmheadfont (Discrete scale-space representation)}
  \label{def-sc-sp-reps-disc}
  A discrete pre-scale-space representation 
  $L \colon \bbbz^D \times \bbbr_+ \rightarrow \bbbr$ of a
  signal $f \colon \bbbz^D \rightarrow \bbbr$ generated by a
  family of discrete kernels $T \colon \bbbz^D \times \bbbr_+ \rightarrow \bbbr$,
  which are discrete scale-space kernels, is said to be
  a discrete scale-space representation of $f$.
\end{definition}

\subsection{Necessity}

\begin{theorem}
  {\tlthmheadfont (Scale-space for discrete signals: Necessity)}
  \label{thm-nD-scale-space-family-nec-disc}
  $\;$ \newline A discrete scale-space representation 
  $L \colon \bbbz^D \times \bbbr_+ \rightarrow \bbbr$ of a
  discrete signal $f \colon \bbbz^D \rightarrow \bbbr$ satisfies the 
  differential equation 
  \begin{equation}
    \label{eq-nD-scsp-undet-par-disc}
    \partial_s L = {\cal A} L
  \end{equation}
  with initial condition $L(\cdot;\;0) = f(\cdot)$ for some
  infinitesimal scale-space generator ${\cal A}$
  characterized by:
  \begin{itemize}
    \item
    the {\em locality\/} condition 
    $a_{\xi} = 0$ if $|\xi|_{\infty} > 1$,
    \item
    the {\em positivity\/} constraint
    $a_{\xi} \geq 0$ if $\xi \neq 0$ and
    \item
    the {\em zero sum\/} condition 
    $\sum_{\xi \in \bbbz^D} a_{\xi} = 0$.
  \end{itemize}
\end{theorem}

\begin{myproof}
  The proof consists of two parts. The first part has already
  been presented in Lemma~\ref{lem-pre-scsp-differentiable-disc},
  where it was shown that the requirements on the kernels imply
  that the family $L$ obeys a linear differential equation. 
  Because of the shift invariance, ${\cal A} L$ can be written
  on the form 
  \begin{equation}
    \label{gen-nD-lin-op-disc}
    (\partial_s L)(x;\; s) 
    = ({\cal A} L)(x;\;s) 
    = \sum_{\xi \in \bbbz^D} a_{\xi} \, L(x+\xi; \;s).
  \end{equation}
  In the second part, counterexamples will be constructed from
  various simple test functions in order to delimit the class 
  of possible operators.

  \paragraph{Step 1}
  
  Let $N_+(x_0)$ denote the set of points around $x_0 \in \bbbz^D$ within
  distance 1 in maximum norm, {\em i.e.\/}, 
  $N_+(x_0) = \{ x \in \bbbz^D \colon \| x - x_0 \|_{\infty} \leq 1 \} $.
  Let furthermore $N(x_0)$ denote the corresponding neighbourhood of
  $x_0$, by excluding the central point.

  The extremum point conditions {\rm (\ref{max-point-cond})} and
  {\rm (\ref{min-point-cond})} combined with
  Definitions~\ref{def-sc-sp-fam-disc}--\ref{def-sc-sp-reps-disc}
  mean that ${\cal A}$ must be {\em local},
  i.e., that $a_{\xi} = 0$ if $\xi \not\in N_+(0)$. This can be easily
  understood by studying the following counterexample:
  First, assume that $a_{\xi_0} > 0$ for some
  ${\xi_0 \not\in N_+(0)}$
  and define a function
  $f_1 \colon \bbbz^D \rightarrow \bbbr$ by
  \begin{equation}
    f_1(x) = \left \{ \begin{array}{rl}
                  \varepsilon > 0 & \mbox{if $x = 0$,} \\
                          0          & \mbox{if $x \in N(\xi_0)$,} \\
                       1            & \mbox{if $x = \xi_0$,} \\
                       0            & \mbox{otherwise.}
                \end{array}
       \right.
  \end{equation}
  Obviously, $\xi_0$ is a local maximum point for $f_1$. 
  From (\ref{gen-nD-lin-op})   one obtains $\partial_s L(\xi_0;\;0) = 
  \epsilon a_{0} + a_{\xi_0}$.
  It is clear that this value can be positive provided that $\varepsilon$ is
  chosen small enough.
  Hence, $L$ cannot satisfy {\rm (\ref{max-point-cond})}.
  Similarly, it can also be shown that $a_{\xi_0} < 0$
  leads to a violation of the non-enhancement property
  {\rm (\ref{min-point-cond})} (let $\varepsilon < 0$). 
  Consequently, $a_{\xi}$ must be zero if $\xi \not\in N_+(0)$.

  \paragraph{Step 2}

  With the assumed weak definitions of local extremum points, it is
  clear that for a constant input function $f_2(x) = C$,
  the point $x = 0$ is both a local maximum point and a
  local minimum point. Hence $\partial_s L(0;\;0)$
  must be zero, which implies that the {\em coefficients must sum to zero}
  \begin{equation}
    \sum_{\xi \in \bbbz^D} a_{\xi} = 0.
  \end{equation}

  \paragraph{Step 3}

  Finally, by observing that due to the zero sum condition, 
  {\rm (\ref{gen-nD-lin-op})} can be written
  \begin{equation}
    \label{gen-nD-lin-op-red}
    \partial_s L =
      ({\cal A}L)(x;\;s) = 
      \sum_{\xi \in N(0)} a_{\xi} (L(x-\xi; \;s) - L(x; \;s)).
  \end{equation}
  By considering the test function
  \begin{equation}
    f_3(x, y) = 
      \left \{ 
         \begin{array}{rl}
     \epsilon > 0 & \mbox{if $x = 0$,} \\
     -1           & \mbox{if $x =\tilde{\xi}$,} \\
     0            & \mbox{otherwise,}
         \end{array}    
      \right.
  \end{equation}
  for some $\tilde{\xi}$ in $N(0)$, one easily realizes
  that $a_{\xi}$ must be {\em non-negative\/} if $\xi \in N(0)$.

  The initial condition $L(\cdot;\;0) = f$ is a direct
  consequence of the definition of the notion of a pre-scale-space kernel.
\end{myproof}

\subsection{Sufficiency}

\begin{theorem}
  {\tlthmheadfont (Scale-space for discrete signals: Sufficiency)}
  \label{thm-nD-scale-space-family-suff-disc}
  $\;$ 
  \newline 
  Given an infinitesimal generator ${\cal A}$ corresponding to
  \begin{equation}
    ({\cal A} L)(x;\;s) 
    = \sum_{\xi \in \bbbz^D} a_{\xi} L(x+\xi; \;s),
  \end{equation}
  where the coefficients $a_{\xi}$ satisfy
  \begin{itemize}
    \item
    the {\em locality\/} condition 
    $a_{\xi} = 0$ if $|\xi|_{\infty} > 1$,
    \item
    the {\em positivity\/} constraint
    $a_{\xi} \geq 0$ if $\xi \neq 0$ and
    \item
    the {\em zero sum\/} condition 
    $\sum_{\xi \in \bbbz^D} a_{\xi} = 0$,
  \end{itemize}
  the solution of the diffusion equation
  \begin{equation}
    \partial_s L = {\cal A} L
  \end{equation}
  with initial condition $L(\cdot;\; 0) = f$ constitutes a discrete
  scale-space representation. Specifically, $L$ obeys 
  \begin{align}
    \begin{split}
      \partial_s L \leq 0 \quad\quad \mbox{at any local maximum},
    \end{split}\\
    \begin{split}
      \partial_s L \geq 0 \quad\quad \mbox{at any local minimum}.
    \end{split}
  \end{align}
\end{theorem}

\begin{myproof}
  From (\ref{gen-nD-lin-op-red}) it follows that the influence of $L$
  at any extremum point $(x_0;\; s_0)$ in scale-space can be written
  \begin{equation}
    \partial_s L(x_0;\; s_0) =
      ({\cal A}L)(x_0;\; s_0) = 
      \sum_{\xi \in N(0)} a_{\xi} (L(x_0-\xi; \;s_0) - L(x_0; \;s_0)),
  \end{equation}
  If $x_0$ is a local maximum, then all differences 
  $L(x_0-\xi; \;s_0) - L(x_0; \;s_0)$ are negative.
  Combined with the positivity of $a_{\xi}$, it follows that 
  $\partial_s L(x_0;\; s_0) \leq 0$.
  If $x_0$ is a local minimum, we apply the same way of reasoning to $-L$.
\end{myproof}

\section{Affine hybrid pyramids}
\label{sec-aff-hybr-pyr}

In this section, we will show how the fully discretized affine
Gaussian scale-space concept presented in
Section~\ref{sec-theory-disc-aff-rec-fields-3x3} in the main article can be
combined with spatial subsampling, to allow for computationally more
efficient smoothing operations at coarser spatial scales.
The motivation for this approach is similar to the motivations
underlying regular spatial pyramids: At coarser scales, when the image
representations become gradually smoother, the image values to also
become gradually more redundant, implying that the loss of accuracy
caused by spatial subsampling will be lower.

Specifically, we will show how repeated filtering with our derived
$3 \times 3$ kernel $k(C_{xx}, C_{xy}, C_{yy}, C_{xxyy}, \Delta_s)$
according to (\ref{eq-disc-3x3-kernel}) can be combined with spatial subsampling
operations to compute the discrete affine Gaussian scale-space
representations at coarser scales in a computationally more efficient
manner compared to using the same resolution at all levels of
scale. Consider the upper bound on the scale step $\Delta s \leq 1/2$ 
as holds for the isotropic diffusion equation 
(Equation~(\ref{eq-upper-bound-scale-step-isotropic-scsp}))
and which can also be extended to the non-isotropic diffusion equation
(Section~\ref{sec-anal-delta-s-non-aniso-diff-eq}).
By applying this upper bound relative to the
current resolution $h = 2^l$ after $l$ spatial subsampling stages 
by a factor of 2, the effective bound on the spatial step will become
\begin{equation}
  \label{eq-scale-step-hybr-pyr}
  \Delta s \leq \frac{h^2}{2} = \frac{2^{2l}}{2}, 
\end{equation}
thus implying that larger scale steps can be taken and at fewer
iterations will thus be needed to reach the coarser scale representations.
In addition, the computations will be done on a substantially lower
number of pixels, with the number of pixels divided by a factor
of $h^2 = 2^{2l}$.
Together, this implies that the computational efficiency will be
improved by two main factors that both increase exponentially 
with the amount of spatial subsampling.
This will lead to an extension of isotropic pyramid 
and hybrid pyramid representations 
(Burt and Adelson \cite{BA83-COM}; Crowley \cite{Cro81}; Crowley and Parker
\cite{Cro84-dolp,Cro84-peaks}; Lindeberg and Bretzner
\cite{LinBre03-ScSp})
to the notion affine hybrid pyramid representations.

\subsection{Reduction operators}
\label{sec-reduct-op-aff-hybr-pyr}

Following Burt and Adelson \cite{BA83-COM}; Crowley \cite{Cro81}; Crowley and Parker
\cite{Cro84-dolp,Cro84-peaks} and Lindeberg and Bretzner
\cite{LinBre03-ScSp}), let us describe the
transformation between two adjacent resolution levels in a
pyramid by a reduction operator.
For simplicity, let us assume that the size $N$ of the smoothing filter is the same over both
the coordinate directions and odd.
Then, the transformation from the representation $L^{(l)}$
at the current resolution level $l$, to the representation
$L^{(l+1)}$ at the next coarser resolution level $l+1$ is for some
set of filter coefficients $c \colon \bbbz \rightarrow \bbbr$
given by
\begin{align}
  \label{eq-reduce}
  \begin{split}
    L^{(l+1)}
      & = \reducecycle(L^{(l)})
  \end{split}\\
  \begin{split}
    L^{(l+1)}(x)
      & = \sum_{m=-(N-1)/2}^{(N-1)/2} \sum_{n=-(N-1)/2}^{(N-1)/2} 
             c(m, n) \, L^{(l)}(S x - m, S y - n).
  \end{split}
\end{align}
where $S = 2$ is the spatial subsampling factor.
Next, let us assume that the smoothing operation can be decomposed
into several smoothing steps:
\begin{equation}
  \label{eq-reduce-cycle}
  \begin{tabular}{cll}
    $\reducecycle$
      & $:=$
      & $\subsample$
      \\
    &
      & $\smooth^+$
      \\
 \end{tabular}
\end{equation}
where the notation $\op^+$ means that several operators
of the form $\op$ may occur. $\reducecycle$ is thus composed of
one or more smoothing operations followed by a subsampling step.
The subsampling operation is here defined by
\begin{align}
  \label{eq-sample}
    \begin{split}
      R
        & = \subsample(L;\; S)
    \end{split}
\\
    \begin{split}
      R(x, y)
        & = L(S x, S y) \quad\quad (s \in \bbbz_+),
    \end{split}
\end{align}
where again we usually choose $S = 2$ and where each smoothing step is of
the form
\begin{align}
  \label{eq-smooth}
    \begin{split}
      R
        & = \smooth(L)
    \end{split}
\\
    \begin{split}
      R(x, y)
        & = \sum_{m=-N}^{N}  \sum_{n=-N}^{N} a(m, n) \, L(x - m, y - n).
    \end{split}
\end{align}
Specifically, we assume
that the coefficients of the smoothing operation originate
from the  $3 \times 3$-kernel (\ref{eq-disc-3x3-kernel}) arising 
from the discretization of the diffusion operator 
\begin{equation}
  \label{eq-deltasmooth-def}
   \smooth(L) = \deltasmooth(L;\; \Delta s, 1, \Sigma),
\end{equation}
which in turn will be repeated $K$ times at each level $l$ of
resolution, corresponding to an affine hybrid pyramid with the
following structure over spatial scales as reflecting variations in
the size in the image domain
\begin{equation}
  \label{eq-reduce-cycle-aff-hybr-pyr}
\begin{tabular}{cll}
    $\reducecycle$
      & $:=$
      & $\subsample$
      \\
    &
      & $\deltasmooth(L;\; \Delta s, K, \Sigma)$
      \\
 \end{tabular}
\end{equation}
and with such a representation
computed for each orientation and eccentricity to reflect 
a grid-based sampling over the three-parameter family of affine Gaussian kernels.
The grid spacing $h$ as function of the resolution level $l$
will be given by
\begin{equation}
  \label{eq-h-fcn-of-l-hybr-pyr}
  h(l) = 2^{l},
\end{equation}
if we adopt the convention that $l = 0$ should correspond to the original resolution.

\subsection{Eccentricity-dependent spatial subsampling}

A particular factor that needs to be taken into account when computing
a hybrid pyramid representation based on affine smoothing kernels
is that amount of smoothing will be different in different spatial directions.
Let us adopt the convention from hybrid pyramids computed from
spatially isotropic binomial kernels (Lindeberg and Bretzner \cite{LinBre03-ScSp}),
where a binomial-$K$ pyramid can be constructed by applying
$K$ spatial smoothing steps with scale step $\Delta s = 1/2$ 
before a spatial subsampling stage is permitted.
Then, the total amount of smoothing before the subsampling operation
is at any level of resolution equal to
\begin{equation}
  \Delta s_{tot} = K \, \Delta s \quad\quad \mbox{for $\Delta s = \frac{1}{2}$}
\end{equation}
in units of a normalized grid spacing with $h = 1$.
Let us next for an affine hybrid pyramid representation
assume that the spatial covariance matrix $\Sigma$ is
normalized such that the maximum eigenvalue $\lambda_{max} = 1$.
Then, the minimum amount of spatial smoothing over all directions in
image space is given by $\lambda_{min} \in [0, 1]$.

To compute an affine hybrid pyramid with the same guaranteed minimum
amount of spatial smoothing as for a regular binomial-$K$ pyramid
based on spatially isotropic smoothing kernels with $C_{xx} = C_{yy}$ and $C_{xy}$,
we then have to stay at the same level of spatial resolution as long
as the iteration index $k$ within any level of resolution satisfies
\begin{equation}
  \label{eq-crit-choose-k-for-subsampl-aff-hybr-pyr}
  k \, \Delta s \, \lambda_2 \leq \frac{K}{2}.
\end{equation}
Thus, the spatial subsampling operation has be applied more
restrictively for highly eccentric/anisotropic spatial smoothing
kernels compared to more isotropic kernels.

\begin{figure}[hbtp]
  \begin{center}
    {\footnotesize\em Equivalent affine hybrid pyramid kernels for $K = 3$}
    
    \medskip

    \begin{tabular}{cccccc}
     \hspace{-4mm}
     \includegraphics[width=0.15\textwidth]{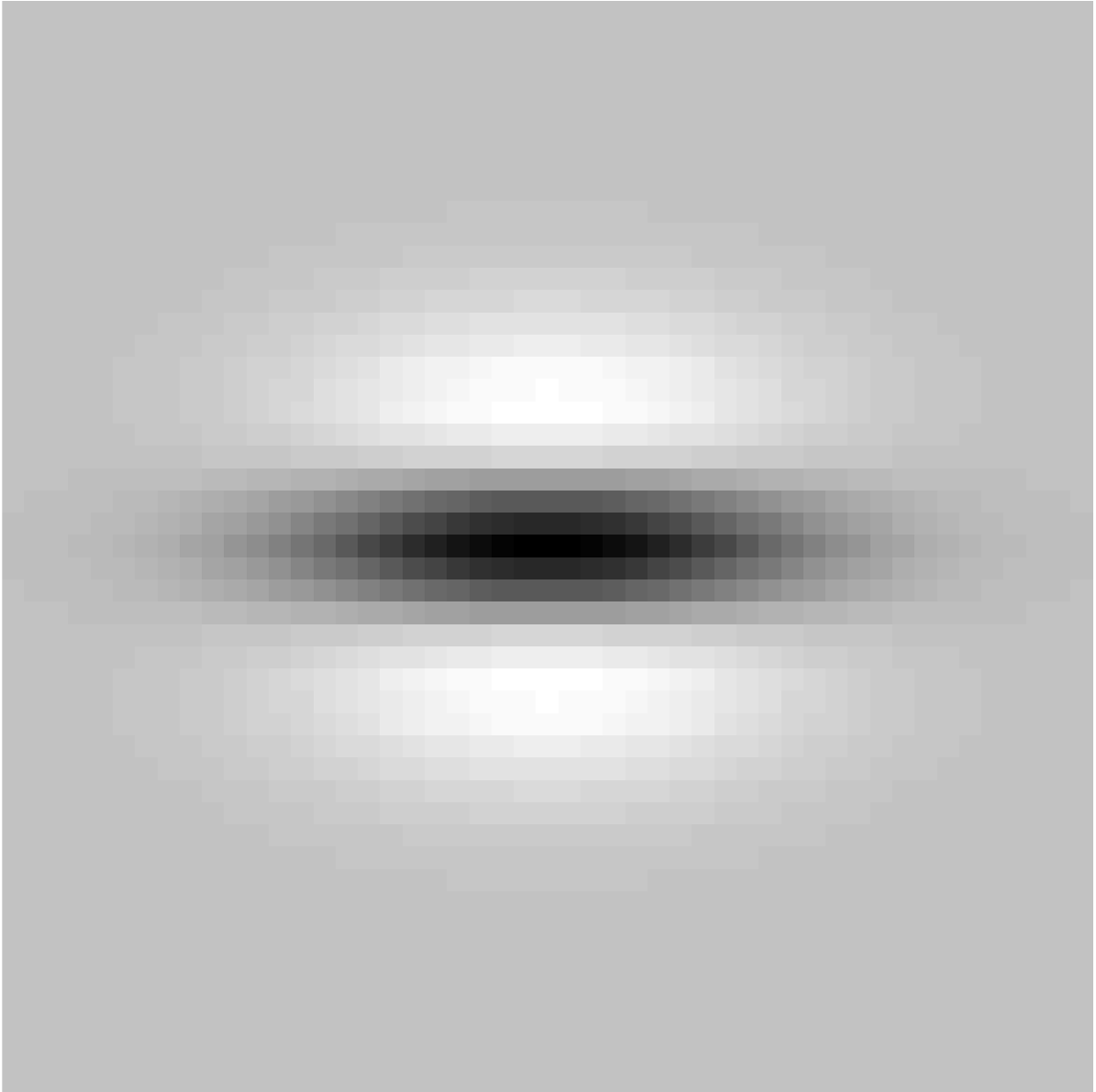} \hspace{-4mm} &
     \includegraphics[width=0.15\textwidth]{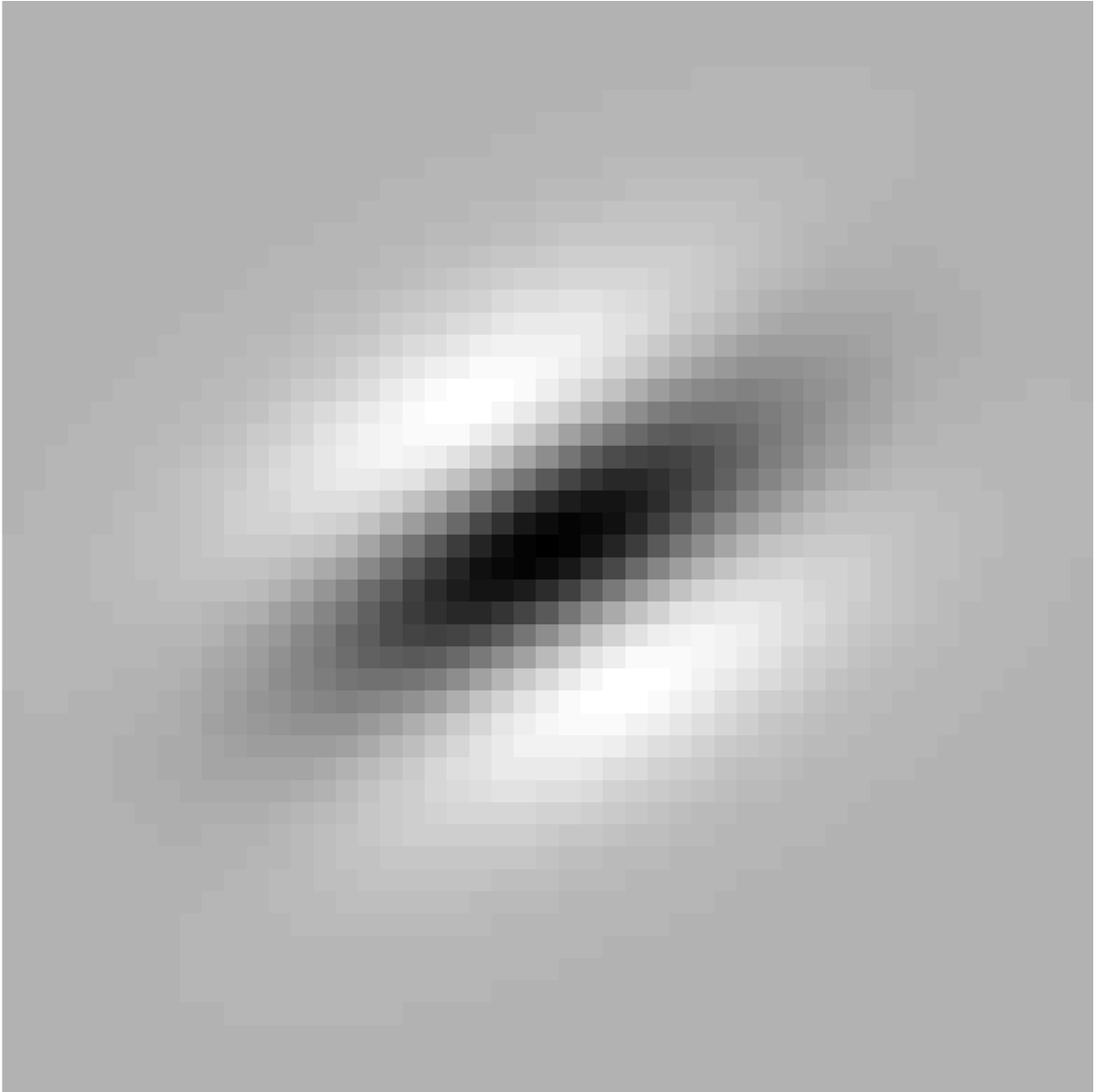} \hspace{-4mm} &
     \includegraphics[width=0.15\textwidth]{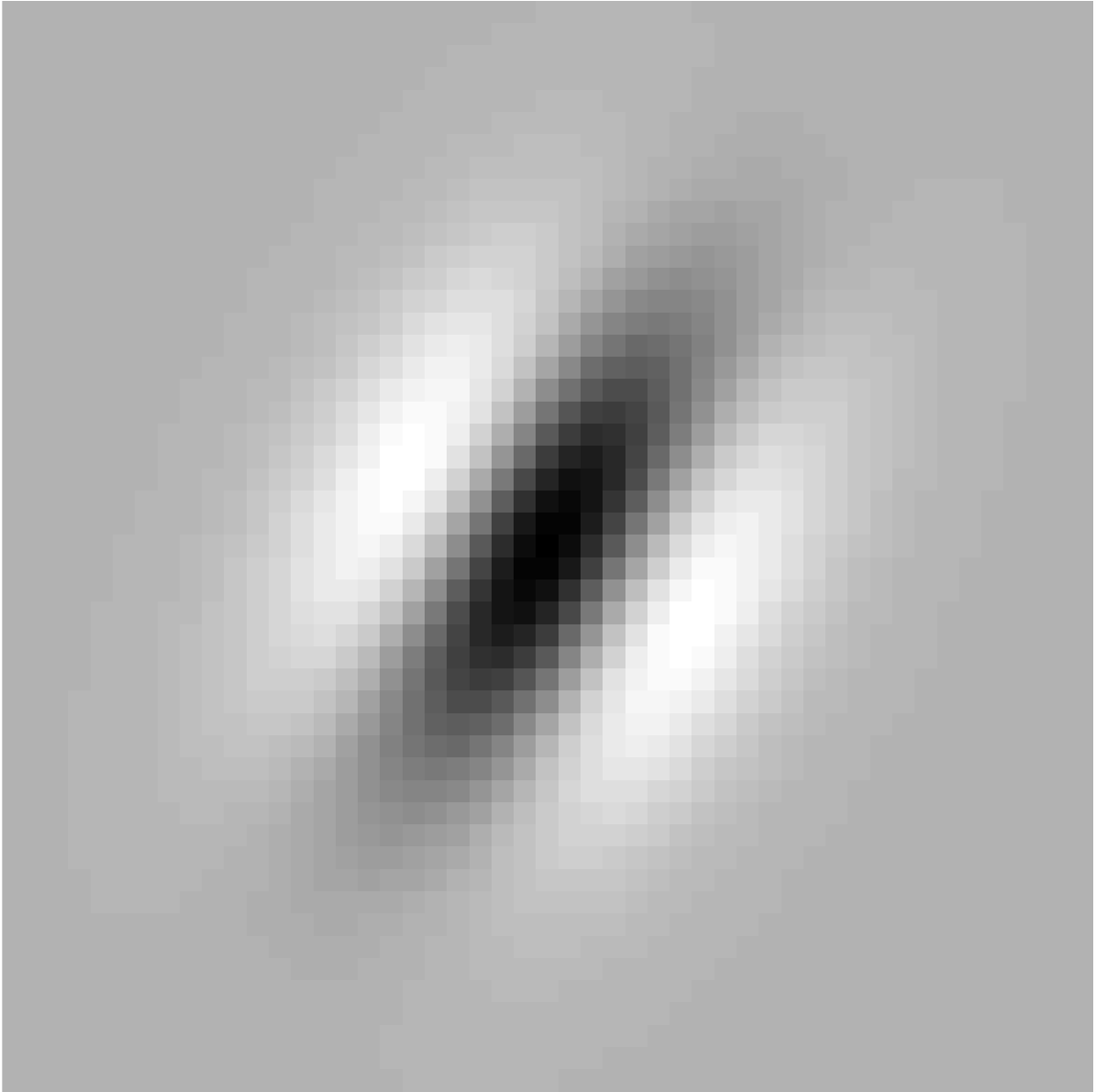} \hspace{-4mm} &
     \includegraphics[width=0.15\textwidth]{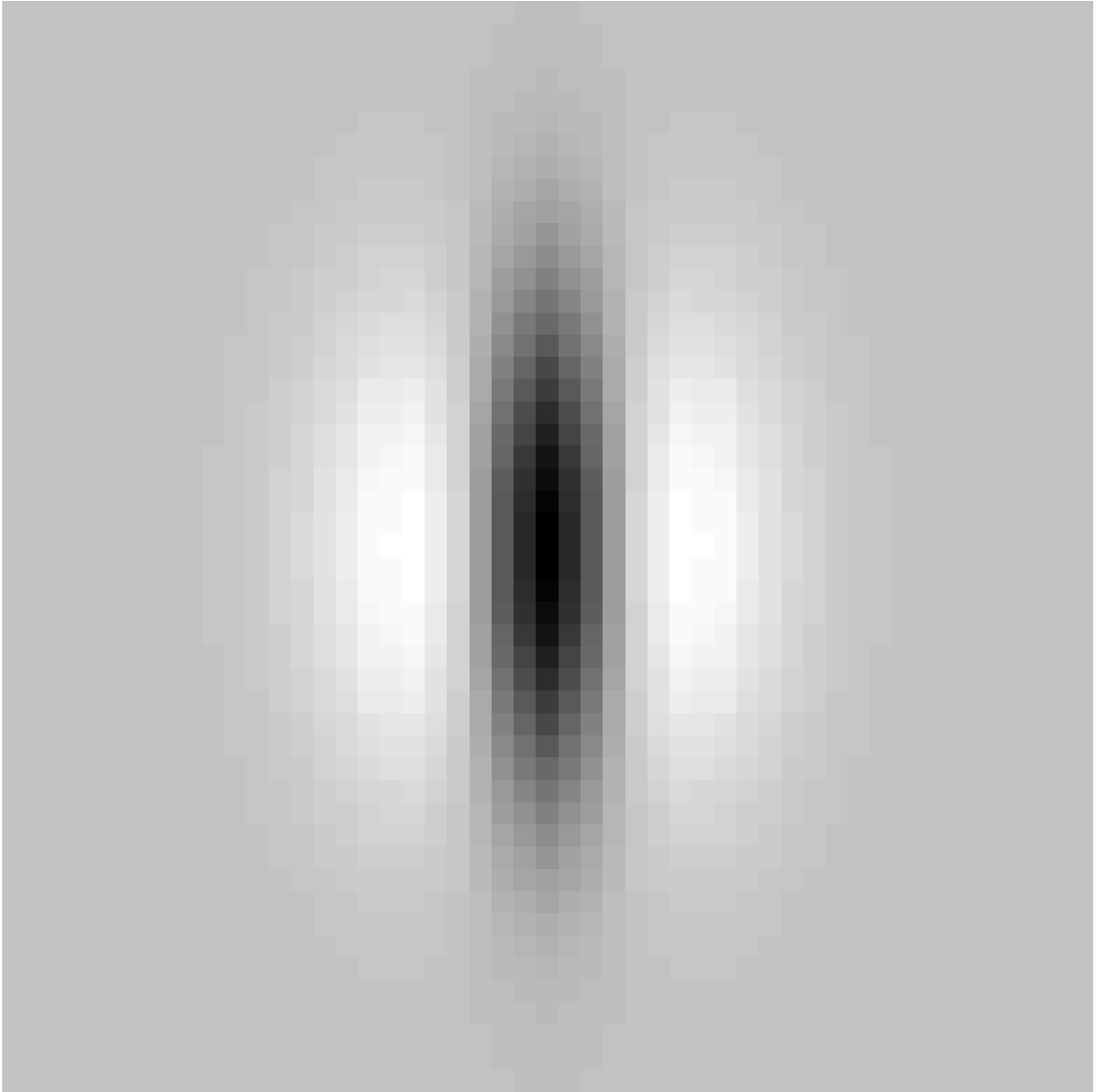} \hspace{-4mm} &
     \includegraphics[width=0.15\textwidth]{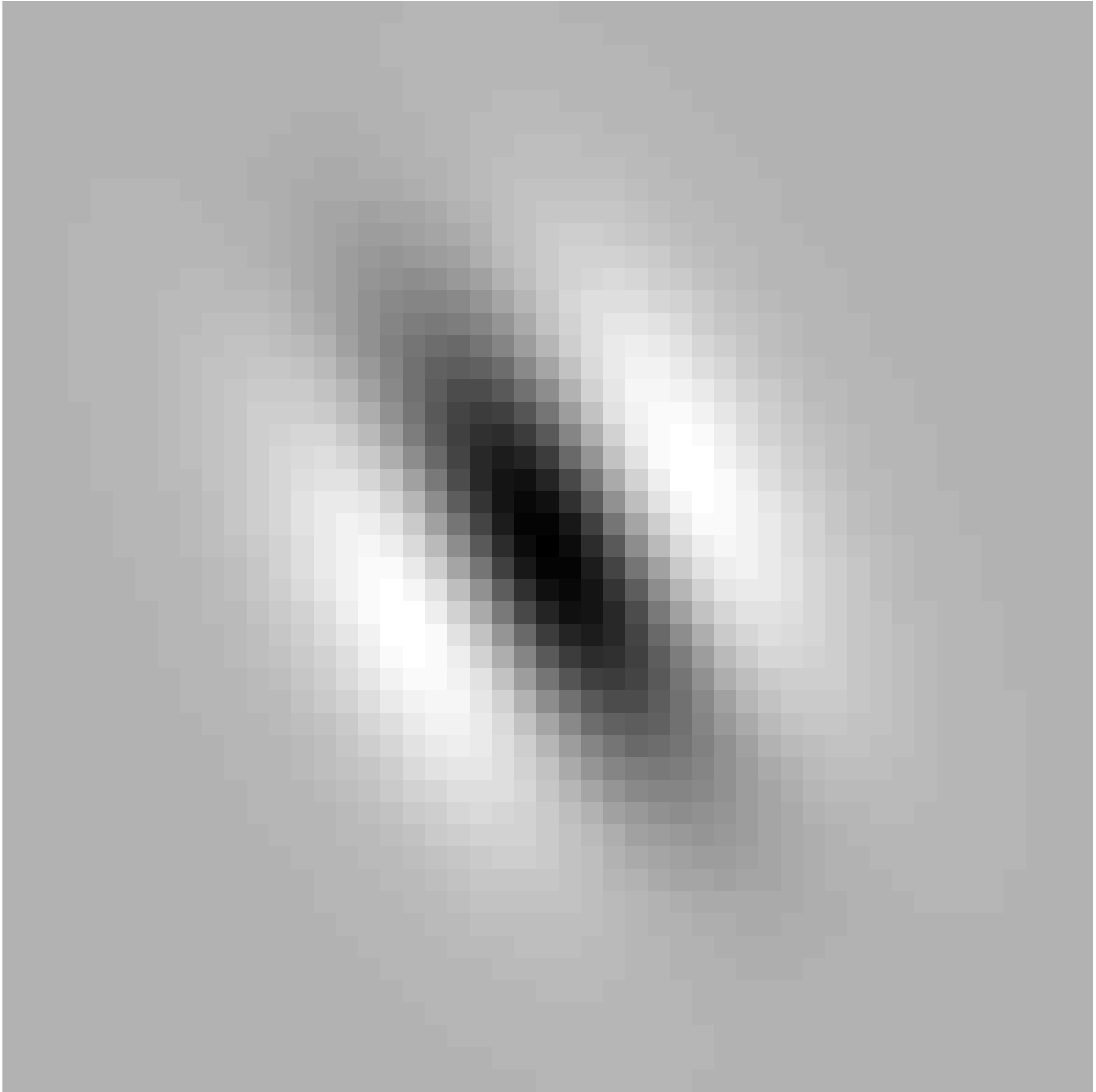} \hspace{-4mm} &
     \includegraphics[width=0.15\textwidth]{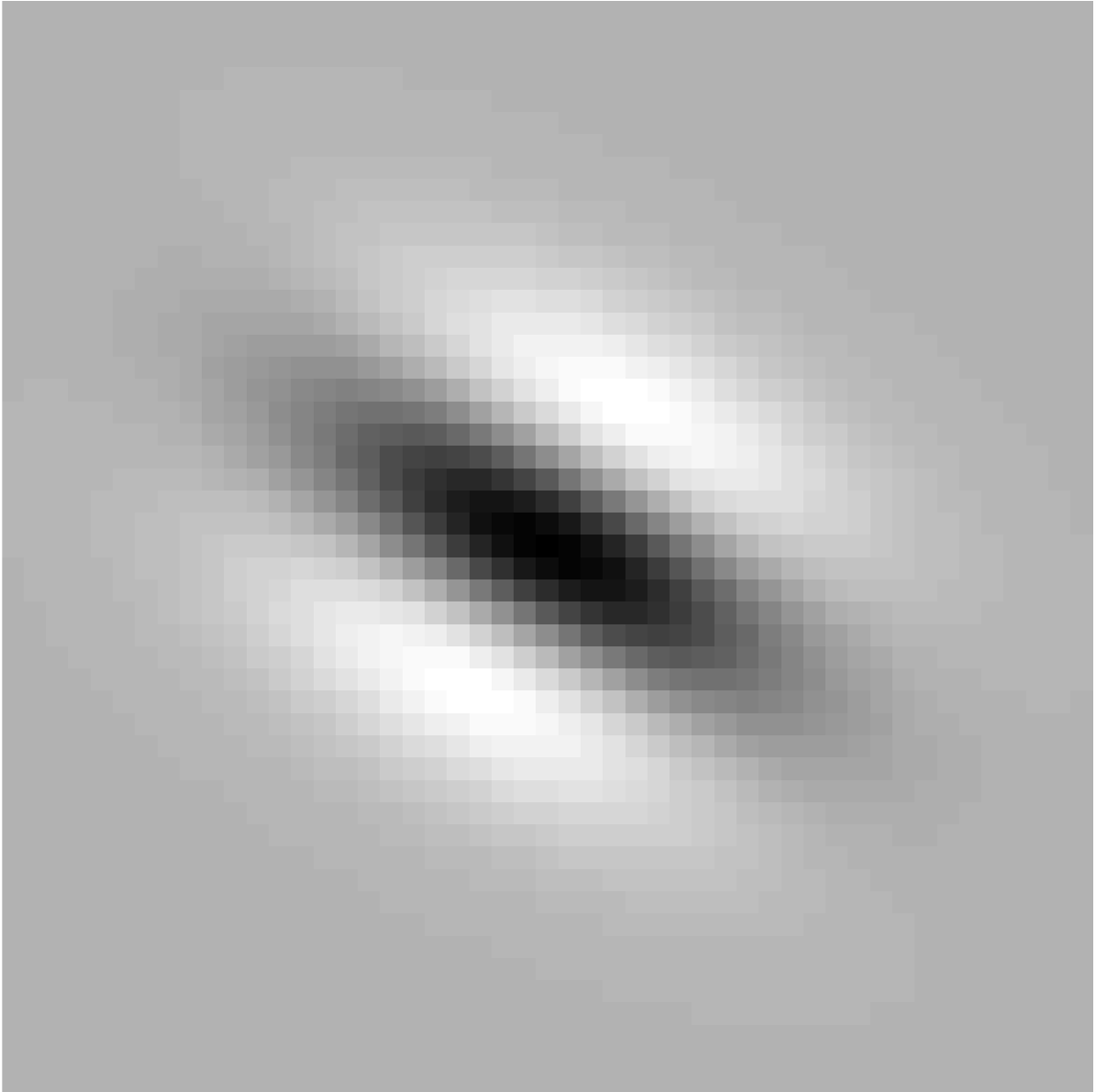} \hspace{-4mm} \\
     \hspace{-4mm}
    \includegraphics[width=0.15\textwidth]{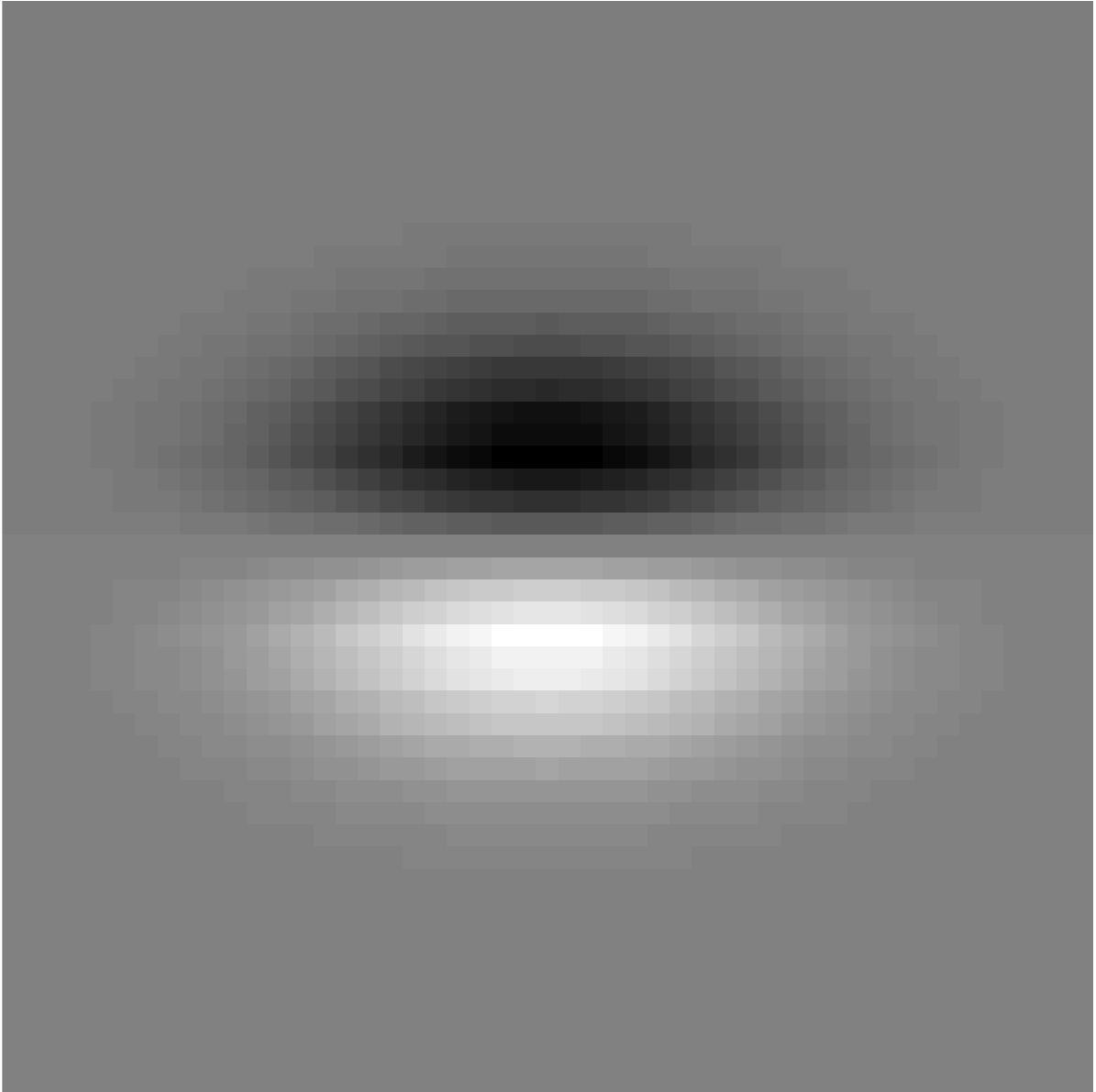} \hspace{-4mm} &
     \includegraphics[width=0.15\textwidth]{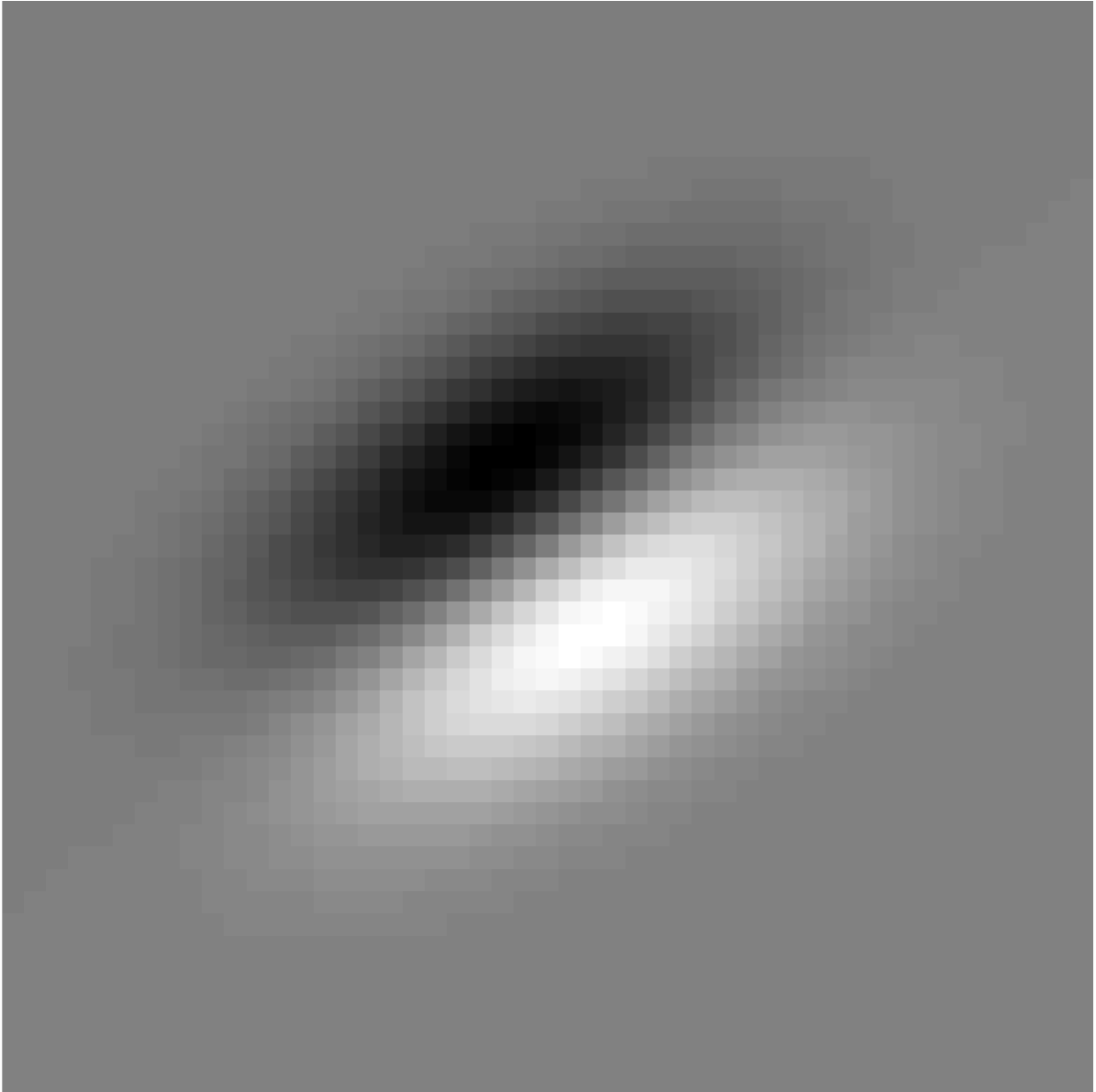} \hspace{-4mm} &
     \includegraphics[width=0.15\textwidth]{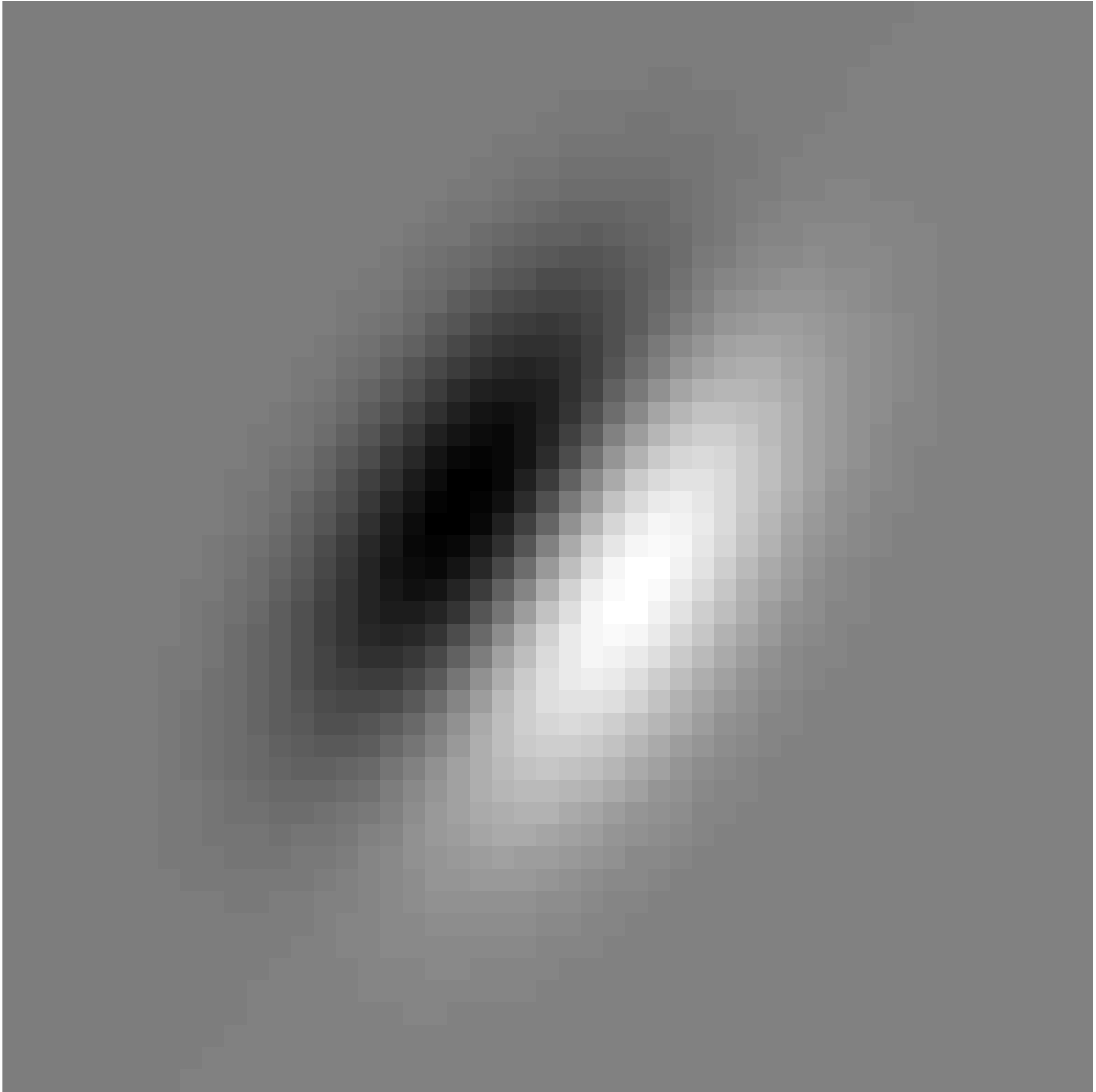} \hspace{-4mm} &
     \includegraphics[width=0.15\textwidth]{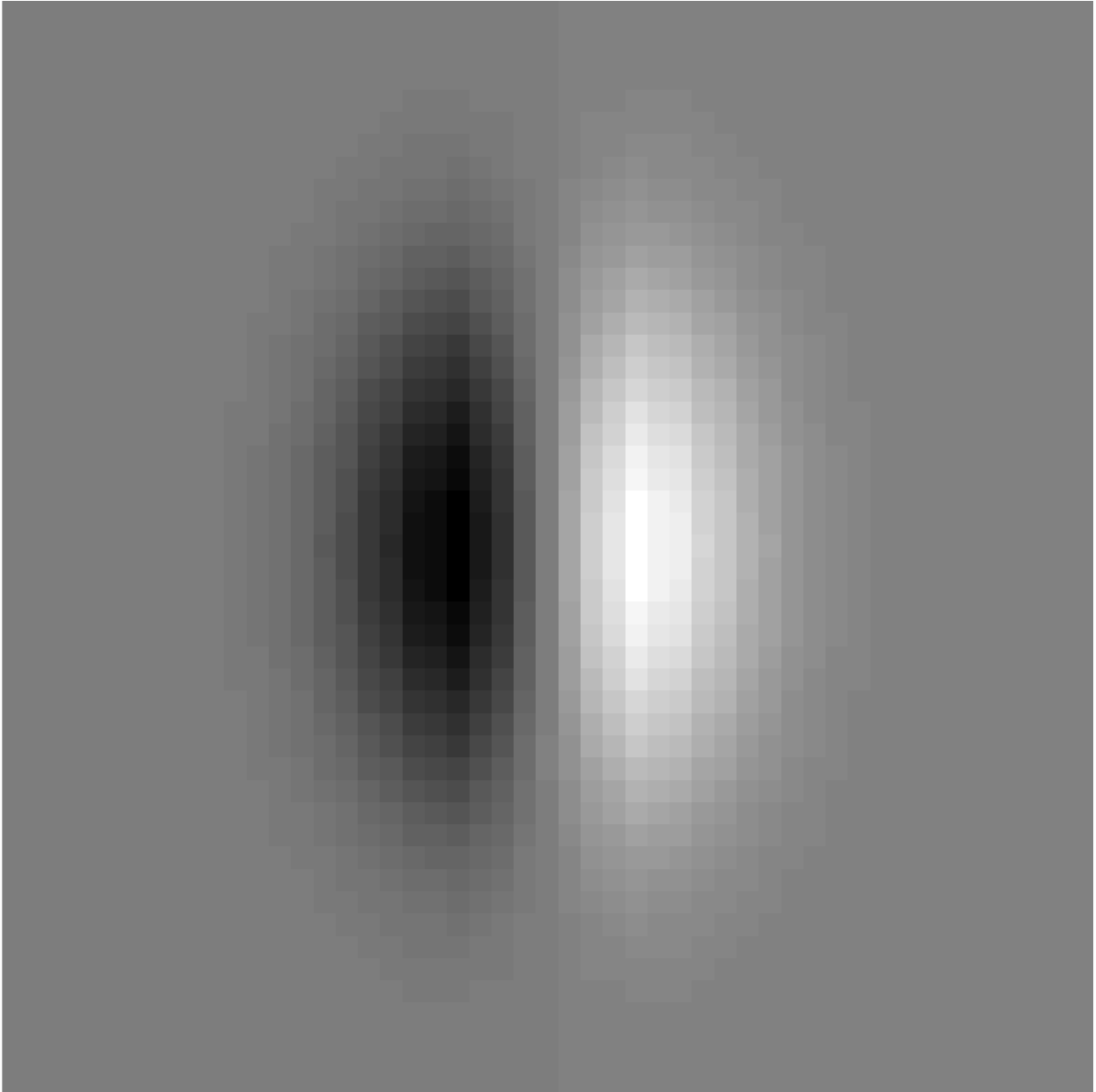} \hspace{-4mm} &
     \includegraphics[width=0.15\textwidth]{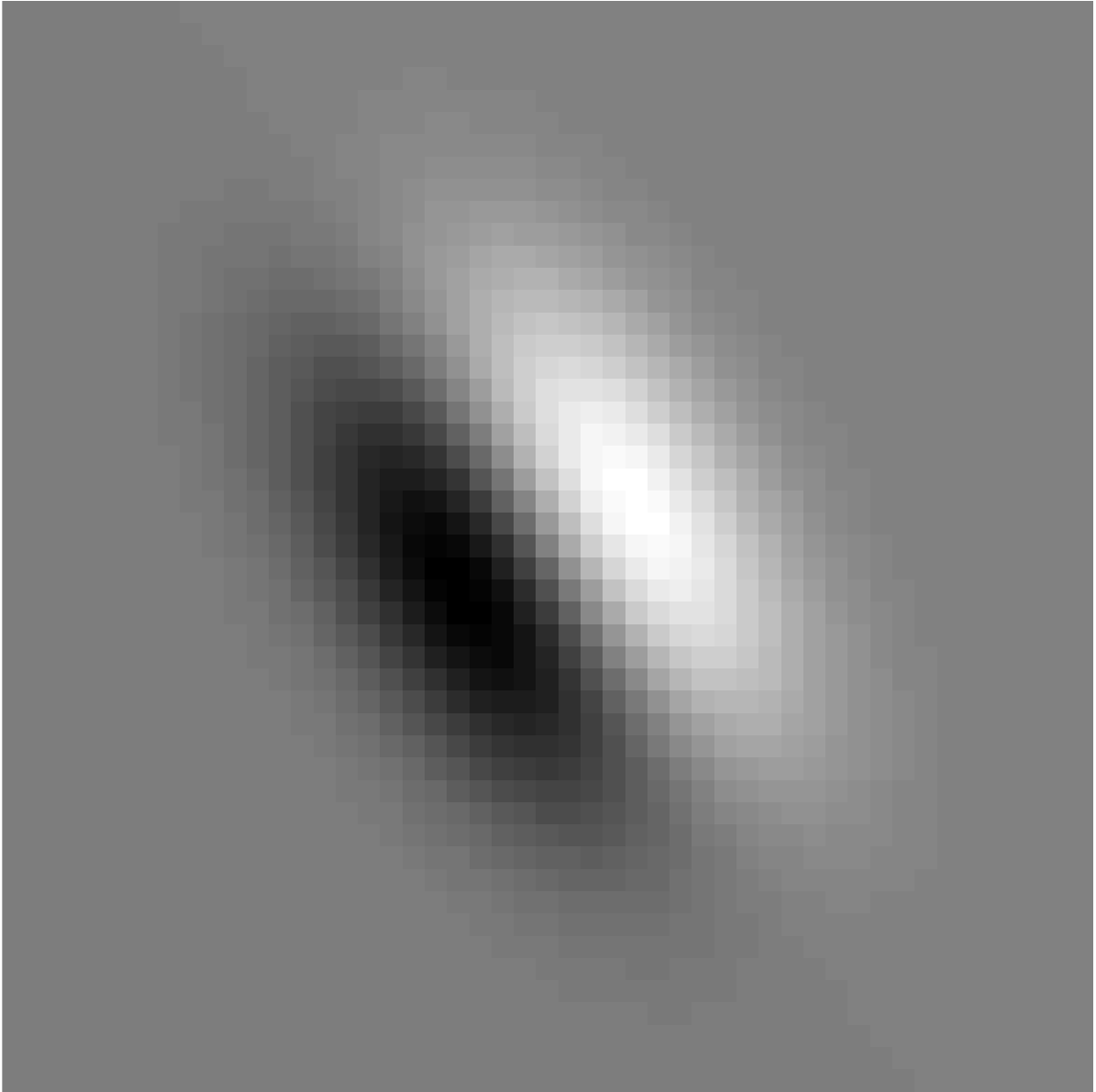} \hspace{-4mm} &
     \includegraphics[width=0.15\textwidth]{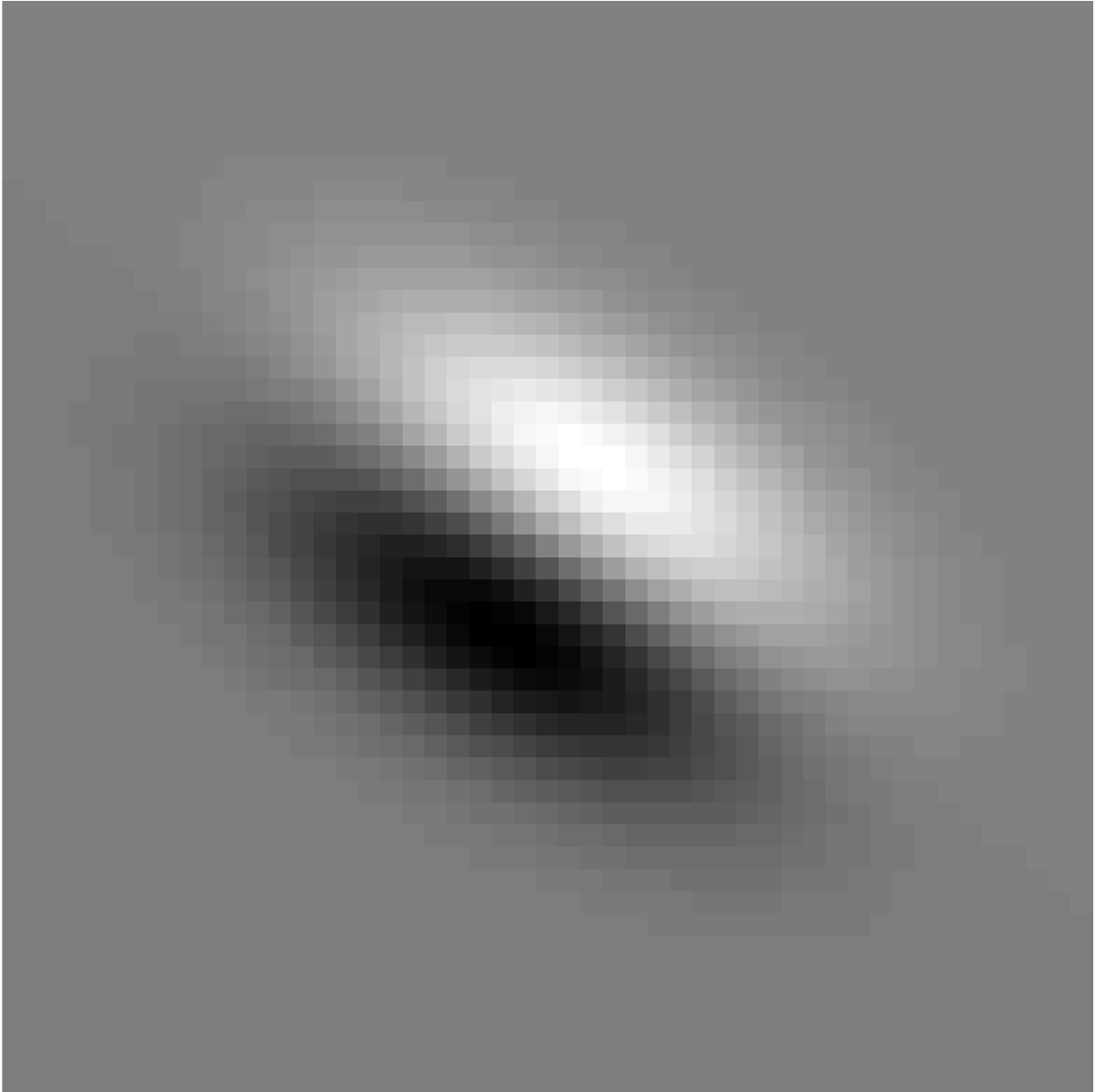} \hspace{-4mm} \\
     \hspace{-4mm}
   \includegraphics[width=0.15\textwidth]{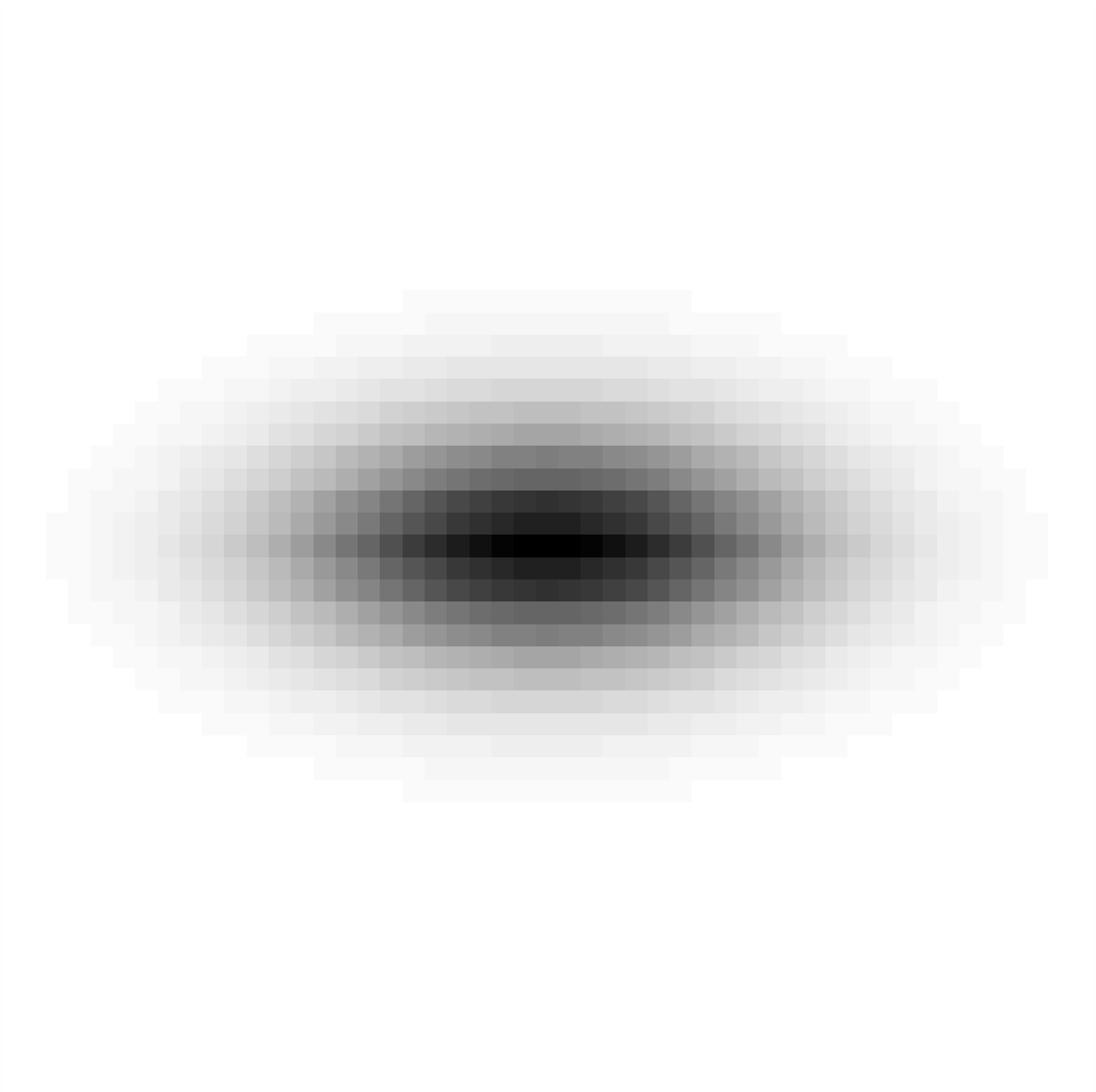} \hspace{-4mm} &
     \includegraphics[width=0.15\textwidth]{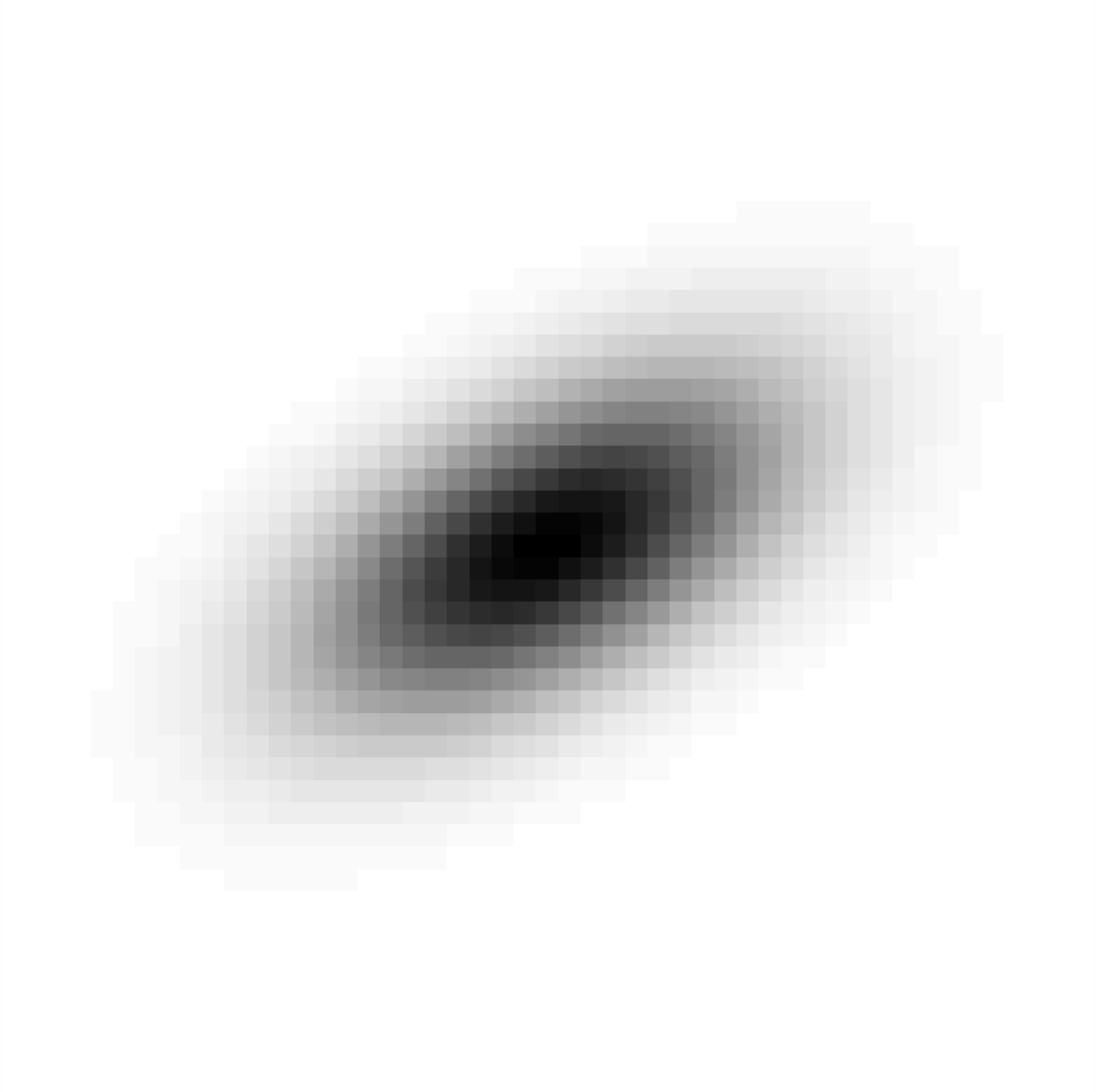} \hspace{-4mm} &
     \includegraphics[width=0.15\textwidth]{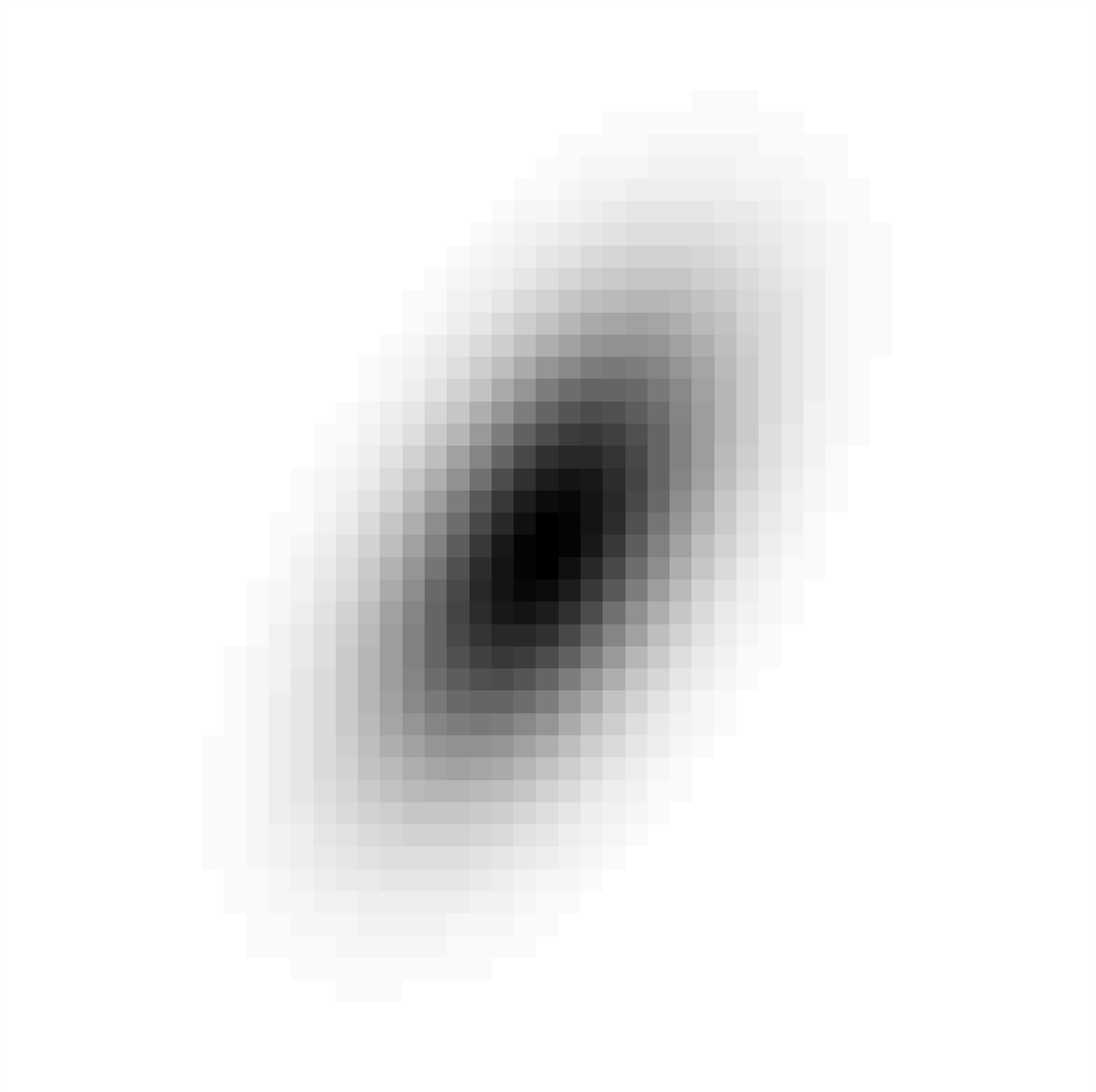} \hspace{-4mm} &
     \includegraphics[width=0.15\textwidth]{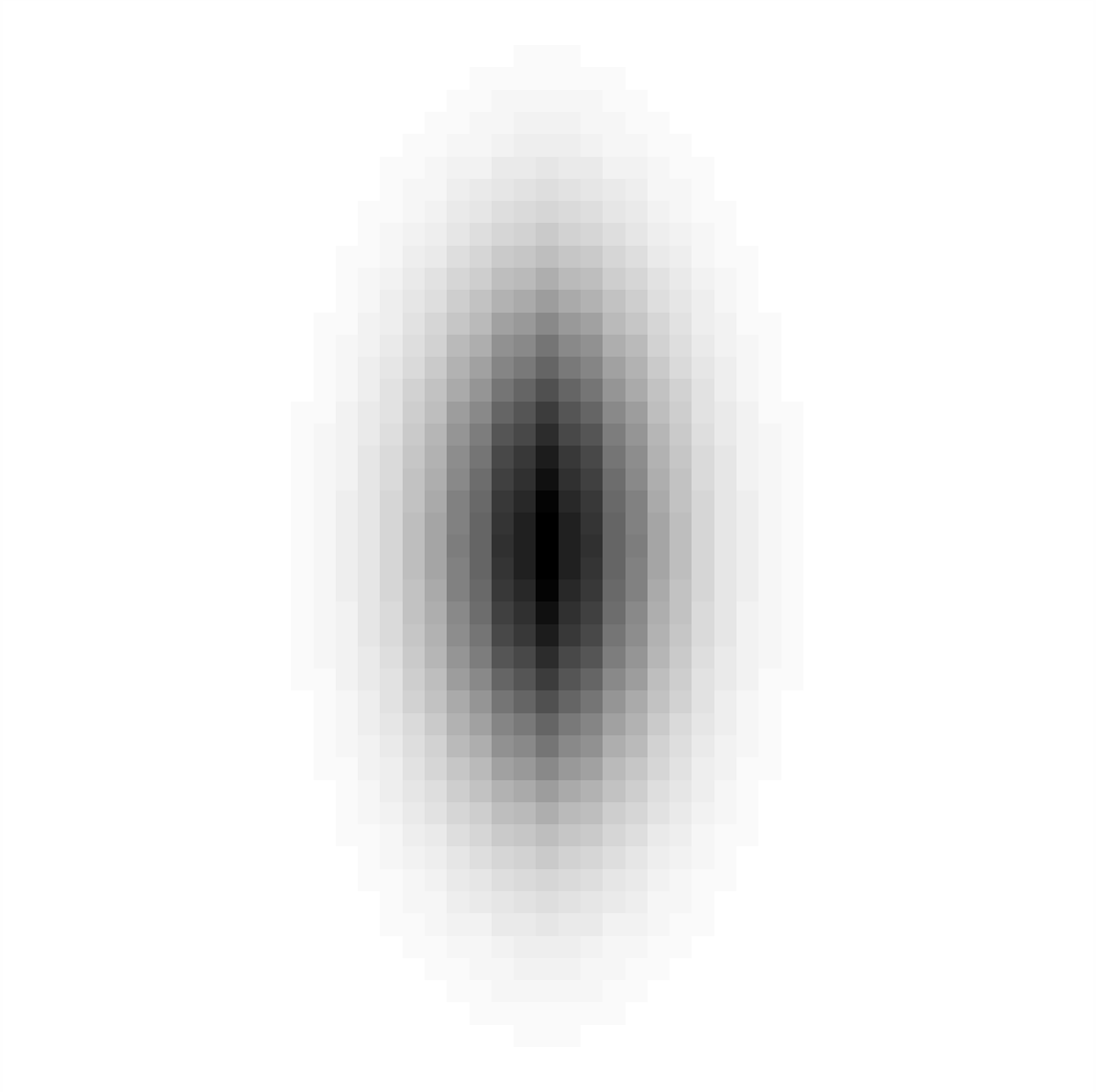} \hspace{-4mm} &
     \includegraphics[width=0.15\textwidth]{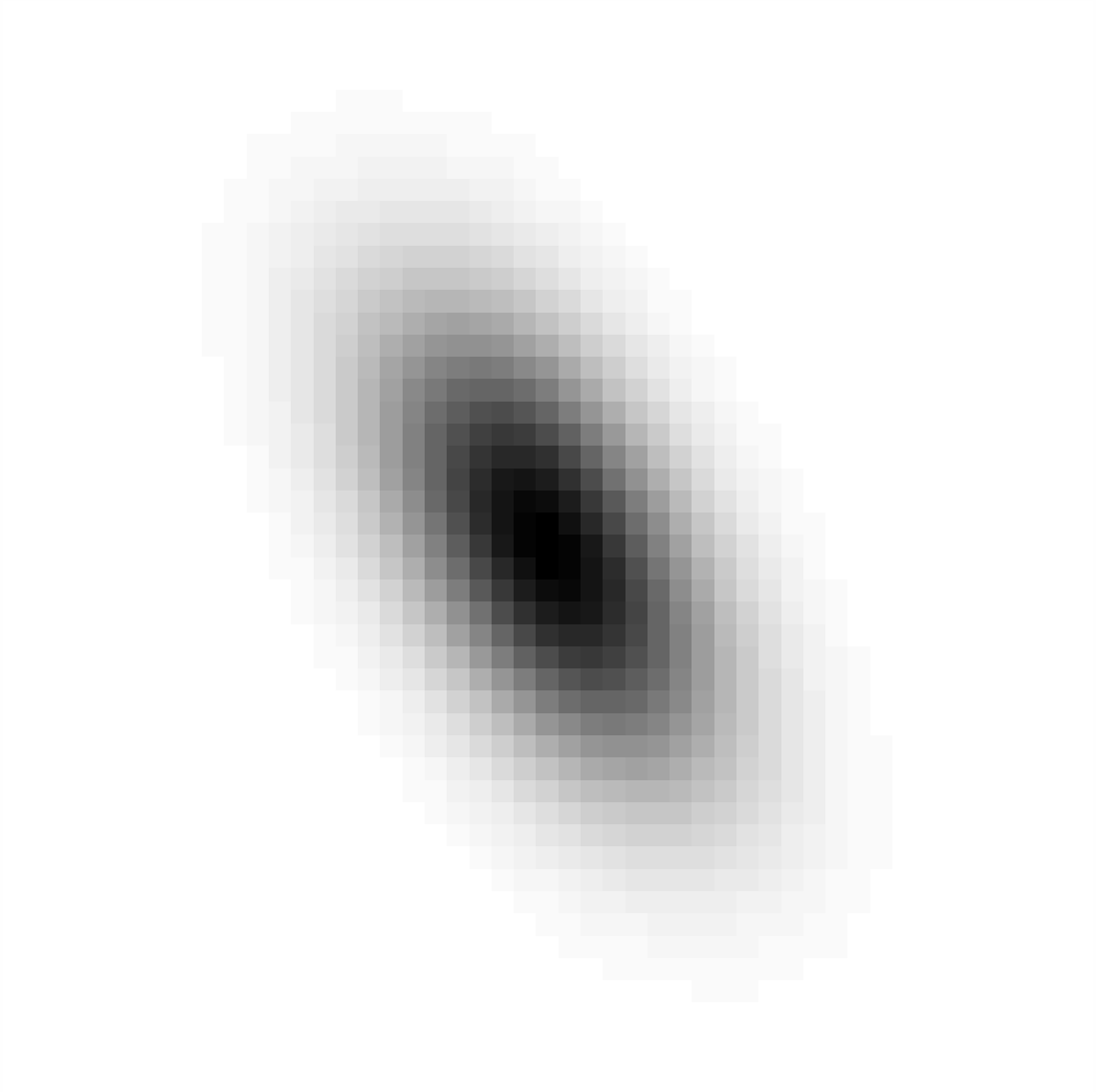} \hspace{-4mm} &
     \includegraphics[width=0.15\textwidth]{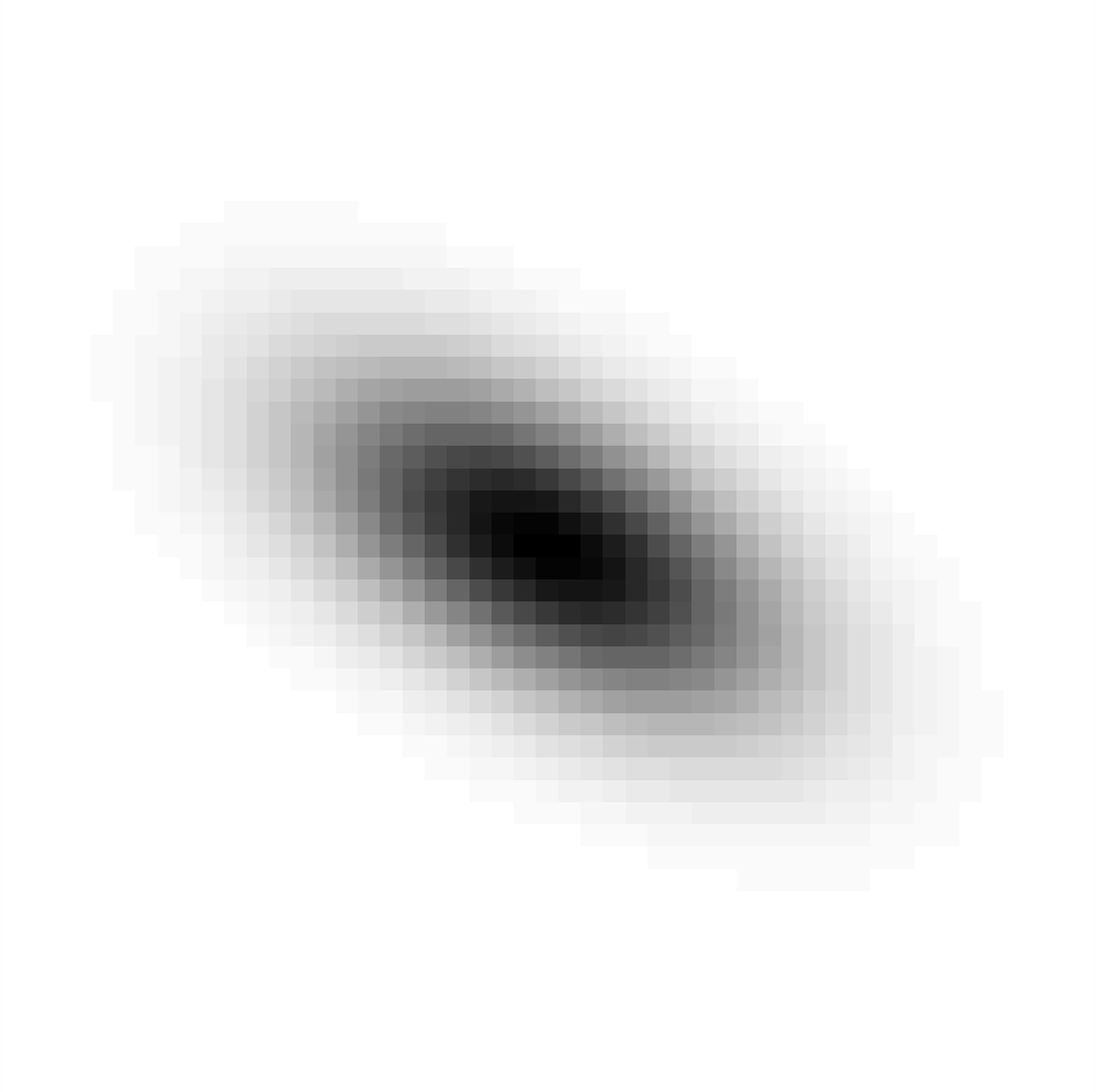} \hspace{-4mm} \\
    \end{tabular} 
  \end{center}
  \vspace{-4mm}
  \caption{Examples of equivalent convolution kernels 
    kernels $\kappa^{(l, k)}(x, y;\; \Sigma)$ with their directional
    derivative approximation kernels $\kappa^{(l, k)}_{\orth \varphi}(x, y;\; \Sigma)$ and
    $\kappa^{(l, k)}_{\orth \varphi \orth \varphi}(x, y;\; \Sigma)$ 
   up to order two in the two-dimensional case with the steepness of
   the pyramid determined by $K = 3$,
   here at resolution level $l = 2$ and iteration level $k = 4$
   corresponding to $\lambda_1 = 62$ for $\lambda_2/\lambda_1=1/4$ and
    $\alpha = 0, \pi/6, \pi/3, \pi/2, 2\pi/3, 5\pi/6$.
   (Horizontal axis: $x \in [-24, 24]$. Vertical axis: $y \in [-24, 24]$.)}
  \label{fig-equiv-aff-hybr-pyr-dir-ders-l2-k4-K3-lambda1-62}

  \bigskip
  \bigskip

  \begin{center}
     {\footnotesize\em Equivalent spatio-chromatic affine hybrid pyramid kernels for $K = 5$}
    
    \medskip

   \begin{tabular}{cccccc}
     \hspace{-4mm}
     \includegraphics[width=0.15\textwidth]{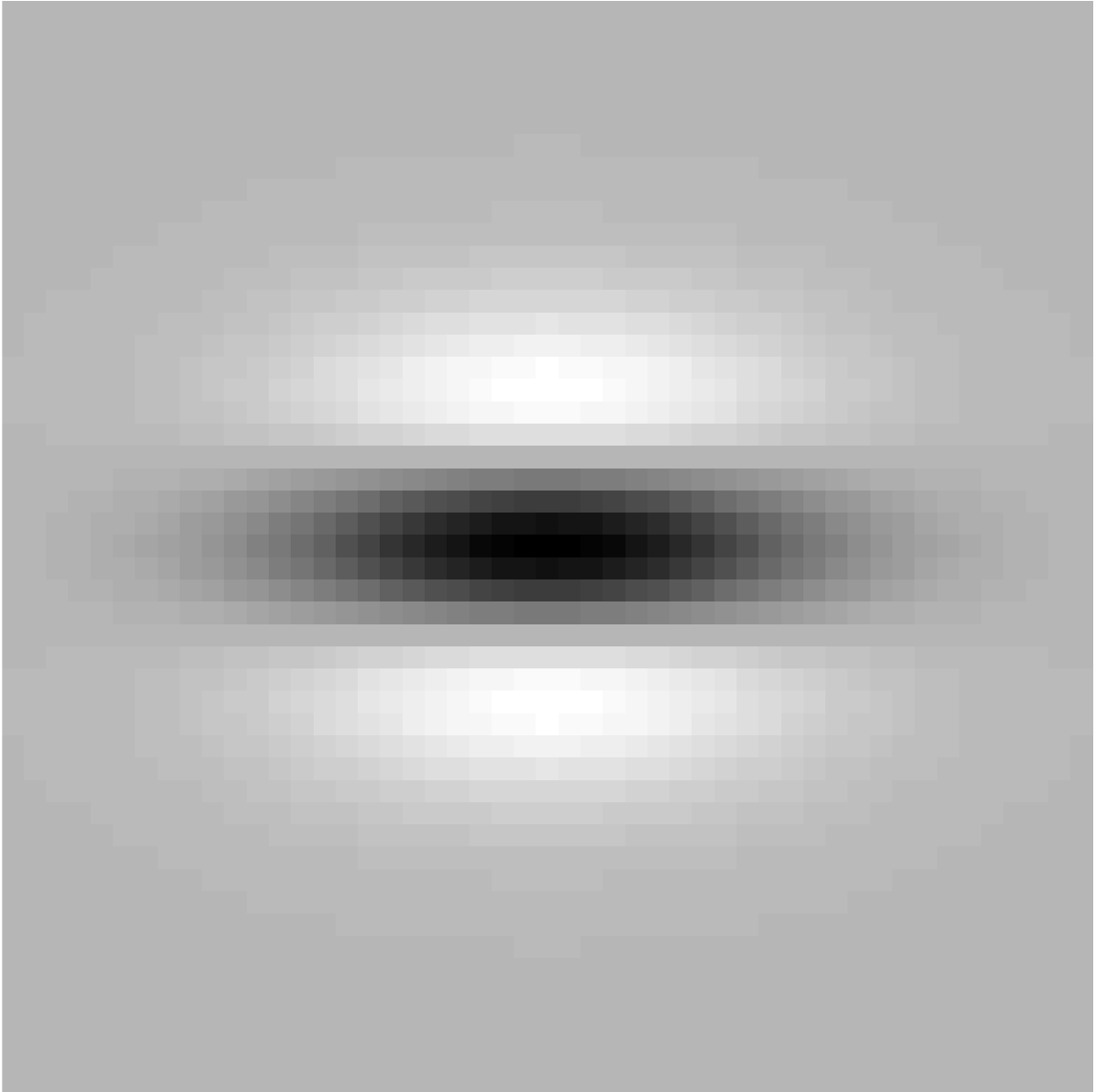} \hspace{-4mm} &
     \includegraphics[width=0.15\textwidth]{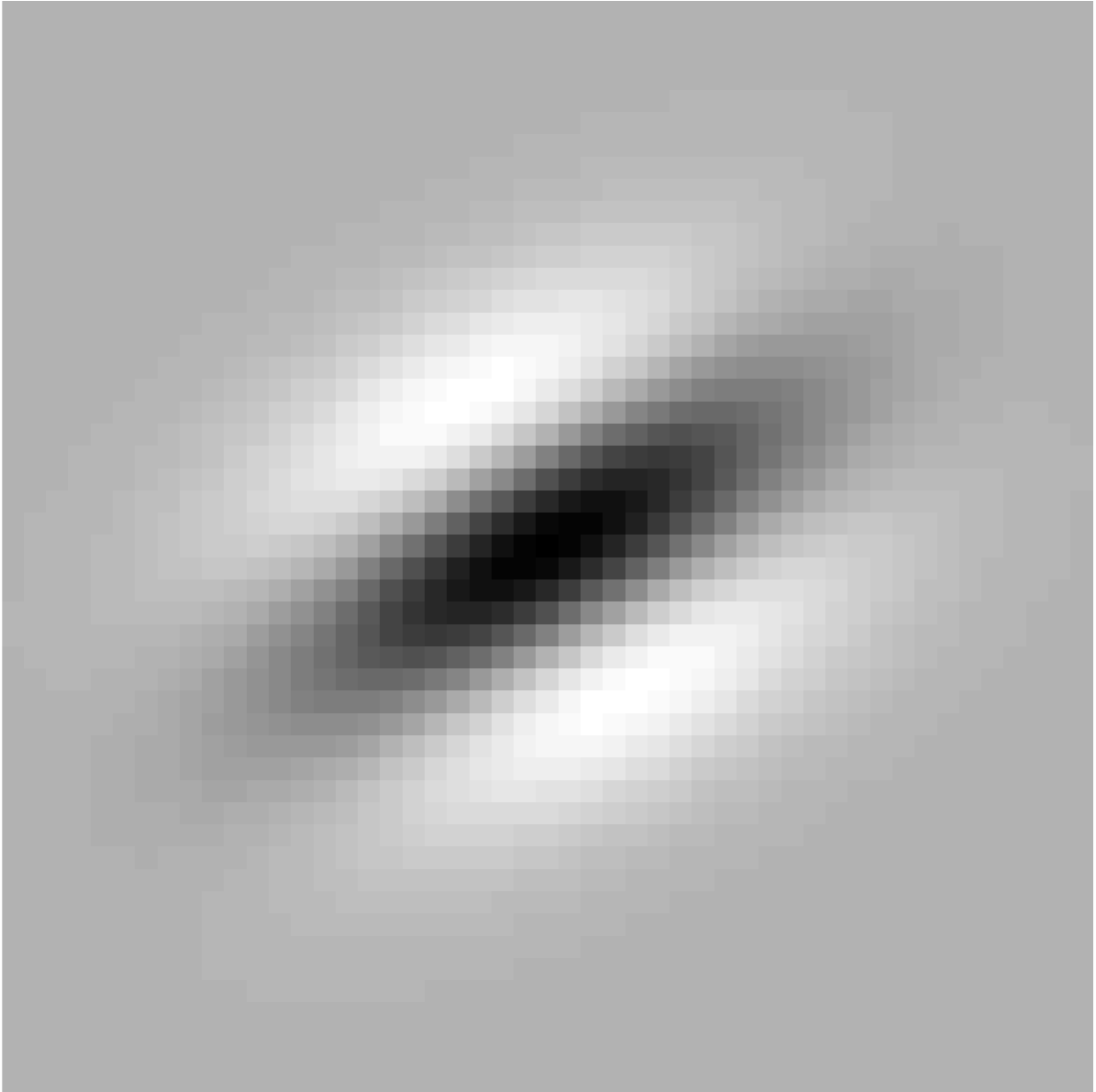} \hspace{-4mm} &
     \includegraphics[width=0.15\textwidth]{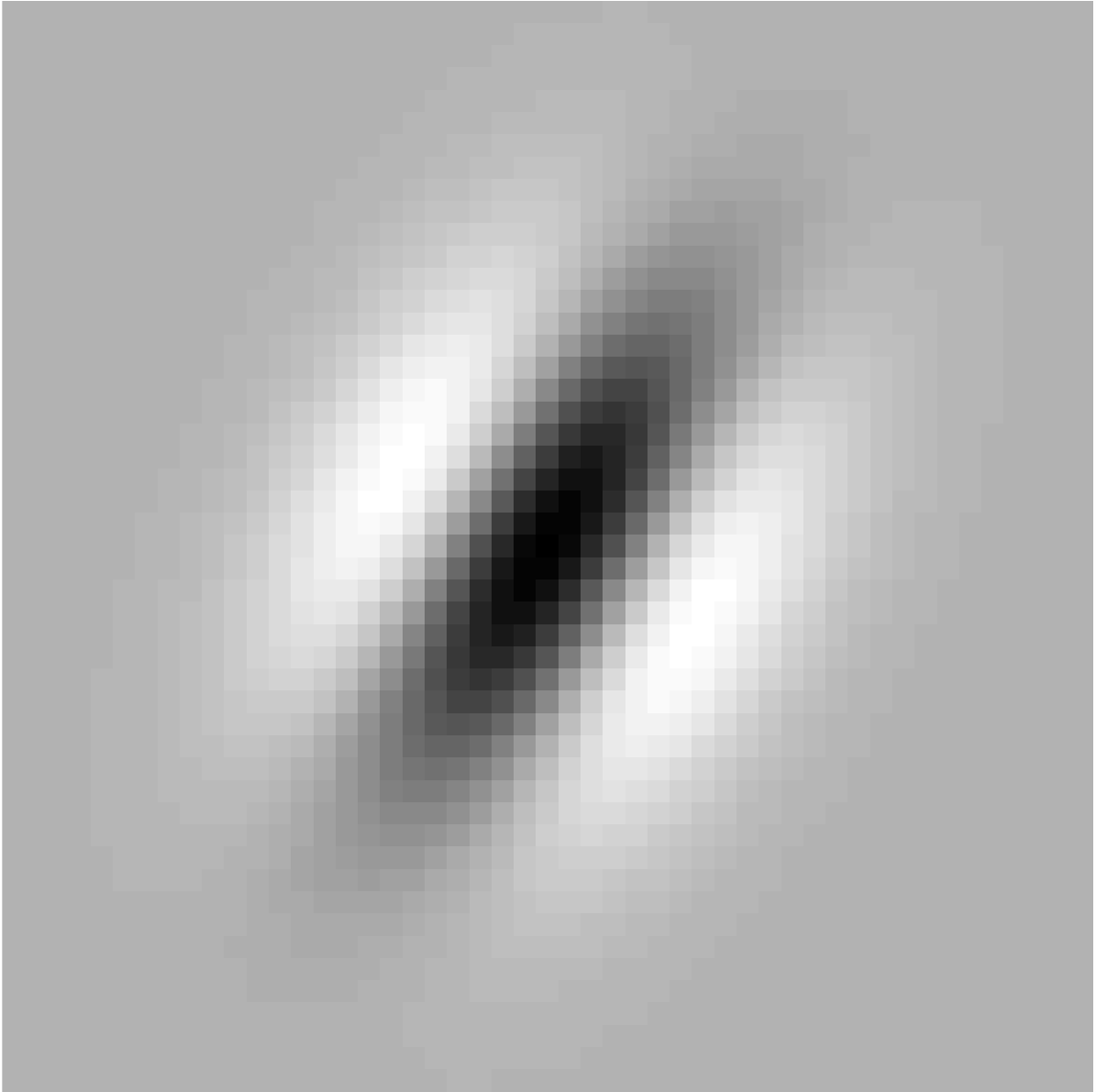} \hspace{-4mm} &
     \includegraphics[width=0.15\textwidth]{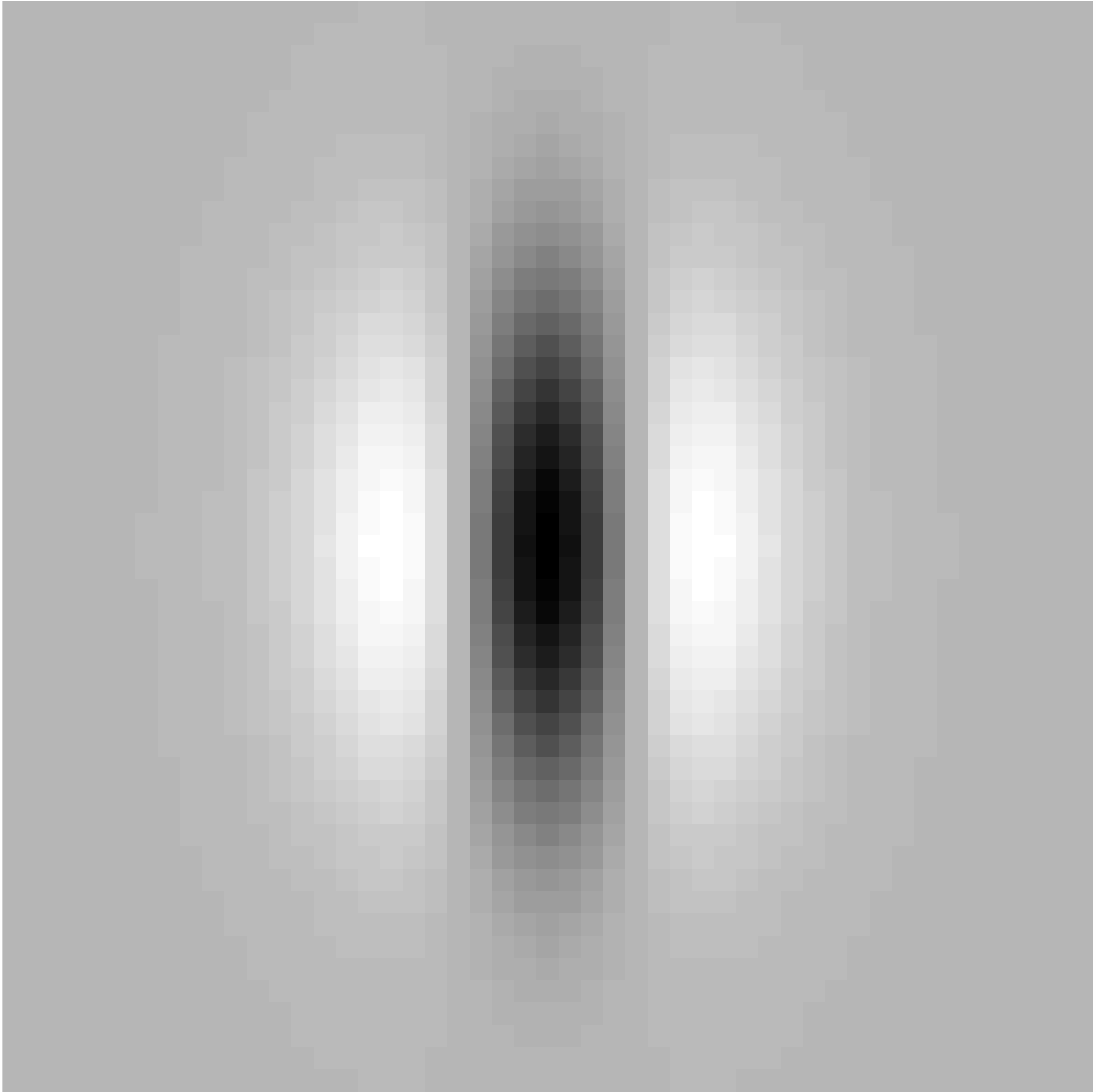} \hspace{-4mm} &
     \includegraphics[width=0.15\textwidth]{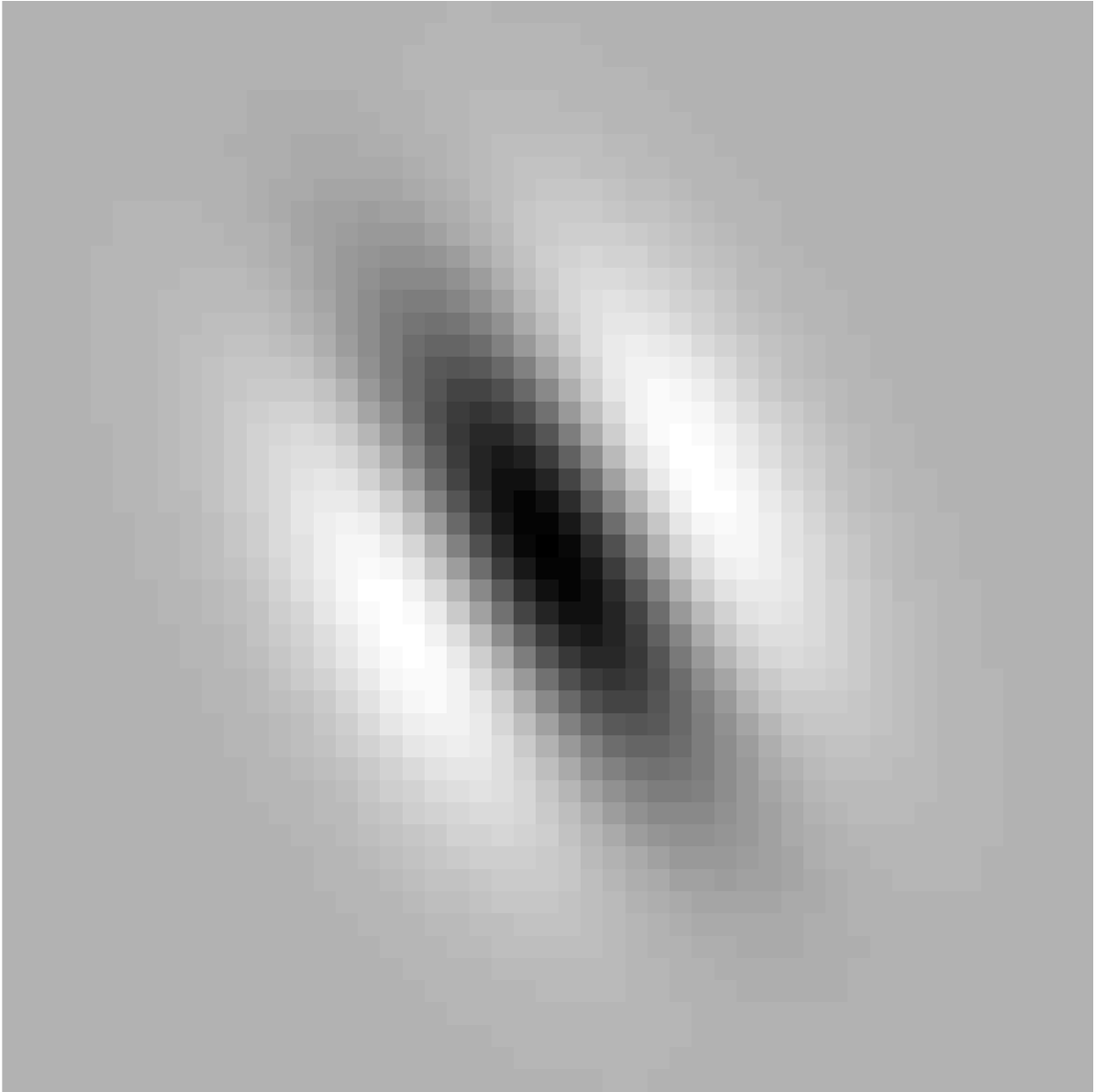} \hspace{-4mm} &
     \includegraphics[width=0.15\textwidth]{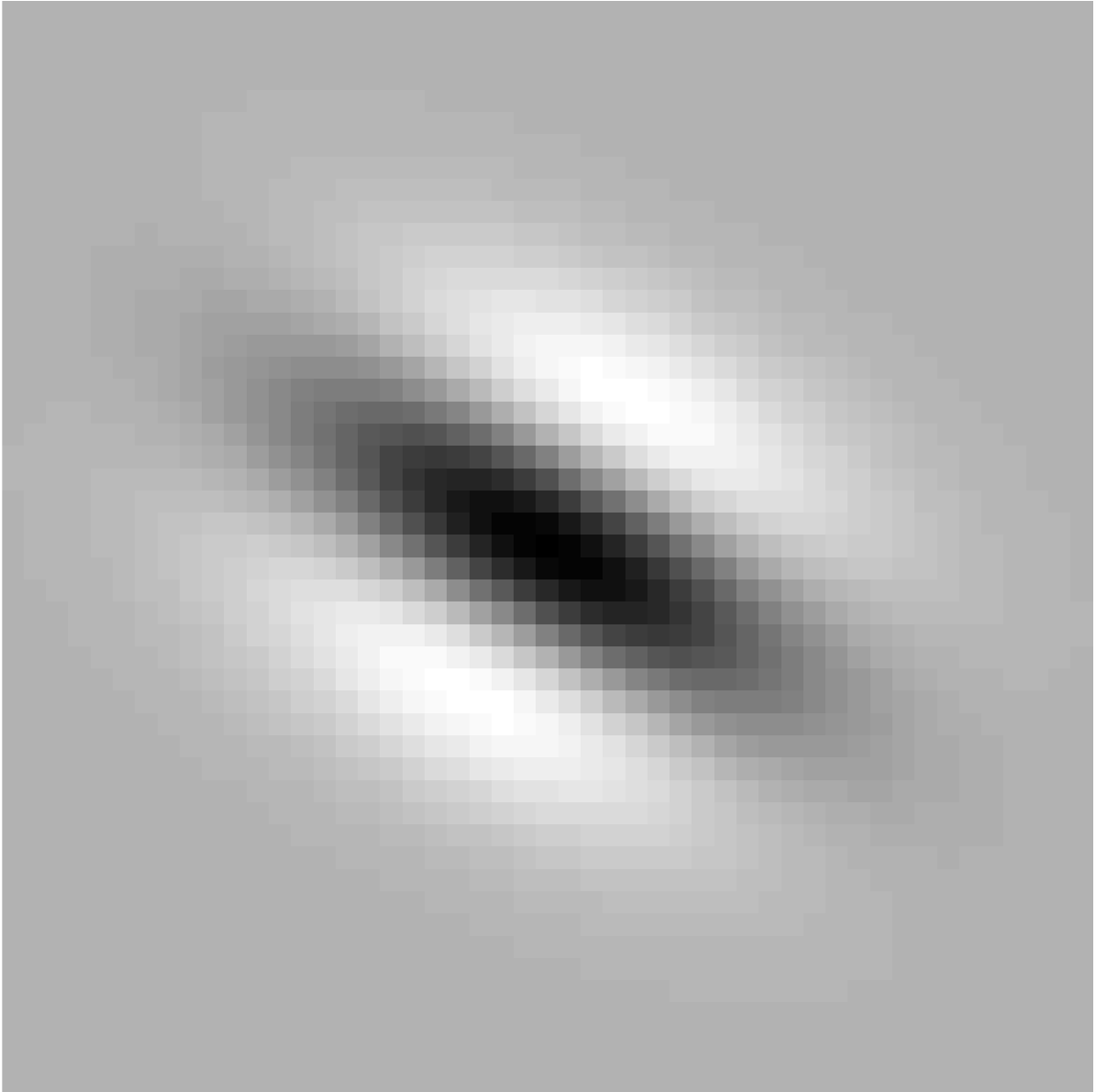} \hspace{-4mm} \\
     \hspace{-4mm}
    \includegraphics[width=0.15\textwidth]{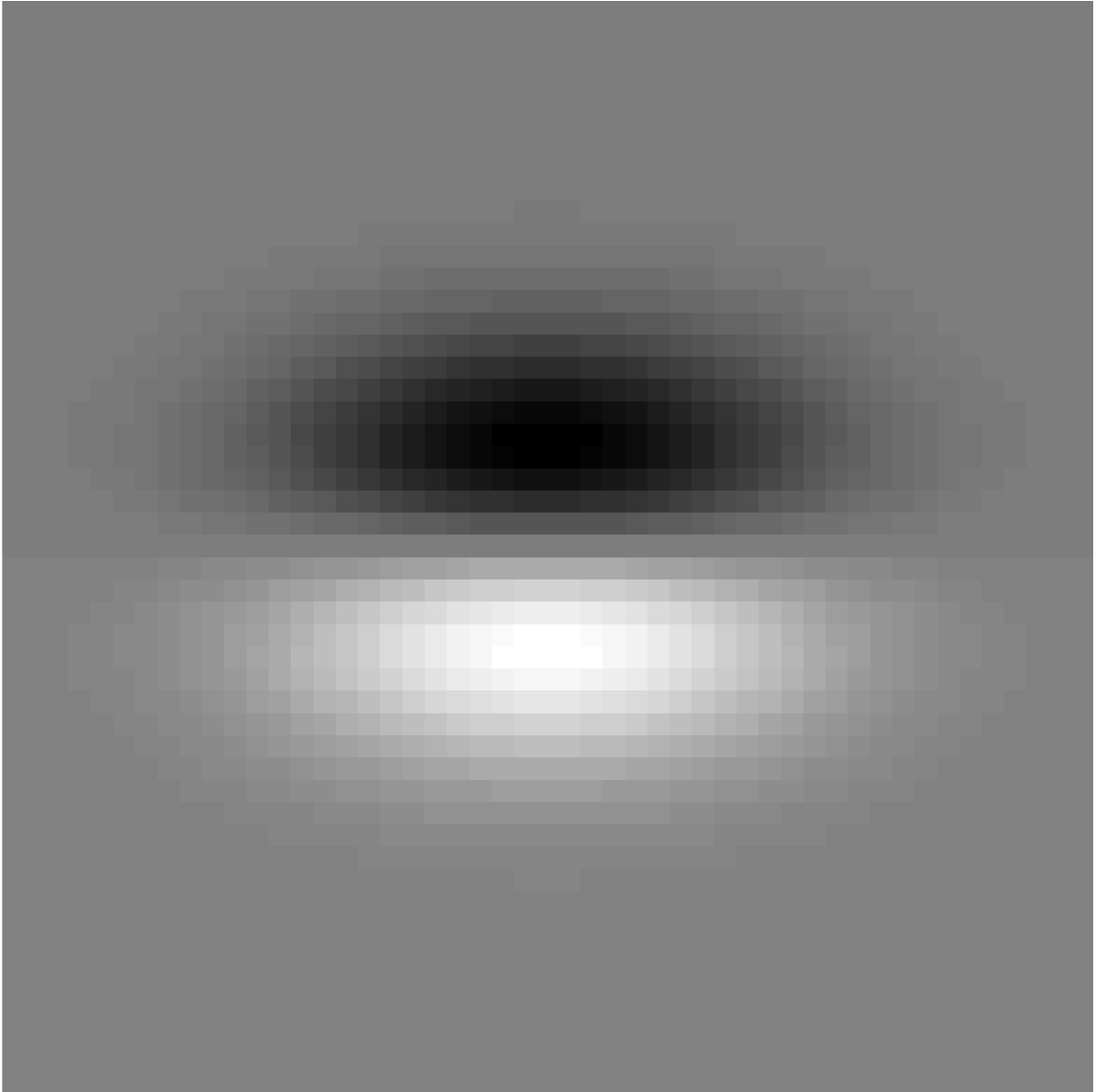} \hspace{-4mm} &
     \includegraphics[width=0.15\textwidth]{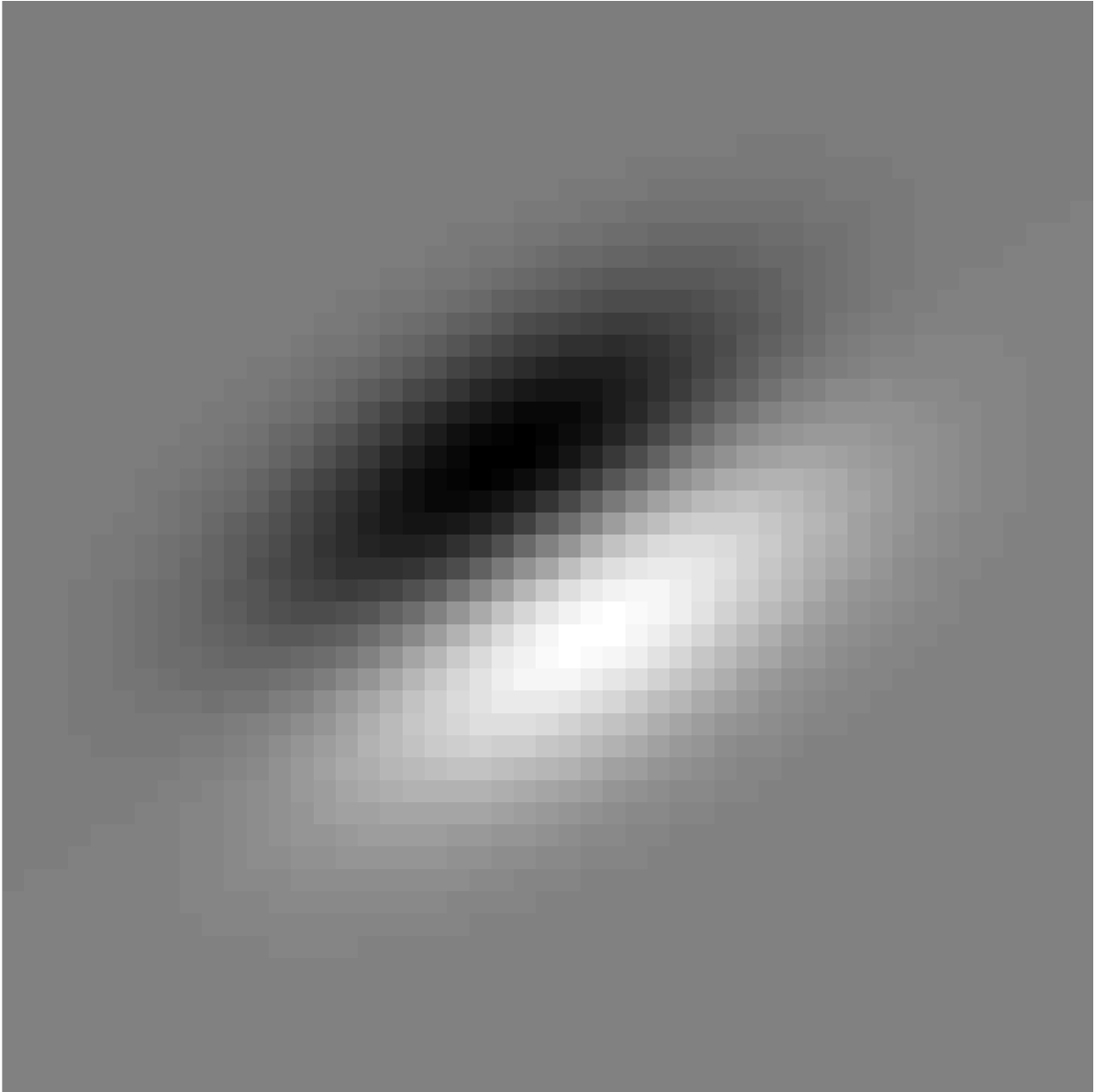} \hspace{-4mm} &
     \includegraphics[width=0.15\textwidth]{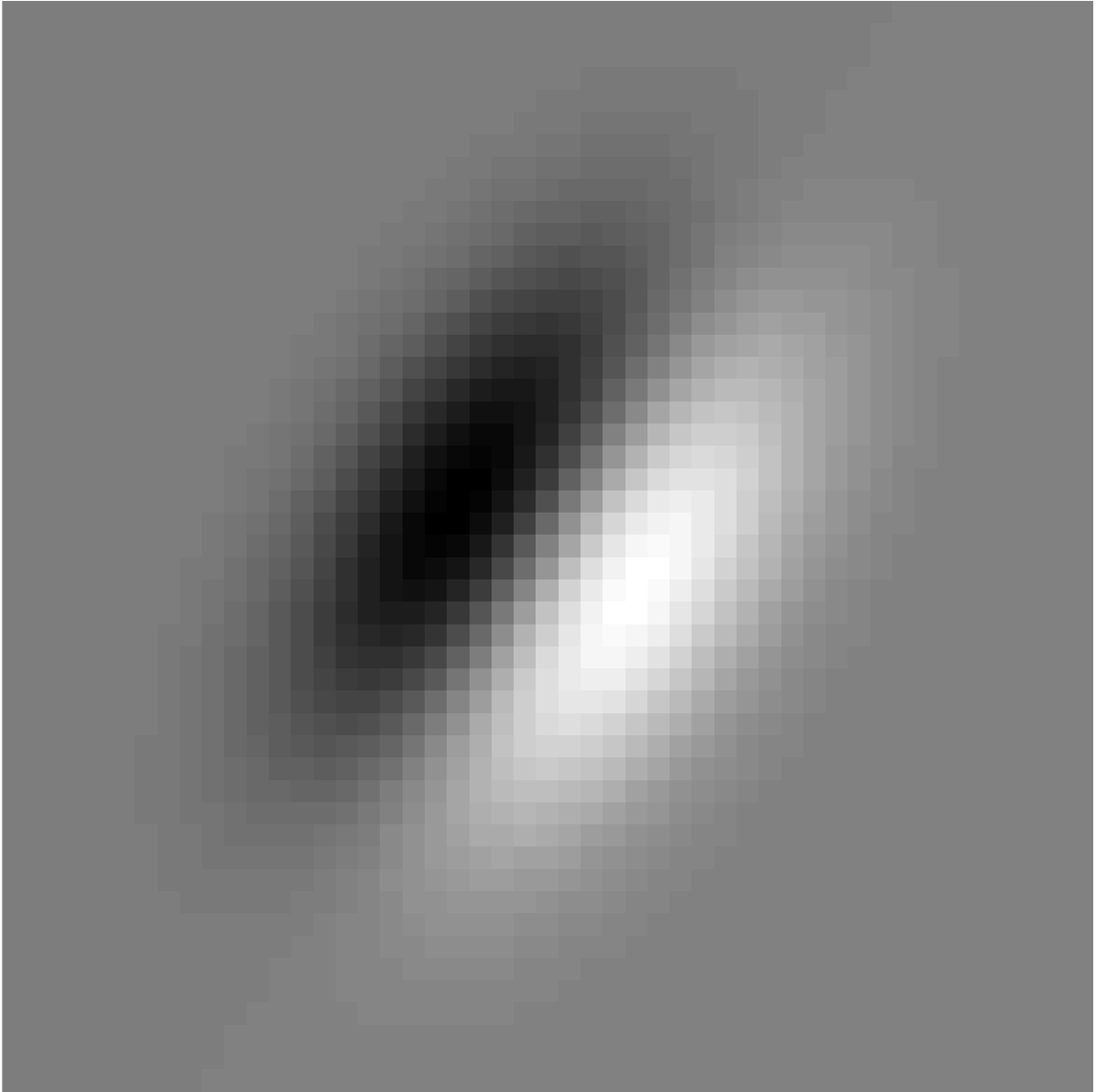} \hspace{-4mm} &
     \includegraphics[width=0.15\textwidth]{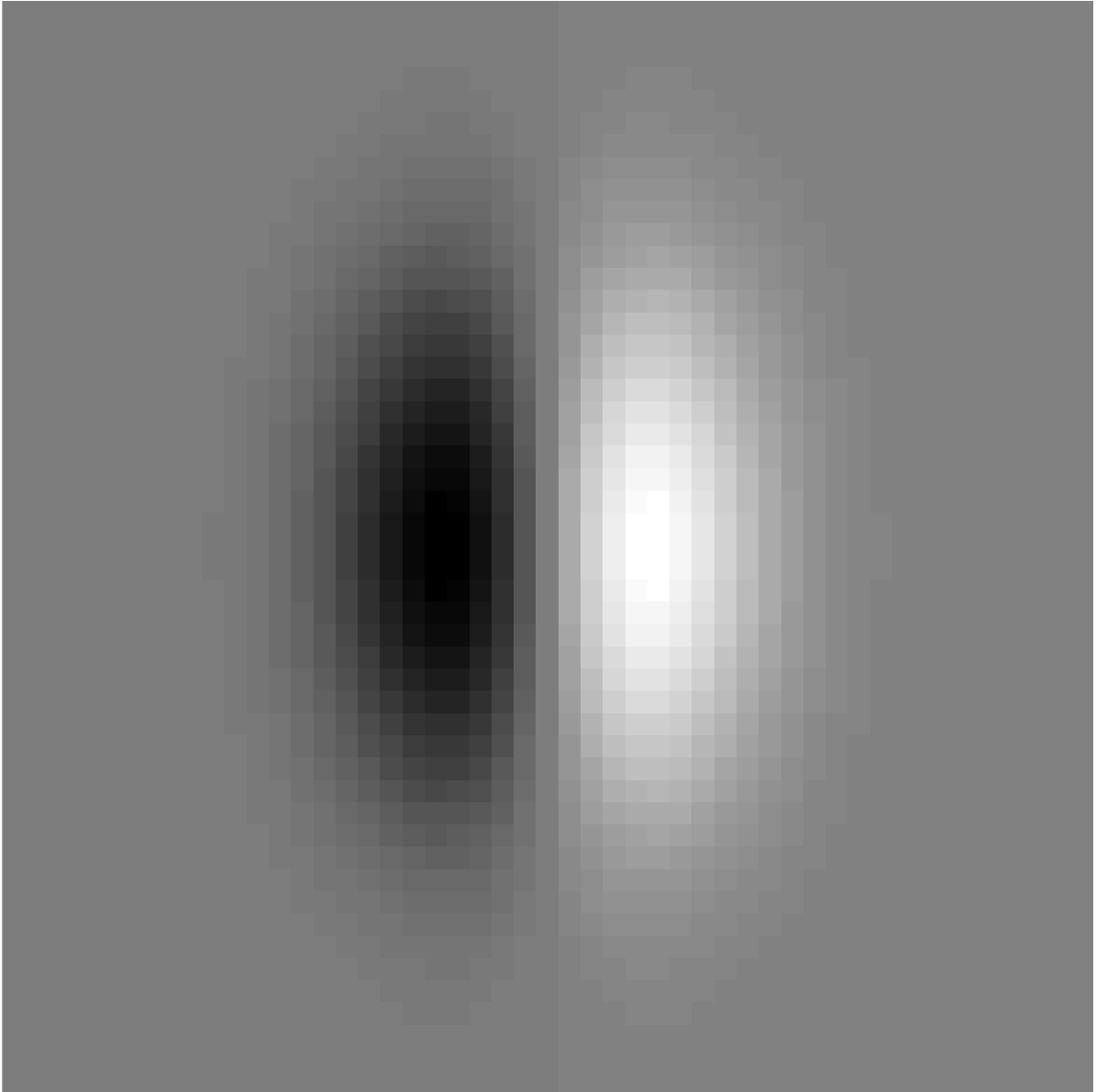} \hspace{-4mm} &
     \includegraphics[width=0.15\textwidth]{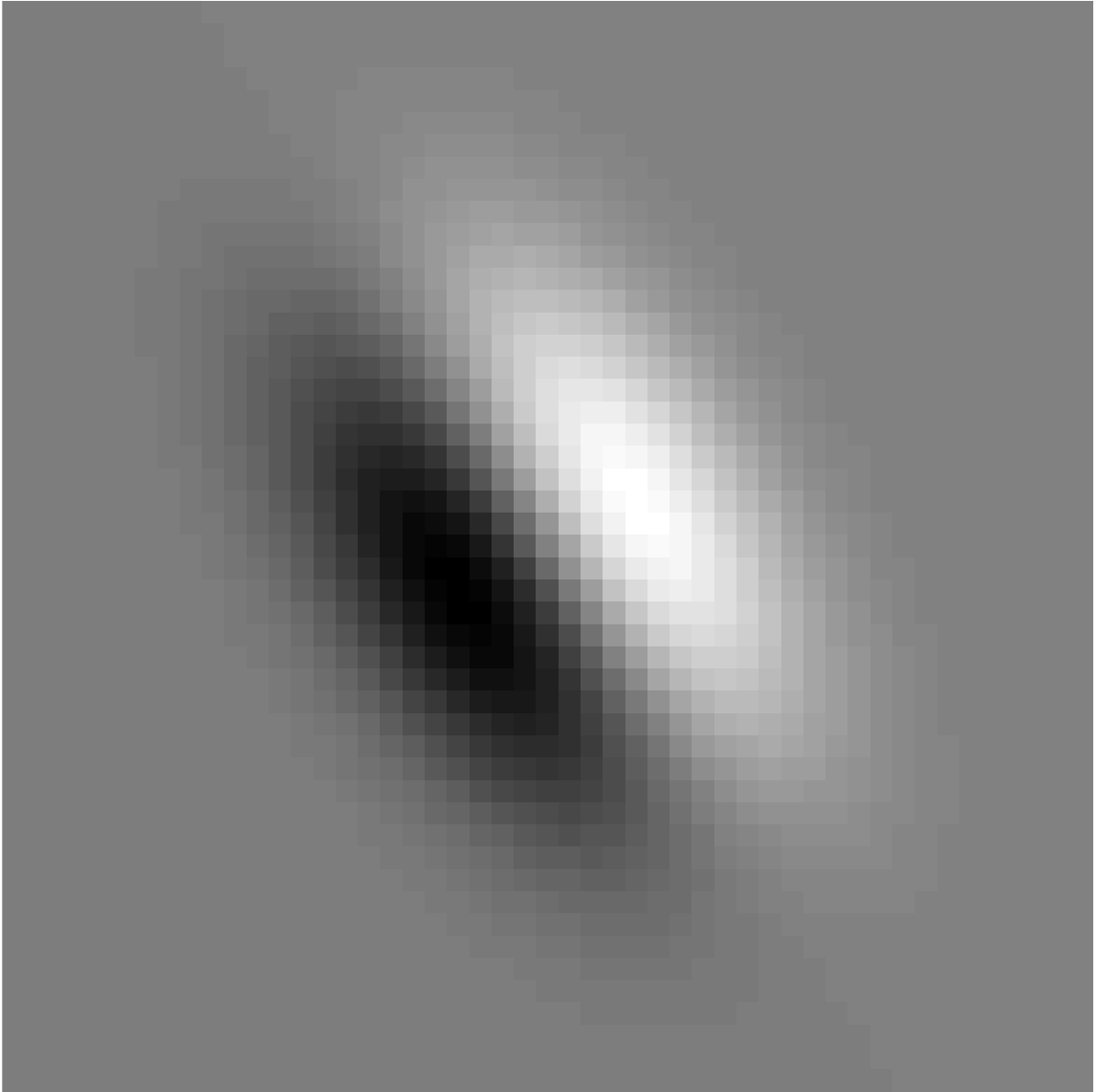} \hspace{-4mm} &
     \includegraphics[width=0.15\textwidth]{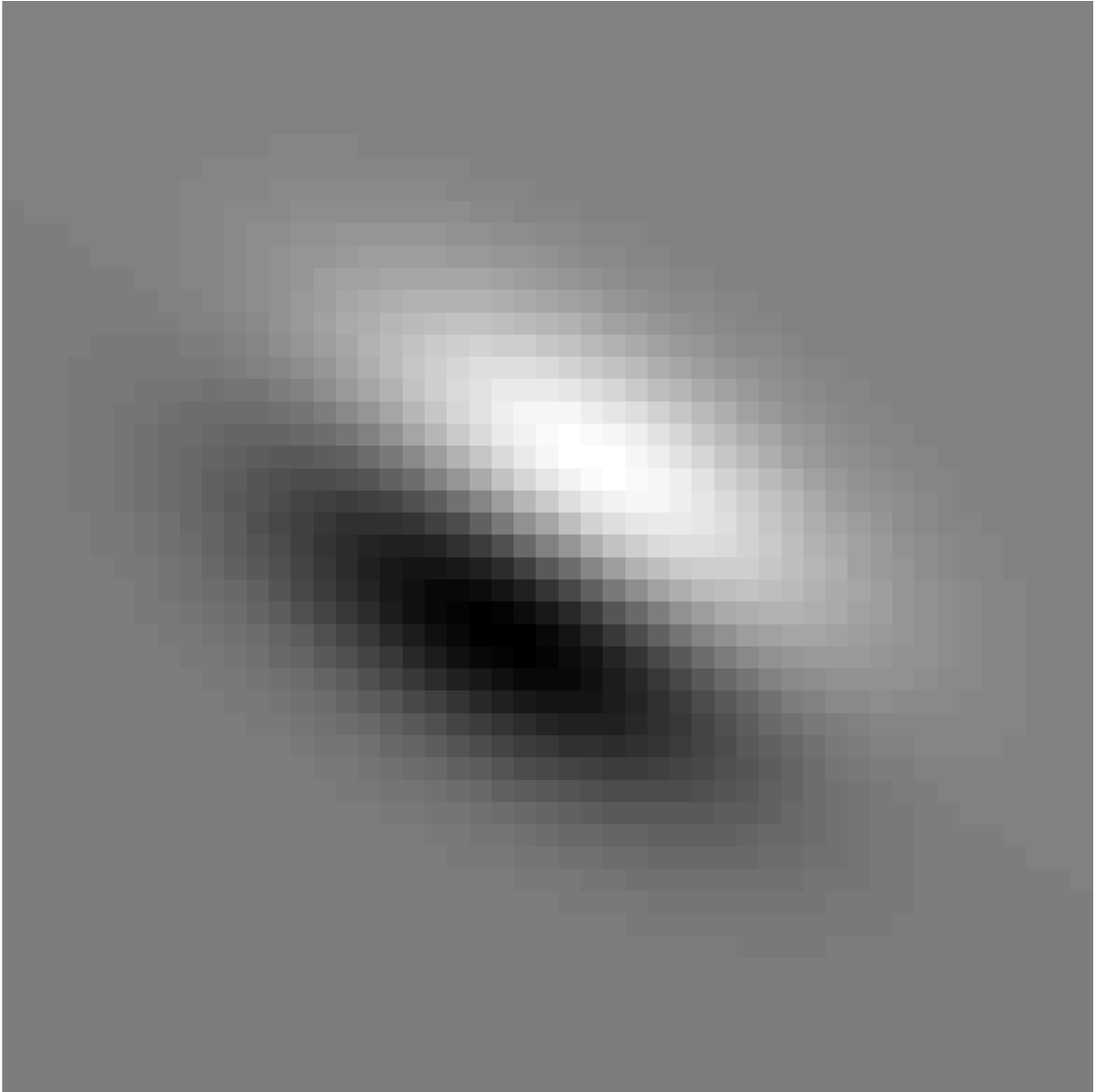} \hspace{-4mm} \\
     \hspace{-4mm}
   \includegraphics[width=0.15\textwidth]{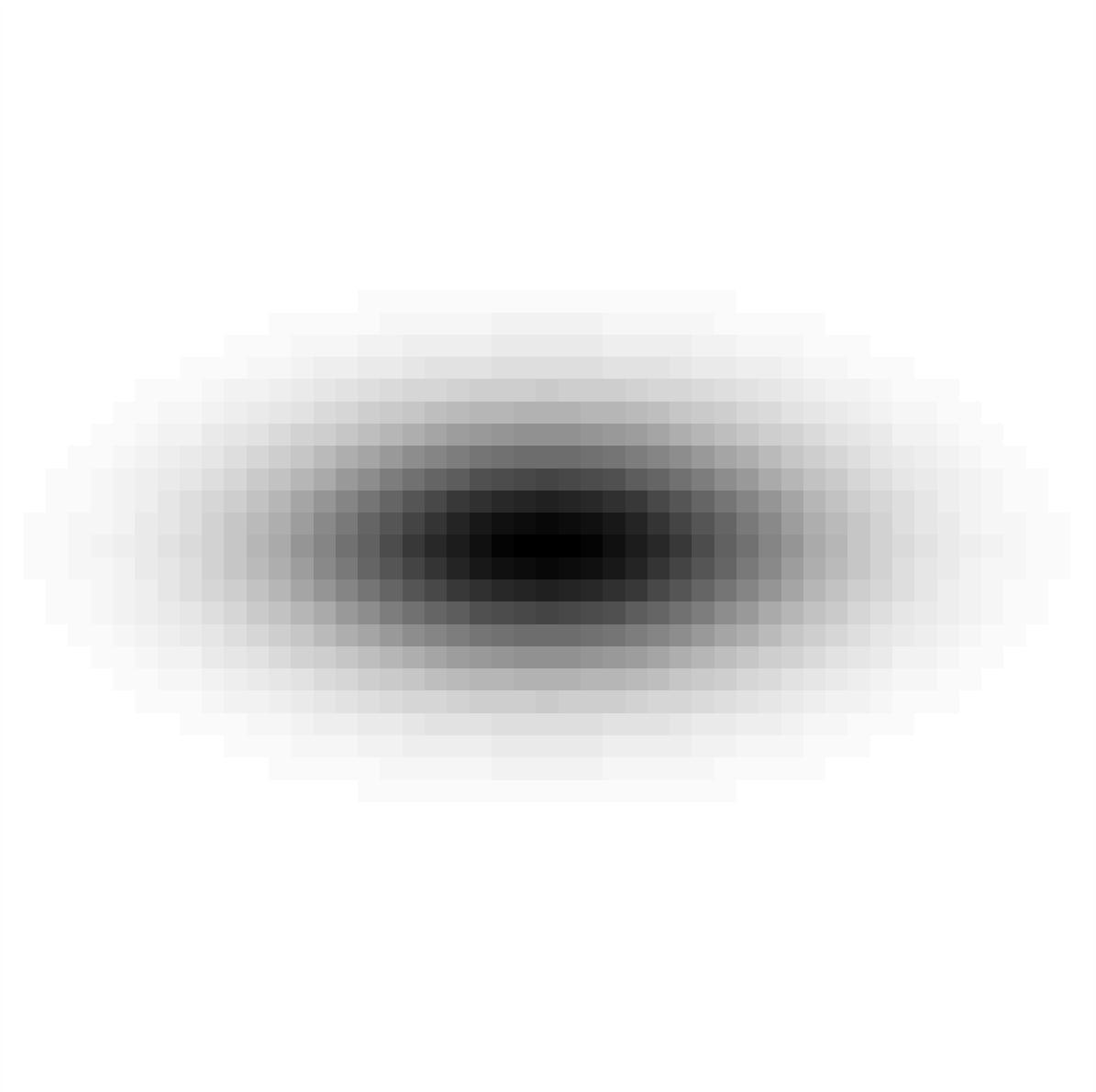} \hspace{-4mm} &
     \includegraphics[width=0.15\textwidth]{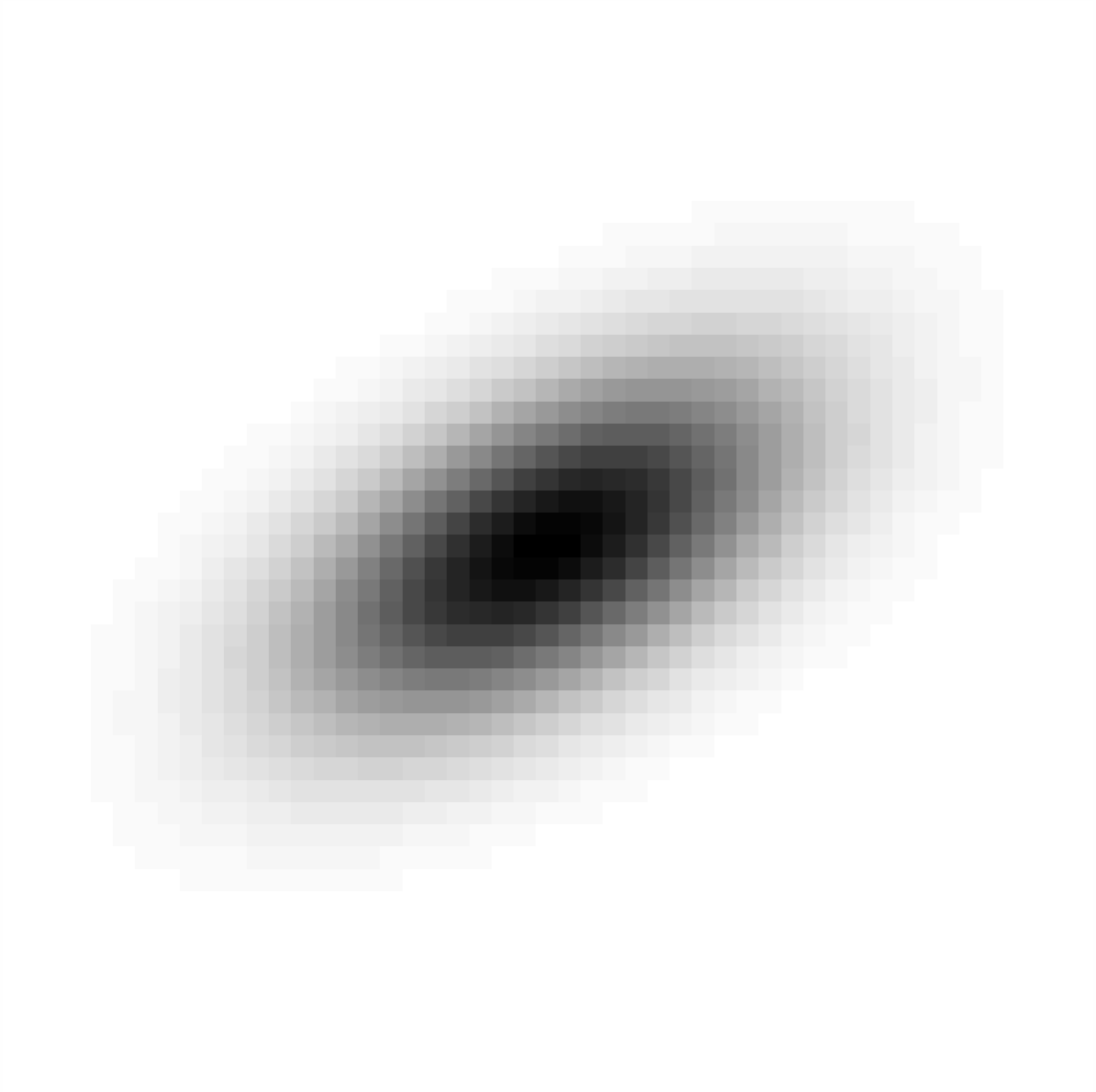} \hspace{-4mm} &
     \includegraphics[width=0.15\textwidth]{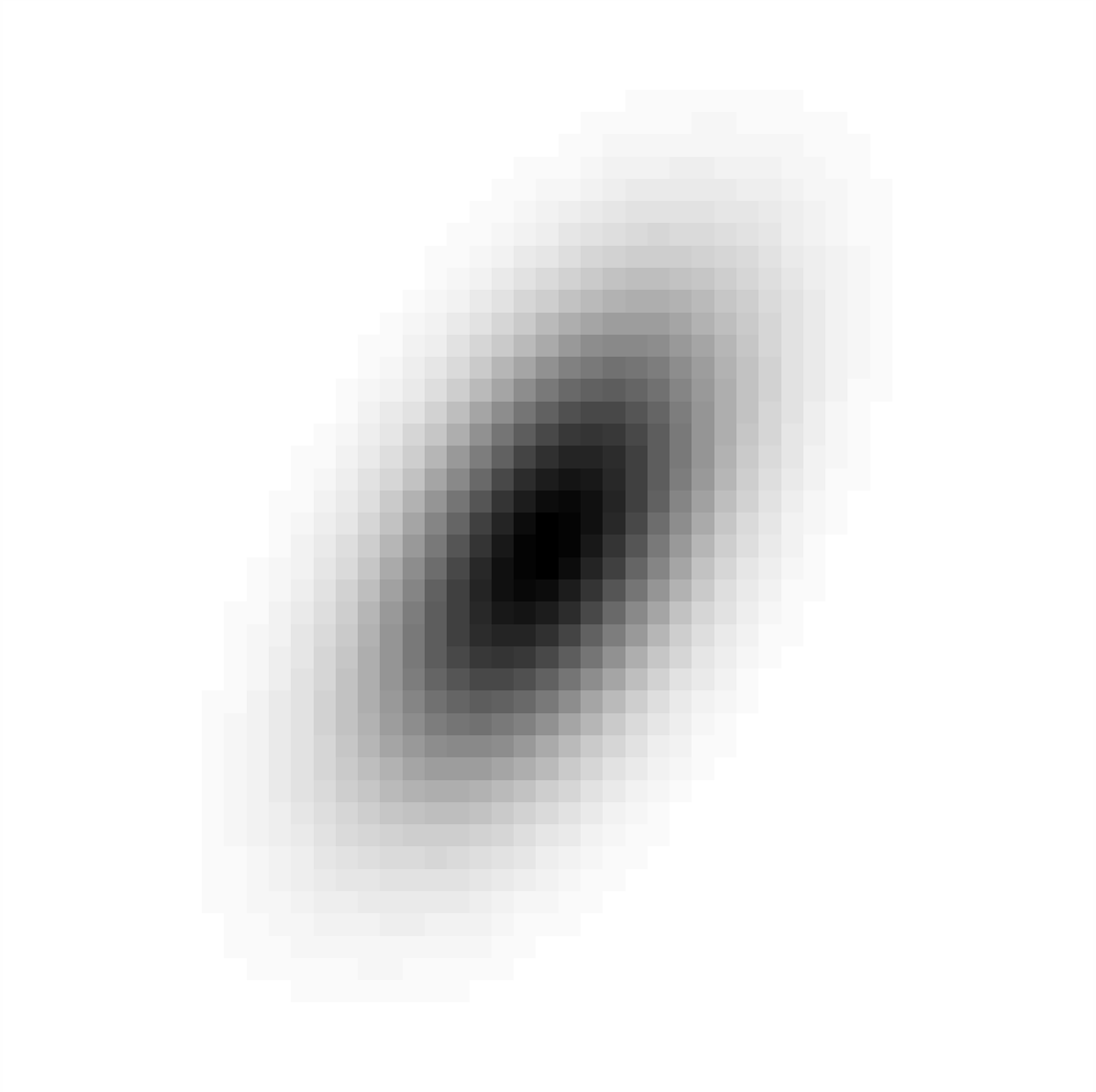} \hspace{-4mm} &
     \includegraphics[width=0.15\textwidth]{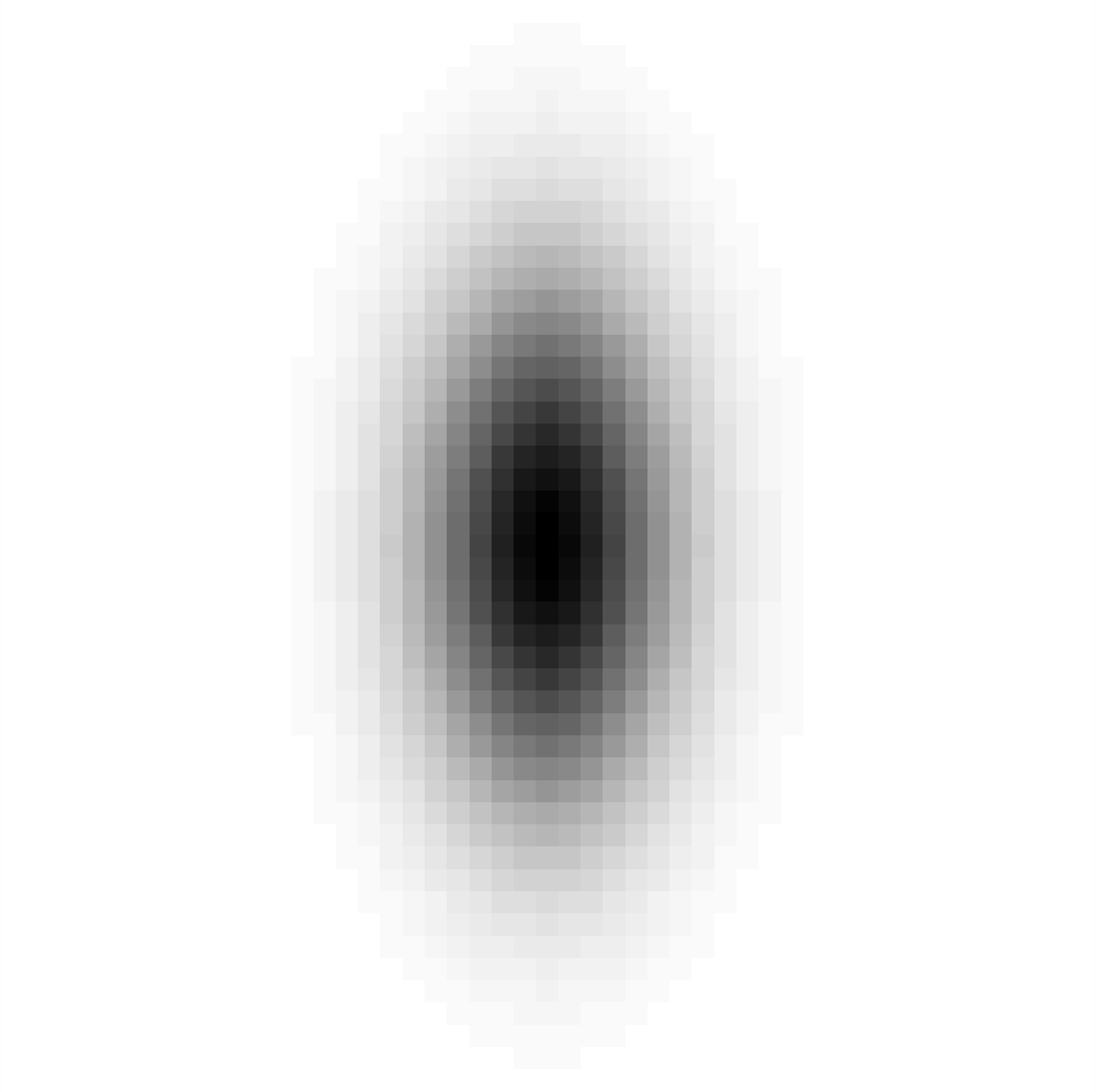} \hspace{-4mm} &
     \includegraphics[width=0.15\textwidth]{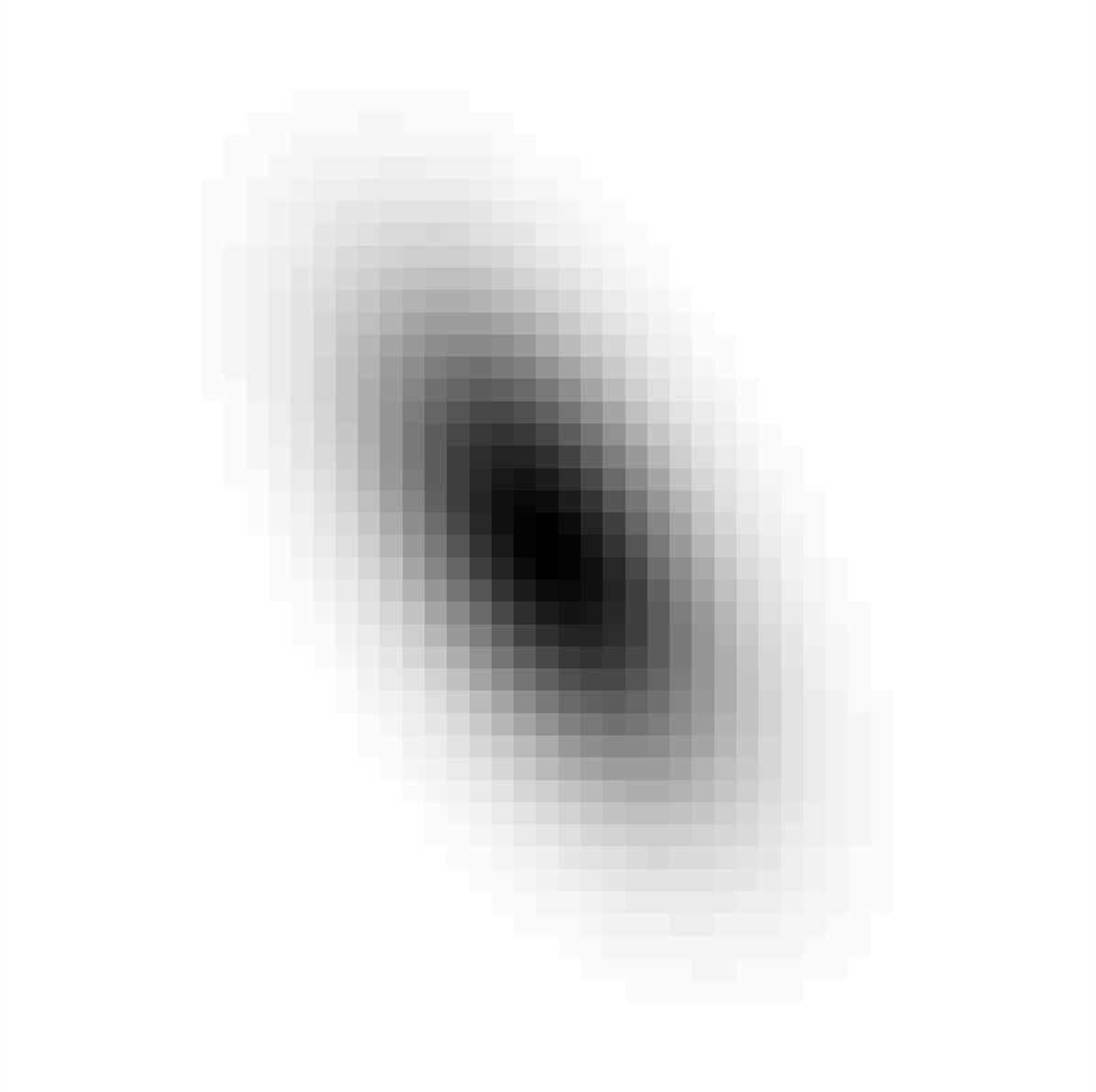} \hspace{-4mm} &
     \includegraphics[width=0.15\textwidth]{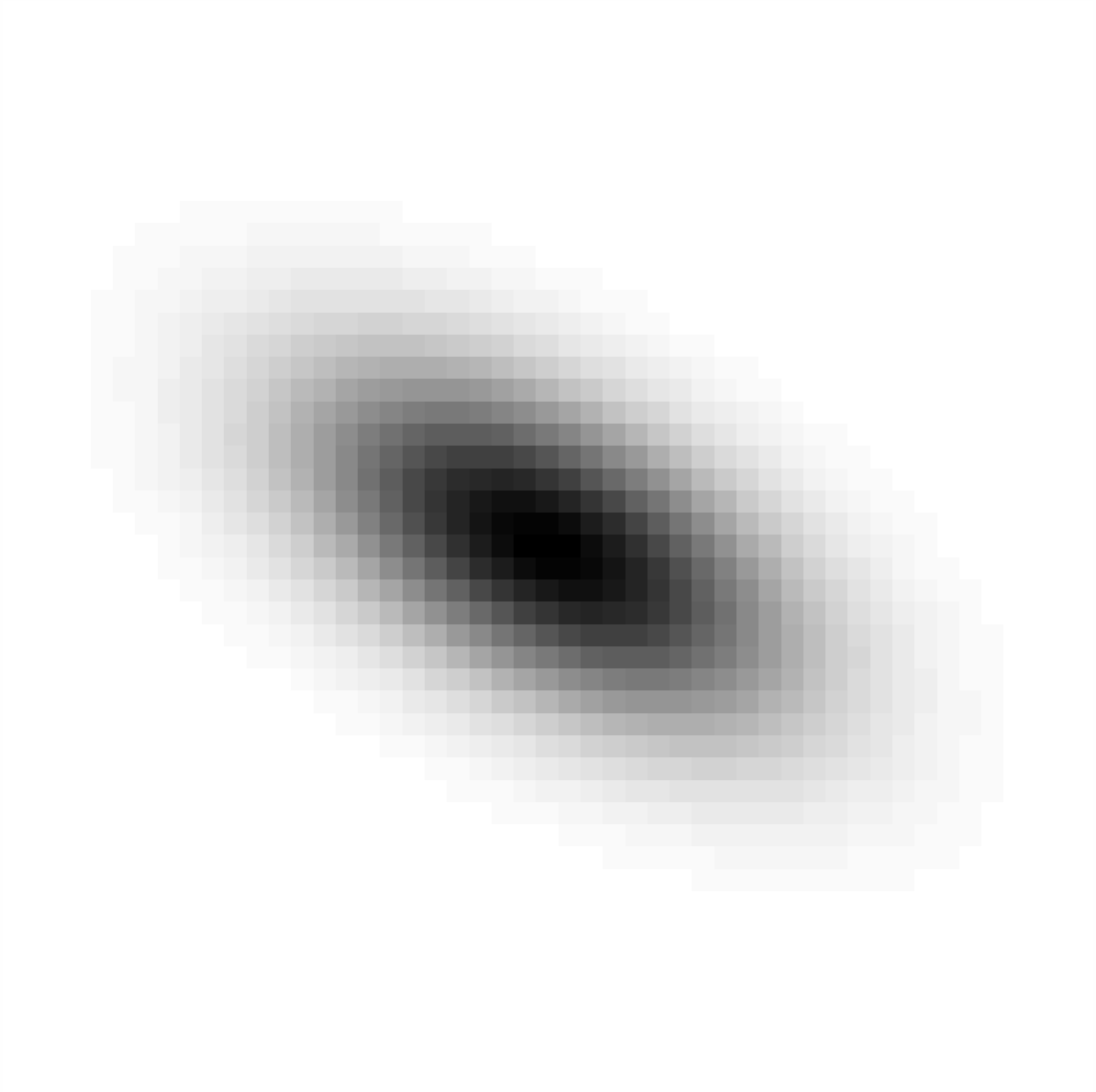} \hspace{-4mm} \\
    \end{tabular} 
  \end{center}
  \vspace{-4mm}
  \caption{Examples of equivalent convolution kernels 
    kernels $\kappa^{(l, k)}(x, y;\; \Sigma)$ with their directional
    derivative approximation kernels $\kappa^{(l, k)}_{\orth \varphi}(x, y;\; \Sigma)$ and
    $\kappa^{(l, k)}_{\orth \varphi \orth \varphi}(x, y;\; \Sigma)$ 
   up to order two in the two-dimensional case with the steepness of
   the pyramid determined by $K = 5$,
   here at resolution level $l = 2$ and iteration level $k = 2$
   corresponding to $\lambda_1 = 66$ for $\lambda_2/\lambda_1=1/4$ and
    $\alpha = 0, \pi/6, \pi/3, \pi/2, 2\pi/3, 5\pi/6$.
   (Horizontal axis: $x \in [-24, 24]$. Vertical axis: $y \in [-24, 24]$.)}
  \label{fig-equiv-aff-hybr-pyr-dir-ders-l2-k2-K5-lambda1-66}
\end{figure}

\begin{figure}[hbtp]
  \begin{center}
    {\footnotesize\em Equivalent spatio-chromatic affine hybrid pyramid kernels for $K = 3$}
    
    \medskip

    \begin{tabular}{cccccc}
     \hspace{-4mm}
     \includegraphics[width=0.15\textwidth]{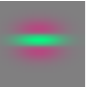} \hspace{-4mm} &
     \includegraphics[width=0.15\textwidth]{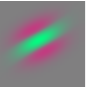} \hspace{-4mm} &
     \includegraphics[width=0.15\textwidth]{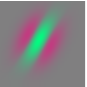} \hspace{-4mm} &
     \includegraphics[width=0.15\textwidth]{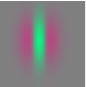} \hspace{-4mm} &
     \includegraphics[width=0.15\textwidth]{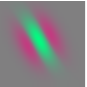} \hspace{-4mm} &
     \includegraphics[width=0.15\textwidth]{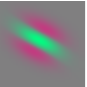} \hspace{-4mm} \\
     \hspace{-4mm}
    \includegraphics[width=0.15\textwidth]{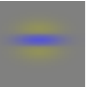} \hspace{-4mm} &
     \includegraphics[width=0.15\textwidth]{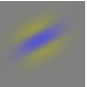} \hspace{-4mm} &
     \includegraphics[width=0.15\textwidth]{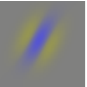} \hspace{-4mm} &
     \includegraphics[width=0.15\textwidth]{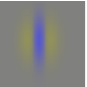} \hspace{-4mm} &
     \includegraphics[width=0.15\textwidth]{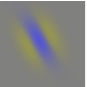} \hspace{-4mm} &
     \includegraphics[width=0.15\textwidth]{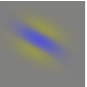} \hspace{-4mm} \\
\hspace{-4mm}
     \includegraphics[width=0.15\textwidth]{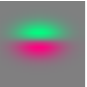} \hspace{-4mm} &
     \includegraphics[width=0.15\textwidth]{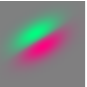} \hspace{-4mm} &
     \includegraphics[width=0.15\textwidth]{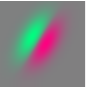} \hspace{-4mm} &
     \includegraphics[width=0.15\textwidth]{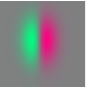} \hspace{-4mm} &
     \includegraphics[width=0.15\textwidth]{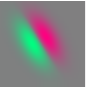} \hspace{-4mm} &
     \includegraphics[width=0.15\textwidth]{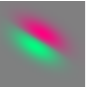} \hspace{-4mm} \\
     \hspace{-4mm}
    \includegraphics[width=0.15\textwidth]{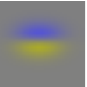} \hspace{-4mm} &
     \includegraphics[width=0.15\textwidth]{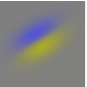} \hspace{-4mm} &
     \includegraphics[width=0.15\textwidth]{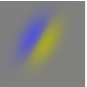} \hspace{-4mm} &
     \includegraphics[width=0.15\textwidth]{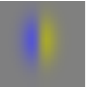} \hspace{-4mm} &
     \includegraphics[width=0.15\textwidth]{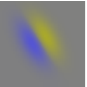} \hspace{-4mm} &
     \includegraphics[width=0.15\textwidth]{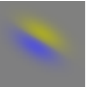} \hspace{-4mm} \\
    \end{tabular} 
  \end{center}
  \vspace{-4mm}
  \caption{Examples of equivalent spatio-chromatic directional
    derivative approximation kernels $\kappa^{(l, k)}_{\orth \varphi}(x, y;\; \Sigma)$ and
    $\kappa^{(l, k)}_{\orth \varphi \orth \varphi}(x, y;\; \Sigma)$ 
   over the red/green colour-opponent channel and
   up to order two in the two-dimensional case with the steepness of
   the pyramid determined by $K = 3$,
   here at resolution level $l = 2$ and iteration level $k = 4$
   corresponding to $\lambda_1 = 62$ for $\lambda_2/\lambda_1=1/4$ and
    $\alpha = 0, \pi/6, \pi/3, \pi/2, 2\pi/3, 5\pi/6$.
   (Horizontal axis: $x \in [-24, 24]$. Vertical axis: $y \in [-24, 24]$.)}
  \label{fig-equiv-aff-hybr-pyr-UV-dir-ders-l2-k4-K3-lambda1-62}

  \bigskip
  \bigskip

  \begin{center}
     {\footnotesize\em Equivalent spatio-chromatic affine hybrid pyramid kernels for $K = 5$}
    
    \medskip

   \begin{tabular}{cccccc}
     \hspace{-4mm}
     \includegraphics[width=0.15\textwidth]{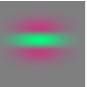} \hspace{-4mm} &
     \includegraphics[width=0.15\textwidth]{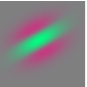} \hspace{-4mm} &
     \includegraphics[width=0.15\textwidth]{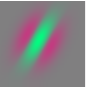} \hspace{-4mm} &
     \includegraphics[width=0.15\textwidth]{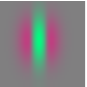} \hspace{-4mm} &
     \includegraphics[width=0.15\textwidth]{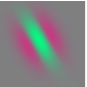} \hspace{-4mm} &
     \includegraphics[width=0.15\textwidth]{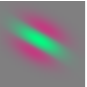} \hspace{-4mm} \\
    \hspace{-4mm}
     \includegraphics[width=0.15\textwidth]{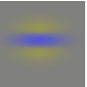} \hspace{-4mm} &
     \includegraphics[width=0.15\textwidth]{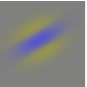} \hspace{-4mm} &
     \includegraphics[width=0.15\textwidth]{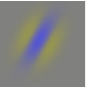} \hspace{-4mm} &
     \includegraphics[width=0.15\textwidth]{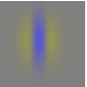} \hspace{-4mm} &
     \includegraphics[width=0.15\textwidth]{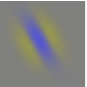} \hspace{-4mm} &
     \includegraphics[width=0.15\textwidth]{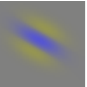} \hspace{-4mm} \\
      \hspace{-4mm}
    \includegraphics[width=0.15\textwidth]{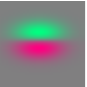} \hspace{-4mm} &
     \includegraphics[width=0.15\textwidth]{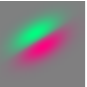} \hspace{-4mm} &
     \includegraphics[width=0.15\textwidth]{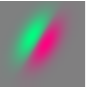} \hspace{-4mm} &
     \includegraphics[width=0.15\textwidth]{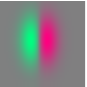} \hspace{-4mm} &
     \includegraphics[width=0.15\textwidth]{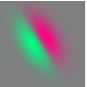} \hspace{-4mm} &
     \includegraphics[width=0.15\textwidth]{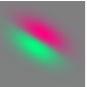} \hspace{-4mm} \\
     \hspace{-4mm}
    \includegraphics[width=0.15\textwidth]{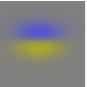} \hspace{-4mm} &
     \includegraphics[width=0.15\textwidth]{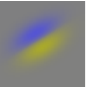} \hspace{-4mm} &
     \includegraphics[width=0.15\textwidth]{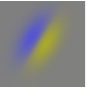} \hspace{-4mm} &
     \includegraphics[width=0.15\textwidth]{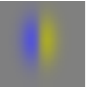} \hspace{-4mm} &
     \includegraphics[width=0.15\textwidth]{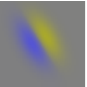} \hspace{-4mm} &
     \includegraphics[width=0.15\textwidth]{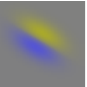} \hspace{-4mm} \\
 \end{tabular} 
  \end{center}
  \vspace{-4mm}
  \caption{Examples of equivalent spatio-temporal directional
    derivative approximation kernels $\kappa^{(l, k)}_{\orth \varphi}(x, y;\; \Sigma)$ and
    $\kappa^{(l, k)}_{\orth \varphi \orth \varphi}(x, y;\; \Sigma)$ 
   over the red/green colour-opponent channel and
   up to order two in the two-dimensional case with the steepness of
   the pyramid determined by $K = 5$,
   here at resolution level $l = 2$ and iteration level $k = 2$
   corresponding to $\lambda_1 = 66$ for $\lambda_2/\lambda_1=1/4$ and
    $\alpha = 0, \pi/6, \pi/3, \pi/2, 2\pi/3, 5\pi/6$.
   (Horizontal axis: $x \in [-24, 24]$. Vertical axis: $y \in [-24, 24]$.)}
  \label{fig-equiv-aff-hybr-pyr-UV-dir-ders-l2-k2-K5-lambda1-66}
\end{figure}

A higher value of $K$ will, in general, lead to equivalent discrete derivative
approximation kernels that constitute numerically better
approximations in relation to the discrete affine Gaussian kernels
expressed over a constant grid spacing $h = 1$ 
(see Figures~\ref{fig-equiv-aff-hybr-pyr-dir-ders-l2-k4-K3-lambda1-62}--\ref{fig-equiv-aff-hybr-pyr-dir-ders-l2-k2-K5-lambda1-66}),
whereas a lower value of $K$ will lead to higher computational efficiency.
To guarantee a sufficient minimum amount of spatial smoothing before
the spatial subsampling operation is performed, we recommend that $K$ should be
chosen at least greater than two:
\begin{equation}
   K > 2.
\end{equation}
In relation to an isotropic regular pyramid representation for 
$C_{xx} = C_{yy}$ and $C_{xy} = 0$, the lower bound on the minimum amount of spatial
smoothing $\Delta s_{tot} = K \, \Delta s = 2 \times 1/2 = 1$
required in this way corresponds to the amount of spatial smoothing
obtained by applying separable smoothing with the binomial
kernel $(1/16, 4/16, 6/16, 4/16, 1/16)$ along each dimension,
which constitutes a common way of generating classic image pyramids
(Crowley \cite{Cro81}; Burt and Adelson \cite{BA83-COM};
 Crowley and Parker \cite{Cro84-dolp,Cro84-peaks};
 Lindeberg \cite{Lin90-PAMI,Lin93-Dis}; Lindeberg and Bretzner \cite{LinBre03-ScSp}).
Since such pyramids may, however, imply undersampling of the image
data, we propose a condition of the form $K > 2$ instead of $K \geq 2$, 
with the complementary recommendation that higher values of $K$ 
will improve the performance of image analysis/computer vision
algorithms further.

\subsection{Equivalent convolution kernels and derivative approximation
  kernels}
\label{eq-equiv-conv-der-kern-aff-hybr-pyr}

Since the representation at each level is constructed from a
set of repeated smoothing and subsampling operations,
which are all linear operations,
the composed operation can equivalently be modeled
as the result of applying one kernel $\kappa^{(l, k)}$,
termed {\em equivalent convolution kernel\/}, to the original image,
followed by a pure subsampling step with 
composed subsampling factor $S = 2^l$.
If we define a dual operator to the $\reducecycle$ operator
according to
\begin{center}
  \begin{tabular}{cll}
    $\expandcycle$
      & $:=$
      & $\smooth^+$
      \\
      & & $\enlarge$
      \\
 \end{tabular}
\end{center}
where the $\enlarge$ operation enlarges any $D$-dimensional image by a
factor $S = 2$
\begin{align}
  \label{eq-enlarge}
    \begin{split}
      E
        & = \enlarge(L)
    \end{split}\\
    \begin{split}
      E(x, y)
        & = \left\{
              \begin{array}{ll}
                L(x/S, y/S)
                  & \mbox{if all indices in $x$ are multiples of $S$} \\
                0
                  & \mbox{if any index in $x$ is not a multiple of $S$}
              \end{array}
            \right.
    \end{split}
\end{align}
then the equivalent convolution kernel corresponding to 
resolution level $l$ and internal iteration level $k \in [0, K]$
within this level of resolution can be written
\begin{equation}
  \label{eq-equiv-conv-kernel}
  \kappa^{(l, k)} = \expandall(\delta^{(l, k)}),
\end{equation}
where $\delta^{(l, k)}$ is a discrete delta function at level $(l, k)$
and $\expandall$ denotes the $\expandcycle$ operators
corresponding to the set of all the $\reducecycle$ operators
used for reaching this level in the affine hybrid pyramid.

Similarly, derivative approximations can be computed by taking
the grid spacing $h$ at the current into explicit account
\begin{equation}
  \label{eq-disc-der-approx-subsampling}
  \partial_{x^r} \approx {\cal D}_{x^r} = \frac{1}{h^{|r|}} \, \delta_{x^r},
\end{equation}
at any level with resolution $h$ in the pyramid.
Then, the corresponding {\em equivalent derivative approximation kernel\/}
will be given by
\begin{equation}
  \label{eq-equiv-der-approx-kernel}
  \kappa_{x^r}^{(l, k)} = \frac{1}{h^{|r|}} \expandall(\delta_{x^r}^{(l, k)}),
\end{equation}
where higher-dimensional difference approximations $\delta_{x^r}=\delta_{{x_1}^{r_1}}\delta_{{x_2}^{r_2}}..\delta_{{\
x_D}^{r_D}}$
are expressed in terms of the one-dimensional $r$th order
difference operator
\begin{equation}
  \delta_{x^r}
  = \left\{
      \begin{array}{ll}
        (\delta_{xx})^{r/2}           & \mbox{if $r$ is even}, \\
        \delta_x \, \delta_{x^{r-1}}  & \mbox{if $r$ is odd}.
      \end{array}
    \right.
\end{equation}
Figures~\ref{fig-equiv-aff-hybr-pyr-dir-ders-l2-k4-K3-lambda1-62}--\ref{fig-equiv-aff-hybr-pyr-dir-ders-l2-k2-K5-lambda1-66}
show examples of equivalent convolution kernels and equivalent
derivative approximation kernels over the grey-level channel for the
cases of $K = 3$ and $K = 5$.
Figures~\ref{fig-equiv-aff-hybr-pyr-UV-dir-ders-l2-k4-K3-lambda1-62}--\ref{fig-equiv-aff-hybr-pyr-UV-dir-ders-l2-k2-K5-lambda1-66}
show corresponding equivalent
derivative approximation kernels over the red/green and yellow/blue
colour-opponent channels.

\subsection{Scale-normalized derivative approximations}

To compute scale-normalized derivatives in an affine hybrid pyramid representation
at iteration level $k$ within resolution level $l$,
we can in a corresponding manner as in
Section~\ref{sec-sc-norm-ders-disc}
define {\em variance-normalized\/} directional derivative approximations
according to
\begin{equation}
  L_{\varphi^m \orth \varphi^n,norm}(x, y;\; \Sigma_s) 
  = \lambda_1^{m\gamma_1/2} \, \lambda_2^{n\gamma_2/2} \, 
      \delta_{\varphi^m \orth \varphi^n} \, L(x, y;\; \Sigma_s),
\end{equation}
where $\delta_{\varphi^m \orth \varphi^n}$ denotes the directional
derivative approximation operator generalizing the definitions in
(\ref{eq-dir-der-phi})--(\ref{eq-dir-der-phiorthphiorth}) 
with the implicit understanding that $\delta_{\varphi^2} = \delta_{\varphi\varphi}$
and $\delta_{\orth\varphi^2} = \delta_{\orth\varphi\orth\varphi}$, etc.
As spatial scale normalization parameters $\lambda_1$ and $\lambda_2$,
we should take the eigenvalues of the spatial covariance matrix
of the equivalent convolution kernel $\kappa^{(l, k)}$ according 
to (\ref{eq-equiv-conv-kernel}).
With the proposed way of defining affine hybrid pyramids according to (\ref{eq-reduce-cycle-aff-hybr-pyr}),
the spatial covariance matrix of the
equivalent derivative approximation kernel will for every resolution
level $l$ and iteration level $k$ be
equal to the spatial covariance matrix predicted by 
the discretization theory, if we multiply the increments
in $C_{xx}$, $C_{xy}$ as $C_{yy}$ as obtained from the iteration
kernel (\ref{eq-disc-3x3-kernel}) with overall scale step $\Delta s$ as used as the primitive
smoothing kernel in the reduction cycle
(\ref{eq-reduce-cycle-aff-hybr-pyr}) by $h^2$ with $h$ according to (\ref{eq-h-fcn-of-l-hybr-pyr}).

The definition of scale-normalized affine Gaussian
derivative approximations from $l_p$-normalization 
is analogous to the corresponding case without spatial subsampling 
(\ref{eq-sc-norm-der-raw-def-alpha-Lp-norm})
\begin{equation}
  L_{\varphi^m \orth \varphi^n,norm}(x, y;\; \Sigma_s) 
  = \alpha(\lambda_1, \gamma_1, \lambda_2, \gamma_2) \,
      \delta_{\varphi^m \orth \varphi^n} \, L(x, y;\; \Sigma_s),
\end{equation}
with the only difference that the expression
$\| \delta_{\varphi}^m \; \delta_{\orth \varphi}^n \, h(x, y;\; \Sigma_s)  \|_p$
in (\ref{eq-sc-norm-der-def-alpha-Lp-norm}) that defines
the scale-normalization factor $\alpha(\lambda_1, \gamma_1, \lambda_2, \gamma_2)$
as function of the spatial scale parameters 
should be replaced by $\| \kappa_{\varphi^m\orth \varphi^n} ^{(l, k)} \|_p$
\begin{equation}
  \label{eq-sc-norm-der-def-alpha-Lp-norm-aff-hybr-pyr}
   \alpha(\lambda_1, \gamma_1, \lambda_2, \gamma_2) \, 
   \| \kappa_{\varphi^m\orth \varphi^n} ^{(l, k)} \|_p
  = \lambda_1^{m\gamma_1/2} \, \lambda_2^{n\gamma_2/2} \,
    \| \partial_{\varphi}^m \partial_{\orth \varphi}^n \; g(x, y;\;  \Sigma_s) \|_p
\end{equation}
with the equivalent discrete
derivative approximation operator $\kappa_{\varphi^m\orth \varphi^n} ^{(l, k)}$ defined by expanding
the discrete directional derivative approximation operator
$\delta_{\varphi^m \orth \varphi^n}/h^{m+n}$ at the corresponding
level of resolution according to
(\ref{eq-equiv-der-approx-kernel})
\begin{equation}
  \kappa_{\varphi^m\orth \varphi^n}^{(l, k)} = \frac{1}{h^{m+n}} \expandall(\delta_{\varphi^m \orth \varphi^n}^{(l,k)})
\end{equation}
and with $h \geq 1$ denoting the grid spacing at the current level of resolution.

\subsection{Measuring the subsampling rate}
\label{sec-sub-sampl-rate-aff-hybr-pyr}

To describe how the grid spacing $h$
depends on the scale parameter in a hybrid pyramid,
we can introduce a {\em subsampling factor\/} $\rho$ from
the relation
\begin{equation}
  \label{eq-def-subsampl-rate-rho}
  h_{max}(\lambda_1, \lambda_2, \rho) 
  = \rho \, \sqrt{s} \, \min(\sqrt{\lambda_1}, \sqrt{\lambda_2}),
\end{equation}
assuming that the overall scale parameter $s$ represents the total
amount of smoothing with the effects of varying grid spacing taken into account,
so that the contribution to $s$ at any resolution level $l$ is given by
\begin{equation}
  \Delta s_{tot} = \left. h(l)^2 \, K \, \Delta s\right|_{h=1} 
\end{equation}
and with the eigenvalues $\lambda_1$ and $\lambda_2$
normalized such that $\lambda_1 = 1$ and $\lambda_2 \in [0, 1]$.

For reasons of computational efficiency, we define the actual grid
spacing as the maximum power of two that does not exceed this
upper bound
\begin{equation}
  \label{eq-def-h-of-rho-t}
  h(s, \Sigma, \rho)
  =
  \left\{
    \begin{array}{ll}
      \max_{h' = 2^{i-1} \colon i \in \bbbz_+ \backslash \{ 0 \} }
        h' : h' \leq  h_{max}(\lambda_1, \lambda_2, \rho)
        & \mbox{if $h_{max} \geq 1$}, \\
      1
        & \mbox{otherwise}.
    \end{array}
  \right.
\end{equation}
Thus, a subsampling factor of $\rho = 0$ corresponds to preserving
the original resolution at all levels of scales, whereas increasing
values of $\rho$ correspond to higher degrees of subsampling at
coarser levels of scale.


\bibliographystyle{spmpsci}
{\footnotesize
\bibliography{bib/defs,bib/tlmac}
}
\end{document}

%% file: discaffkernels-siam-subm-shared.tex
\usepackage{amsfonts}
\ifpdf
  \DeclareGraphicsExtensions{.eps,.pdf,.png,.jpg}
\else
  \DeclareGraphicsExtensions{.eps}
\fi

\numberwithin{theorem}{section}

\newcommand{\TheTitle}{Discrete approximations of the affine Gaussian 
              derivative model for visual receptive fields} 
\newcommand{\TheAuthors}{Tony Lindeberg}

\headers{Discrete affine Gaussian receptive fields}{\TheAuthors}

\title{{\TheTitle}\thanks{
\funding{The support from the Swedish Research Council 
              (contract 2014-4083) is gratefully acknowledged.}}}

\author{
  Tony Lindeberg\thanks{Computational Brain Science Lab,
        Department of Computational Science and Technology,
        School of Computer Science and Communication,
        KTH Royal Institute of Technology,
        SE-100 44 Stockholm, Sweden 
    (\email{tony@kth.se}, \url{https://www.kth.se/profile/tony}).}
}

\usepackage{amsopn}

\newcommand{\bbbr}{{\mathbb R}} 
\newcommand{\bbbz}{{\mathbb Z}}

\newcommand{\orth}{\bot}

\newcommand\tlthmheadfont{}

\newenvironment{myproof}{\noindent{\em Proof:} }{\hfill $\square$ \par}

\newcommand{\subsample}{\operatorname{\mbox{\sc SubSample}}}
\newcommand{\smooth}{\operatorname{\mbox{\sc Smooth}}}
\newcommand{\deltasmooth}{\operatorname{\mbox{\sc DeltaSmooth}}}

\newcommand{\reducecycle}{\operatorname{\mbox{\sc ReduceCycle}}}
\newcommand{\expandcycle}{\operatorname{\mbox{\sc ExpandCycle}}}

\newcommand{\op}{\operatorname{\mbox{\sc Op}}}
\newcommand{\enlarge}{\operatorname{\mbox{\sc Enlarge}}}
\newcommand{\expandall}{\operatorname{\mbox{\sc ExpandAll}}}
